\documentclass[twoside,12pt]{Latex/Classes/PhDthesisPSnPDF_fGrande}
\usepackage[boxruled,vlined,linesnumbered,spanish]{algorithm2e}
\renewcommand*{\bibfont}{\footnotesize}

\ifpdf  
    \pdfcatalog { /PageMode (/UseOutlines)
                  /OpenAction (fitbh)  }
\fi
       
\begin{document}
\selectlanguage{spanish}


\title{
Diseño e implementación de una meta-heurística multi-poblacional de optimización combinatoria enfocada a la resolución de problemas de asignación de rutas a vehículos.}

\crest{\includegraphics[width=2cm]{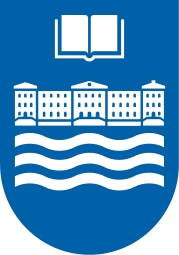}}
\university{Universidad de Deusto}
\degree{Tesis doctoral por}
\author{ ENEKO OSABA ICEDO} 
\collegeordept{dentro del Programa de Doctorado en Ing. Informática y Telecomunicación}
\textadvisor{Director }
\advisor{FERNANDO DÍAZ MARTÍN }
\textsignaturecandidate{El doctorando}
\textsignatureadvisor{El director}
\cityofbirth{Bilbao}

\degreedate{Julio de \the\year}

\renewcommand\baselinestretch{1.2}
\baselineskip=18pt 

\maketitle  


\thispagestyle{empty}

\hfill

\vfill

\medskip

\noindent

\textit{
Diseño e implementación de una meta-heurística multi-poblacional de optimización combinatoria enfocada a la resolución de problemas de asignación de rutas a vehículos.}

\vspace{5mm}

Autor: Eneko Osaba

Director: Fernando Díaz

\vfill
\vfill
\noindent
\noindent
Texto impreso en Bilbao

\noindent
Primera edición, 
Julio de \the\year

\vspace{1cm}
\hrule
\bigskip

\cleardoublepage

\frontmatter


\begin{dedication} 

\textit{Para todos los que algún día}
 
\textit{no se vieron capaces de hacer algo.}

\end{dedication}



\begin{abstracts}        
\selectlanguage{british}

Transportation is an essential area in the nowadays society, both for business sector and citizenry. There are different kinds of transportation systems, each one with its own characteristics. In the same way, various areas of knowledge can deal efficiently with the transport planning, whether entrepreneurial, or urban. Concretely, this thesis is focused in the area of artificial intelligence and optimization problems.

The majority of the problems related with the transport and logistics have common characteristics. This means that they can be modeled as optimization problems, being able to see them as special cases of other generic problems. These problems fit into the combinatorial optimization field. Much of the problems of this type have an exceptional complexity, requiring the employment of techniques for its treatment. There are different sorts of these methods. Specifically, this work will be focused on meta-heuristics.

A great amount of meta-heuristics can be found the literature, each one with its advantages and disadvantages. Due to the high complexity of combinatorial optimization problems, there is no technique able to solve all these problems optimally. This fact makes the fields of combinatorial optimization and vehicle routing problems be a hot topic of research.

Therefore, this doctoral thesis will focus its efforts on developing a new meta-heuristic to solve different kind of vehicle routing problems. The presented technique offers an added value compared to existing methods, either in relation to the performance, and the contribution of conceptual originality.

With the aim of validating the proposed model, the results obtained by the developed meta-heuristic have been compared with the ones obtained by other four algorithms of similar philosophy. Four well-known routing problems have been used in this experimentation, as well as two classical combinatorial optimization problems. In addition to the comparisons based on parameters such as the mean, or the standard deviation, two different statistical tests have been carried out: the normal \textit{z}-test, and the Friedman test. Thanks to these tests it can be affirmed that the proposed meta-heuristic is competitive in terms of performance and conceptual originality.
\end{abstracts}


\begin{resumen}        
\selectlanguage{spanish}

El transporte es un factor crucial para la sociedad actual, tanto para la ciudadanía como para el sector empresarial. Existe una gran variedad de sistemas de transporte, cada uno con sus ventajas e inconvenientes. Del mismo modo, diversas áreas de conocimiento están capacitadas para hacer frente a la planificación eficiente del transporte, ya sea de carácter empresarial, o urbano. En este trabajo se centrará la atención en una de esas áreas: la inteligencia artificial y los problemas de optimización.

La gran mayoría de los problemas surgidos en el área del transporte y la logística tienen características comunes, lo que hace que puedan modelizarse como problemas de optimización y que puedan ser vistos como casos particulares de otros problemas genéricos. Estos problemas se encuadran en el ámbito de la optimización combinatoria. Gran parte de los problemas de este tipo cuentan con una complejidad excepcional, haciendo necesaria la utilización de técnicas para su tratamiento. Existen diferentes métodos de esta índole, aunque este trabajo se centrará en las meta-heurísticas.

Puede encontrarse un vasto número de meta-heurísticas en la literatura actual, cada una con sus propias características. Debido a la alta complejidad de los problemas a tratar, no hay una técnica definitiva que consiga resolver de forma óptima todos los problemas de optimización. Este hecho hace que el campo tanto de la optimización combinatoria, como el de los problemas de asignación de rutas de transporte, sea un tema de investigación candente.

Con todo esto, el presente trabajo de tesis doctoral centrará su esfuerzo en la creación de una nueva meta-heurística que ayude a la resolución de problemas de asignación de rutas a vehículos. La técnica presentada ofrece un valor añadido frente a los métodos existentes, ya sea en relación al rendimiento, o a la aportación de originalidades conceptuales.

Con la intención de validar el modelo, se han comparado los resultados de la meta-heurística propuesta con los obtenidos por otros cuatro algoritmos de filosofía similar. Para esta experimentación se han utilizado cuatro famosos problemas de asignación de rutas a vehículos, y dos problemas clásicos de la optimización combinatoria. En la comparativa se ha realizado un completo análisis estadístico sobre la calidad de las soluciones obtenidas y la convergencia temporal de cada técnica, utilizando para ello pruebas estadísticas parámetricas, como el test normal \textit{z}, y pruebas no paramétricas, como el test de Friedman. Gracias a estos tests se ha podido afirmar de forma rigurosa que la técnica propuesta es competitiva en cuanto a rendimiento se refiere, y original conceptualmente hablando.

\end{resumen}


\begin{laburpena}        
\selectlanguage{basque}

Garraioek garrantzi handia daukate gaur egungo gizartean, bai herritarrentzat, bai enpresentzat. Hainbat garraio sistema existitzen dira, bakoitzak bere abantaila eta eragozpenekin. Era berean, hainbat ezaguera eremuk gaitasuna daukate garraioen plangintzari modu eraginkor batean aurre egiteko. Lan honen arreta adimen artifizialean eta optimizazio problemetan ardaztuko da.

Garraio eta logistika eremuetan sortutako problema gehienek ezaugarri komunak dauzkate. Gertaera honek optimizazio problemak bezala modelatzea ahalbidetzen du, horrela, beste optimizazio problema generiko batzuen kasu bereziak bezala ikusi ahal dira. Problema hauek optimizazio konbinatorioaren arloan sartu ahal dira. Optimizazio mota honetan azaldutako problemek ohiz kanpoko konplexutasuna daukate, teknika berezien erabilpena behartuz modu egoki batean aurre egiteko. Honen ildoan, era honetako hainbat metodo existitzen dira, hala ere, lan hau meta-heuristiketan zentratuko da.

Meta-heuristika ugari aurkitu ahal dira gaur egungo literaturan, bakoitzak bere ezaugarriekin. Aipatutako arloan sortutako problemen konplexutasun handia dela eta, ezinezkoa da optimizazio problema guztiak modu ezin hobean ebazten duen teknika bat aurkitzea. Gertakari honek optimizazio konbinatorioaren eta ibilgailuentzako ibilbideen esleipenaren problemen eremuak ikerketa arlo erakargarriak izatea eragiten du.

Honekin guztiarekin, doktore-tesi hau ibilgailuentzako ibilbideen esleipenaren problemak ebazteko meta-heuristika berri baten sorkuntzan ardaztuko da. Aurkeztutako algoritmoak balio erantzi bat eskainiko du existitzen diren metodoen aurrean, bai errendimenduaren aldetik, bai originaltasun-konzeptualaren ikuspegitik.

Planteatutako hipotesia egiaztatzearen asmoarekin, proposatutako meta-heuristikak eskuratutako emaitzak filosofia bereko beste lau algoritmo desberdinak lortutakoekin konparatuko dira. Esperimentazio hau burutzeko ezagunak diren ibilgailuentzako ibilbideen esleipenaren lau problema diferente erabili dira, hala nola optimizazio konbinatorio arloan egokitutako bi problema klasiko ezberdin. Emaitzen konparaketa hainbat parametroen arabera egin da, batezbesteko aritmetikoa eta desbideratze tipikoaren arabera, adibidez. Horrez gain, bi azterketa estatistikoak egin dira lortutako emaitzekin: \textit{z}-normalaren proba, eta Friedmanen proba. Gauzatutako esperimentazioari esker, proposatutako meta-heuristika etorkizun handiko teknika bat dela zentzuz baieztatu ahal da.

\end{laburpena}


\selectlanguage{spanish}
\begin{acknowledgements}      

Siendo sinceros, tras embarcarme en esta aventura jamás pensé que el apartado de agradecimientos sería uno de los que más quebraderos de cabeza me iba a generar. Esto es debido a la claustrofóbica necesidad de congregar en este par de páginas a todas las personas que merecen ser mencionadas. Soy consciente de que lo típico de este apartado es comentar en un primer momento lo difícil, tortuoso y escabroso que ha sido la elaboración de la tesis, para después dar paso a los meritorios agradecimientos. Lo siento, en mi caso no ha sido así. No puedo decir que hayan sido cuatro años caminando sobre un sendero cubierto de rosas, pero si puedo asegurar que he disfrutado de todos y cada uno de los momentos y que, realmente, no he tenido en ningún momento la sensación de estar trabajando. Quizá sea por eso por lo que me gustaría que mis agradecimientos suenen sinceros, debido a que, en gran parte, esto ha sido gracias a toda esta gente. 

Me gustaría comenzar dedicando un agradecimiento a mi familia. A mi ama \textit{Mari Feli} y a mi aita \textit{Paco}, los cuales me han apoyado siempre, en todo, de manera incondicional, y que no solo me han dado una educación que considero excelente, si no que me han brindado la oportunidad y los medios para recibir una educación académica de nivel. Gracias a ellos y a mi admirado hermano \textit{Borja} soy quien soy hoy en día. Por supuesto, a mi novia y compañera incondicional \textit{Nahikari Abaunza}, que ha seguido más de cerca todo el desarrollo de la tesis desde el primer día hasta el último, y cuyo apoyo he notado a lo largo de estos cuatro años de forma continua. Eskerrik asko Nitxu!

Comenzando con los agradecimientos ``laborales", en primer momento agradeceré a \textit{Asier Perallos} el haberme brindado la oportunidad de trabajar todos estos años en este maravillo grupo de trabajo que es Mobility, y por haberme ofrecido la oportunidad para dar mis primeros pasos en el mundo de la docencia. Gracias también a \textit{Enrique Onieva}, por todo el conocimiento que me ha aportado y por darme la oportunidad de trabajar con el.

A todos mis compañeros de departamento, gracias a los cuales el día a día en la universidad ha sido realmente maravilloso. Para ellos debe ser uno de los agradecimientos más sinceros, por haber creado en mi memoria muchísimos momentos inolvidables, risas, fiestas, alegrías y alguna que otra gamberrada... \textit{Sanni}, \textit{Pablo}, \textit{Itziar}, \textit{Browner}, \textit{Ander}, \textit{Gorka}, \textit{Lucio}, \textit{Luis}, \textit{Bruno}, \textit{Hugo}, \textit{Unai H.}, \textit{Laura}, \textit{Txertu}, \textit{Isabel}, \textit{Alfonso}, \textit{Nacho},\textit{Pili}... Muchos de vosotros habéis traspasado las fronteras de \textit{compañero de trabajo} para convertiros en algo más. A todos vosotros, muchísimas gracias, de corazón. Entre ellos, me gustaría realizar un agradecimiento personalizado a dos grandes personas: \textit{Idoia} y \textit{Peru}. Gracias a los dos por ayudarme en la revisión de esta tesis, vuestros comentarios han sido realmente valiosos.

Por otro lado, me gustaría agradecer a \textit{Xin-She Yang} toda la atención y amabilidad brindada durante mi maravillosa estancia en la universidad de Middlesex, en Londrés. 

Gracias también a \textit{Roberto Carballedo}, el cual podría considerar de forma honesta mi \textit{padre} en todo este mundo laboral-científico. Él fue uno de los partícipes para que comenzara mi andadura en Mobility, y fue quien me introdujo en el mundo de la inteligencia artificial. Por eso y por muchas cosas más, ¡gracias!

Para terminar, a esa persona que ha estado junto a mi todas las semanas y que siempre me ha recibido con una cordialidad y agrado admirables, mi director, \textit{Fernando Díaz}. Podría decir multitud de cosas, pero voy a tratar de condensar mi agradecimiento en una escueta frase que resume mi sentimiento a la perfección: si volviera atrás en el camino y comenzase mi tesis de nuevo, no dudaría ni un solo segundo en volver a elegir a Fernando como mi director. Muchas gracias por todo.

\end{acknowledgements}


\setcounter{secnumdepth}{5} 
\setcounter{tocdepth}{5}    


\renewcommand{\tablename}{Tabla}

\tableofcontents            
\listoffigures
\renewcommand{\listtablename}{Índice de tablas}	
\listoftables  





\mainmatter

\pagestyle{fancy}



\begin{savequote}[40mm]
Siempre que te pregunten si puedes hacer un trabajo, contesta que sí y ponte enseguida a aprender cómo se hace.
\qauthor{Franklin D. Roosevelt}
\end{savequote}

\chapter{Introducción y motivación}
\label{cha:Intro}
\graphicspath{ {1_introduccion/figures/} }

\lettrine{H}{}{oy} en día el transporte es un factor crucial para la sociedad, tanto para la ciudadanía, como para el sector empresarial. El transporte público, por ejemplo, es un medio utilizado por gran parte de la población actual y afecta directamente a la calidad de vida de las personas. Existen multitud de sistemas de transporte público, algunos regulares y otros bajo demanda. Estos últimos nacieron con el objetivo de cubrir los nichos que deja el transporte público regular, atendiendo la demanda concreta de los usuarios. Este es un tipo de transporte imprescindible para zonas rurales o con poca demanda, en las cuales no existen líneas de transporte convencionales por motivos económicos.

En cuanto al transporte en el ámbito empresarial, el rápido avance de las tecnologías ha hecho que la logística cobre una gran importancia en esta área. El hecho de que todo el mundo esté permanentemente conectado ha conducido a que las redes de transporte sean cada vez más exigidas, hecho que no tenía relevancia en tiempos pasados, en los que una empresa tenía una clientela dispersa en un territorio geográfico no muy amplio. En tiempos actuales, el hecho de abrir las miras logísticas a un nivel más amplio es un factor indispensable si se quiere sobrevivir en un mundo empresarial que no da pie al error. Esta necesidad hace esencial la existencia de una red de logística competitiva, ya que esta puede marcar las diferencias entre unas empresas y otras. 

Por citar algunos datos que resalten la importancia de la logística en el mundo empresarial, en algunos negocios, como el de los alimentos o las bebidas, los costes de distribución de las mercancías pueden suponer un incremento en el precio del producto de hasta un 70\% \cite{golden1987or}. Gracias a ejemplos como este, es sencillo cerciorarse de la relevancia que posee este sector, el cual tiene que ser debidamente abordado desde los ámbitos geográficos más reducidos (a nivel municipal, o provincial), hasta los más extensos (a nivel continental, o mundial).

\section{Planteamiento del problema}
\label{sec:planteamiento}

Existen diversas áreas de conocimiento capacitadas para hacer frente a la planificación eficiente de la logística, ya sea de carácter urbano o empresarial. En este trabajo se centrará la atención en el campo de la inteligencia artificial y los problemas de optimización.  

En relación con esto último, la gran mayoría de los problemas surgidos en el área del transporte y la logística guardan varias características comunes, lo que hace que puedan modelizarse como problemas de optimización y que puedan ser vistos como casos particulares de otros problemas genéricos. Varios ejemplos de estos problemas genéricos pueden ser el problema del viajante comercial \cite{TSP1}, o el problema de la asignación de rutas a vehículos básico \cite{VRP3}. Estos problemas han sido tratados infinidad de veces desde la visión de la inteligencia artificial, y tanto ellos como muchas de sus variantes más conocidas serán descritos en capítulos posteriores en este mismo documento (Sección \ref{sec:problemas_asignacion_rutas}).

En general, los problemas de optimización resultantes de la modelización de situaciones reales de transporte son denominados \textit{problemas de asignación de rutas a vehículos}, o \textit{problemas de enrutado de vehículos}. Estos problemas se encuadran en el ámbito de la optimización combinatoria, rama que genera una amplia producción científica anualmente. La mayoría de los problemas de asignación de rutas a vehículos cuentan con una complejidad computacional elevada, y pueden catalogarse como problemas NP-Duros o NP-Difíciles \cite{NP-completo}. Un problema es clasificado como NP-Duro cuando no existe una técnica capaz de encontrar en un tiempo aceptable una solución óptima para todos los casos posibles de este.

Es esta complejidad computacional la que hace atractiva para la comunidad científica la resolución de este tipo de problemas. Para enfrentarse a este reto científico existen varias alternativas propias de la inteligencia artificial. En este trabajo se centrará la atención en una de estas alternativas: los algoritmos meta-heurísticos. A grandes rasgos, las meta-heurísticas son técnicas de optimización que enfocan su esfuerzo en buscar una solución a un problema específico utilizando tan solo información general y conocimiento común a todos los problemas de optimización. Para ello, estos algoritmos trabajan con una o varias soluciones, las cuales modifican progresivamente con la intención de mejorarlas poco a poco. Es de crucial importancia entender que el objetivo principal de estas técnicas no es el de encontrar la solución óptima al problema que estén abordando, debido a la complejidad de estos. En su lugar, las meta-heurísticas tratan de encontrar una buena solución que se acerque en lo posible al óptimo global.

Existe un vasto número de meta-heurísticas en la literatura actual, siendo el algoritmo genético \cite{genetico}, el recocido simulado \cite{SA2} o la búsqueda tabú \cite{tabu1} algunas de las más utilizadas a lo largo de la historia. Todas y cada una de las meta-heurísticas existentes en la comunidad científica cuentan con sus ventajas e inconvenientes, y algunas trabajan con mayor efectividad que otras. Aun así, el hecho de que no exista una técnica definitiva que consiga resolver de forma óptima todos los problemas de optimización hace que el campo tanto de la optimización combinatoria, como el de los problemas de asignación de rutas a vehículos, sea un tema de investigación candente, el cual produce una gran cantidad de trabajos científicos año tras año.

Con todo esto, el presente trabajo de tesis doctoral centrará su esfuerzo en la creación de una técnica meta-heurística que ayude a la resolución de problemas de optimización combinatoria, y más concretamente a la de problemas de asignación de rutas a vehículos. La técnica presentada tratará de ofrecer un valor añadido frente a las técnicas ya existentes, ya sea en relación a la mejora de resultados, o a la aportación de originalidades conceptuales. Trabajos como el que se realizará en esta tesis se tornan necesarios debido a la importancia que ha cobrado la planificación logística, y a la necesidad de abordarla de una forma rápida y eficaz.



\section{Hipótesis y objetivos}
\label{sec:Hipotesis}
Una vez introducida la problemática, y con la intención de abordarla de una manera eficiente, se ha planteado la siguiente hipótesis a validar durante el desarrollo de la tesis doctoral:


\definecolor{lightBlue}{HTML}{CCE5FF}
\definecolor{strongBlue}{HTML}{003366}
\setlength{\fboxrule}{2pt}

\begin{center}
\fcolorbox{strongBlue}{lightBlue}{
	\setlength{\fboxrule}{2pt}
	\begin{minipage}[c][1.2\height][c]{0.8\textwidth}
	   	<<Es posible encontrar una novedosa meta-heurística que aporte un valor añadido y mejore las técnicas ya existentes para la resolución de problemas de asignación de rutas a vehículos.>>
     \end{minipage}
}
\end{center}

Para lograr la validación de esta hipótesis se ha propuesto el siguiente objetivo general de la tesis. Este objetivo dimana directamente de la hipótesis, y su consumación es estrictamente necesaria para la certificación de la misma.

	<<\textit{Diseñar e implementar una meta-heurística de carácter multi-poblacional para la eficiente resolución de problemas de optimización combinatoria en general, y problemas de asignación de rutas a vehículos en particular, aportando a su vez un valor añadido en comparación con los métodos ya existentes}>>

Este objetivo ha sido el que se ha intentado alcanzar a lo largo del desarrollo de este trabajo, y cuya consecución da por ratificada la hipótesis arriba planteada. Además de esto, y con el designio de alcanzar un mayor nivel de detalle, se han establecido los siguientes objetivos específicos, cuyo cumplimiento concluirán en el alcance del objetivo general.

\begin{itemize}	
	\item OE1: Definir la filosofía y el modo de operar deseado para la meta-heurística. Este objetivo se logra realizando un examen exhaustivo de las técnicas existentes en la actual literatura, con la pretensión de identificar aquellas características que puedan desembocar en un buen rendimiento por parte del método a diseñar.	

	\item OE2: Seleccionar los problemas que van a emplearse en la experimentación. Para seleccionar el conjunto de problemas que van a utilizarse para este propósito es indispensable conocer en profundidad el estado del arte relacionado. Es importante contar con un buen cúmulo de problemas, de diversas características, y que, además, cuenten con un interés científico y social importante y actual. Este objetivo ha de establecerse antes de comenzar con la implementación de la técnica, ya que esta dependerá de los problemas seleccionados para la fase de pruebas.
	
	\item OE3: Diseño e implementación de la técnica. Esta debe seguir la filosofía establecida en el objetivo OE1, y debe contar con varios objetivos de obligatorio cumplimiento, como por ejemplo, evitar un excesivo consumo de recursos computacionales, resolver los problemas en un tiempo competitivo, o carecer de un diseño excesivamente complejo.
	
	\item OE4: Configurar el entorno de pruebas. Este objetivo está compuesto por dos sub-objetivos. El primero de ellos consiste en conseguir identificar las técnicas existentes con las que se comparará el rendimiento del algoritmo propuesto. Es importante que estos algoritmos sean de interés actual, tengan una reputación demostrada, y guarden cierta similitud en cuanto a filosofía y concepto con la técnica presentada. Para este propósito es necesario un exquisito conocimiento de la literatura disponible. El segundo sub-objetivo es establecer la parametrización adecuada para cada meta-heurística. Es de crucial importancia que todos los algoritmos cuenten con una configuración similar, pues solo de esta forma el cotejo posterior resultará justo y objetivo.
	
	\item OE5: Análisis y evaluación de los resultados logrados en la experimentación. Una vez realizada la experimentación pertinente se analizarán los resultados obtenidos. Este análisis de los resultados se llevará a cabo tanto para la calidad de los mismos, como para el tiempo de ejecución empleado y comportamiento de convergencia demostrado. Para realizar una comparación apropiada se utilizarán variables propias de la estadística descriptiva, como la media y la desviación típica. Además de esto, con la intención de realizar una equiparación objetiva, rigurosa y fiable, se llevarán a cabo varios tests estadísticos, como el test de Friedman, o el test normal $z$.
\end{itemize}

La Tabla \ref{tab:1actividades} recoge varias de las actividades más representativas que se llevarán a cabo para alcanzar la consecución del objetivo general. Estas actividades poseen un vínculo con alguno, o varios, de los objetivos específicos arriba mencionados. Esta relación también queda representada en la misma tabla.

\begin{table}[tb]
\centering
\definecolor{normalBlue}{HTML}{99CCFF}
\definecolor{lightBlue}{HTML}{CCE5FF}
\rowcolors{2}{lightBlue}{normalBlue}
\begin{tabular}{>{\centering}m{0.7\textwidth} >{\centering\arraybackslash}m{0.2\textwidth}}
\hline
\textbf{Actividad} & \textbf{Objetivos} \\
\hline
\hline 
Análisis de los problemas de optimización combinatoria existentes, tanto clásicos como actuales & OE1 y OE2 \\
Análisis de los problemas de asignación de rutas a vehículos existentes, tanto clásicos como actuales & OE1 y OE2 \\
Análisis de las técnicas existentes en la literatura para la resolución de los problemas arriba mencionados & OE1 y OE4 \\
Selección de los problemas a tratar, y diseño de bancos de pruebas para cada uno de ellos & OE2 y OE4 \\
Diseño e implementación de la técnica & OE3 \\
Elección e implementación de las técnicas a utilizar en la experimentación para la comparación de resultados & OE4 \\
Ejecución de la experimentación diseñada & OE4 y OE5 \\
Análisis y comparación de los resultados mediante variables de estadística descriptiva & OE5 \\
Análisis y comparación de los resultados mediante tests estadísticos & OE5 \\

\hline
\end{tabular}

\caption{Actividades más destacables y objetivo específicos con los que se relacionan}
\label{tab:1actividades}
\end{table}

Además de todos estos objetivos, los cuales son considerados necesarios para lograr alcanzar el objetivo general, que a su vez servirá para corroborar la hipótesis planteada, el autor de esta tesis se ha impuesto dos objetivos personales, los cuales tendrá en cuenta a lo largo del desarrollo completo de la misma. Estos objetivos son los siguientes:

\begin{itemize}
	\item Maximizar, en la medida de lo posible, la contribución a la comunidad científica mediante la publicación de artículos en congresos y revistas, tanto nacionales como internacionales, y de carácter tanto científico como divulgativo.
	\item Maximizar la claridad y reproducibilidad de los algoritmos planteados para que puedan ser estudiados, utilizados y/o modificados posteriormente por cualquier investigador o desarrollador, ya sea con el objetivo de aportar algún valor añadido o aplicarlos en alguna herramienta.
\end{itemize}



\section{Intereses científicos y sociales}
\label{sec:interes}  

Una vez introducido el marco de la tesis, y después de haber destacado la hipótesis y los objetivos que han marcado el camino hacia la validación de la misma, en este apartado se pretende destacar los intereses inmanentes a la realización de este trabajo. El objetivo de esta sección, por lo tanto, es destacar porqué es necesario que una tesis de esta índole sea escrita, y describir los aspectos que han motivado la elaboración de la misma. Con todo esto, el interés social del tema que se aborda podría destacarse en los siguientes puntos:

\begin{itemize}
	\item En la actualidad, en la gran mayoría de empresas la logística juega un papel clave, ya sea de forma directa o indirecta. Es por esto por lo que una gran cantidad de empresas tienen en cuenta el transporte de productos como una parte más de sus estrategias de negocio. Con la utilización de las técnicas que van a desarrollarse en este trabajo los gastos relacionados con el transporte, tanto temporales como económicos, van a verse reducidos permitiendo un ahorro monetario y una mayor productividad, con los beneficios que esto acarrea (abaratamiento de los costes de cara al público, mayor calidad en los productos\dots).
	
	\item Con la técnica que va a ser implementada durante esta tesis doctoral se va a conseguir que los vehículos sean más eficientes a la hora de seleccionar las rutas que tienen que tomar para realizar sus desplazamientos. Esto va a hacer que los tiempos de trayecto disminuyan considerablemente y que la congestión en las carreteras se vea minimizada. Esta disminución tiene una gran cantidad de ventajas entre las cuales podrían destacarse las siguientes:
	
	\begin{itemize}
		\item Decremento del tiempo en la carretera, lo que minimiza el riesgo de accidentes.

		\item	Disminución del consumo energético y consumo de carburante. Por todos es conocido que el carburante es un bien escaso. El hecho de ver su consumo reducido no solo permitirá a los usuarios de las carreteras ahorrar en combustible, si no que permitirá que sus altos costes puedan verse reducidos.

		\item Reducción de la emisión de gases CO2. Hoy en día es una obligación moral y social para toda la ciudadanía tener un fuerte compromiso con el medioambiente. La anteriormente mencionada disminución del consumo de combustible conllevará la ventaja de reducir también la emisión de este tipo de gases contaminantes.
	\end{itemize}
		
\end{itemize}

En relación al interés científico que reviste esta área, se puede aseverar que es amplio, ya que la mayoría de los problemas de optimización combinatoria y de asignación de rutas a vehículos pertenecen a la clase NP-Duros. Este tipo de problemas tienen un gran interés académico ya que, a día de hoy, no es posible encontrar técnicas que sean capaces de resolver cualquier instancia de una forma óptima y en un tiempo polinomial. A lo largo de la historia se han desarrollado innumerables estudios en este campo, y actualmente sigue siendo un tema importante en la investigación, el cual es objeto de una gran cantidad de estudios en los distintos congresos y revistas científicas mundiales año tras año. Este tema será tratado con mayor detalle en el siguiente capítulo de esta tesis.



\section{Principales aportaciones del trabajo presentado}
\label{sec:aportacionesIntro}  

En este apartado se expondrán las principales aportaciones que ofrece el modelo que se presentará en esta tesis para validar la hipótesis anteriormente planteada. Para ello, en primer lugar conviene describir de manera escueta las características más reseñables de la meta-heurística desarrollada.

Es imprescindible comenzar esta descripción mencionando que la técnica propuesta consiste en una meta-heurística multi-poblacional, la cual se basa en diversos conceptos futbolísticos para guiar su proceso de búsqueda. Es por esta razón por la que el nombre escogido para designar al algoritmo ha sido Golden Ball. Al igual que otras muchas técnicas existentes, el Golden Ball trabaja con un conjunto de soluciones aleatoriamente creadas. Estas soluciones son agrupadas en diferentes conjuntos, o equipos, los cuales componen una liga. Los diferentes equipos evolucionan de forma autónoma, es decir, sus soluciones se modifican individualmente con la intención de ser mejoradas. A este proceso se le llama ``entrenamiento". Es interesante mencionar que cada equipo \textit{entrena} a sus soluciones de diferente manera. A su vez, los equipos se enfrentan unos con otros completando una liga convencional. Una vez finaliza la temporada liguera, comienza el periodo de fichajes, en el que los equipos intercambian sus jugadores, saliendo beneficiados los equipos punteros. Por otro lado, los equipos que han obtenido malos resultados cambian su entrenador, es decir, cambian la manera en la que modifican sus soluciones, con la intención de encontrar un método de modificación más eficiente.

Estos son los conceptos básicos en los que se basa la técnica Golden Ball, explicados todos ellos de una forma abreviada y sencilla. Las principales aportaciones de la técnica presentada respecto a las meta-heurísticas que pueden verse en la literatura son las siguientes:

\begin{itemize}

	\item Como se especificará más adelante, el Golden Ball basa la modificación de sus soluciones en dos operadores distintos, uno de carácter individual, y el otro de carácter cooperativo. Muchas meta-heurísticas siguen esta filosofía, como los algoritmos genéticos, por ejemplo. Aun así, el Golden Ball ofrece un enfoque raramente visto en la literatura, dando mayor importancia a la mejora individual, y utilizando los operadores cooperativos en un segundo plano.
	
	\item El Golden Ball utiliza un mecanismo que permite a las sub-poblaciones cambiar en múltiples ocasiones la forma en la que modifican (o entrenan) las soluciones. Es justo mencionar que existen algunas meta-heurísticas en la literatura, como la búsqueda de vecindario variable \cite{mladenovic1997variable}, que cuentan con recursos similares a este. Aun así, en el caso del Golden Ball, no solo las subpoblaciones varían durante el transcurso de la ejecución. Los jugadores, en concreto, también pueden cambiar su forma de entrenar, alternando sus equipos en situaciones en las que se estima que puede ser provechoso para el proceso de búsqueda.

	\item Como se ha mencionado anteriormente, el Golden Ball es una técnica multi-poblacional. Esto quiere decir que la población completa de soluciones es dividida en diferentes conjuntos. Un tema que ha generado multitud de debates en la literatura es la forma en la que las diferentes subpoblaciones que conforman una técnica multi-poblacional comparten sus soluciones entre ellas. Para ello, el Golden Ball ofrece un esquema de intercambio de soluciones fijo e invariable, el cual se fundamenta en la calidad general de cada uno de los conjuntos.
	
	\item El constante crecimiento de la literatura relacionada muestra cómo las meta-heurísticas bío-inspiradas son un tema de interés para la comunidad científica. Al hilo de esto, el Golden Ball es la primera técnica cuyo proceso de búsqueda mimetiza diversos conceptos futbolísticos. Siendo el futbol un deporte de masas, practicado y seguido en todo el planeta, es de esperar que los conceptos en los que se apoya la técnica sean familiares para cualquier lector, lo que facilitará su comprensión e instigará a investigadores noveles a adentrarse en el campo de estudio.
	
\end{itemize}

Estas son varias de las contribuciones que ofrece la meta-heurística propuesta en este trabajo doctoral. Todas ellas serán tratadas con mayor detalle en secciones posteriores (Sección \ref{sec:aportaciones}). Por otro lado, además de las aportaciones ofrecidas por la meta-heurística propuesta, este trabajo también pretende proporcionar las siguientes contribuciones genéricas: 

\begin{itemize}

	\item Un contratiempo muy común cuando se intenta trabajar en el campo de las meta-heurísticas y optimización es la falta de transparencia a la hora de presentar las técnicas. Esto hace que la replicabilidad y la reproducción de las técnicas que se encuentran en la literatura se vean realmente comprometidas. Con este inconveniente en mente, el autor de este trabajo se ha propuesto proporcionar una meta-heurística clara y con una estructura correcta, la cual estará a disposición de cualquier investigador o desarrollador para su posterior estudio, empleo y modificación.
	
	\item Otra adversidad a la hora de afrontar los diferentes retos que se presentan en este campo de estudio es la falta de una metodología adecuadamente estructurada para aplicarla a la implementación y comparación de diferentes técnicas. Teniendo este hecho en cuenta, el autor de esta tesis propone en este mismo trabajo un conjunto de buenas prácticas. Con esto, se pretende establecer un procedimiento para el correcto desarrollo y posterior comparación de meta-heurísticas diseñadas para la resolución de problemas de optimización combinatoria.
	
\end{itemize}



\section{Metodología de investigación}
\label{sec:Metodologia}  

Como ya se ha indicado en apartados anteriores, el campo de la optimización combinatoria y las técnicas para su resolución es un campo en constante crecimiento, generando multitud de producción científica año tras año. Esta es la principal razón por la que se ha hecho imprescindible acogerse a una metodología de investigación de ciclo continuo, en la que la actualización del conocimiento y la identificación constante de mejoras sean dos de sus principales pilares.  
 
Con esta importante necesidad se ha planteado un procedimiento de filosofía iterativa en el que el cumplimiento de cada ciclo contribuye al refinamiento de la solución planteada para validar la hipótesis. La idea principal de este proceso cíclico es que los conocimientos adquiridos en su fase inicial ayuden a diseñar una técnica cada vez más prometedora, capaz de competir cara a cara contra métodos actuales, ya sea en cuanto a resultados se refiere, o en cuanto a concepto, ofreciendo originalidades y aportaciones reseñables. 

\begin{figure}[tb]
	\centering
	\includegraphics[width=1.0\textwidth]{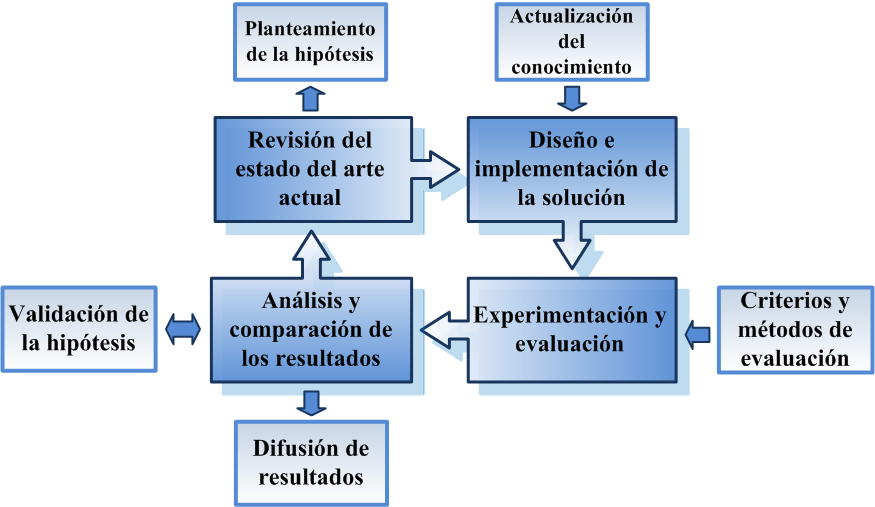}  
	\caption{Metodología de investigación empleada}
	\label{fig:1metodologia}
\end{figure}

En la Figura \ref{fig:1metodologia} se muestra la metodología de investigación empleada. En esta misma imagen puede observarse el carácter iterativo del procedimiento, el cual puede sintetizarse de la siguiente manera: 

\begin{itemize}
	\item \textit{Revisión del estado del arte actual}: El objetivo principal de esta fase es el de investigar el estado del arte relacionado con el campo en el que se está trabajando. Para lograr esto, se hará uso de la bibliografía relacionada, tomando publicaciones de la comunidad científica publicadas en revista y actas de congresos nacionales e internacionales. Los conocimientos adquiridos en esta fase deben desembocar en el planteamiento de la hipótesis, después de haber encontrado, y analizado, posibles nichos de mejora.
		
	\item \textit{Diseño e implementación de la solución}: Después de haber adquirido o actualizado el conocimiento necesario, y teniendo siempre en mente la hipótesis a validar, en esta fase se debe diseñar (o modificar, siempre y cuando se esté en un ciclo posterior al primero) e implementar la solución que se va a proponer para intentar probar la hipótesis previamente planteada.
	
	\item \textit{Experimentación y evaluación}: La meta de esta fase es la de someter a la solución resultante del paso anterior a un proceso de experimentación y evaluación. Para realizar este procedimiento es de vital importancia aportar unos criterios y métodos de evaluación, en los que se incluyen los problemas empleados, la parametrización utilizada o las técnicas con las que se compararán los resultados en la fase posterior. Todos estos criterios y métodos se han de construir haciendo uso del conocimiento adquirido en la primera etapa de esta metodología.
			
	\item \textit{Análisis y comparación de los resultados}: Tras la realización de la experimentación pertinente, los resultados obtenidos tienen que ser analizados y contrastados con los obtenidos por otras técnicas punteras de la literatura. Después de realizar este examen, es conveniente comprobar si la hipótesis planteada ha quedado validada. En tal caso, se podría dar por finalizado el desarrollo de la tesis, teniendo siempre en cuenta potenciales actualizaciones en el estado del arte. En caso de que la hipótesis aun no quede comprobada, el ciclo vuelve a comenzar por su punto inicial. A su vez, y siendo conscientes de que es uno de los aspectos más importantes del desarrollo de la tesis, esta etapa tiene que dar como resultado la difusión de resultados, materializada en producción científica, ya sea en congresos, como en revistas nacionales o internacionales.
		
\end{itemize}



\section{Estructura de la tesis}
\label{sec:Estructura}   
En el presente apartado se introduce la estructura de este trabajo doctoral. En concreto, la tesis cuenta con siete capítulos:

\begin{itemize}
	\item El primero, el presente, se trata de la introducción, y es el capítulo en el que se trata de asentar los conceptos básicos y el ámbito de trabajo de esta tesis. También se han introducido los intereses científicos y sociales que posee la tesis, además de sus aportaciones y contribuciones. Además de esto, se han presentado tanto la hipótesis, como los objetivos que guiarán su desarrollo.

	\item El segundo capítulo del documento proporciona al lector una visión general de la optimización combinatoria. Se describen en un primer momento varios problemas clásicos de esta clase, introduciendo tras esto los conceptos de los problemas de asignación de rutas a vehículos. Además de esto, se hará un esfuerzo en detallar diferentes variantes de este último tipo de problemas, cuyo interés científico hoy en día es indudable, y los cuales tendrán gran importancia en la experimentación de esta tesis.
	
	\item El tercer capítulo bien podría considerarse una continuación del anterior. Después de haber introducido el campo de la optimización combinatoria y haber descrito en detalle varios de los problemas más conocidos e interesantes dentro de este ámbito, en este tercer apartado se presentan los conceptos de heurística y meta-heurística. El grueso de este capítulo trata de presentar varias de las alternativas más utilizadas a lo largo de la historia para abordar los diferentes problemas de optimización surgidos en la literatura.
	
	\item El cuarto capítulo es el que recoge la pormenorizada explicación del modelo presentado en esta tesis para validar la hipótesis planteada en la introducción. También hay lugar en este apartado para exponer diversas reflexiones llevadas a cabo durante el desarrollo del trabajo, claves a la hora de decidir la dirección del mismo. Adicionalmente, en este capítulo se explicarán las principales aportaciones y contribuciones del modelo propuesto, así como las diferentes similitudes y originalidades respecto a las principales técnicas de la literatura.

	\item El quinto capítulo aborda la experimentación llevada a cabo para validar el modelo presentado en este trabajo. Múltiples apartados componen este capítulo, en los que se describen las técnicas utilizadas, se introducen las configuraciones empleadas para los problemas utilizados, y se muestran y analizan los resultados logrados. Además de esto, en este punto de la tesis se introduce una propuesta de buenas prácticas para la implementación y comparación de meta-heurísticas enfocadas a la resolución de problemas de optimización combinatoria.

	\item En el sexto capítulo se describen varias posibles implementaciones prácticas de la meta-heurística presentada en este trabajo. Para cada una de ellas, primeramente se propone una situación propia del mundo real relacionada con el transporte, la cual se modelizará como un problema complejo de asignación de rutas a vehículos, para después ser resuelto por la técnica propuesta.

	\item Finalmente, el séptimo y último capítulo del trabajo expone las conclusiones de la tesis, así como las líneas futuras de trabajo y mejora.  
  
\end{itemize}



\begin{savequote}[40mm]
Todos somos muy ignorantes. Lo que ocurre es que no todos ignoramos las mismas cosas.
\qauthor{Albert Einstein}
\end{savequote}

\chapter{Optimización combinatoria y problemas de asignación de rutas a vehículos}
\label{cha:optimizacion}

\graphicspath{ {2.1_optimizacion/figures/} }

\lettrine{H}{}{oy} en día, la optimización es un campo que recibe mucha atención dentro de la inteligencia artificial. Existen varios tipos de optimización, como la optimización numérica \cite{numerical}, lineal \cite{linear}, continua \cite{continuous} o la optimización combinatoria \cite{combinatorial}, entre muchas otras. Normalmente, los problemas surgidos en este campo nacen a partir de situaciones reales, las cuales se tratan de modelizar como problemas de optimización. La resolución de estos problemas plantea grandes retos científicos debido a que, habitualmente, cuentan con una complejidad elevada. Esta complejidad y su facilidad inherente de ser aplicable al mundo real son las principales razones que hacen que la optimización sea un ámbito atractivo para la comunidad científica.

Esta tesis se centrará únicamente en problemas de optimización combinatoria, cuyo objeto es el estudio y la resolución algorítmica de problemas con restricciones en los que las variables constituyentes son de tipo discreto y finito. Más concretamente, y pese a que se hará uso de diferentes problemas de otra índole, los problemas de asignación de rutas a vehículos serán los que ocupen el grueso de la experimentación ofrecida en este trabajo.

La optimización combinatoria es ampliamente estudiada en tiempos actuales por la comunidad científica. Prueba de ello son la multitud de artículos y libros que son publicados año tras año enfocados en este tema \cite{combinatorialBook,combinatorialPaper}. Por otro lado, congresos como el ``Integer Programming and Combinatorial Optimization", el cual se encuentra en su decimoctava edición, convirtiéndose en un congreso de gran prestigio a nivel internacional, pueden considerarse otra prueba de este interés, así como el creciente número de ediciones especiales en revistas científicas internacionales centradas en este tipo de optimización, como el ``Special Issue on Combinatorial Optimization: Theory of Algorithms and Complexity", publicado anualmente en la revista Theoritecal Computer Science (F.I. 0.516). 

El interés suscitado por los problemas planteados dentro de la optimización combinatoria puede diferenciarse en dos aspectos completamente diferentes. Por un lado, se puede destacar el inmanente interés social que poseen estos problemas, ya que surgen a partir de problemas cotidianos del mundo real. Para ser más precisos, esto quiere decir que diferentes situaciones reales pueden ser modelizadas como problemas de optimización combinatoria para ser tratadas y resueltas con mayor eficiencia. Por otro lado, una gran parte de estos problemas poseen una gran complejidad de resolución, por lo que encontrar soluciones eficientes a ellos constituye un reto atractivo para los investigadores de hoy en día. Siendo más concretos, una gran cantidad de estos problemas son catalogados como NP-Duro \cite{NP-completo,NP-completo1}. Según la teoría de la complejidad computacional, un problema es considerado NP-Duro cuando no existe una técnica capaz de encontrar una solución óptima para todas las instancias\footnote{instancia: una instancia de un problema es un ejemplo concreto de ese problema.} en un tiempo polinomial.

En la literatura puede encontrarse una abundante cantidad de problemas de optimización combinatoria. En esta sección se describirán varios de ellos, con el propósito de que todo lector pueda hacerse una idea de la naturaleza de este tipo de problemas, y siendo conscientes de que la cantidad total de ellos es muchísimo más amplia. Con todo esto, en la Sección \ref{sec:probclasicos} se presentan varios problemas clásicos de optimización combinatoria. Tras esto, en la Sección \ref{sec:problemas_asignacion_rutas} se hará un breve repaso a varios de los problemas más importantes de asignación de rutas a vehículos. 

\section{Ejemplos de problemas clásicos de optimización combinatoria}
\label{sec:probclasicos}

Como ya se ha aludido con anterioridad, en esta sección se describirán varios problemas clásicos de optimización combinatoria. Este apartado pretende servir de introducción para aquellos lectores que no se sientan familiarizados con el campo de estudio. Los problemas descritos en esta parte han sido seleccionados ya sea por su ajuste a problemas interesantes del mundo real (Sección \ref{sec:JSP}), o porque van a ser empleados en este mismo estudio para la validación de la técnica presentada (Apartados \ref{sec:NQP} y \ref{sec:BPP}). En cualquier caso, todo ellos han sido ampliamente referenciados y utilizados a lo largo de la historia, y por ello son considerados como problemas clásicos de optimización. 

\subsection{Problema de la programación de la producción discreta}
\label{sec:JSP}

\begin{sloppypar}
El primero de los problemas descrito es el problema de la programación de producción discreta (\textit{Job-shop Scheduling Problem}, JSP). En este caso, se podría clasificar al JSP dentro de los problemas relacionados con la organización industrial. Desde que fue propuesto \cite{JSS1}, el JSP se ha convertido en uno de los problemas de optimización más estudiados, siendo el foco de muchos estudios aun a día de hoy \cite{JSS2,JSS3}. Existen muchas variaciones del JSP \cite{JSS4,JSS5}, aunque a continuación tan solo se introducirá el JSP más básico. Para este problema se dispone de un conjunto de máquinas ($M$), las cuales realizan diversas tareas. Con estas máquinas se pretende elaborar un producto concreto a partir de una materia prima inicial. Para materializar dicho producto, hay que seguir una serie de pasos, donde cada uno de ellos consiste en la aplicación de una determinada máquina durante un periodo de tiempo. A cada uno de los pasos de este proceso se le llama operación. Del mismo modo, se denominará orden de trabajo a la secuencia de operaciones requerida para la finalización del producto.
\end{sloppypar}

De esta manera, dado $M$ y un conjunto de órdenes de trabajo, un programa es una asignación que fija a cada operación una máquina y un intervalo de tiempo para ser efectuada. El objetivo del JSP es encontrar un programa que minimice el tiempo empleado para realizar todas las operaciones. 

\subsection{Problema de las N reinas}
\label{sec:NQP}

El segundo problema que se tratará en este apartado es el problema de las N reinas (\textit{N-Queens Problem}, NQP) \cite{NQP1}. El NQP es la generalización del clásico problema de colocar 8 reinas en un tablero convencional de ajedrez, con el requisito de que ninguna de ellas sea atacada por cualquier otra \cite{NQP2}. Esta primitiva versión del NQP fue propuesta por M. Bezzel en 1848 \cite{NQP3}. De esta forma, el NQP consiste en situar $N$ reinas en un tablero de ajedrez $N$x$N$, de manera que no exista ningún ataque entre ellas. Este es un clásico problema de diseño combinatorio (problema de satisfacción de restricciones), el cual puede ser también formulado con un problema de optimización combinatoria \cite{NQP4}. En el trabajo que se presenta en esta tesis, se tratará al NQP de este modo, donde una solución $X$ es codificada como una tupla de $N$ valores $(q_1,q_2,...,q_n)$, la cual es una permutación del conjunto de valores $(1, 2, ..., N)$. Cada $q_i$ representa la fila ocupada por la reina emplazada en la columna $i$-ésima. Utilizando esta representación las colisiones verticales y horizontales son evitadas, y la complejidad del problema pasa a ser $O(N!)$. Así, el objetivo del problema será minimizar el número de colisiones diagonales entre reinas. Más específicamente, una reina $i$ atacará a otra reina $j$ si, y solo si:

\[
	|i-q_i| = |j-q_j| \ \ \ \ \ \forall i,j: \{1,2, ..., N\}; i \neq j 
\]

\begin{figure}[tb]
	\centering
		\includegraphics[width=0.4\textwidth]{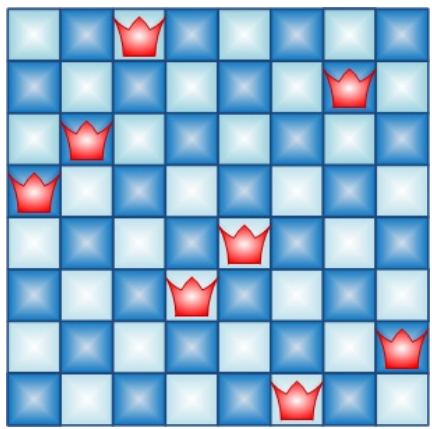}
	\caption{Ejemplo de instancia de un NQP compuesto por 8 reinas}
	\label{fig:NQP}
\end{figure}

Una posible solución a un NQP compuesto por 8 reinas es mostrada en la Figura \ref{fig:NQP}. De acuerdo a la codificación explicada, la solución interpretada en esta figura sería codificada como $f(X)=(4, 3, 1, 6, 5, 8, 2, 7)$. Como puede observarse, en esta posible solución pueden encontrarse 3 diferentes colisiones (4-3, 6-5, y 6-8). Esta misma formulación ha sido anteriormente empleada en la literatura en innumerables ocasiones \cite{NQP5,NQP6}.

\subsection{Problema del empaquetado unidimensional}
\label{sec:BPP}

Finalmente, el problema del empaquetado unidimensional (\textit{one dimensional Bin Packing Problem}, BPP), será el último de los problemas mencionados en esta sección introductoria. El empaquetado de productos o paquetes en diferentes recipientes o contenedores es una tarea diaria y crucial en el ámbito de la producción y distribución. Dependiendo tanto del tamaño de los paquetes a embalar, como del tamaño y capacidad de los contenedores, se pueden formular múltiples problemas de empaquetado. Este tipo de cuestiones han sido ampliamente tratadas en la literatura desde hace varias décadas. En el estudio presentado en \cite{BBP1} se realiza una extensa introducción a este tipo de problemas. En la experimentación desarrollada para la validación de esta tesis se hará uso del BPP, considerado como el problema más simple de embalaje, y el cual ha sido utilizado en la literatura en infinidad de ocasiones como problema de benchmarking\footnote{Problema de benchmarking: problema de simple formulación, generalmente utilizado para la validación de una técnica, o para la comparación entre varias técnicas.} \cite{BPP2,BPP3,BPP4}. Concretamente, el BPP consiste en un conjunto de paquetes $I=(i_1, i_2, \dots , i_n)$, cada uno con un tamaño $s_i$ asociado, y un suministro ilimitado de contenedores con una misma capacidad $q$. El objetivo del BPP consiste en empaquetar todos los paquetes en el menor número de recipientes posible. De esta manera, el propósito es minimizar la cantidad de contenedores utilizados.

Como ya se ha mencionado anteriormente, además de los problemas descritos, existe una abundante cantidad de problemas dentro del campo de la optimización combinatoria. A su vez, existe un conjunto de problemas de este tipo que, gracias a su indiscutible interés científico y social, goza de gran popularidad en la comunidad científica. Estos problemas son los de asignación de rutas a vehículos, y son los que ocuparán el grueso de la experimentación presentada en este trabajo doctoral. Si bien la técnica propuesta puede ser aplicada a cualquier tipo de problema de optimización combinatoria, se ha decidió centrar el esfuerzo en los problemas de asignación de rutas a vehículos. En la sección que comienza a continuación se realizará una introducción a los problemas de este tipo.

\section{Problemas de asignación de rutas a vehículos}
\label{sec:problemas_asignacion_rutas}

Como ya se ha mencionado en apartados anteriores, los problemas de asignación de rutas a vehículos gozan de gran popularidad dentro de la comunidad científica. La razón de esta popularidad puede desgranarse en dos factores distintos. Por un lado, debido a su complejidad, supone un reto tratar de resolver este tipo de problemas. De esto que el atractivo científico inherente a estos problemas sea irrevocable. Por otro lado, como ya se ha mencionado en la introducción de este trabajo, el beneficio empresarial que constituye una eficiente logística, y las ventajas sociales que supondrían este hecho, hace que el tratamiento y resolución de estos problemas posea un gran interés social. Evidencia de este interés es el creciente número de publicaciones científicas que se añaden a la literatura año tras año. Citar en esta tesis todas estas publicaciones sería inviable. Aun así, a lo largo de este documento se citarán una abundante cantidad de artículos y libros enfocados a este campo.

En esta sección se van a describir de forma detallada varios de los problemas de asignación de rutas a vehículos más conocidos y estudiados, comenzando por los más básicos y continuando con las variantes más famosas de éstos. Sin embargo, antes de profundizar en los distintos problemas conviene describir de forma breve cada uno de los \textit{actores} que forman parte de este tipos de problemas.

\subsection{Actores de los problemas de asignación de rutas a vehículos}
\label{sec:actores}

Explicado grosso modo, los problemas de asignación de rutas a vehículos consisten en un conjunto de clientes a servir, uno o varios depósitos y una flota de vehículos con los cuales hay que abastecer la demanda de los consumidores. El objetivo de estos problemas suele ser el de minimizar tanto el número de vehículos utilizados, como la distancia recorrida, o los gastos generados. Todo esto depende del tipo de problema que se esté abordando.

La naturaleza y características de los clientes, depósitos y vehículos, junto a diferentes restricciones a la hora de gestionar las rutas, dan lugar a multitud de variantes del problema que se verán en este mismo capítulo. 

\subsubsection{Clientes}
\label{sec:Clientes}

Los clientes, o consumidores, son los auténticos protagonistas de estos tipos de problemas. Cada cliente se encuentra situado en una posición geográfica dentro del territorio operativo\footnote{Territorio operativo: Territorio geográfico en el que el vehículo ofrece su servicio}. Cada consumidor tiene una demanda que tiene que ser satisfecha por alguno de los vehículos del sistema. Normalmente los clientes solicitan la entrega de cierta cantidad de materiales, convencionalmente servida por un solo vehículo y en una sola visita, aunque en ocasiones el cliente puede tratarse de un proveedor, o de un establecimiento, el cual solicita una demanda de recogida de bienes.

Existen casos en los que los clientes no solicitan la recogida o entrega de ciertos materiales. En su lugar, éstos solicitan la utilización de un servicio, como por ejemplo el transporte desde un punto hasta otro \cite{DRT}.

En otros casos los consumidores pueden imponer ciertas restricciones al sistema. Una de ellas son las restricciones temporales \cite{VRPTW3}, las cuales tienen que ser respetadas obligatoriamente, aunque existen casos en los que estas restricciones pueden traspasarse a cambio de una penalización para el proveedor \cite{VRPSTW}. Otra de las restricciones que pueden imponer los clientes es el tipo de vehículo que puede servirles. Existen problemas en los que la flota de vehículos es heterogénea, y puede suceder que ciertos consumidores solo puedan ser servidos por un tipo de unidad móvil \cite{VRPHV}. Esta situación puede darse por el hecho de que el cliente viva en alguna zona inaccesible para algún tipo de vehículo.

\subsubsection{Vehículos}
\label{sec:vehiculos}

Normalmente, los vehículos cuentan con una capacidad limitada, la cual no puede ser sobrepasada en ningún momento. A su vez, puede darse el caso en el que cada vehículo tenga asociados ciertos costes fijos por su utilización, aunque lo más normal es aplicar estos costes tan sólo en problemas en los que se tiene una flota de vehículos heterogénea, de forma que los vehículos más grandes serán más costosos, mientras que los pequeños supondrán un menor desembolso económico.

Como ya se ha explicado en el apartado de los clientes, puede ser que las unidades móviles no sólo tengan que entregar mercancías, si no que tengan también la necesidad de recogerlas en ciertos puntos de su ruta. Esto hace que la cantidad total de mercancías recogidas y mercancías aún por entregar no pueda superar la capacidad máxima de los vehículos. Dentro de este tipo de problemas, existen instancias en las que es obligatorio realizar primero la fase de entrega, seguida de la fase de recogida \cite{VRPB}, o instancias en las que los distintos tipos de clientes pueden alternarse \cite{VRPSPD}.

La cantidad de unidades móviles disponibles es una variable de decisión, de forma que está en manos de los usuarios el limitar la flota, o utilizar un número de vehículos ilimitado.

\subsubsection{Depósitos o almacenes}
\label{sec:depositos}

Por lo general, en un problema de asignación de rutas a vehículos, las mercancías a distribuir se encuentran almacenadas en un depósito, o almacén. Aún así, existen instancias, como ya se ha mencionado anteriormente, en las que los bienes a repartir se encuentran en posesión de distintos clientes, los cuales actúan como proveedores. El almacén puede ser también un punto de entrega de materiales, ya que en los sistemas en los que se recoge material de los consumidores, éste tiene que ser almacenado en el depósito.

Por norma general, todos los vehículos del sistema se encuentran en el depósito inicialmente. Asimismo, las rutas trazadas tienen que comenzar y terminar en el depósito central, aunque existen algunas variantes en las que el vehículo puede finalizar su ruta en otro punto.

Por otro lado, en ciertos problemas el número de depósitos puede ser mayor que uno, de forma que el sistema tiene que elegir qué clientes tienen que ser servidos por los vehículos pertenecientes a cada depósito \cite{MDVRP}.

Finalmente, pueden encontrarse variantes en las que haya almacenes intermedios, en los cuales los vehículos tienen la opción de realizar una parada para reponer las mercancías, o repostar combustible \cite{VRPSF}. De este modo, las rutas pueden ser más largas, y las unidades móviles pueden atender a un número mayor de clientes.

Como se ha explicado en la introducción de este capítulo, las diferentes características de los tres actores descritos, junto a diferentes restricciones adicionales, hacen que la cuantía de problemas de asignación de rutas a vehículos existentes sea realmente amplia. A continuación se detallarán varios de estos problemas.

\subsection{El problema del viajante comercial}
\label{sec:TSP}

El problema del viajante comercial, \textit{Traveling Salesman Problem}, o simplemente TSP, es uno de los problemas más famosos y ampliamente estudiados a lo largo de la historia dentro de la investigación operativa y las ciencias de la computación. La definición del problema es la siguiente:

\textit{Dado un conjunto de clientes o nodos y una matriz de distancias entre ellos, el objetivo es encontrar una ruta que visite todos y cada uno de los clientes una sola vez y que minimice la distancia total recorrida.}

\begin{figure}[tb]
	\centering
		\includegraphics[width=1.0\textwidth]{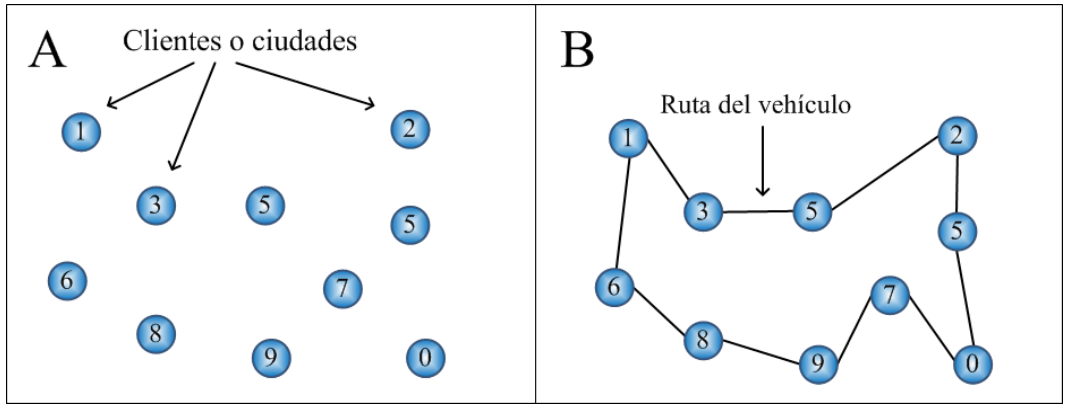}
	\caption{Ejemplo de instancia de un TSP de 10 nodos, y posible solución}
	\label{fig:TSP}
\end{figure}

En la Figura \ref{fig:TSP} puede observarse el ejemplo de una instancia compuesta por 10 nodos, en la que se ofrece una posible solución.

En el TSP, al contrario que en la mayoría de los problemas que se describirán después, no existe un depósito (y en caso de haberlo no se distingue de los clientes), los clientes no tienen demanda asociada y todos han de ser visitados por un mismo vehículo, creando una única ruta. Es por esto por lo que podría decirse que es el problema de asignación de rutas más sencillo de describir y formular.

El primer TSP fue definido en 1800 por los matemáticos Hamilton y Kirkman \cite{TSP1}, aunque su primera formulación se presentó en 1930, por Karl Menger. Desde entonces, a lo largo de los años, el TSP ha sido objeto de multitud de estudios y se le considera como uno de los problemas fundamentales de la optimización combinatoria. Existen varias razones por las cuales esto ha sido así \cite{TSP2}. La primera de ellas es la facilidad a la hora de describirlo y formularlo, junto con su dificultad para resolverlo. La segunda razón es la amplia aplicabilidad que tiene a una gran variedad de problemas de planificación y asignación de rutas. Una razón más es la gran cantidad de información existente relacionada con el TSP, lo que hace que se convierta en un buen problema de benchmarking. En numerosas ocasiones, nuevas técnicas de optimización combinatoria se prueban en primer lugar con el TSP para verificar su efectividad \cite{TSP3,TSP4,TSP5}.

El único objetivo del TSP es el de minimizar la distancia total recorrida por el vehículo. Por esta razón, en instancias compuestas por pocos nodos, el enfoque más intuitivo puede ser el de analizar todas y cada una de las posibles rutas, para seleccionar después de entre todas ellas aquella que minimice la distancia. Esta técnica, conocida como fuerza bruta \cite{BruteForce}, es capaz de obtener la solución óptima, e incluso se podría decir que en instancias muy reducidas es un enfoque correcto. Aun así, según aumenta el número de ciudades o clientes esta técnica se hace inviable e imposible de utilizar. Esto es debido a que la complejidad computacional del problema es amplia, y aumenta de forma explosiva según incrementa el número de nodos a visitar. Esta complejidad es del orden de $O(n!)$, lo que supone que, por ejemplo, en una instancia con tan solo 10 ciudades, el número de permutaciones posibles es igual a 3,62$\times$$10^6$.

Debido a la elevada complejidad computacional, el enfoque óptimo para este tipo de problemas es el de utilizar técnicas como las que se verán en la siguiente sección, las cuales se centran en explorar la mayor cantidad de soluciones posibles, intentando obtener una solución cercana a la óptima.

\subsubsection{Definición del problema}
\label{sec:TSPform}

El TSP puede ser definido como un grafo completo $G= (V,A)$, donde $V$ contiene el conjunto de clientes que tienen que ser visitados por el vehículo, $V= \{v_1,v_2,\dots,v_n\}$, mientras que el conjunto $A$ representa los arcos que interconectan cada uno de los nodos, $A= \{(v_i,v_j): v_i,v_j  \in V,i\neq j \}$. A su vez, cada arco tiene un coste $d_{ij}$, el cual denota la distancia que supone ir desde un cliente $i$ a un cliente $j$. La versión mas básica del TSP cuenta con unos costes simétricos, es decir, la distancia es la misma para el arco $(i,j)$, que para el arco $(j,i)$, por lo que el conjunto $A$, podría transformarse en $E= \{(v_i,v_j): v_i,v_j  \in V,i< j \}$. Con todo esto, el objetivo del TSP consiste en encontrar una ruta tal que, comenzando y finalizando en el mismo cliente, visite cada $v_i$ una sola vez, y minimice la distancia total recorrida. De esta manera, la función objetivo\footnote{La función objetivo es esa función que se trata de minimizar o maximizar. En el caso de los problemas de asignación de rutas suele consistir en la distancia total recorrida, o el coste total que supone realizar todas las rutas planeadas. Para este tipo de problemas, el objetivo es minimizar esta función lo máximo posible} es la distancia total de la ruta. 


\subsubsection{Formulación matemática}
\label{sec:TSPmathform}

El problema puede formularse de manera formal, y matemática, de la siguiente manera \cite{TSPmath}:

\begin{equation}
	\sum_{(i,j)\in A}{d_{ij}x_{ij}} \label{TSPeq1}
\end{equation}
\begin{equation}
	\sum_{i\in \Delta + (i)}{x_{ij}=1}, \ \ \ \forall i \in V \label{TSPeq2}
\end{equation}
\begin{equation}
	\sum_{i\in \Delta - (j)}{x_{ij} \geq 1}, \ \ \ \forall j \in V \label{TSPeq3}
\end{equation}
\begin{equation}
	\sum_{i\in S, j \in \Delta + (i)\backslash S}{x_{ij} \geq 1}, \ \ \ \forall S \subset E \label{TSPeq4}
\end{equation}
\begin{equation}
	x_{ij} \in {0,1}, \ \ \ \forall {i,j} \in E \label{TSPeq5}
\end{equation}

\begin{figure}[tb]
	\centering
		\includegraphics[width=0.6\textwidth]{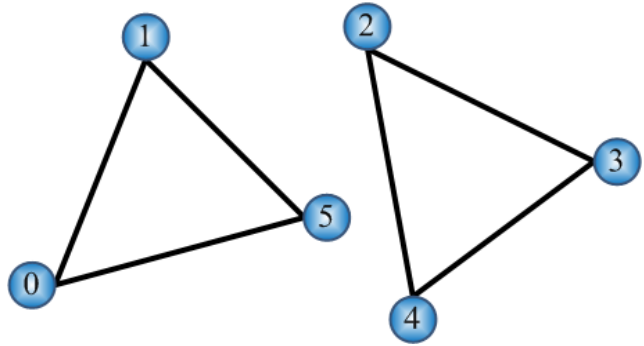}
	\caption{Ejemplo de un posible ciclo en una instancia del TSP compuesta por 6 nodos}
	\label{fig:ciclo}
\end{figure}

Siendo la variable $x_{ij}$ una variable binaria que toma el valor 1 si el arco es utilizado en la solución, y 0 en caso contrario. La función objetivo, \ref{TSPeq1}, es el sumatorio de todos los arcos utilizados en la solución, o lo que es lo mismo, la distancia total de la ruta. Como ya se ha explicado, esta función tiene que ser minimizada. Las restricciones \ref{TSPeq2} y \ref{TSPeq3} indican que cada nodo tiene que ser visitado y abandonado una sola vez, mientras que la fórmula \ref{TSPeq4} garantiza la no existencia de sub-tours, e indica que todo subconjunto de nodos $S$ tiene que ser abandonado al menos 1 vez. Esta restricción es de vital importancia, ya que si no estuviera presente la solución podría tener más de un ciclo. En la Figura \ref{fig:ciclo} se muestra un ejemplo de una instancia de 6 nodos en la que se dibuja una solución incorrecta. Esta solución  viola la sentencia \ref{TSPeq4} para el subconjunto  S=\{0,1,5\}. Esta restricción puede ser expresada de otras muchas formas \cite{TSPdifForm}.

\subsection{Variantes del TSP}
\label{sec:TSPvariantes}

\begin{sloppypar}
Además del TSP convencional, existe una gran cantidad de variantes de este problema, surgidas todas ellas con el designio de adaptarlo con mayor fidelidad a entornos reales. Ejemplos de esta afirmación pueden ser el TSP con recogidas y entregas, o TSPPD \cite{TSPPD}, el TSP múltiple o m-TSP \cite{VRP5}, o el TSP dinámico \cite{DVRP}.
\end{sloppypar}

De entre todas las variantes propuestas en la literatura, en este trabajo se describirán únicamente el TSP asimétrico (Sección \ref{sec:ATSP}), y el m-TSP (Apartado \ref{sec:mTSP}). No se describirán más variantes del TSP ya que, posteriormente y relacionadas con el problema asignación de rutas de vehículos, se presentará una mayor cantidad de variantes, similares en concepto a las del TSP, pero con una formulación más compleja y una aplicabilidad e interés social más amplios.

\subsubsection{El problema del viajante comercial asimétrico}
\label{sec:ATSP}

El problema del viajante comercial asimétrico, \textit{Asymmetric Traveling Salesman Problem}, o simplemente ATSP, es una variante simple del TSP en el que la distancia entre cada par de nodos-clientes, es diferente dependiendo de la dirección en el que se ejecute. De esta forma, la igualdad $d_{ij}=d_{ji}$, la cual es característica intrínseca del TSP básico, no se verifica en este caso. Además de esto, la transformación de $A= \{(v_i,v_j): v_i,v_j  \in V,i\neq j \}$, a $E= \{(v_i,v_j): v_i,v_j  \in V,i< j \}$, válida para la versión del TSP convencional, no es aplicable para el ATSP. Por otro lado, por lo que respecta a la formulación del problema, esta es exactamente igual a su versión básica.

Esta simple variación dota al problema de mayor realismo, y hace que la complejidad computacional de tratarlo aumente ligeramente. Es por esto por lo que el ATSP goza también de gran popularidad en la comunidad científica, aportando numerosos trabajos a la literatura año tras año \cite{ATSP1,ATSP2}.

\subsubsection{El problema del viajante comercial múltiple}
\label{sec:mTSP}

El problema del viajante comercial múltiple, TSP múltiple, o m-TSP, es una generalización del TSP en la que entra en juego el concepto de depósito. En esta variante, en lugar de un solo vehículo, existe un número finito y fijo de éstos, con los que hay que conseguir visitar a todos los clientes. Para ello, cada vehículo tiene que trazar una ruta en la cual atiende a un cierto número de clientes. Además, todas las rutas tienen que empezar y terminar en un mismo sitio: el depósito central. 

Con todo esto, el objetivo del m-TSP es el de encontrar exactamente $m$ rutas, una para cada vehículo, de modo que cada cliente sea visitado exactamente una vez por alguno de esos vehículos, teniendo en cuenta que cada ruta no puede estar compuesta por más de $q$ nodos, y que el objetivo es minimizar la distancia total recorrida por todas las rutas. En este caso, la distancia total recorrida es igual al sumatorio de las distancias de cada una de las rutas de la solución.

Como puede deducirse, esta variante es más adaptable al mundo real que el propio TSP. Un ejemplo de ello es la planificación de imprentas de prensa, la cual es una de las primeras aplicaciones del m-TSP \cite{mTSP1}. Otro ejemplo es la asignación de ruta a autobuses escolares \cite{mTSP2}. A pesar de ello, este problema no ha recibido tanta atención como el TSP, y es menos aplicable que el VRP y sus variantes, por lo que los estudios realizados en el m-TSP son limitados.

\subsection{El problema de la asignación de rutas a vehículos}
\label{sec:VRP}

El problema de la asignación de rutas a vehículos, \textit{Vehicle Routing Problem}, o simplemente VRP, es uno de los problemas de optimización combinatoria más estudiados a lo largo de la historia. Desde que fue propuesto han sido centenares los estudios realizados centrados en el tratamiento de este problema. Estos esfuerzos pueden comprobarse en la literatura actual \cite{VRP1,VRP2}. Además, existen multitud de variantes del VRP básico. Algunas de estas variantes serán descritas posteriormente en esta tesis.

Realizando una mirada histórica, el primer VRP fue propuesto por Dantzing y Ramser \cite{VRP3} en el año 1959. En esta primera presentación los autores describieron una aplicación real para la entrega de gasolina a estaciones de servicio, y propusieron la formulación matemática para este problema.

Este primer estudio pretendía dar respuesta a una problemática muy extendida en el ámbito del transporte, aunque fue cinco años después cuando Clarke y Wright \cite{VRP4} propusieron el primer algoritmo efectivo para la resolución del problema: el algoritmo de ahorros. A partir de este momento, el mundo de la asignación de rutas a vehículos creció (y continúa creciendo) de forma importante, introduciendo diferentes variantes aplicadas a problemas reales y añadiendo a la formulación inicial distintas restricciones o características.

Como ya se ha mencionado anteriormente en este mismo trabajo, el interés en esta área no es puramente práctico. Al igual que los demás problemas de optimización combinatoria, el VRP es NP-Duro. Esto quiere decir que la elevada complejidad computacional hace que no sea posible la aplicación de métodos exactos que obtengan la solución óptima para todas las instancias del problema. Esto hace que posean un gran interés académico, y que a lo largo de la historia se hayan planteado multitud de alternativas para encontrar soluciones aproximadas a la óptima. Varias de estas técnicas se describirán en secciones posteriores a este capítulo.

En pocas palabras, podría decirse que el VRP es una extensión del m-TSP en la que cada cliente tiene asociada una demanda conocida. Además, el VRP no tiene por qué tener un número fijo de agentes viajeros ni un número máximo de clientes por ruta. Las diferencias y relaciones entre estos dos problemas pueden verse en \cite{VRP5}. Una definición sencilla del problema VRP podría ser la siguiente:

\textit{Dada la existencia de un depósito central, un número de clientes con una demanda finita y una flota de vehículos, el objetivo del problema es encontrar un conjunto de rutas, las cuales tienen que ser recorridas por la flota de vehículos que atiendan la demanda de los distintos clientes, minimizando la distancia total recorrida. Como restricciones de obligatorio cumplimiento, las rutas tienen que comenzar y finalizar en el depósito central, y los clientes tan solo pueden ser atendidos por un solo vehículo.}

\begin{figure}[tb]
	\centering
		\includegraphics[width=1.0\textwidth]{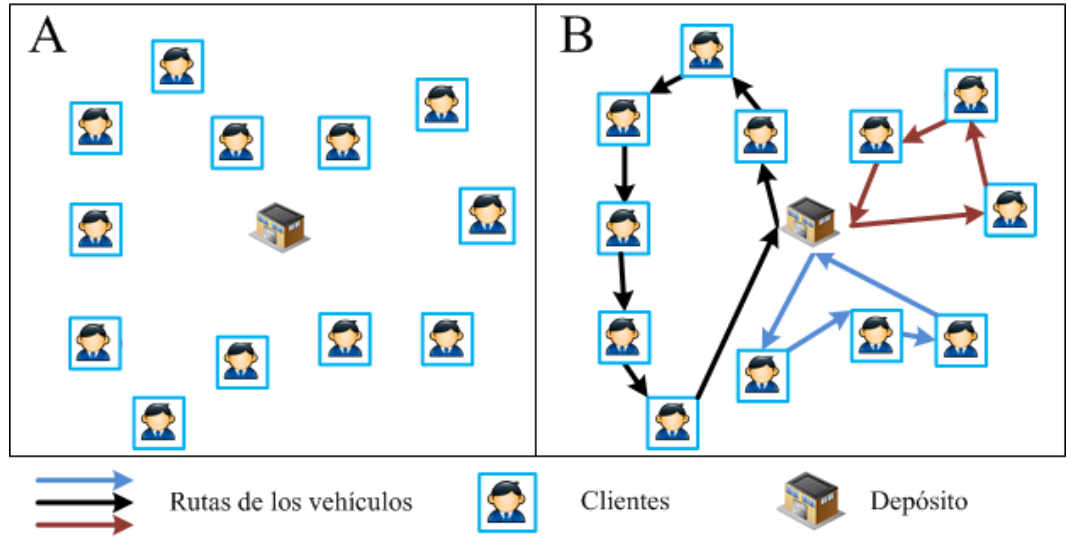}
	\caption{Ejemplo de una pequeña instancia del VRP básico y posible solución}
	\label{fig:VRP}
\end{figure}

En la Figura \ref{fig:VRP} se muestra una ilustración en la que se puede ver una pequeña instancia de un problema VRP y una posible solución a ésta.

\subsubsection{Definición del problema}
\label{sec:VRPform}

A continuación se muestra la formulación del problema VRP básico. A medida que se vayan presentando las distintas variantes de este problema, se agregarán nuevas restricciones y variables a esta primera formulación.

La red de transporte se modela mediante un grafo ponderado $G= (V,A)$, al igual que el TSP. Los nodos del grafo vienen representados por el conjunto $V=\{v_0,v_1,…,v_n\}$, donde $v_0$ representa el depósito y $V'=V-\{v_0\}$ representa al conjunto de los clientes. Por otro lado,	el conjunto $A= \{(v_i,v_j): v_i,v_j  \in V,i\neq j \}$ representa los arcos del grafo. Cada arco $(v_i,v_j)$ tiene asociado un coste $d_{ij}$ dentro de una matriz de costes $C=(d_{ij})$. En muchos contextos este coste es igual al tiempo del viaje entre $v_i$ y $v_j$. En caso contrario, existirá otra variable $t_{ij}$  que representa el tiempo que tarda la unidad de transporte en hacer el viaje. Al igual que el TSP, el VRP básico es un problema simétrico, de forma que $d_{ij}$ = $d_{ji}$ para cualquier valor de $(v_i,v_j )\in A$. Con todo esto, el conjunto $A$ puede ser remplazado por un conjunto de arcos $E= \{(v_i,v_j): v_i,v_j  \in V,i< j \}$.

Aparte de estas variables, existen otras destacables:

	\begin{itemize}
		\item Todos los clientes tienen que tener una demanda fijada. Esta demanda se representa mediante el conjunto $d=\{q_1,q_2,…,q_n\}$. 
		\item De forma opcional, los clientes pueden tener una variable $s_i$ asignada, igual al tiempo de servicio del cliente.
		\item Se asume que existen $m$ vehículos disponibles, donde $m_L < m < m_U$. Cuando $m_L = m_U$ se dice que $m$ tiene un valor fijo. Por el contrario, cuando $m_L$=1 y $m_U$ = $\infty$ se dice que es un valor libre. En los casos en los que $m$ no sea un valor previamente fijado, puede que el uso de un vehículo tenga un coste $f$ asociado, de forma que se penalice la excesiva utilización de estos.
		\item Cada ruta viene definida por la variable $r_i$, siendo $i$ el vehículo que la ejecuta.
	\end{itemize}

	\subsubsection{Formulación matemática}
\label{sec:VRPmathform}

A continuación se muestra una posible formulación matemática del VRP básico \cite{exactos4}:
	
\begin{equation}
	\sum_{(i,j)\in A}{d_{ij}x_{ij}} \label{VRPeq1}
\end{equation}
\begin{equation}
	\sum_{i}{x_{ij}=1}, \ \ \ \forall j \in V \label{VRPeq2}
\end{equation}
\begin{equation}
	\sum_{j}{x_{ij}=1}, \ \ \ \forall i \in V \label{VRPeq3}
\end{equation}
\begin{equation}
	\sum_{i}{x_{ij} \geq |S|-v(S)}, \ \ \ \forall S : S\subseteq V\backslash {1}, |S|\geq 2 \label{VRPeq4}
\end{equation}
\begin{equation}
	x_{ij} \in {0,1}, \ \ \ \forall {i,j} \in E; i \neq j \label{VRPeq5}
\end{equation}

Como se puede comprobar, la formulación básica del VRP es muy similar a la del TSP. Aun así, en algunas ocasiones el número de vehículos es considerado un valor fijo que no puede ser superado. Esto hace que las siguientes dos restricciones se unan a las anteriores, las cuales indican que $m$ es la cantidad de vehículos utilizados y que todos tienen que regresar al depósito central.	

\begin{equation}
	\sum_{i}{x_{0j}\leq m}, \ \ \ \forall j \in V \label{VRPeq6}
\end{equation}
\begin{equation}
	\sum_{j}{x_{i0}\leq m}, \ \ \ \forall i \in V \label{VRPeq7}
\end{equation}

Como ya se ha visto en formulaciones anteriores, la variable $x_{ij}$ es una variable binaria que determina si el arco $(i,j)$ es utilizado o no en la solución. La fórmula \ref{VRPeq1} es la función objetivo, la cual hay que minimizar en la mayor medida posible. Por otro lado, las fórmulas \ref{VRPeq2}y \ref{VRPeq3} aseguran que todo cliente es visitado por alguna ruta exactamente una vez. Para finalizar, la restricción \ref{VRPeq4} sirve para la eliminación de sub-tours.

\subsection{Variantes del VRP}
\label{sec:VRPvariantes}

Es de interés remarcar que la formulación descrita en el apartado anterior corresponde a la versión más básica del VRP, la cual fue la primera en ser formulada. Pese a esto, este problema no suele ser muy estudiado actualmente en su versión más convencional, ya que, aunque es más aplicable que el TSP o el m-TSP, resulta insuficiente para abordar gran parte de los problemas surgidos hoy en día en el mundo real. Es por esto por lo que a lo largo de la historia se han ido introduciendo diferentes modificaciones de este primer problema, con el propósito de adaptarlo a las distintas necesidades reales. De esta manera, y como norma general, cuando se trata el VRP en la literatura, ya sea utilizándolo como problema de benchmarking o para intentar resolverlo con la mayor eficacia posible, se realiza sobre una de las innumerables variantes de éste. A continuación se van a describir varias de las variantes más famosas. Algunas de éstas serán utilizadas para la validación de la técnica propuesta en esta tesis doctoral.

\subsubsection{Problema de la asignación de rutas a vehículos con capacidad limitada}
\label{sec:CVRP}

El problema de la asignación de rutas a vehículos con capacidad limitada, \textit{Capacitated Vehicle Routing Problem} o CVRP \cite{CVRP1}, es una variante del VRP básico en la que cada uno de los vehículos cuenta con una capacidad máxima limitada. En este problema concreto todos los vehículos de la flota tienen el misma capacidad. De esta manera, el objetivo del CVRP es encontrar los recorridos que deben realizar los distintos vehículos utilizados, de forma que se minimice la distancia total recorrida y se satisfaga la demanda de los clientes, añadiendo la restricción obligatoria de que el sumatorio de las demandas de los clientes de una misma ruta no puede superar la capacidad máxima del vehículo que la ejecuta. Actualmente, el CVRP es una de las variantes del VRP con más reconocimiento, siendo numerosos los estudios que utilizan este problema en su experimentación \cite{CVRP2,CVRP3}.

Al igual que en el VRP básico, en el CVRP no existe un numero definido de unidades de transporte, por lo que, en caso de no estar fijado, el problema tiene que buscar el mínimo número de vehículos a utilizar.

En este caso, a la formulación vista en la Sección \ref{sec:VRPform} se debe añadir una nueva variable $Q$, siendo ésta la capacidad total de los vehículos. Por otro lado, las siguientes restricciones han de ser agregadas a la formulación matemática descrita en la Sección \ref{sec:VRPmathform}:

\begin{equation}
	\sum_{i\in S}{q_{i}y^r_i\leq Q}, \ \ \ \forall r \in K \label{CVRPeq1}
\end{equation}
\begin{equation}
	y^r_i\in \{0,1\}, \ \ \ \forall r \in K \label{CVRPeq2}
\end{equation}

Donde $K$ es un conjunto de unidades móviles suficientemente grande como para atender la demanda de todos los clientes, y la variable $y_i^r$ es una variable binaria la cual es igual a 1 si el vehículo $r$ atiende la demanda del cliente $i$, y 0 en caso contrario. De esta forma, la restricción \ref{CVRPeq1} permite que el sumatorio de las demandas de una ruta no supere la capacidad máxima de los vehículos.



\subsubsection{Problema de la asignación de rutas a vehículos con ventanas de tiempo}
\label{sec:VRPTW}

El problema de la asignación de rutas a vehículos con ventanas de tiempo, \textit{Vehicle Routing Problem with Time Windows}, o VRPTW, es una extensión del CVRP, en el que aparte de las restricciones de capacidad de cada uno de los vehículos, cada cliente tiene una ventana temporal asociada. Este rango temporal posee un límite inferior y un límite superior que las unidades móviles tienen que respetar. Dicho de otra manera, cada cliente tiene que ser atendido en un momento que se encuentre dentro de su ventana temporal.

Por lo tanto, una ruta no será factible si un vehículo llega a la posición de un cliente después del límite superior del intervalo. Por el contrario, puede que la planificación haga que una unidad móvil llegue a un consumidor antes de su límite inferior. En este caso, el cliente no puede ser servido antes de este límite, por lo que el vehículo tendrá que esperar hasta que llegue el momento para poder abastecerlo. Además, el depósito central también posee una ventana temporal, la cual restringe el periodo de actividad de cada unidad móvil con el fin de adecuarlo a este intervalo.

\begin{sloppypar}
Este problema ha sido ampliamente estudiado tanto en el pasado \cite{VRPTW1,VRPTW2,VRPTW3}, como en la actualidad \cite{VRPTW4,VRPTW5}. Una de las razones por las que ha suscitado tanto interés ha sido su doble naturaleza, ya que podría considerarse como un problema de dos fases, una fase referente al problema de la asignación de rutas a vehículos, y otra fase referente a la planificación o \textit{scheduling} de los consumidores.
\end{sloppypar}

Respecto a la formulación matemática del VRPTW, ésta puede representarse de varias formas, haciendo uso de más o menos variables \cite{VRPTW6,VRPTW7}. Una de las formulaciones más interesantes puede encontrarse en \cite{VRPTW8}.

\subsubsection{Problema de la asignación de rutas a vehículos con ventanas de tiempo flexibles}
\label{sec:VRPSTW}

El problema de la asignación de rutas a vehículos con ventanas de tiempo flexibles, \textit{Vehicle Routing Problem with Soft Windows}, o VRPSTW es una variante del problema anteriormente descrito. Haciendo un análisis a la literatura se podría afirmar que es menos conocido que el VRPTW simple. Aun así, han sido varios los estudios que lo han tratado a lo largo de la historia \cite{VRPSTW2,VRPSTW3}. En este problema, las restricciones que suponen las ventanas de tiempo pueden traspasarse, acarreando con ello  una penalización en la función objetivo. Normalmente esta penalización es un valor abstracto, con el que se intenta cuantificar la ``insatisfacción" del cliente por no haber logrado respetar sus condiciones. Es por esto por lo que la función de penalización cambia según el estudio, teniendo en cuenta que cuanto mayor sea la penalización, se actuará con menor tolerancia frente a soluciones que incumplan las restricciones, y viceversa.

\subsubsection{Problema de la asignación de rutas a vehículos heterogéneos}
\label{sec:HVRP}

El problema de la asignación de rutas a vehículos heterogéneos, \textit{Heterogeneous Vehicle Routing Problem}, o HVRP, es una variante del CVRP en la que la flota de vehículos disponible está compuesta por unidades móviles de diferentes características. De esta manera, existirán diferentes tipos de vehículos, los cuales tendrán distintas capacidades, y diferentes costes de utilización. Teniendo en cuenta esto, uno de los retos que plantea este problema consiste en encontrar un equilibrio entre el coste de las unidades móviles utilizadas y las características de estas para atender las demandas de los clientes. Dentro de este tipo de problemas existe una gran cantidad de sub-variantes, cada una de las cuales con particularidades que las diferencia de las demás. Se recomienda la lectura del trabajo \cite{HVRPvariantes} para obtener más información acerca de estas variantes.

La primera formulación para un problema de este tipo se propuso en 1957, en un problema relacionado con el aprovisionamiento de aceite \cite{HVRP1}. Desde entonces se han planteado distintas variantes de este problema, cada una con su diferente manera de formularlo, como por ejemplo, la presentada por Gheysens et al \cite{HVRP2} en 1984.

\subsubsection{Problema de la asignación de rutas a vehículos con retorno de mercancías}
\label{sec:VRPB}

El enrutado de vehículos con retorno de mercancías, \textit{Vehicle Routing Problem with Backhauls}, o simplemente VRPB \cite{VRPB1}, es una variante del CVRP en la que los clientes pueden o bien demandar la entrega, o la recogida de cierta mercancía. En este problema la capacidad de los vehículos se convierte en un factor de extrema importancia, ya que hay que tener en cuenta que según qué cliente se visite habrá que recoger materiales que, obligatoriamente, tienen que caber en la unidad móvil.

En esta variante concreta los clientes no podrán solicitar la recogida y la entrega de forma simultánea. Además, de forma obligatoria, se realizarán primero las entregas de los materiales para dar paso después a las recogidas. Esto es así ya que de lo contrario supondría una circulación de materiales dentro del vehículo de transporte que podría ser contraproducente, como, por ejemplo, ocupar la parte exterior del maletero cuando al fondo de este aun queden mercancías por sacar.

\begin{figure}[tb]
	\centering
		\includegraphics[width=0.8\textwidth]{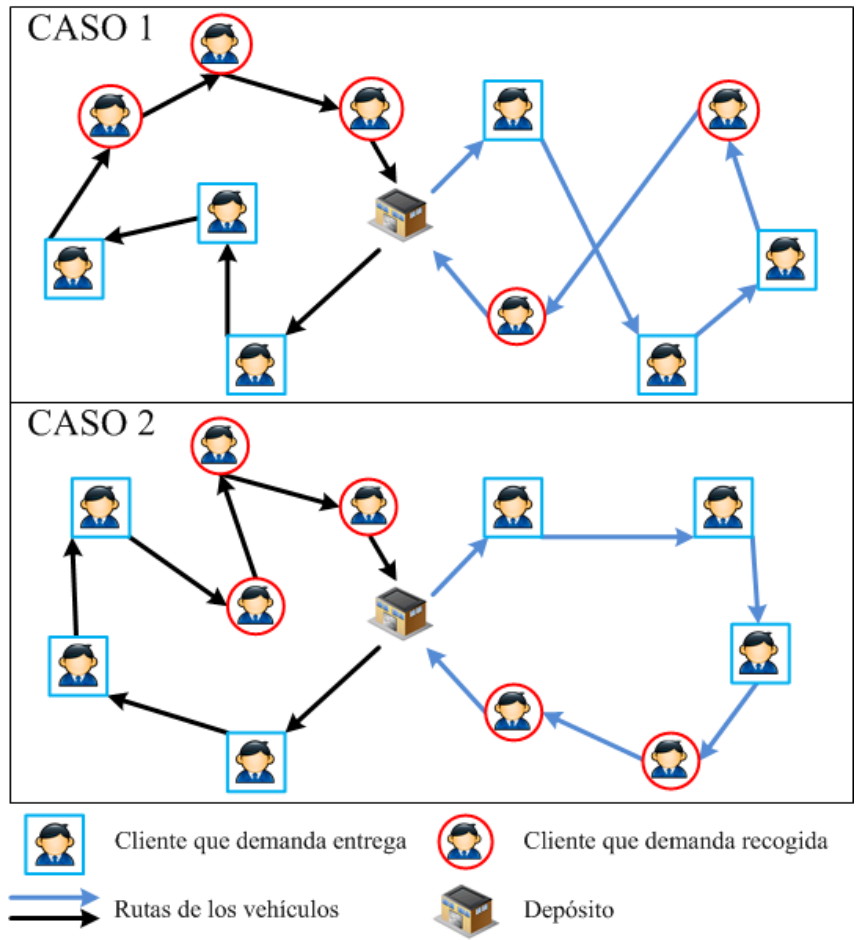}
	\caption{Dos ejemplos de instancias sencillas del VRPB}
	\label{fig:VRPB}
\end{figure}

En la Figura \ref{fig:VRPB} se muestran dos ejemplos de rutas en las que se da esta característica. La situación geográfica de los clientes en ambos es la misma, pero su naturaleza difiere. De esta forma se puede comprobar el cambio que supone en una ruta que unos clientes demanden entrega o recogida.

Una aplicación ampliamente conocida para este tipo de problema es la industria de comestibles, donde las tiendas o supermercados serían los clientes que solicitan entrega y los proveedores serían los solicitantes de recogida. En los últimos años se ha comprobado que supone un gran ahorro el hecho de combinar los dos servicios, visitando tanto a los proveedores como a los consumidores dentro de la misma ruta. La Interstate Commerce Commission estimó que, gracias a la introducción del \textit{backhauling}, en Estados Unidos el ahorro dentro de la industria de comestibles había alcanzado los 160 millones de dólares \cite{VRPB2}. Desde su formulación, el VRPB ha sido frecuentemente referenciado en la literatura \cite{VRPB3,VRPB4,VRPB5}.

%

\subsubsection{Problema de la asignación de rutas a vehículos con entregas y recogidas simultáneas}
\label{sec:VRPSPD}

El enrutado de vehículos con entregas y recogidas simultáneas, \textit{Vehicle Routing Problem with Pick Up and Deliveries}, o VRPSPD es una variante del VRP en la cual los clientes no solo demandan la entrega de cierta cantidad de materiales, si no que pueden también requerir la recogida de mercancía, al igual que en el VRPB. En este caso, el VRPSPD cuenta con la particularidad de que las entregas y recogidas pueden realizarse de forma simultánea. Esta variante fue introducida en 1989 \cite{VRPSPD1} y cuenta con una importancia vital, ya que es fácilmente aplicable al mundo real en casos como la industria de la bebida, por ejemplo, donde la botellas vacías en muchas ocasiones tienen que ser devueltas.

Esta variante es similar a la vista en la sección anterior, solo que en esta ocasión, tanto la recogida como la entrega pueden ser simultáneas en el mismo cliente, mientras que en el VRPB cada cliente puede, o bien requerir una recogida o una entrega, pero en ningún caso las dos a la vez.

Existen varios problemas similares al VRPSPD aparte del VRPB. Uno de ellos es el EDP \cite{VRPSPD2}, en el que los clientes pueden requerir de entrega y recogida, aunque en este caso no se realiza de forma simultánea, se realiza en dos fases, la primera de recogida y la segunda de entrega.

Otro tipo de problema similar es el \textit{Vehicle Routing Problem with Pick-up and Delivering}. En esta variante, al igual que en el VRPB, los clientes tan solo pueden solicitar entrega o recogida, pero a diferencia a este último, no hay prioridad sobre qué servicio es el que hay que realizar primero, pudiendo alternar recogidas y entregas.

\subsubsection{Problema de la asignación de rutas a vehículos dinámico}
\label{sec:DVRP}

En el VRP básico y en todas las variantes que se han visto hasta el momento, toda la información relacionada con las demandas y con el enrutado (tiempos, distancias, costes...) está disponible desde el principio, antes de construir las rutas y antes del día de la ejecución de éstas. En el problema de la asignación de rutas a vehículos dinámico, \textit{Dynamic Vehicle Routing Problem}, o DVRP, parte de esta información relevante no está disponible en el momento de diseñar las rutas, de forma que esta información puede ser conocida, o incluso alterada, mientras se están construyendo o ejecutando dichas rutas.

La primera referencia de un problema de este tipo pertenece a Wilson y Colvin \cite{DVRP1}, quienes plantearon una variación dinámica del \textit{arc routing problem} \cite{ARP} donde las peticiones de los clientes consistían en viajes que iban desde un origen hasta un destino concreto y que podían aparecer de forma dinámica. 

En sistemas de este tipo, los vehículos tienen que servir dos tipos de peticiones o solicitudes: las avanzadas y las inmediatas. Las primeras son las que los clientes han realizado antes de que comience el proceso de asignación de rutas, por lo que son conocidas desde un inicio. Por otra parte, las inmediatas, o \textit{immediate requests}, introducidas por Psaraftis \cite{DVRP2}, son aquellas que son recibidas de forma dinámica y que aparecen en tiempo real, en cualquier momento en el que la ruta está siendo ejecutada por la unidad móvil. Frecuentemente, el hecho de gestionar las peticiones inmediatas supone una gran complejidad, ya que requieren de una planificación en tiempo real de las rutas. De esta forma, según la flexibilidad del problema \cite{DVRP3}, la inserción de nuevas peticiones puede ser más o menos compleja. En sistemas en los que existan ventanas de tiempo, por ejemplo, la inserción se vuelve más compleja que en entornos donde no las hay.

En este tipo de problemas, la fuente más común de dinamismo viene dada por la llegada \textit{on-line} de peticiones por parte de los consumidores durante la ejecución de las rutas. Normalmente estas peticiones pueden requerir el abastecimiento de materiales \cite{DVRP4}, o bien la solicitud de un servicio \cite{DVRP5}. El dinamismo del tiempo de trayecto, el cual es factor muy común en el mundo real, también se ha tenido en cuenta es una gran cantidad de estudios \cite{DVRP6}. Finalmente, algunos trabajos recientes aportan dinamismo a la cantidad de la demanda de ciertos clientes ya conocidos \cite{DVRP7}, o a la disponibilidad del vehículo \cite{DVRP8}, en la que una posible avería de la unidad móvil es lo que aporta el dinamismo al sistema.

Como ha podido comprobarse a lo largo de esta extensa sección, el número de variantes del problema VRP es considerablemente amplio, signo inequívoco de su interés científico (tanto actual como histórico). Además de las presentadas, existen innumerables variantes adicionales. Con el objetivo de no extender en exceso este capítulo, estas variante no serán descritas en este documento, ya que se ha considerado que quedan fuera del alcance del trabajo presentado en esta tesis. 

Pese a esto, en la literatura se han propuesto otro tipo de problemas de asignación de rutas a vehículos: los problemas ricos de asignación de rutas a vehículos. Los aspectos básicos de estos problemas serán descritos de manera breve en el siguiente apartado.

\subsubsection{Problemas ricos de asignación de rutas a vehículos}
\label{sec:RVRP}

Los problemas que se han descrito hasta el momento, pese a intentar adaptarse con la mayor fidelidad posible, no son perfectamente aplicables a situaciones complejas del mundo real. Es por esto por lo que estos problemas anteriormente mencionados trabajan sobre situaciones y escenarios ideales, y son considerados como problemas académicos. Pese a esto, como ya se ha visto a lo largo de este capítulo, son problemas poseedores de una gran importancia e interés, tanto académico como científico.

Como ya se ha mencionado anteriormente, dada la gran ventaja económica y social que reporta una eficiente planificación de rutas, los problemas del tipo VRP reciben mucha atención dentro de la comunidad científica. Es por esta razón por la que recientemente se han propuesto unos modelos diferentes de VRP. Estos nuevos problemas son los problemas ricos de asignación de rutas a vehículos, \textit{Rich Vehicle Routing Problems}, o R-VRPs. Estos problemas son casos especiales del VRP, con la distinción de poseer un gran conjunto de restricciones y una formulación extraordinariamente compleja. El objetivo principal de los R-VRPs es el de considerar esas restricciones que se encuentran en situaciones reales, proponiendo soluciones que sean perfectamente ejecutables en entornos del mundo real.

Existen numerosos trabajos en la literatura centrados en este tipo de problemas \cite{RVRP1,RVRP2}. Como ejemplos concretos, en el trabajo presentado por Ceselli et al. se presenta un R-VRP con una gran cantidad de restricciones, entre las que se pueden mencionar la existencia de múltiples capacidades, ventanas de tiempo asociadas con los clientes y depósitos, restricciones de distancia y duración en las rutas, la opción de realizar rutas que no finalicen en un deposito o incompatibilidades entre diferentes productos \cite{RVRP3}. Otro ejemplo puede ser el R-VRP propuesto por Pellegrini et al. en \cite{RVRP4}, donde presentan una variante con vehículos heterogéneos, ventanas de tiempo múltiples, duración máxima de rutas y la posibilidad de que los clientes sean visitados en más de una ocasión (por un mismo vehículo o por más de uno).

Como ejemplos más recientes se pueden destacar los trabajos \cite{de2015gvns} y \cite{amorim2014rich}. En el primero de estos artículos se propone un RVPR con ventanas de tiempo estrictas y flexibles, flota heterogénea, prioridades en los clientes y restricciones vehículo-cliente. La solución propuesta por los autores de tal estudio ya ha sido integrada en una herramienta de gestión de flotas en el archipiélago canario. el segundo de los trabajos hace frente a un problema de gestión de alimentos perecederos. En este artículo se presenta un VRP con flota heterogénea, dependiente del sitio y con ventanas de tiempo múltiples. Un ejemplo reciente adicional podría ser el presentado por Lahyani et al. en \cite{lahyani2015multi}, en el que desarrollan un R-VRP para la recolección de aceite de oliva en Túnez. En esta ocasión, el R-VRP propuesto tiene en cuenta restricciones relacionadas con el multi-producto, multi-periodo y multi-compartimento.

Dicho esto, resulta de interés mencionar que los R-VRPs poseen ciertas ventajas y desventajas frente a las variantes del VRP analizadas con anterioridad. Es cierto que los R-VRPs gozan de una gran adaptabilidad a situaciones del mundo real, hecho que es realmente valorado a nivel industrial y social. Aun así, estos problemas son muy específicos, y cada uno de ellos es aplicable a un único caso real. Es por esto por lo que la replicabilidad de los R-VRP es muy inferior a las variantes descritas anteriormente en este trabajo. Además de esto, estos problemas requieren un gran esfuerzo computacional, ya sea para diseñarlos, implementarlos, o para ejecutarlos.

Pese a estas desventajas, el autor de este trabajo es consciente del gran interés que suscita este tipo de problemas. Por ello, en esta tesis se dedicará un capítulo a la descripción e implementación de un R-VRP, el cual servirá para validar la aplicabilidad del modelo presentado en este trabajo a problemas de esta índole. Toda esta información se encuentra recogida en el Capítulo \ref{cha:aplicacion}.



\begin{savequote}[40mm]
Lo que sabemos es una gota de agua; lo que ignoramos es el océano.
\qauthor{Isaac Newton}
\end{savequote}

\chapter{Técnicas para la resolución de problemas}
\label{sec:tecnicas_de_resolucion}

\graphicspath{ {2.2_tecnicas/figures/} }

\lettrine{E}{}{xisten} varias formas de enfocar la resolución de los problemas que se han visto hasta ahora en esta tesis. A lo largo de la historia las técnicas de resolución más utilizadas han sido las heurísticas y las meta-heurísticas. Estos métodos son capaces de abordar cualquier tipo de problema de optimización combinatoria ofreciendo soluciones aproximadas, ya que, como se ha explicado anteriormente, para estos problemas no existe algoritmo alguno que permita encontrar en un tiempo factible la solución óptima para todas las instancias. En este apartado se van a describir varias de estas técnicas, concretamente, las más utilizadas dentro de la optimización de rutas.

Antes de comenzar, merece la pena mencionar los enfoques exactos, los cuales rastrean todo el espacio de soluciones\footnote{Espacio de soluciones: conjunto de soluciones posibles de un problema} para encontrar siempre la solución óptima. Estas técnicas son válidas tan solo para instancias con pocos clientes a los que atender o para relajaciones de problemas más complejos. En caso contrario, los tiempos de ejecución de estos métodos son impermisibles. Ejemplos de estos algoritmos son el propuesto por Feillet et al. en \cite{exactos1}, el presentado por Hirabayashi et al. para el problema de rutas de arcos capacitados \cite{exactos2}, o el desarrollado por Baldacci et al. para una variante del problema CVRP \cite{exactos3}. Para más información acerca de estos métodos, es conveniente consultar \cite{exactos4} o \cite{exactos5}.

Como ya se ha mencionado, los métodos más utilizados para la resolución de los problemas que se están tratando en este trabajo son las heurísticas y las meta-heurísticas. Una heurística es una técnica de optimización que trata de encontrar solución a un problema haciendo uso de información específica y conocimiento concreto del problema que esté tratando. De esta forma, una heurística explora el espacio de soluciones haciendo una intensificación de la búsqueda dentro de aquellas áreas que puedan ser más prometedoras, con la intención de encontrar resultados de forma rápida. Generalmente, estas técnicas son aplicadas sobre problemas conocidos y con una formulación simple, como por ejemplo el TSP o el VRP, ya que la dificultad de encontrar heurísticas adaptables a problemas más complicados se hace difícil en muchas ocasiones.

\begin{sloppypar}
Por otro lado, una meta-heurística es un método de optimización que busca solución a un problema específico utilizando tan solo información general y conocimiento común a todos los problemas de optimización. Las meta-heurísticas exploran un área mucho mayor del espacio de soluciones, con el propósito de encontrar una buena solución independientemente del problema que estén tratando. Por esta razón, estas técnicas son más apropiadas para su aplicación en problemas con una compleja formulación, ya que no utilizan información específica a la hora de explorar el espacio de soluciones factibles del problema. Existe un gran número de meta-heurísticas en la literatura. Algunas de estas están basadas en una búsqueda simple, como el recocido simulado \cite{SA,SA2} o la búsqueda tabú \cite{TS1,TS2}. Otras muchas meta-heurísticas se basan en búsquedas múltiples, como los algoritmos genéticos \cite{genetic1,genetic2}, las colonias de hormigas \cite{ant1,ant2} o el método imperialista \cite{imperialist}. Las meta-heurísticas pueden también ser clasificadas como algoritmos trayectoriales y algoritmos constructivos. Los primeros comienzan con una o varias soluciones iniciales completas, las cuales son modificadas hasta llegar a la solución final. Los segundos, por otro lado, parten de una o varias soluciones iniciales parciales, las cuales van construyéndose hasta dar con las soluciones finales.
\end{sloppypar}

De esta forma, las heurísticas se centran en la resolución de problemas con una formulación simple, intentando encontrar la solución óptima de una forma rápida. Las meta-heurísticas, por el contrario, pueden ser aplicadas a una amplia variedad de problemas reales, o con una complejidad elevada, para los que sería muy complicado implementar una heurística específica. Las técnicas que se verán en este apartado, y aquellas con las que se va a trabajar en esta tesis doctoral se encuadran en el grupo de las meta-heurísticas. 

Con todo esto, este capítulo está dividido en tres grandes bloques. El primero de ellos, sección \ref{sec:trayectoriales}, corresponde a los métodos trayectoriales de búsqueda simple . El segundo de ello, apartado \ref{sec:poblacionales}, tratará sobre las técnicas poblacionales, mientras que el último de ellos hablará sobre las meta-heurísticas multi-poblacionales (sección \ref{sec:multi}).

\section{Métodos trayectoriales de búsqueda simple}
\label{sec:trayectoriales}

Los algoritmos trayectoriales de búsqueda simple son técnicas de optimización aproximada en los que se parte de una solución inicial completa, que se transforma progresivamente en una nueva solución mediante pequeñas modificaciones parciales. Estas modificaciones se realizan hasta conseguir la mejor solución posible en un tiempo factible. Pese a que existen muchos tipos de algoritmos de este tipo, en este trabajo tan solo se describirán tres, los cuales son considerados como los más utilizados en la literatura a lo largo de la historia: las búsquedas locales simples, el recocido simulado, y la búsqueda tabú.

\subsection{Búsquedas locales simples}
\label{sec:locales}

Los métodos de búsqueda local son una clase de algoritmos de aproximación basados en mejorar de forma iterativa la solución de un problema. Su forma de funcionar puede explicarse brevemente de la siguiente forma: la búsqueda comienza generando una solución inicial, llamada \textit{estado inicial}. A partir de ella, y haciendo uso de una función generadora de sucesores, se rastrea su vecindario\footnote{Cada solución tiene un conjunto de soluciones asociadas, llamadas vecindario o entorno. Estas soluciones pueden ser alcanzadas realizando una sola modificación en la solución actual} en busca de la mejor solución posible. Al encontrar una mejor solución, el algoritmo se mueve\footnote{Cada solución del vecindario puede obtenerse directamente a partir de la solución actual realizando una operación llamada movimiento} hasta ella, designándola como \textit{estado actual}. Este proceso se ejecuta de forma iterativa, y la ejecución de la búsqueda finaliza en el momento en el que la función generadora de sucesores no es capaz de encontrar una solución que mejore la actual, es decir, cuando no exista una solución mejor dentro de su vecindad.

El punto débil de las búsquedas locales es la facilidad que tienen para caer en un óptimo local sin posibilidad de salir de él. Por esta razón los algoritmos de búsqueda local son técnicas que devuelven óptimos locales, y cuyo rendimiento depende completamente de la solución inicial con la que comiencen el proceso. Dicho de otra forma, si la solución inicial queda colocada en una posición favorable, la búsqueda alcanzará la solución óptima; si esto no ocurre, el algoritmo devolverá irremediablemente un óptimo local. A lo largo de la historia se han desarrollado un gran número de técnicas para solucionar esta desventaja. La búsqueda tabú y el recocido simulado son dos de estos algoritmos, los cuales serán descritos más adelante. 

A continuación se van a pormenorizar dos procesos importantes para este tipo de algoritmos. Estas nociones van a ser aplicables también a las meta-heurísticas que se verán en apartados posteriores. Estos dos procesos son la creación de la solución inicial por un lado, y la función generadora de sucesores por el otro.

\subsubsection{Solución inicial}
\label{sec:solinic}

Como ya se ha indicado anteriormente, este tipo de algoritmos comienzan con una solución inicial, denominada \textit{estado inicial}, que se va modificando a lo largo de la ejecución con el objetivo de encontrar una solución aceptable. La solución inicial puede ser generada de forma aleatoria o bien puede utilizarse una función heurística para comenzar el proceso con una solución mejorada. Lógicamente la solución inicial no debe ser la solución óptima, ni siquiera tiene porqué acercarse, pero en algunos casos puede significar un buen punto de partida para comenzar con la búsqueda. Varias de las funciones heurísticas más utilizadas en este ámbito pueden ser el nearest neighbor, o vecino cercano  \cite{NN1}, o la heurística I1 de Solomon \cite{solomon}. En relación con esto, en los trabajos \cite{osaba1} y \cite{osaba2} pueden encontrarse sendos estudios realizados por el autor de esta tesis en relación al impacto que produce el inicializar las soluciones iniciales para el TSP y el NQP utilizando un algoritmo genético.

\subsubsection{Generador de sucesores}
\label{sec:gensuc}

La función generadora de sucesores es la encargada de realizar los movimientos sobre el estado actual con el objetivo de hacer avanzar al algoritmo hacia nuevas soluciones. Estos operadores pueden ser clasificados en dos tipos: operadores intra-ruta y operadores inter-ruta \cite{savelsbergh}. Los primeros realizan modificaciones dentro de una misma ruta, sin que las demás se vean afectadas. Este dato carece de relevancia dentro de los problemas del tipo TSP, ya que, al contrario que en la familia de problemas VRP, solo se cuenta con una ruta. Los segundos se caracterizan por realizar intercambios de nodos o aristas entre diferentes rutas, y son muy efectivos para problemas del tipo VRP. Aun así, estos operadores también pueden ser utilizados como operadores intra-ruta, como bien se mostrará más adelante.

\begin{figure}
	\centering
		\includegraphics[width=0.7\textwidth]{2opt.png}
	\caption{Ejemplo de un movimiento 2-opt en una ruta compuesta por ocho nodos}
	\label{fig:2opt}
\end{figure}

Un ejemplo de operador intra-ruta es el \textit{2-opt}. Este fue definido por Lin en 1965 \cite{exactos5}, y su forma de trabajar consta de dos pasos: el primero consiste en eliminar dos arcos dentro de una ruta existente, mientras que el segundo se ocupa de crear dos nuevos arcos evitando la generación de subtours. Este operador cuenta con gran popularidad y con un alto grado de efectividad. Un gran número de implementaciones hacen uso de este generador de sucesores, como por ejemplo \cite{2opt1} y \cite{2opt2}. En la Figura \ref{fig:2opt} se muestra un ejemplo de este tipo de movimiento.

\begin{figure}
	\centering
		\includegraphics[width=0.7\textwidth]{3opt.png}
	\caption{Ejemplo de un movimiento 3-opt en una ruta compuesta por ocho nodos}
	\label{fig:3opt}
\end{figure}

El \textit{3-opt} es otro operador de similares características. También definido por Lin, en este caso en lugar de dos se eliminan tres aristas, para generar después tres nuevas evitando la generación de subtours. Como puede resultar lógico, la complejidad que supone utilizar este operador es mucho mayor que el \textit{2-opt} simple. En la Figura \ref{fig:3opt} se representa un ejemplo del funcionamiento de este operador.

\begin{figure}
	\centering
		\includegraphics[width=0.7\textwidth]{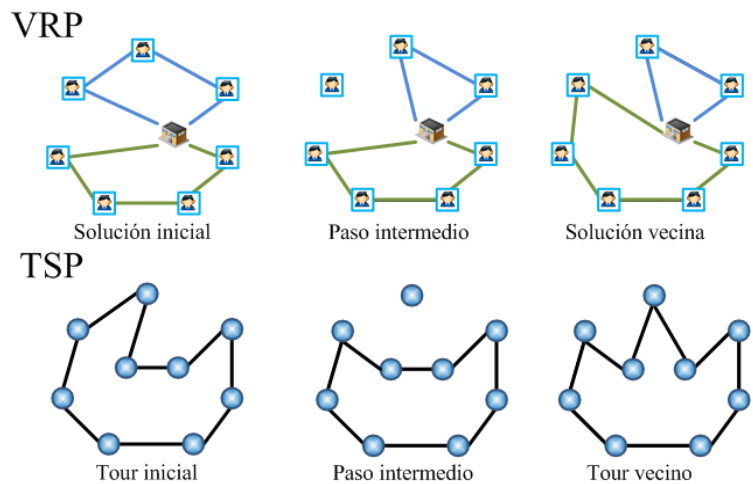}
	\caption{Ejemplo del Vertex Insertion para el VRP y TSP}
	\label{fig:vertex}
\end{figure}

En cuanto a los operadores inter-ruta, los más utilizados son el \textit{Vertex Insertion} y el \textit{Swapping}, también utilizados frecuentemente como intra-ruta. En el primero de ellos, en primer lugar se selecciona un nodo al azar de la ruta, o de una de las rutas. Este nodo se extrae para después ser insertado en otra posición aleatoria, generando con ello una nueva solución. En el caso del VRP, se trata de un movimiento de nodo de una ruta a otra. Este operador es uno de los más utilizados en la literatura, ya que es sencillo de implementar y obtiene buenos resultados \cite{vertex1} \cite{vertex2}. En la Figura \ref{fig:vertex} se muestran dos ejemplos de este tipo de movimiento.

\begin{figure}
	\centering
		\includegraphics[width=0.7\textwidth]{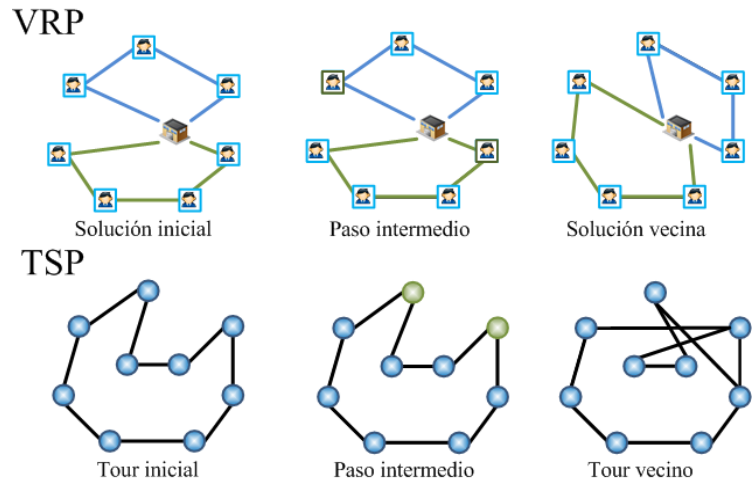}
	\caption{Ejemplo del Swapping para el VRP y TSP}
	\label{fig:swapping}
\end{figure}

El segundo de los operadores mencionados, el \textit{Swapping}, selecciona dos nodos al azar para después intercambiar sus posiciones. En el caso del VRP el proceso de intercambio se realiza entre dos rutas, de forma que el operador es similar al anterior, con la diferencia de que son dos los clientes que se ven afectados, uno por cada ruta implicada. Dos ejemplos en los que se usa este tipo de movimiento son los trabajos \cite{swapping1} y \cite{swapping2}. En la Figura \ref{fig:swapping} se muestran dos posibles ejemplos para este operador.

Como ya se ha comentado con anterioridad, el número de operadores generadores de sucesores disponibles en la literatura es inmenso. En este trabajo se han descrito varios de los más utilizados, siendo estos, además, los que van a ser utilizados para la experimentación posterior. Para encontrar información adicional sobre este tema, se recomienda la lectura de \cite{VRPTW3}, \cite{generadores1}, o \cite{TSP2}.

\subsection{Búsqueda Tabú}
\label{sec:tabu}

Los orígenes de la búsqueda tabú datan de finales de los años 70 y principios de los 80, de la mano de Fred Glover. Durante estos primeros años algunas de las ideas fueron utilizadas en varios problemas de planificación y otros dominios. El éxito de estas aplicaciones y el creciente interés por las heurísticas en general empujaron al mismo Glover a formalizar los conceptos fundamentales de la búsqueda tabú \cite{tabu1}. Este tipo de búsquedas se crearon en un intento de dotar de inteligencia a los algoritmos de búsqueda local. Según Glover ``la búsqueda tabú guía un procedimiento de búsqueda local para explorar el espacio de soluciones más allá del óptimo local". Desde su creación, esta meta-heurística se ha convertido en una de las técnicas más utilizadas para la resolución de problemas de optimización combinatoria \cite{tabu2}, en gran parte gracias a que los resultados obtenidos al emplearla son prometedores. Aplicaciones actuales de esta técnica pueden verse en \cite{tabu3} o \cite{tabu4}. El autor de esta tesis también ha realizado investigaciones centradas en esta técnica \cite{tabu5}.

Al basarse en las búsquedas locales, los algoritmos tabú comienzan su ejecución con una solución inicial, la cual puede ser aleatoria o bien generada por cualquiera de las funciones que se han visto en el apartado anterior. Después, haciendo uso de una función generadora de sucesores, también vistas anteriormente, la solución evoluciona de forma iterativa.

Hasta aquí la filosofía tabú no aporta ninguna novedad con respecto a las búsquedas anteriormente descritas. La verdadera novedad viene dada por el mecanismo implementado para huir de los óptimos locales. Este mecanismo es llamado \textit{memoria}, y está ideado con el objetivo de dirigir la búsqueda teniendo en cuenta el historial de ésta. Para esto, los algoritmos tabú mantienen una lista con los movimientos recientemente ejecutados, los cuales son considerados como \textit{prohibidos}. De esta forma, cada vez que la función generadora de sucesores propone un nuevo movimiento, el algoritmo lo analiza, y en caso de no estar prohibido, lo ejecuta y lo inserta en la lista tabú. En caso contrario lo evita, eligiendo en su lugar otro movimiento permitido. Otro factor indispensable en esta búsqueda es la posibilidad de aceptar soluciones peores a la actual, para así poder escapar de los óptimos locales y continuar la búsqueda por otras regiones del espacio de soluciones. 

Dentro de la memoria tabú, también llamada \textit{Short Term Memory}, se puede almacenar distinta información dependiendo del criterio de prohibición que se elija. Existen multitud de criterios para decidir si un movimiento es tabú o no, algunos más restrictivos que otros. En relación a esto se han realizado un gran número de estudios a lo largo de la historia \cite{tabu6}. Hay que tener en cuenta que, además, los diferentes criterios están estrechamente relacionados con el tipo de generador de sucesores que se esté aplicando.

Otro de los temas ámpliamente estudiados a lo largo de la historia, y aún sin determinar, es el tamaño ideal que debe tener la lista tabú \cite{iterations1}. Hay que tener en cuenta que si el tamaño es muy pequeño puede caerse con facilidad en un óptimo local. En cambio, si el tamaño es muy grande, restringe en exceso el proceso de búsqueda. En la literatura pueden encontrarse multitud de estrategias para determinar el tamaño adecuado de la lista tabú. El hecho de que existan tantas estrategias hace suponer que no existe un tamaño ideal para cada problema. Como bien se menciona en \cite{iterations1}, el tamaño de la lista óptimo dependerá directamente de varios factores, como el tamaño de la instancia o el generador de sucesores utilizado.

Finalmente, es conveniente indicar que existen otros tipos de memoria aparte de la ya referida \textit{Short Term Memory}. Esta memorias son la denominada \textit{Long Term Memory}, y la conocida como \textit{Intermediate Term Memory}. Se recomienda la lectura de los trabajos \cite{tabu2} o \cite{VRPSPD5} para obtener más información acerca de estos tipo de memoria. Otro concepto ampliamente extendido y que merece la pena mencionar es el criterio de aspiración. Cuando este criterio se satisface, permite al algoritmo que ejecute movimientos prohibidos. Se aconseja la lectura del artículo \cite{tabu6} para ampliar la información en este aspecto.

\subsection{Recocido simulado}
\label{sec:recocido}

El primer algoritmo de recocido simulado, o \textit{simulated annealing}, fue propuesto en 1953 por Metropolis et al. \cite{SA1}, aunque fue años más tarde, a mediados de la década de los 80, cuando, por un lado, Kirkpatric et al. \cite{SA2}, y por otro, Cerny \cite{SA3} introdujeron este concepto en el campo de la optimización combinatoria. Esta meta-heurística es una de las más utilizadas a lo largo de la historia gracias a su capacidad de converger hacia soluciones de alta calidad. Aun así, la gran desventaja de este método de búsqueda es su elevado coste computacional, mayor que otras meta-heurísticas como, por ejemplo, la búsqueda tabú.

A día de hoy, son multitud los estudios que se presentan anualmente los cuales hacen uso de esta técnica para resolver algún problema de optimización combinatoria, o de asignación de rutas a vehículos \cite{SA4,SA5}. El propio autor de esta tesis ha realizado varios estudios relacionados con el recocido simulado aplicado a problemas de planificación de rutas a vehículos \cite{SA6,SA7,SA8}.

El funcionamiento del recocido simulado se basa en el principio físico de enfriamiento de los metales, el cual les hace pasar de una situación liquida a una sólida. Los algoritmos de este tipo se implementan de la misma forma que se implementan los algoritmos de búsqueda local, es decir, con su correspondiente función de inicialización de la solución y generador de sucesores. La diferencia con las búsquedas locales radica en que el criterio de sustitución de una nueva solución no es determinista, sino que está basado en el criterio de Metropolis. De esta forma, cuando la función generadora de sucesores genera un nuevo estado, si éste es de mejor calidad que el actual, se acepta. En caso contrario, la solución es aceptada de forma probabilística en función de la temperatura del sistema y del aumento que supone en la función objetivo. 

La característica más importante del recocido simulado, por lo tanto, es el hecho de que no solo acepta soluciones mejores a la actual, sino que también permite aceptar un número decreciente de soluciones que supongan una perdida en la calidad de la solución. De esta forma, permite al proceso de búsqueda huir de los óptimos locales en caso de quedarse atrapado en ellos y, así, poder continuar la búsqueda por otras áreas del espacio de soluciones en busca del óptimo global.

La temperatura del metal se considera dentro de la técnica como un parámetro de control, el cual va a actuar como juez a la hora de aceptar o no nuevas soluciones. Al tratarse de un proceso de enfriamiento, en un principio el valor de temperatura será muy elevado, de forma que el algoritmo tendrá una gran tolerancia para admitir soluciones peores a la actual, permitiendo a la solución moverme libremente por el espacio de soluciones de la misma forma que las partículas del metal se desplazan a su antojo en su fase líquida. Después, progresivamente, la temperatura desciende poco a poco, permitiendo cada vez menos soluciones peores a la actual. Finalmente, cuando la temperatura se aproxime a cero, no se aceptará ninguna pérdida de calidad, aceptando tan solo soluciones que mejoren a la actual. En este momento el algoritmo se habrá transformado en un proceso de búsqueda local convencional y finalizará su ejecución de la misma forma que un algoritmo de este tipo.

\section{Algoritmos poblacionales}
\label{sec:poblacionales}

Al igual que los algoritmos trayectoriales de búsqueda simple, los algoritmos poblacionales son técnicas de optimización aproximada, los cuales operan sobre soluciones completas que se van transformando durante el proceso. Estos métodos, a diferencia de los algoritmos anteriormente descritos, operan en cada iteración con un conjunto de soluciones (población) que interactúan entre sí. De esta forma, proporcionan de forma intrínseca un proceso de búsqueda múltiple paralela en el espacio de soluciones del problema. Estas técnicas gozan actualmente de una gran popularidad, lo que conlleva a la creación y perfeccionamiento de una gran cantidad de técnicas año tras año. En esta sección se describirán varios algoritmos de este tipo, cuya popularidad está más que contrastada en la comunidad científica.

\subsection{Algoritmos genéticos}
\label{sec:genetico}

Siendo probablemente la técnica poblacional más utiliza a lo largo de la historia, los algoritmos genéticos (GA) se basan en el proceso genético de los organismos vivos y en la ley de la evolución de las especies, propuesta por Darwing. En el mundo real, a lo largo de las generaciones, las poblaciones evolucionan acorde a la selección natural y la supervivencia de los más fuertes. En un intento por imitar este proceso natural se crearon los algoritmos genéticos. Los principios básicos de esta técnica fueron propuestos por Holland en 1975 \cite{genetico}, aunque su uso práctico para la resolución de problemas complejos fue demostrado años más tarde por De jong \cite{genetic2} y Goldberg \cite{genetic1}. El poder de este tipo de algoritmos viene dado por su robustez y por poder tratar con éxito una gran variedad de problemas provenientes de diferentes áreas, como el transporte \cite{transport,transport1}, la industria \cite{industry}, o la ingeniería del software \cite{software1,software2}. Desde su creación, han sido muchos los estudios realizados, y los artículos y los libros publicados acerca de estos métodos.

Al tratarse de técnicas poblacionales, los algoritmos genéticos trabajan con una población de individuos, cada uno de los cuales, generalmente, representa una solución factible del problema a resolver. Cada individuo tiene asociado un valor, o puntuación, llamado \textit{fitness}, el cual se le asigna mediante una función objetivo. Este valor está directamente relacionado con la calidad o la bondad de dicha solución. 

El primer paso del proceso es generar la población inicial y evaluarla. Después, por cada iteración, se seleccionan los individuos más adecuados para el proceso de cruce, en el que se generan nuevos individuos mediante el cruce de los cromosomas seleccionados. Tras esto, se aplica el proceso de mutación, en el que se seleccionan ciertos individuos al azar para aplicar sobre ellos una pequeña modificación. Finalmente, la población se reduce hasta su tamaño inicial, eliminando de ésta los individuos menos interesantes, en un proceso denominado selección de supervivientes. La estructura básica de un algoritmo genético puede verse en el Algoritmo \ref{alg:GA}.

\begin{algorithm}[tb]
	 \SetAlgoLined
	
	 Inicialización y evaluación de la población inicial
	
	 \Repeat{hasta que el criterio de terminación se satisfaga}{

	   Proceso de selección de padres
		
		 Fase de cruce
		
		 Fase de mutación
		
		 Proceso de selección de supervivientes
		
		}
	 El algoritmo devuelve el fitness del mejor individuo de la población 
   \caption{Pseudocódigo de un algoritmo genético básico}
	 \label{alg:GA}
\end{algorithm}

A lo largo de la historia, han sido infinidad los trabajos enfocados en el estudio de los algoritmos genéticos. Estos estudios pueden agruparse en tres diferentes categorías:

\begin{itemize}
	\item Aplicación práctica: Estas investigaciones se centran en la aplicación de algoritmos genéticos a problemas de optimización concretos. De entre las tres categorías que van a listarse, esta es la que más trabajos aporta a la literatura anualmente. Dos subcategorías pueden identificarse en este primer grupo: trabajos que presentan variaciones o implementaciones del algoritmo genético clásico \cite{geneticEj1,geneticEj2,geneticEj3}, o algoritmos genéticos hibridados  \cite{hybrid1,hybrid2,hybrid3}.
	
	\item Implementación de nuevos operadores: Estos trabajos presentan implementaciones novedosas de diferentes operadores específicos, como operadores de cruce \cite{cross1,cross2}, o de mutación \cite{mutation1,mutation2}. Generalmente, estos operadores son de carácter heurístico, son aplicados a un problema concreto, y poseen un gran rendimiento.
	
	\item Análisis del comportamiento del algoritmo: Este tipo de investigaciones se basan en el análisis de ciertos aspectos teóricos o prácticos de los algoritmos genéticos. Estos estudios examinan, entre otros temas, diversas características acerca del comportamiento del algoritmo, como la convergencia \cite{convergence}; la eficiencia de ciertas fases del algoritmo, como el cruce \cite{study3,effcross1} o la mutación \cite{effmut1,effmut2}; o la influencia de adaptar ciertos parámetros dinámicamente, como la probabilidad de cruce o de mutación \cite{parameter1,parameter2,parameter3}. Gracias a estos análisis se ha logrado un profundo conocimiento de los algoritmos genéticos, dando como resultado el nacimiento de nuevas técnicas, como los algoritmos genéticos adaptativos \cite{adapt1,adapt2}, o los algoritmos genéticos paralelos \cite{parallelGA1,parallelGA2}, los cuales superan las desventajas inherentes a los algoritmos genéticos básicos. 
\end{itemize}

En lo que resta de sección se explicarán en detalle los componentes más importantes de este tipo de meta-heurísticas. Para ello, se mostrarán ejemplos aplicados a los problemas que se han descrito en la Sección \ref{sec:problemas_asignacion_rutas}: El TSP, VRP y sus extensiones.

\subsubsection{Selección de los padres}
\label{sec:selpad}

Como se verá posteriormente, el cruce es el proceso en el que los organismos de una población interactúan entre sí para generar nuevos individuos. Los individuos generados son llamados hijos, y, como en las leyes naturales, cada hijo tiene que tener un padre y una madre. Es por esto por lo que para realizar este proceso hay que seleccionar de toda la población aquellos individuos que van a actuar de padres.

Existen muchos criterios para la selección de los padres, siendo uno de los más conocidos el \textit{torneo binario}. En este criterio se van seleccionando de forma iterativa un par de organismos al azar. Estos dos individuos se enfrentan entre sí, y el mejor de los dos formará parte de la lista de individuos seleccionados para el proceso de cruce. Este proceso se repite hasta que se tenga el número de padres deseado. Otro criterio ampliamente utilizado es el denominado \textit{elitista}, en el que se seleccionan como padres los \textit{n} mejores individuos de la población según su valor de \textit{fitness}.

\subsubsection{Proceso de cruce}
\label{sec:crossover}

Como ya se ha introducido previamente, en el proceso de cruce, o recombinación, dos o más individuos de la población existente interactúan entre sí para formar nuevos individuos. Existen multitud de funciones de cruce, y es un campo en el cual se han realizado un gran número de estudios a lo largo de la historia \cite{study3,effcross1,trabajoscruce3}. El propio autor de esta tesis ha llevado a cabo varios estudios en este campo \cite{enekocruce1,enekocruce2,enekocruce3}. Para problemas de fácil formulación, como el TSP y sus variantes, el BBP, o el NQP, la implementación de diferentes funciones de cruce es un proceso trivial, debido a la sencilla formulación y codificación de las posibles soluciones del problema. Para los problemas más complejos, como los pertenecientes a la familia VRP, el número de funciones de este tipo son menores, ya que las restricciones que han de cumplir cada una de las rutas hacen muy difícil que los hijos resultantes del proceso sean factibles. A continuación se muestran varias funciones de cruce, las cuales han sido frecuentemente referenciadas a lo largo de la historia, y cuya utilización es adecuada para problemas en los que la representación de los individuos se realiza mediante la codificación por permutaciones. A grandes rasgos, la codificación por permutación se encarga de representar las soluciones de un problema como cadenas de números. En el caso del TSP, por ejemplo, esta cadena representa la ruta que el vehículo debe tomar. En esta tesis se dedicará un apartado completo a las codificaciones empleadas para cada problema utilizado a lo largo de la experimentación \ref{sec:problemasexp}.

\begin{itemize}
	\begin{sloppypar}
  \item \textit{Cruce por orden (OX)}: Este cruce fue propuesto por Davis en 1985 \cite{OX}. Este operador construye los hijos eligiendo pequeños sub-conjuntos de uno de los padres, y manteniendo el orden de los nodos del otro progenitor. El primer paso de este proceso consiste en seleccionar dos puntos de corte, idénticos para los dos padres. Los segmentos que queden encuadrados entre los puntos de corte se conservarán en los hijos manteniendo el mismo orden y posición. Tras esto, y empezando por el segundo punto de corte, los nodos restantes se insertan en el mismo orden en que aparecen en el otro progenitor, teniendo en cuenta que los elementos ya insertados son omitidos. Cuando el final de la cadena se ha alcanzado, se continúa por el comienzo de ésta. A continuación se muestra un ejemplo de este proceso utilizando dos individuos compuestos por 8 nodos.
	\[P_1= (1 2 | 3 4 5 | 6 7 8) \rightarrow H_1= (* * | 3 4 5 | * * *) \rightarrow H_1= (8 7 | 3 4 5 | 1 2 6)\]
	\[P_2= (2 4 | 6 8 7 | 5 3 1) \rightarrow H_2= (* * | 6 8 7 | * * *) \rightarrow	H_2= (4 5 | 6 8 7 | 1 2 3)\]
	\end{sloppypar}

 \item \textit{Cruce del punto medio (HX)}: Este cruce es uno de los más utilizados, al igual que uno de los más sencillos de implementar. El primer paso del HX consiste en realizar un punto de corte en el punto central de cada individuo. Los nodos situados antes del punto de corte son copiados en el mismo orden en el hijo. A continuación, los elementos restantes son insertados manteniendo el orden del otro pariente. Un ejemplo de este proceso de cruce puede ser el siguiente:
\[P= (1 2 3 4 | 5 6 7 8)  \rightarrow H1= (1 2 3 4 | * * * *) \rightarrow H1= (1 2 3 4 | 6 8 7 5)\]
\[M= (2 4 6 8 | 7 5 3 1)  \rightarrow H1= (2 4 6 8 | * * * *) \rightarrow H2= (2 4 6 8 | 1 3 5 7)\]

\end{itemize}

Respecto a los problemas relacionados con la familia de problemas VRP, como ya se ha referido al comienzo de esta sección, su alta complejidad y las estrictas restricciones de estos hacen que el diseño e implementación de cruces sea una tarea más compleja. Por esta razón, el número de operadores para estos problemas es inferior a los anteriormente descritos. 

Dos posibles ejemplos de cruce para el CVRP podrían ser el cruce de rutas cortas (SRX), o el cruce de rutas largas (LRX). Estos operadores son un caso particular del cruce tradicional, y podrían equipararse al HX visto anteriormente. Con todo esto, el funcionamiento del SRX podría describirse de la siguiente manera: el primero de los pasos consiste en insertar en el primer hijo la mitad de las rutas (las más cortas) de uno de los padres tal y como aparecen en éste. Después de esto, los nodos no seleccionados se insertan en la misma posición en la que se encuentran en el pariente contrario (teniendo en cuenta la capacidad del vehículo). Asumiendo una posible instancia de 17 nodos (depósito incluido), un ejemplo de este cruce podría ser el siguiente:

\[P= (1,2,3,4,\textbf{0},9,10,11,12,\textbf{0},13,14,15,16,\textbf{0},5,6,7,8)\]
\[M= (1,12,6,3,\textbf{0},2,4,7,11,\textbf{0},5,14,16,9,\textbf{0},8,13,10,15)\]

Los hijos resultantes serían los mostrados a continuación:

\[H1= (1,2,3,4,\textbf{0},9,10,11,12,\textbf{0},6,7,5,14,\textbf{0},16,8,13,15)\]
\[H2= (1,12,6,3,\textbf{0},2,4,7,11,\textbf{0},9,10,13,14,\textbf{0},15,16,5,8)\]

El LRX opera de forma similar al SRX, con la pequeña particularidad de que, en este caso, las rutas escogidas para ser copiadas en los hijos son las más largas de cada pariente.

Estos son varios de los cruces más utilizados para la resolución de problemas basados en la codificación por permutaciones. Existe una ingente cantidad de operadores adicionales, los cuales no serán descritos en esta tesis por no aumentar en exceso su extensión. Para conocer más tipos de cruce para los problemas de tipo VRP, conviene revisar \cite{crucesVRP1} o \cite{crucesVRP2}, donde pueden encontrarse varios ejemplos útiles. Por otro lado, para obtener más información referente a cruces aplicables al TSP, se aconseja la lectura de \cite{TSP2}.

\subsubsection{Proceso de mutación}
\label{sec:mut}

En una primera instancia, el proceso de mutación surgió con el objetivo de huir de óptimos locales. Al igual que sucede con los algoritmos trayectoriales de búsqueda simple, es posible que durante la ejecución de un algoritmo genético la población vaya dirigiéndose hacia regiones del espacio de soluciones que no interesa explorar. Con todo esto, la filosofía principal del proceso de mutación es la de realizar pequeños cambios en los individuos, con el objetivo de ampliar el espectro de exploración de la población. Es un proceso que se realiza después de los cruces, y el cual se ejecuta de forma probabilística con un valor de probabilidad, normalmente, muy bajo.

Existen diferentes formas de mutar, aunque por lo general es un proceso sencillo, y los operadores utilizados no tienen gran complejidad. Esto es así ya que el propósito de este procedimiento es el de realizar pequeños cambios en los individuos. De la misma forma que ocurre con los operadores de cruce, existen muchos operadores implementados para el TSP y los problemas de codificación por permutaciones, y muchos otros para el VRP y sus variantes.

Para los problemas del tipo VRP los operadores de mutación pueden ser de dos tipos distintos: intra-ruta e inter-ruta. Como ya se ha explicado, los primeros son aquellos en los que los cambios se realizan entre diferentes rutas. Por lo general, estos operadores suelen seguir la misma filosofía que los operadores de generación de sucesores que se han visto en la Sección \ref{sec:gensuc} de búsquedas locales, como el Swapping, o el Vertex Insertion \dots De hecho, estos dos son utilizados a menudo como operadores de mutación para algoritmos genéticos aplicados a problemas de la familia VRP.

En cuanto a los operadores intra-ruta, los cambios se realizan dentro de una misma ruta, por lo que los operadores pueden ser utilizados tanto en el VRP como en el TSP. Un operador muy común es el basado en el cambio \cite{mutacion1}, en el que la función de mutación genera varios intercambios de nodos dentro del mismo cromosoma. El número de intercambios viene dado por un atributo del algoritmo que se introduce de antemano, el cual podría llamarse factor de mutación. 

A lo largo de la historia se han formulado multitud de operadores de mutación, algunos de ellos, como dicta la filosofía del proceso de mutación, de formulación y ejecución simples \cite{TSP2}, otros, por otro lado, mucho más sofisticados \cite{mutation1}. 

\subsubsection{Selección de supervivientes}
\label{sec:sup}

Los procesos de cruce y mutación generan una cantidad de organismos que han de ser insertados en la población actual. La población existente en un algoritmo poblacional tiene un tamaño finito, y es por esto por lo que se hace necesario reducir el número de individuos después de los procesos anteriormente mencionados,  desechando aquellos que sean menos interesantes. Por un lado, se pueden destacar los algoritmos genéticos generacionales. En estos algoritmos, durante cada generación se crea una población completa de nuevos individuos, y esta nueva población reemplaza directamente a la predecesora. 

Por otro lado, existen algoritmos en los que los individuos creados durante el proceso de cruce y mutación son mezclados con los provenientes de la generación anterior. En este caso, se hace necesario realizar una criba, con el objetivo de mantener el tamaño fijo de la población. De esto se encarga la función de selección de supervivientes. 

Un ejemplo de este tipo de funciones es la denominada \textit{elitista-aleatoria}. Esta función confecciona la población superviviente en dos pasos: por un lado, una parte de la población estará compuesta por los mejores individuos de toda la población. La otra parte, por el contrario, se seleccionará de forma aleatoria entre los elementos aun no insertados. Por ejemplo, si se tiene una población de 50 cromosomas, y se quiere reducir esta hasta los 30 individuos, una función 50\% elitista - 50\% aleatoria elegirá los 15 mejores cromosomas según su \textit{fitness}, y los 15 restantes los seleccionará de forma aleatoria de entre los 35 sobrantes.

Un aspecto de crucial importancia en los algoritmos genéticos es el mantenimiento de la diversidad dentro de la población. Es por esto por lo que elegir una parte de la población sucesora de forma aleatoria ayuda a mantener cierta diversidad dentro de ella, evitando caer fácilmente en algún óptimo local.

Al igual que con el criterio de selección de los padres, existen multitud de criterios de selección de supervivientes. Estos criterios pueden ser los mismos que los expuesto en la sección de selección de padres (Sección \ref{sec:selpad}), con la particularidad de estar enfocados a la función de supervivencia.

\subsubsection{Criterio de parada}
\label{sec:parada}

Los pasos que se han descrito hasta ahora (selección de reproductores, función de cruce, mutación y selección de supervivientes) se ejecutan de forma iterativa, donde cada iteración es denominada generación. Estas iteraciones se repiten hasta que se cumple el criterio de parada estipulado para el algoritmo. Tras esto, el algoritmo devolverá su mejor individuo a modo de solución final. Pese a que existen diferentes alternativas para establecer el criterio de parada, son dos los principales mecanismos utilizados. El primero de ellos se fundamenta en marcar un número fijo de generaciones a ejecutar. En el momento en el que algoritmo llega a la generación fijada la ejecución finaliza y se devuelve el mejor individuo de la última generación. Por otro lado, el segundo se basa en fijar un número de generaciones sin mejoras. En este caso el algoritmo finalizará su ejecución cuando no se produzcan mejoras en los individuos de una población durante un número fijado de generaciones consecutivas.

\subsection{Algoritmos híbridos}
\label{sec:hibridos}

Los algoritmos híbridos, o meméticos, son aquellos que combinan varias técnicas en una misma con el objetivo de superar las desventajas de ambas, y conseguir unos mejores resultados que los que se obtendrían ejecutando los métodos por separado. Gran parte de las meta-heurísticas híbridas combinan un algoritmo genético con uno trayectorial simple. Por lo tanto, como se puede deducir, las posibilidades son abundantes. Este campo es realmente amplio, y goza de gran popularidad en la literatura. Es preciso reconocer a Pablo Moscato como el ``padre" de este concepto \cite{memetic1}, siendo los trabajos de este investigador muy valorados por la comunidad científica \cite{memetic2}. 

Existe, por ejemplo, una técnica llamada \textit{seeding}, la cual combina un algoritmo genético con uno de búsqueda local. El objetivo de este proceso es el de aplicar la técnica de búsqueda local a la hora de generar la población inicial, con lo que la ejecución del algoritmo poblacional comenzaría con una población con cierta calidad, en lugar de comenzar con una aleatoria. Varios estudios, como el realizado por Lawler et al. en 1985 \cite{TSP1} y Johnson en 1990 \cite{memetico1}, demuestran cómo crear una población con una calidad media para un TSP. Hay que tener en cuenta que el proceso de \textit{seeding} es un proceso delicado, ya que la población tiene que tener cierta diversidad para evitar que el algoritmo converja hacia un óptimo local. El propio autor de esta tesis ha realizado varios estudios en relación a esta problemática \cite{osabainic1,osaba1}. Un algoritmo muy utilizado para estos casos es el \textit{2-opt}, ya mencionado en la Sección \ref{sec:gensuc}.

El enfoque más utilizado en este tipo de algoritmos es el de integrar un algoritmo trayectorial simple dentro del proceso evolutivo del algoritmo genético, bien sea sustituyendo alguno de los operadores como el de mutación o cruce, o como un paso más del proceso. Existen multitud de estos tipos de algoritmos híbridos, combinando de distinta forma una gran variedad de técnicas.

En \cite{memetico2}, por ejemplo, se muestra un algoritmo híbrido que combina la búsqueda tabú con un algoritmo genético para resolver el TSP. Para este mismo problema, en \cite{memetic3} y en \cite{memetic4} se pueden ver dos algoritmos genéticos combinados con una búsqueda local, la cual se aplica a todos los hijos resultantes del proceso de cruce.

\begin{sloppypar}
En referencia a los problemas de tipo VRP, en el trabajo \cite{VRPB7} se presenta un algoritmo genético con búsqueda local para VRPB, el cual ejecuta la búsqueda local a todos los hijos generados en el proceso de reproducción. De esta misma forma, en \cite{memetic5}, para el problema PVRPTW, Nguyen et al. crean un proceso llamado ``educación", que consiste en la aplicación de una búsqueda local de la misma forma que los anteriormente mencionados. Finalmente, en el trabajo de Nagata y Braysy \cite{memetic6}, se utiliza un operador de búsqueda local para reparar y optimizar aquellos hijos no factibles resultantes del proceso de cruce. 
\end{sloppypar}

\subsection{Optimización por enjambre de partículas inteligentes}
\label{sec:PSO}

El método de optimización de enjambres de partículas, o \textit{Particle Swarm Optimization} (PSO), es una de las meta-heurísticas más populares para la resolución de problemas de optimización. Fue propuesto por James Kennedy y Russell C. Eberhart en 1995 \cite{PSO1,PSO2}, con la intención de que emulase el comportamiento que toman los conjuntos de animales, como los bancos de peces o bandadas de pájaros, o enjambres de insectos, como las abejas. En la mayoría de los casos, son estas últimas las que se ponen como enjambre ejemplo para explicar el método. 

En lo que se refiere a la meta-heurística en cuestión, se trata de una técnica poblacional, la cual cuenta con una población (enjambre) de individuos (partículas) que son mejorados de forma iterativa con el propósito de encontrar el óptimo global. Cada partícula guarda información acerca de la mejor posición obtenida por ella y por cualquier individuo del entorno. Esta información influirá directamente en su comportamiento. De esta forma, cada individuo lleva asociados los siguientes atributos: posición actual, velocidad actual, mejor posición encontrada por ella y mejor posición obtenida por cualquier partícula del entorno.

En el funcionamiento de un PSO pueden distinguirse tres pasos diferentes. El primero de ellos es la inicialización de las soluciones. Inicialmente, las partículas comienzan en una posición aleatoria dentro del espacio de soluciones y se les asigna un valor de velocidad aleatorio. La solución inicial de cada individuo pasa a ser la mejor solución, ya que hasta el momento es la única. De forma análoga se hace lo mismo con la mejor solución del enjambre.

Después de esto, y en cada iteración, la velocidad de cada insecto se recalcula en función de la mejor solución de toda la población, y se actualiza la posición de cada partícula en base a ella. Finalmente, la actualización de las mejores posiciones se considera el tercer paso de este algoritmo. Después de cada iteración, si la solución obtenida por cada partícula supera a la mejor conocida hasta el momento, se actualiza el parámetro de mejor solución encontrada. Estos pasos se repiten hasta que el criterio de parada del algoritmo se cumpla.

Pese a que en un primer momento el PSO no fue diseñado para ser aplicado a problemas de optimización discreta, o problemas de optimización combinatoria, en los últimos años han sido multitud las implementaciones de esta técnica para problemas de este tipo. Varios artículos de relevancia científica son el propuesto en 2003 por Wang et al. en \cite{PSOTSP1}, el realizado por Clerc en 2004 \cite{PSOTSP2} o el publicado en 2007 por Shi et al. \cite{PSOTSP3}. 

También se han publicado innumerables trabajos en relación a los problemas de la familia TSP y VRP. Ejemplos de estos estudios son el presentado por Marinakis et al. en 2013 \cite{PSOVRP1}, en el que propone un versión discreta del PSO para resolver el CVRP con demandas estocásticas, o el publicado por Belmecheri et al. el mismo año \cite{PSOVRP2}, en el que se presenta un PSO aplicado al CVRP con flota heterogénea, retorno de mercancías y ventanas temporales. 

\subsection{Optimización por colonias de abejas artificiales}
\label{sec:ABC}

La optimización por colonias de abejas artificiales, o \textit{artificial bee colony} (ABC), es un algoritmo propuesto en 2005 por Karaboga para problemas numéricos multi-modales y multi-dimensionales \cite{ABC2005,ABC2006}. El ABC es un algoritmo basado en enjambres que emula el comportamiento de recolección de polen de las abejas melíferas. En esta técnica, la población consiste en una colonia en la que cohabitan tres tipos diferentes de abejas, las cuales reciben los nombres de trabajadoras, espectadoras y exploradoras. Cada tipo de abeja tiene un comportamiento diferente. 

En el ABC, cada posible solución al problema de optimización es representada por una fuente de comida, y el \textit{fitness} de cada solución es representado mediante la cantidad de néctar de esa fuente. Cada abeja trabajadora tiene una fuente de comida asignada, y en cada generación, este tipo de abejas busca una nueva fuente en su vecindario. Si la nueva fuente encontrada tiene una cantidad de néctar superior a la anterior, la abeja se mueve hacia ella. Tras esto, cada abeja espectadora selecciona al azar una fuente de comida entre todas las existentes, y realiza un movimiento idéntico al que realizan las abejas del tipo anterior.

Una fuente de comida puede ser abandonada por una abeja trabajadora en caso de estar un número predeterminado de iteraciones sin experimentar mejoras. En este caso, la abeja pasará a ser exploradora, y se desplazará a una posición aleatoria sin importar si es peor o mejor que su actual posición. De esta forma, en el ABC, la exploración del espacio de soluciones es llevada a cabo por las abejas exploradoras, mientras que las trabajadoras y las espectadoras se encargan de la explotación. 

En \cite{ABCSurvey} puede encontrarse un estudio detallado acerca del ABC. En ese estudio se destaca cómo este tipo de algoritmo ha sido aplicado a problemas de asignación de rutas a vehículos en contadas ocasiones hasta el momento, ya que el ABC no fue diseñado para ello en un primer momento. Aun así, esta tendencia está cambiando en los últimos años \cite{ABCrouting1,ABCrouting2,ABCrouting3}.

\subsection{Optimización por colonias de hormigas}
\label{sec:Ant}

La técnica de optimización por colonias de hormigas (ACO) es una de las meta-heurísticas más empleadas en la literatura a lo largo de la historia. Este algoritmo fue propuesto por Marco Dorigo en 1992 a modo de tesis doctoral, y desde entonces han sido infinidad los trabajos realizados sobre esta técnica \cite{ACO1,ACO2}. También han sido multitud las implementaciones del ACO propuestas para abordar diferentes problemas de asignación de rutas a vehículos \cite{ant2,ACO4,ACO5}.

La idea original del ACO se inspira en el comportamiento de las colonias de hormigas para el rastreo y explotación de fuentes de alimentos. En estos algoritmos la población está compuesta por diferentes agentes, o individuos, llamados hormigas, los cuales trabajan en paralelo con el fin de encontrar buenas soluciones a los problemas con los que se trabaje. Siendo más minuciosos, por cada iteración cada hormiga construye una solución, la cual se genera teniendo en cuenta experiencias pasadas, tanto propias como grupales (las soluciones buenas conseguidas en el pasado influirán positivamente, y viceversa). Siguiendo esta filosofía, cuando una hormiga busta una solución, esta propaga cierta cantidad de feromonas. Estas feromonas atraerán al resto de hormigas a adoptar los mismos movimientos llevados a cabo por la hormiga que las depositó. Como puede deducirse, las soluciones con mayor calidad y las más interesantes provocarán que la hormiga deposite una cantidad de feromonas mayor, con el objetivo que el resto de las compañeras tengan mayor facilidad para seguirlas.

Además de la versión clásica, existen diferentes extensiones del ACO con reconocimiento en la comunidad científica, como pueden ser el sistema de hormigas MAX-MIN \cite{MMAS}, el basado en ranking \cite{RBAS}, o la colonia de hormigas ortogonal continua \cite{COAC}.

\subsection{Otros tipos de algoritmos poblacionales}
\label{sec:others}

Como ya se ha mencionado con anterioridad, en la literatura existen multitud de algoritmos poblacionales para la resolución de todo tipo de problemas de optimización. En esta sección se han descrito varios de los más utilizados en la literatura desde su propuesta. Además de estos existen muchos más, los cuales también han demostrado ser técnicas prometedoras. Un ejemplo de estos es el algoritmo inspirado en murciélagos. Esta meta-heurística, propuesta por Yang en 2010 \cite{bat}, está basada en el comportamiento ecolocalizador de los micromurciélagos, los cuales encuentran a sus presas discriminando ciertos tipos de insectos, incluso en la más completa oscuridad. Este mismo autor propuso en el año 2009 el algoritmo basado en luciérnagas \cite{fire}. Esta técnica goza de gran popularidad hoy en día, y está fundada en el comportamiento de alumbrado de estos insectos, mecanismo con el que atraen a otras luciérnagas. Para finalizar con esta sección, merece la pena mencionar el algoritmo basado en cucos \cite{cuckoo1}. Esta meta-heurística se fundamenta en el comportamiento parasitario de algunas especies de cucos en combinación con la conducta de vuelo de algunos pájaros y moscas de la fruta.

\section{Algoritmos multi-poblacionales}
\label{sec:multi}
Como ya se ha mencionado en la sección anterior, los algoritmos poblacionales trabajan con un conjunto de individuos los cuales suelen ser soluciones completas del problema que se esté tratando. Una evolución de este tipo de meta-heurísticas son las técnicas que se introducirán en esta sección: los algoritmos multi-poblaciones. Es interesante mencionar que la meta-heurística propuesta en este trabajo doctoral forma parte de este conjunto de técnicas.

La base de los algoritmos multi-poblacionales es la de trabajar con múltiples poblaciones de individuos, las cuales, generalmente, trabajan de forma individual y se relacionan mediante estrategias de comunicación. Estas estrategias pueden ser de lo más variadas, como bien se podrá ver a lo largo de esta sección. Además, este último factor es algo que nunca puede descuidarse a la hora de diseñar una meta-heurística de este tipo, ya que estas tácticas de comunicación son un factor crucial a la hora de obtener mejores o peores resultados. Asimismo, el hecho de tener múltiples poblaciones hace que los algoritmos multi-poblacionales sean unas técnicas prometedoras para realizar una exploración exhaustiva del espacio de soluciones. 

En este apartado se describirán varios métodos de este estilo, los cuales han sido citados y utilizados en la literatura en numerosas ocasiones. Finalmente, al igual que los algoritmos poblacionales, estas meta-heurísticas son un tema de actualidad en la comunidad científica, produciendo multitud de artículos y libros anualmente y descubriendo nuevas y eficientes técnicas regularmente.

\subsection{Algoritmos genéticos paralelos}
\label{sec:PGA}

Como ya se ha detallado en el apartado \ref{sec:genetico} de este trabajo, desde su propuesta en la década de los 70, los GA se han convertido en una de la técnicas más utilizadas para la resolución de una amplia variedad de problemas. Pese a esta gran aceptación por parte de la comunidad científica, y al igual que todas las meta-heurísticas existentes, el GA tiene ciertas desventajas, siendo las más conocidas su rápida convergencia a un óptimo local, o la dificultad que supone equilibrar la capacidad de explotación con la de exploración.

Con la intención de sobreponerse a estos inconvenientes se propusieron los algoritmos genéticos paralelos (PGA). Estas meta-heurísticas son particularmente fáciles de implementar, y generalmente ofrecen mejoras sustanciales en los resultados. Realizando una revisión de la literatura actual puede verse cómo existen diferentes maneras de paralelizar un GA. Por lo general se utiliza una clasificación la cual divide los PGA en tres categorías principales: PGAs de grano fino \cite{fine}, modelos panmícticos o PGAs uni-poblacionales maestro-esclavo, \cite{panmitic} y los PGAs basados en islas \cite{Island}.

Los PGAs de grano fino consisten en una población estructurada espacialmente, y son apropiados para sistemas con múltiples computadoras trabajando en paralelo. Los procesos de selección y cruce se limitan a pequeños subconjuntos de la población, denominados barrios, que se solapan unos con otros, lo cual permite la interacción entre ellos. El segundo de los tipos es el modelo panmíctico. En este tipo de técnicas existe tan solo una población panmíctica (al igual que en los GAs convencionales), pero la evaluación del \textit{fitness} de los individuos se divide entre varios procesadores. Debido a que en este PGA concreto el proceso de selección y cruce se realiza considerando la población completa, también pueden ser denominados como algoritmos genéticos paralelos globales.

Por último, de entre los tipos de PGAs existentes, los basados en islas son los que han gozado de mayor popularidad en la comunidad científica. Estas técnicas consisten en múltiples poblaciones cuya evolución se realiza por separado la mayoría del tiempo, e intercambian individuos entre sí de manera ocasional. Estos algoritmos también pueden ser nombrados como \textit{multi-hogar}, \textit{distribuidos} o de \textit{grano grueso}. Muchos estudios comparten la opinión de que esta forma de implementar los PGAs es la más adecuada, pese a que las estrategias y frecuencia de comunicación entre las distintas subpoblaciones es una cuestión compleja. 

En la literatura pueden encontrarse un sinfín de trabajos describiendo innumerables aspectos y detalles en la implementación de los PGAs basados en islas \cite{Island,Island2,Island3}. Por otro lado, en el artículo publicado en \cite{DGAstudy} se presenta un completo y extenso estudio acerca de los PGAs. En este trabajo pueden encontrarse varios enfoques de migración y comunicación entre diferentes poblaciones. El propio autor de esta tesis ha realizado investigaciones centradas en las estrategias de migración de los PGAs \cite{enekomigracion}.

Un ejemplo de PGAs aplicados a problemas de asignación de rutas a vehículos puede encontrarse en \cite{PGAVRP1}, donde un PGA es utilizado para tratar un problema con flota heterogénea. Por otro lado, en \cite{PGAVRP2} se propone otra implementación de un algoritmo paralelo para abordar un VRPTW. En último lugar, en el artículo \cite{PGATSP} se presenta un estudio sobre enfoques paralelos y distribuidos aplicados al TSP convencional.

\subsection{Algoritmo Imperialista}
\label{sec:imperialista}

Esta meta-heurística multi-poblacional fue propuesta por Atasphaz-Gargari y Lucas en el año 2007 \cite{imperialist}. En este método, los integrantes de la población son denominados países con el objetivo de ser más fiel a la metáfora del imperialismo.

En una primera fase, el algoritmo genera su población inicial de países. De todos ellos, los mejores son seleccionados como países imperialistas, mientras que los restantes son designados como simples colonias. Una vez seleccionados los países imperialistas y basándose en su calidad, o en su fuerza (lo que en otras técnicas se ha denominado \textit{fitness}), las colonias se reparten entre ellos, formando diferentes imperios. Cada imperio tendrá su fuerza imperial, la cual se calcula en función de la calidad de su país imperialista sumada a la de sus colonias.

Una vez generada la población inicial, y creados los diferentes imperios, comienza el proceso iterativo. En un primer paso, todas las colonias de cada imperio realizan un pequeño movimiento en base a su país imperialista. Este movimiento es llamado \textit{movimiento de acercamiento} y es un proceso en el que la colonia se acerca en el espacio de soluciones a su país imperialista. Después de este proceso de movimiento se da paso a la competición imperialista. En este paso, el imperio más débil de todos los existentes pierde a su colonia más débil, la cual es conquistada por cualquiera de los otros imperios restantes. La conquista se realiza de forma probabilística con respecto a la fuerza de los imperios, es decir, que los imperios más potentes tienen mayor capacidad de conquista. Este proceso hace que los imperios más débiles queden destruidos. Cuando un imperio se debilita hasta cierto punto se destruye, y todas sus colonias pasan a formar parte de los otros grupos. Continuando con este proceso, llegará un momento en que solo exista un imperio y todas las colonias estén bajo su control. Alcanzado este momento, el algoritmo convergerá hacia un estado llamado ``mundo ideal", en el que todos los países serán igual de poderosos y se encontrarán en la misma posición.

Esta meta-heurística ha sido aplicada a varios campos de optimización en los últimos años. En el año 2011, por ejemplo, Nemati, Shamsuddin y Kamarposhti demostraron cómo esta técnica obtiene unos resultados excelentes al aplicarlo al TSP, comparándolo, además, con otras técnicas aplicadas al mismo problema \cite{imperialistTSP1}. Yousefikhoshbakht y Sedighpour ratificaron que esta meta-heurística obtiene resultados prometedores para el TSP en su trabajo publicado en 2013 \cite{imperialistTSP2}. Por otro lado, Wang et al. implementaron en el año 2011 una versión de esta técnica para la resolución del VRPTW \cite{imperialistVRPTW}. Otras aplicaciones de este algoritmo se pueden ver en \cite{imperialist3} y \cite{imperialist4}.

\subsection{Optimización por múltiples colonias de abejas artificiales}
\label{sec:multiABC}

La primera versión del algoritmo de optimización por múltiples colonias de abejas artificiales (PABC) fue propuesto en el año 2009 en \cite{ABC2009}, con el designio de realizar una búsqueda más exhaustiva que las realizadas por los ABC simples. En lo relativo a la manera de trabajar de los PABC, la filosofía principal descansa en la ejecución en paralelo de múltiples instancias del ABC convencional (descrito en la Sección \ref{sec:ABC}), con la intención de aumentar la capacidad de exploración de la técnica. Otro factor crucial en los PABC, el cual ayuda a la obtención de buenos resultados, es la migración de abejas o el intercambio de información entre diferentes colonias que forman el sistema. 

A continuación se presentarán algunas de las implementaciones del PABC más interesantes. En el trabajo presentado en el año 2010 en \cite{PABC1} se presentó un enfoque en el que las variables del problema se dividen en diferentes instancias del mismo algoritmo. Todas las instancias trabajan de manera independiente, y la solución completa se obtiene mediante la colección de las mejores soluciones de cada una de ellas. 

Por otro lado, en la investigación presentada en \cite{PABC2} las abejas son divididas en diferentes subgrupos, los cuales exploran el mismo espacio de soluciones de la misma forma. En cada iteración, dos colonias son seleccionadas de manera aleatoria para compartir sus mejores soluciones entre ellas. Esta transferencia de información se realiza mediante el reemplazo de las nuevas soluciones por la peor solución de la colonia destino. 

\begin{sloppypar}
Finalmente, en el artículo publicado por Parpinelli et al. en el año 2010 \cite{PABC3} se presenta una versión mejorada del PABC, en la que la capacidad de búsqueda de la técnica se incrementa mediante búsquedas locales. En este trabajo se presentan tres tipos de modelos paralelos: el típico maestro-esclavo, el multi-colmena con migraciones, y el jerárquico.
\end{sloppypar}

\subsection{Optimización por múltiples enjambres de partículas inteligentes}
\label{sec:multiPSO}

Con el objetivo de superar los inconvenientes del PSO convencional (rápida convergencia hacia un óptimo local \cite{PSODrawback1,PSODrawback4}, dificultad para mantener el equilibrio entre explotación y exploración \cite{PSODrawback2}, o la llamada ``maldición de la dimensionalidad" \cite{PSODrawback3} se ideó el primer algoritmo de optimización por múltiples enjambres de partículas inteligentes (PPSO). Al igual que en el caso de los PABCs, en la literatura pueden encontrarse múltiples versiones para este tipo de meta-heurísticas. El fundamento básico de esta meta-heurística es el de ejecutar en paralelo múltiples versiones del PSO convencional (visto en la Sección \ref{sec:PSO}), ya sea con las mismas o diferentes parametrizaciones, y con el añadido de implementar una red de migraciones de partículas, bien básica o compleja.

En el trabajo presentado en \cite{PSODrawback3}, por ejemplo, se presenta un PPSO basado en enjambres cooperativos. La técnica presentada en dicho trabajo divide el espacio de soluciones completo es sub-espacios de dimensiones menores. Para realizar esta partición se hace uso de la conocida técnica $k-medias$ y de esquemas regulares de división. Después de esto, los enjambres se encargan de optimizar los diferentes componentes del espacio de soluciones de manera cooperativa.  

Otro ejemplo de PPSO se encuentra en el artículo publicado en \cite{PPSO2}, donde se presentan dos enfoques diferentes (uno competitivo y otro cooperativo) basados en el modelo maestro-esclavo. En la versión competitiva, el enjambre maestro realza sus propias partículas utilizando los mejores ejemplares de sus enjambres sometidos. Por otro lado, en el enfoque cooperativo el enjambre maestro actualiza sus partículas ayudándose de sus súbditos mediante esquemas de colaboración. 

Finalmente, en el ejemplo propuesto en \cite{PPSO3} la población completa se divide en diferentes enjambres, los cuales se comunican entre ellos haciendo uso de tres diferentes estrategias de comunicación. En la primera estrategia múltiples copias de la mejor partícula de cada subpoblación son mutadas, para después migrar y reemplazar a las peores partículas del resto de comunidades. En la segunda táctica de comunicación, una copia de la mejor partícula es enviada a los enjambres vecinos de cada subpoblación. La última estrategia es una hibridación de las dos primeras.

\subsection{Otras técnicas multi-poblacionales}
\label{sec:otherMulti}

De la misma forma que se ha indicado en el caso de los algoritmos poblacionales, en la literatura existe una cantidad reseñable de técnicas multi-poblacionales. En este trabajo se han descrito algunas de ellas, las cuales cuentan con el apoyo de la comunidad científica, y su utilización ha sido extendida a varios campos dentro de la optimización. Aun así, pueden encontrarse en la literatura meta-heurísticas adicionales, como la propuesta por el autor de esta tesis doctoral \cite{osaba2013multi}.

Otro ejemplo de este tipo es el algoritmo de optimización buscador (SOA). Esta meta-heurística, propuesta por Dai et al. en 2006 \cite{SOA2006}, modela el comportamiento del ser humano a la hora de realizar una búsqueda, en la que hace uso de su memoria, experiencia, razonamiento incierto y comunicación con otros seres humanos. En esta técnica los individuos son llamados buscadores, o agentes, y son divididos en $k$ diferentes subpoblaciones. Todos los buscadores de una subpoblación forman juntos una comunidad, la cual posee su propio comportamiento. Los agentes se mueven a través del espacio de soluciones siguiendo una dirección de búsqueda, fundamentada en la posición actual y las posiciones históricas, tanto suyas como las de sus compañeros de comunidad. Para profundizar en los fundamentos de esta meta-heurística se recomienda la lectura de trabajos como el arriba mencionado \cite{SOA2006}. En la literatura pueden encontrarse múltiples estudios enfocados en esta técnica \cite{SOA1,SOA2,SOA3}. Pese a esto, el SOA nunca ha sido aplicado a problemas de enrutado de vehículos.


\begin{savequote}[40mm]
En los momentos de crisis, sólo la imaginación es más importante que el conocimiento.
\qauthor{Albert Einstein}
\end{savequote}

\chapter{Explicación del modelo}
\label{cha:Modelo}

\graphicspath{ {3_golden_ball/figures/} }


\lettrine{U}{}{na} vez introducidos los conceptos de problemas de asignación de rutas a vehículos y técnicas para su resolución, en este capítulo se procederá a describir el modelo que se presenta en esta tesis doctoral para conseguir validar la hipótesis previamente planteada. A grandes rasgos se puede decir que la aportación principal de esta tesis consiste en un nuevo método multi-poblacional basado en conceptos futbolísticos enfocado a la resolución de problemas de optimización combinatoria. Debido a su interés científico y social (descrito en capítulos anteriores), en este trabajo doctoral se hará especial hincapié en la aplicación de esta novedosa técnica a problemas de asignación de rutas a vehículos.

Por lo tanto, en este capítulo se dará protagonismo al modelo presentado en esta tesis. En un primer momento se plasmarán en el apartado \ref{sec:Ref} unas reflexiones acerca de diferentes aspectos sobre algoritmia y optimización combinatoria, las cuales han encaminado el diseño y la elaboración de la meta-heurística propuesta hacia su versión final. Después de esto, en el Apartado \ref{sec:GB} se describirán los aspectos fundamentales de la técnica con la mayor minuciosidad posible. Finalmente, a fin de encuadrar el algoritmo presentado en la literatura actual, en la sección \ref{sec:aportaciones} se introducirán las principales aportaciones y contribuciones del modelo propuesto, y se mencionarán las originalidades y similitudes con las principales técnicas existentes. 

\section{Reflexiones llevadas a cabo para el diseño de la meta-heurística propuesta}	
\label{sec:Ref}

En esta sección se van a argumentar varias de las decisiones tomadas a la hora de diseñar la técnica propuesta en esta tesis. Para ello, en un primer momento, en el apartado \ref{sec:CrossVSmut}, se va a explicar cómo se ha llegado a la conclusión de priorizar la evolución individual de los elementos de la población, dejando la evolución cooperativa en un segundo plano. Después de esto, en la sección \ref{sec:islas} se argumentará la razón de haber diseñado una técnica multi-poblacional en lugar de emplear una sola población. Finalmente, en un apartado breve como es el \ref{sec:futbol} se mencionará el porqué de haber utilizado una metáfora como la del fútbol para desarrollar la meta-heurística presentada.

\subsection{Optimización individual vs. optimización cooperativa}
\label{sec:CrossVSmut}

Como se ha visto en secciones anteriores, gran parte de los algoritmos poblacionales, tanto los uni-poblacionales como los multi-poblacionales, intentan encontrar un equilibrio aceptable entre la explotación y la exploración del espacio de soluciones mediante la utilización de dos tipos de funciones. Las primeras son las operaciones cooperativas, en las cuales dos o más individuos de una o más poblaciones cooperan de alguna manera para generar nuevos individuos, o para modificar los ya existentes. Ejemplos de estos procedimientos pueden ser los conocidos procesos de cruce de los algoritmos genéticos, o los movimientos dirigidos de los algoritmos imperialistas. En el lado opuesto se encuentran las operaciones individuales, mediante las cuales los individuos evolucionan, o varían, de forma autónoma, sin verse influidos por otros factores poblacionales. Dentro de este tipo de procedimientos podrían destacarse los procesos de mutación de los algoritmos genéticos, o los movimientos realizados por las abejas trabajadoras dentro de la optimización por colonias de abejas artificiales.

Como bien se ha podido ver en la Sección \ref{sec:tecnicas_de_resolucion}, la mayoría de los algoritmos poblacionales y multi-poblacionales priorizan los procesos cooperativos, dejando en un segundo plano los individuales. Esto puede apreciarse, por ejemplo, en algoritmos como el genético, el imperialista, o en las técnicas de optimización por enjambre de partículas inteligentes. Siguiendo esta filosofía, estas meta-heurísticas han obtenido buenos resultados aplicados a todo tipo de problemas de optimización, lo que hace incuestionable el acierto que supone este enfoque.

Pese a esto, el método propuesto en este trabajo invierte la filosofía convencional de muchos algoritmos poblacionales, dando prioridad absoluta a la optimización individual, y relegando a un segundo plano los movimientos cooperativos entre individuos. Este atípico enfoque nace después de validar una hipótesis planteada por el autor de esta tesis junto a varios investigadores. Esta hipótesis concierne a uno de los principales procesos de los algoritmos genéticos, el proceso de cruce, y podría enunciarse de la siguiente manera:

\

\textit{"`El operador de cruce de un algoritmo genético convencional es menos eficiente para el proceso de búsqueda y la capacidad de optimización que el operador de mutación, cuando el algoritmo genético es aplicado a problemas de optimización combinatoria basados en codificación por permutaciones"'}

\

Esta hipótesis ha sido validada durante el desarrollo de esta tesis en varios trabajos científicos, publicados tanto en congresos internacionales, como en revistas científicas \cite{enekocruce1,enekocruce2,enekocruce3,enekocruce4}. En estos trabajos se demuestra cómo algoritmos evolutivos, tan solo basados en optimización individual, pueden rendir al mismo nivel o incluso a un nivel superior que algoritmos genéticos básicos, los cuales priorizan las funciones cooperativas. Para verificar la hipótesis arriba descrita se han realizado numerosas pruebas con varias funciones de cruce y de mutación, aplicadas a varios problemas de optimización combinatoria, cada uno con dispares características. Estos problemas son el TSP, ATSP, CVRP, VRPB, NQP y BPP.

Todas las funciones de cruce y mutación utilizadas en estas investigaciones son funciones ciegas, las cuales no emplean información específica del problema. De esta manera, estas funciones se encargan de generar soluciones factibles, cumpliendo en todo momento las restricciones impuestas por los problemas que se estén tratando. Varias de las funciones empleadas son las anteriormente descritas OX y HX. Además de esto, todas las técnicas implementadas para validar la hipótesis propuesta han sido parametrizadas de la misma manera, con el objetivo de que las conclusiones sean rigurosas. Todos estos métodos forman parte de un conjunto de buenas prácticas, las cuales se detallarán en este mismo trabajo, en la Sección \ref{sec:practicas}

Como comentario final acerca de estos estudios llevados a cabo, los autores de los mismos clarifican que esta hipótesis es planteada sobre problemas de optimización combinatoria cuya representación se base en la codificación por permutaciones. Como ya se ha explicado en la introducción de esta tesis, estos problemas, concretamente, son el foco de aplicación de la técnica presentada en esta tesis. Además de esto, los autores son conscientes de que los problemas de optimización combinatoria existentes en la literatura son innumerables. Del mismo modo, también son conscientes de que existen infinidad de funciones ciegas de cruce. Por estas razones, puede ser pretencioso generalizar las conclusiones vertidas en dichos estudios a todos los problemas de optimización combinatoria. Pese a que en las pruebas realizadas se han utilizado problemas de diferentes índole, y funciones de cruce con un reconocimiento contrastado por la comunidad científica, es propio concluir que las deducciones de los citados estudios son objetivas y rigurosas, pero tan solo para las experimentaciones llevadas a cabo.

\subsection{Algoritmos basados en islas vs. uni-poblacionales}
\label{sec:islas}

Como ya se ha visto en capítulos previos, la técnica presentada en este trabajo podría encuadrarse en los algoritmos denominados ``multi-poblacionales", debido a que, como se verá en el Apartado \ref{sec:GB}, la población generada al comienzo de la ejecución es repartida en igual número entre un conjunto prefijado de subpoblaciones.

A lo largo del desarrollo de esta tesis se ha realizado una revisión de la literatura relacionada, pudiendo destacar artículos como \cite{DGAstudy} o \cite{parallelGA1}. Además de esto, se han desarrollado varias investigaciones sobre algoritmos uni-poblacionales y multi-poblaciones, las cuales han servido para decidir la naturaleza de la técnica presentada. En relación a esto último, aparte de las propias investigaciones realizadas con la meta-heurística propuesta, varios trabajos adicionales han contribuido en este aspecto \cite{osaba2013multi,osaba2014adaptive}.

Llevada a cabo la investigación pertinente, las principales razones que han conducido el desarrollo de la técnica presentada al campo de las meta-heurísticas multi-poblaciones son las siguientes:

\begin{itemize}
	\item Un enfoque apropiado para este tipo de técnicas es el de hacer que cada población tenga características y funciones diferentes. Esto hace que a lo largo de la ejecución los individuos sean tratados con funciones de naturaleza distinta y exploren el espacio de soluciones de un modo diferente. Este hecho aumenta la capacidad de exploración y explotación de la técnica.
	
	\item Gracias a las migraciones entre poblaciones es posible actuar con mayor eficacia sobre los individuos atrapados en óptimos locales, obligándolos a migrar en el momento de su detección. Asimismo, teniendo en cuenta lo explicado en el punto anterior, existe una mayor probabilidad de que los individuos atrapados en óptimos locales puedan escapar de ellos.	
\end{itemize}

Sin olvidar las principales desventajas de implementar un algoritmo de estas características:

\begin{itemize}
	\item El reducido tamaño de las subpoblaciones puede hacer que los individuos caigan rápidamente en óptimos locales, debido a la escasez de compañeros con los que cooperar.
	
	\item Las migraciones de individuos entre distintas poblaciones hace que la complejidad de la técnica aumente.
	
	\item El hecho de tener que diseñar diferentes funciones para cada subpoblación puede suponer un aumento en la complejidad de la implementación de la técnica en caso de trabajar con problemas complejos, debido a la dificultad de encontrar diferentes funciones que satisfagan las restricciones de los mismos. 
\end{itemize}

La técnica propuesta en este trabajo no se ve afectada por el primero de los inconvenientes, ya que, como se ha explicado en el apartado previo, prioriza la optimización individual, dejando relegada a un segundo plano la cooperativa. Por otro lado, se ha conseguido encontrar una solución al segundo de los inconvenientes, prefijando de antemano un sistema de migración cuyo funcionamiento recompensa a las subpoblaciones que muestran mejor rendimiento. Esta estrategia se describirá en detalle en la Sección \ref{sec:GB}. Finalmente, la tercera de las desventajas puede suponer un problema real a la hora de implementar cualquier técnica multi-poblacional. Pese a esto, en este trabajo se ha decidido dar mayor peso a las ventajas propias de implementar múltiples poblaciones, aceptando el sacrificio que supone el último de los inconvenientes listados.

\subsection{La metáfora del fútbol}
\label{sec:futbol}

El fútbol es el deporte más popular a nivel mundial. Una prueba de esto fue el estudio llevado a cabo por la FIFA (el órgano que gobierna las federaciones futbolísticas) en el año 2006, donde concluyó que existen, aproximadamente, 270 millones de personas alrededor del mundo envueltas activamente en el mundo del fútbol, incluyendo jugadores, árbitros, y mánagers. De esta vasta cantidad, 265 millones de personas practican el deporte regularmente a modo profesional o amateur, considerando hombres y mujeres de todas las edades. Además de ser el deporte más practicado, también es considerado el deporte más visto y el que cuenta con un mayor número de seguidores en todo el planeta. Como ejemplo, en España, la final del mundial 2010 fue seguida por más de 16 millones de personas, obteniendo un share televisivo del 91\%. En relación a este evento deportivo, Jerome Valcke, secretario general de la FIFA afirmó que la audiencia global acumulada de dicha competición excedió la cantidad de 2600 millones de televidentes. Este seguimiento masivo ha hecho de este deporte un auténtico negocio, el cual genera multitud de trabajos, ya sean de manera directa o indirecta, moviendo así una ingente cantidad de dinero en todo el mundo. Centrándose únicamente en los clubes, en el año 2010 el Manchester United generó unos ingresos que superaban los 251 millones de euros, mientras que otros equipos, como la Juventus, AC Milán o Real Madrid consiguieron un beneficio total conjunto superior a los 450 millones de dólares.

Todos los datos introducidos en este apartado dejan patente que el fútbol es un deporte que suscita un gran interés en la población mundial. Esta es una de las razones que hace atractivo el combinar este mundo con el de la optimización combinatoria, creando una meta-heurística que puede resultar atractiva para la comunidad científica, no solo por su efectividad, si no por su naturaleza y filosofía. Además de esto, el juego del fútbol es un deporte ampliamente conocido por una gran parte de la población mundial, considerándose un deporte de fácil acceso y multicultural. Teniendo esto en cuenta, una meta-heurística basada en estos conceptos puede resultar ser de comprensión sencilla, ya sea para investigadores experimentados o investigadores noveles. Añadido a esto, y debido al gran interés que despierta este deporte entre la juventud mundial, una técnica de esta índole puede ser un aliciente adicional para introducir a jóvenes talentos al mundo de la optimización combinatoria, problemas de asignación de rutas a vehículos y meta-heurísticas para su resolución.

Por último, este deporte es fácilmente aplicable al mundo de las meta-heurísticas, ya que tiene procesos de mejora individual y de cooperación entre jugadores, algo que sucede en los entrenamientos, y procesos competitivos, como son los partidos y los torneos ligueros. Por otro lado, la cooperación entre equipos también ocurre mediante la transferencia de jugadores entre ellos. En el siguiente apartado se detallará cómo todos estos conceptos han sido adoptados por el modelo propuesto en esta tesis doctoral.	

\section{Descripción del modelo propuesto}	
\label{sec:GB}

En esta sección se describirá en profundidad el modelo propuesto en esta tesis doctoral. Como ya se ha mencionado, en este trabajo se presenta una meta-heurística multi-poblacional, la cual se basa en diversos conceptos futbolísticos para guiar su proceso de búsqueda. El nombre designado para esta meta-heurística es Golden Ball (GB).

En un primer momento la técnica emprende su ejecución con la fase de inicialización (Sección \ref{sec:init}), en la que se generan las soluciones (llamadas jugadores) que conformarán la población completa, la cual se distribuye a continuación en diferentes subpoblaciones. (llamados equipos) que completarán el sistema. Es conveniente destacar que cada equipo posee su propia estrategia de entrenamiento (llamada entrenador). 

Una vez terminada la fase inicial da comienzo la etapa principal del algoritmo: la fase de competición (Sección \ref{sec:comp}). Esta segunda etapa se divide en temporadas, cada cual dividida, a su vez, en semanas, en las que los equipos entrenan de manera independiente y se enfrentan unos con otros creando una competición liguera. Al final de cada temporada se efectúan los procesos de transferencia, en los que los jugadores y entrenadores intercambian sus equipos.

\begin{figure}[htb]
	\centering
		\includegraphics[width=1.0\textwidth]{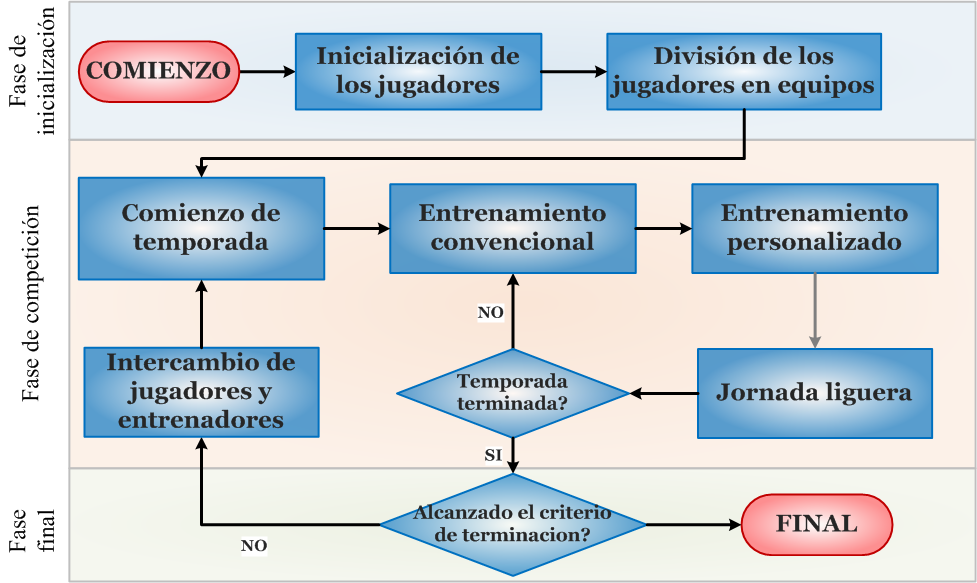}
	\caption{Diagrama de flujo de la meta-heurística GB}
	\label{fig:GBFlow}
\end{figure}

La fase de competición se repite de forma iterativa hasta que el criterio de terminación se satisface. (Sección \ref{sec:term}). El procedimiento completo de la meta-heurística puede verse de manera gráfica en la Figura \ref{fig:GBFlow}, mientras que en el algoritmo \ref{alg:pseudocode1} se presenta un breve pseudocódigo del mismo. En lo que resta de sección se pormenorizarán las diferentes fases del GB.

\begin{algorithm}[tb]
	\SetAlgoLined
	Inicialización de los jugadores (Sección \ref{sec:init})\;
	Distribución de los jugadores entre los distintos equipos (Sección \ref{sec:init})\;
	\Repeat{Criterio de terminación alcanzado \upshape (Sección \ref{sec:term})}{
	  Se ejecuta una temporada (Este paso de detalla en la Sección \ref{sec:comp})\;
	}
	La ejecución termina devolviendo la mejor solución (Sección \ref{sec:term})\;
  \caption{Pseudocódigo del $GB$}
	\label{alg:pseudocode1}
\end{algorithm}

\subsection{Fase de inicialización}
\label{sec:init}

\begin{figure}[tbh]
	\centering
		\includegraphics[width=0.75\textwidth]{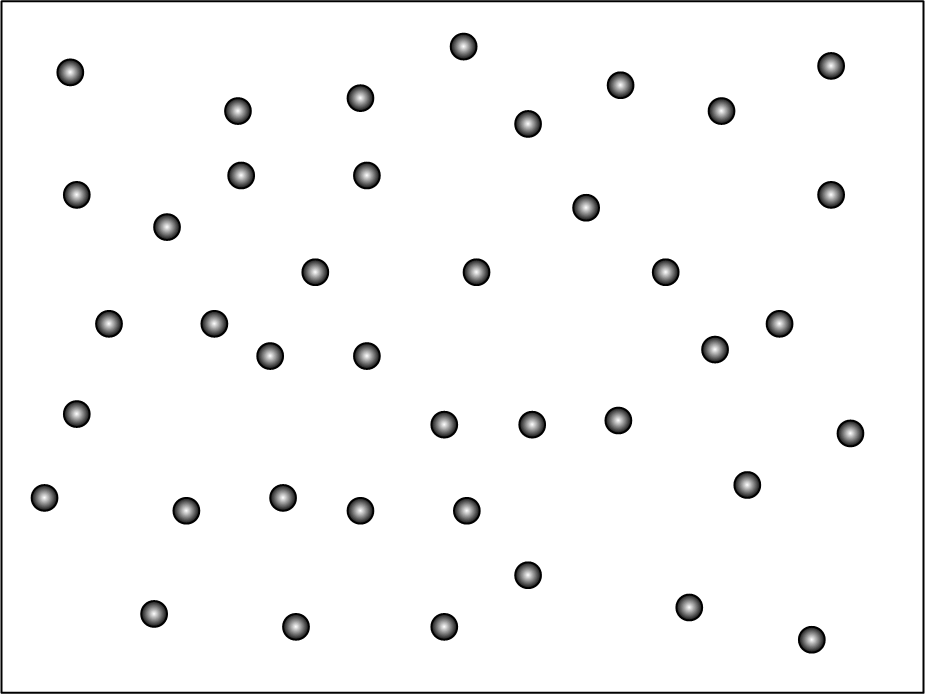}
	\caption{Espacio de soluciones sin la división por equipos realizada}
	\label{fig:emptySpace}
\end{figure}

El primero de los pasos en la ejecución del GB es la creación del conjunto completo de soluciones, denominado \emph{P}, el cual compondrá la población inicial. Todas las soluciones son creadas aleatoriamente y cada una de ellas es denominada como \emph{jugador} $p_i$. En un primer momento los jugadores no tienen ningún equipo asignado, por lo que están dispersos en el espacio de soluciones de forma anárquica, como puede verse en la Figura \ref{fig:emptySpace}. Con todo esto, el conjunto \emph{P} podría ser representado de la siguiente manera:
\[P: \{p_1, p_2, p_3, p_4, p_5, \dots ,p_{TN \ast PT}\}\]
\[\textrm{donde:}\]
\[TN = \textrm{Número total de equipos en el sistema}\]
\[PT = \textrm{Número de jugadores por equipo}\]

Después de generar \emph{P}, los jugadores son distribuidos aleatoriamente entre los diversos equipos $t_i$ que conforman la liga. Este reparto se realiza de forma iterativa, obteniendo un jugador cualquiera de \emph{P} e insertándolo en $t_i$ hasta alcanzar \emph{TN}. Una vez finaliza la distribución, los  jugadores pasan a ser representados por la variable $p_{ij}$, cuyo significado es ``jugador $j$ del equipo $i$". El grupo completo de equipos es representado como \emph{T}, y consiste en un número $TN$ de equipos, siendo \emph{TN}$\geq$2. Con todo esto, a modo de ejemplo, los equipos podrían ser formados de la siguiente forma:
\[\textrm{Equipo} \ t_1= \{p_{11}, p_{12}, p_{13}, \dots ,p_{1PT}\}\]
\[\textrm{Equipo} \ t_2= \{p_{21}, p_{22}, p_{23}, \dots ,p_{2PT}\}\]
\[\dots\]
\[\textrm{Equipo} \ t_{TN}= \{p_{TN1}, p_{TN2}, \dots ,p_{TN PT}\}\]

\begin{figure}[tbh]
	\centering
		\includegraphics[width=1.0\textwidth]{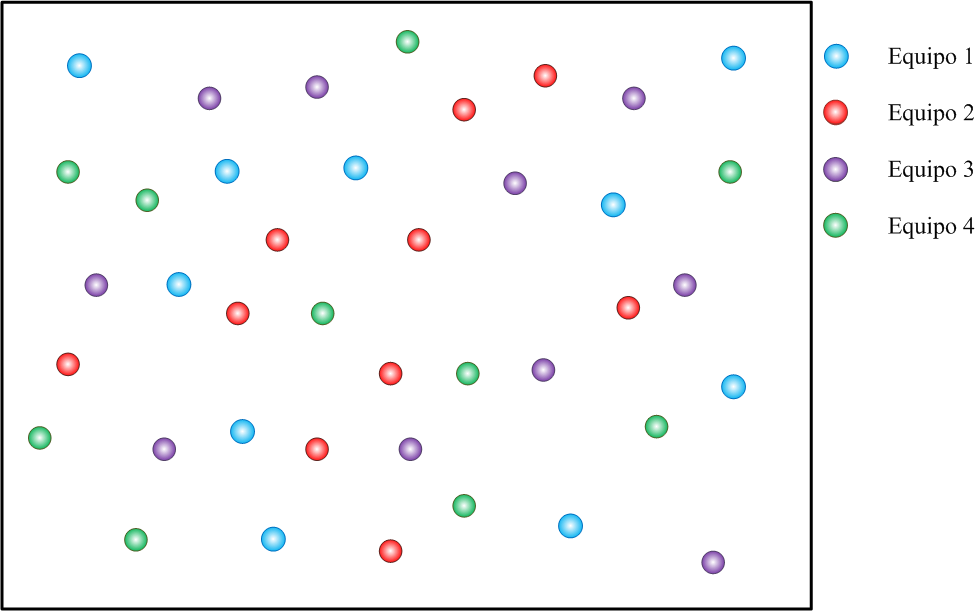}
	\caption{Espacio de soluciones con la división por equipos realizada}
	\label{fig:spaceTeams}
\end{figure}

De esta forma, cada jugador del espacio del soluciones forma parte de un equipo, como puede verse en la Figura \ref{fig:spaceTeams}. El conjunto de equipos \emph{T} puede representarse del siguiente modo:

\[T= \{t_1, t_2, t_3, t_4, \dots ,t_{TN}\}\]

Por otra parte, es lógico pensar que en el mundo real la fuerza o potencia de un equipo depende directamente de la calidad de los jugadores que lo completan. De esta manera, cuanto mejores sean los jugadores, más fuerte será el equipo. Como puede resultar evidente, cuanto más fuerte sea un equipo, podrá ganar más partidos y conseguirá una posición más elevada en la clasificación liguera.

La calidad de un jugador $p_{ij}$ se representa mediante un número real $q_{ij}$. Este número viene determinado por una función objetivo $f(p_{ij})$, dependiente del problema que se esté abordando, de manera que $q_{ij}$ = $f(p_{ij})$ si el objetivo es maximizar, y $q_{ij}$ = 1/$f(p_{ij})$ si el objetivo es minimizar. Por ejemplo, en varios problemas de asignación de rutas a vehículos, como el CVRP, esta función consiste en el sumatorio de las distancias de las rutas que componen una solución. En otros casos, esta función puede ser de una complejidad mayor, y puede tener en cuenta factores como la distancia, el coste de utilización de ciertos vehículos pesados, o la penalización por el incumplimientos de alguna restricción del problema. Además de esto, cada equipo cuenta con un jugador capitán ($p_{icap}$), que es el jugador con el mayor $q_{ij}$ de su equipo. Expresado formalmente:

\[p_{icap} =  p_{ik} \in t_i \Leftrightarrow \forall{j}\in \{1, \dots, PT \}: q_{ik} \geq q_{ij} \]

Es conveniente tener en cuenta que, al igual que sucede en la vida real, es posible que varios jugadores $p_{ij}$ de un mismo equipo $t_i$ posean la misma calidad $q_{ij}$ en el mismo momento temporal. Estas igualdades son transitorias, ya que cada $p_{ij}$ evoluciona de manera individual en los procesos de entrenamiento. Relacionado con este hecho, cuando más de un jugador posee la misma $q_{ij}$, y esa $q_{ij}$ es la más alta de equipo, el $p_{icap}$ es seleccionado de forma aleatoria entre esos jugadores. 

Para calcular la fuerza $TQ_i$ de un equipo, la meta-heurística tiene en cuenta la calidad de todos los jugadores $p_{ij}$ que componen tal equipo. La variable $TQ_i$ puede representarse matemáticamente mediante la siguiente formula: 

\[TQ_i = \sum_{j=1}^{PT}{q_{ij}}/PT\]

Una vez se completa esta fase inicial, la fase de competición da comienzo. Esta segunda etapa se repite de manera iterativa hasta que se cumple el criterio de terminación. 

\subsection{Fase de competición}
\label{sec:comp}

La fase de competición es la etapa principal de la meta-heurística. En esta, los equipos entrenan de forma independiente y cooperativa, mejorando así sus fuerzas progresivamente (Sección \ref{sec:TrainingPhase}). Mientras tanto, los equipos se enfrentan unos contra otros creando una competición liguera, cuyo desarrollo ayuda a la toma de decisiones para el traspaso de jugadores y entrenadores entre equipos (Apartado \ref{sec:PeriodTransfers}). Este proceso está dividido en \textit{temporadas} ($S_i$), cada una de las cuales posee dos periodos de transferencias. Además de esto, en una $S_i$ se suceden tantos partidos como sean necesarios para completar una liga convencional, en la que cada equipo se enfrenta en dos ocasiones a cada uno de los conjuntos restantes (Sección \ref{sec:Matches}). Por esta razón, cada temporada está dividida en dos partes de igual duración. En cada una de estas fracciones, cada equipo disputa un solo enfrentamiento con cada uno de los conjuntos restantes. Con todo esto, cada equipo participa en un número total de $2NT-2$ partidos por cada $S_i$. Finalmente, una temporada tiene tantas sesiones de entrenamientos como jornadas ligueras, o dicho de otra forma, tantas sesiones como partidos jugados por cada $t_i$. Todo este proceso se encuentra esquematizado en el algoritmo \ref{alg:season}.  

	\begin{algorithm}[tb]
	 \SetAlgoLined
		Los puntos de cada equipo $t_i$ son reseteados a 0\;
		\For{$j=1,2$ (Cada temporada se divide en dos partes de igual duración, al igual que en la vida real)}{
			\For{cada jornada liguera \upshape (Sección \ref{sec:Matches}}{
				\For{cada $t_i$ en el sistema}{
					Sesión de entrenamiento para $t_i$ (Sección \ref{sec:TrainingPhase})\;
					Entrenamientos personalizados para $t_i$ (Sección \ref{sec:TrainingPhase})\;
					Calculo de la fuerza $TQ_i$ de $t_i$ (Sección \ref{sec:init})\;
				}
				Jornada liguera, en la que los partidos son jugados (Sección \ref{sec:Matches})\;
			}
			Periodo de fichajes (Sección \ref{sec:PeriodTransfers})\;
			Periodo de terminación de entrenadores (Apartado \ref{sec:PeriodTransfers})\;			
		}
   \caption{Pseudocódigo de una temporada}
	 \label{alg:season}
  \end{algorithm}
	
\subsubsection{Sesiones de entrenamiento}
\label{sec:TrainingPhase}
Una sesión de entrenamiento es ese proceso en el que todos los jugadores de un equipo realizan los ejercicios necesarios para intentar mejorar su calidad. En la vida real, cada equipo tiene su propio método de entrenamiento, el cual depende del entrenador que esté dirigiendo al equipo en ese momento. Algunos métodos de entrenamiento producen mejores resultados que otros, lo que hace que algunos conjuntos progresen mejor que otros. Este hecho se refleja a posteriori en la tabla de clasificación, donde los equipos que realizan un entrenamiento más eficaz consiguen alcanzar posiciones más altas, ya que están más capacitados para ganar partidos. 

Para capturar adecuadamente esta situación en la técnica propuesta, cada $t_i$ contará con su propio método de entrenamiento, el cual consistirá en un procedimiento de búsqueda local con una función de sucesores particular para realizar la búsqueda en el espacio de soluciones. Estas funciones de sucesores tendrán la misma filosofía que las vistas en apartados anteriores de este trabajo (Sección \ref{sec:gensuc}). Para ciertos problemas de asignación de rutas a vehículos, una función apropiada para este proceso podría ser la previamente descrita 2-opt, o la 3-opt. El método de entrenamiento de cada $t_i$ es asignado aleatoriamente durante el proceso de inicialización. Por cada sesión, la función de entrenamiento es aplicada un cierto número de veces (hasta que se alcance su propio criterio de terminación) sobre cada $p_{ij}$. El jugador $p'_{ij}$ generado tan solo será aceptado si, y solo si, $q'_{ij} > q_{ij}$. En tal caso, el nuevo jugador sustituirá al anterior. De este modo, cada equipo rastrea de una manera diferente el vecindario de cada uno de los $p_{ij}$ que posee, haciendo que la evolución de cada jugador sea completamente distinta en función del equipo al que pertenezca en cada momento. Este hecho contribuye a la adecuada exploración y explotación del espacio de soluciones. Estas características se ven realzadas debido a que los jugadores pueden alternar equipos en múltiples ocasiones. 

 \begin{algorithm}[tb]
	 \SetAlgoLined
	 \While{$contador < Termination Criterion$}{
		Se crea un nuevo jugador ($p'_{ij}$) a partir de $p_{ij}$ haciendo uso de la función de entrenamiento\;
		\eIf{$q'_{ij}>q_{ij}$ (la calidad de $p_{ij}$ es mejorada)}{
			$p_{ij}=p'_{ij}$ ($p_{ij}$ es reemplazado $p'_{ij}$)\;
			contador=0\;
		}{
			contador++\;
		}		
	 }
   \caption{Pseudocódigo del proceso de entrenamiento}
	 \label{alg:training}
  \end{algorithm}

Hay que tener en cuenta que cuantas más veces se aplique la función de entrenamiento, más tiempo computacional consumirá. Además de esto, el hecho de aplicar esta función una mayor cantidad de veces no implica una mejora en el rendimiento del algoritmo, ya que el $p_i$ puede caer en un óptimo local. Por esta razón, como ya se ha aludido previamente, cada proceso de entrenamiento tiene su propio criterio de terminación. Con todo esto, cada sesión finalizará cuando se genere un número concreto de sucesores sin experimentar ninguna mejora en la $q_{ij}$ del jugador entrenado. Este número es variable, y está estrictamente relacionado con el vecindario de la función de sucesores utilizada. Haciendo uso del conocido 2-opt como ejemplo, una sesión de entrenamiento se dará por concluida cuando se produzcan $n+\sum_{k=1}^n{k}$ (el tamaño del vecindario) sucesores sin mejora, siendo $n$ el tamaño de la instancia del problema que se esté abordando.

\begin{figure}[tb]
	\centering
		\includegraphics[width=0.8\textwidth]{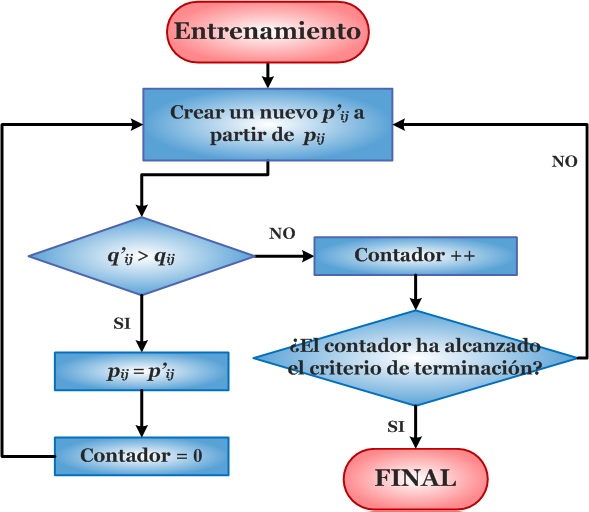}
	\caption{Flujo de trabajo de una sesión de entrenamiento}
	\label{fig:training}
\end{figure}

Con todo esto, el algoritmo \ref{alg:training} describe esquemáticamente este proceso. Por otra parte, en la figura \ref{fig:training} se representa el flujo de trabajo de una sesión de entrenamiento. 

\begin{figure}[htb]
		\centering
			\includegraphics[width=1.0\textwidth]{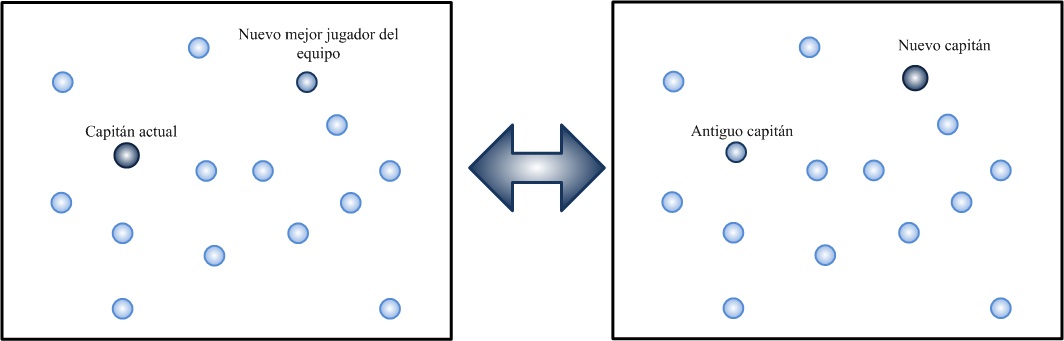}
		\caption{Ejemplo de cambio de capitán}
		\label{fig:changeCap}
	\end{figure}

Es interesante mencionar que este proceso puede suponer un cambio en el $p_{icap}$ de un equipo $t_i$. Este acontecimiento ocurre cuando un jugador $p_{ij}$, después de haber finalizado su sesión de entrenamiento, se encuentra en una situación en la que su calidad es superior a la del capitán de su correspondiente equipo. La figura \ref{fig:changeCap} muestra un ejemplo visual de este intercambio.

Otro tipo de entrenamiento que comprende el GB es el denominado \textit{entrenamiento personalizado}. Es posible que un jugador $p_{ij}$ se encuentre en un periodo en el que, pese a recibir las correspondientes sesiones de entrenamiento, no experimente ninguna mejora en su $q_{ij}$. Desde un punto de vista deportivo, esto puede suceder cuando un jugador se centra en exceso en mejorar ciertas cualidades que no puede mejorar debido a diversos factores, como, por ejemplo, un pobre estado físico. Desde el punto de vista de la optimización, esto sucede cuando un $p_{ij}$ se encuentra en un óptimo local. Con todo esto, el entrenamiento personalizado surge con la idea de ayudar a que dicho $p_{ij}$ consiga huir de tal \textit{obstáculo}. Este método de entrenamiento es llevado a cabo por el $p_{ij}$ con la colaboración del capitán de su equipo. Gracias a este proceso, el $p_{ij}$ podrá verse capacitado para evadirse del óptimo local en el que se encuentra atrapado, pudiendo desplazarse a otras regiones o puntos del espacio de soluciones, los cuales pueden ser prometedores para la búsqueda. 

Desde un punto de vista algorítmico, estos entrenamientos consisten en la combinación de las características de ambos compañeros de equipo, resultando en un jugador que, probablemente, haya realizado un salto en el espacio de soluciones. Este salto puede ser beneficioso para el proceso de búsqueda de la técnica, colaborando con ello a la capacidad de exploración de la misma.

A continuación, se describirá un posible ejemplo de este tipo de ejercicio, en el que un jugador $p_{ij}$ recibe un entrenamiento personalizado con la ayuda de su $p_{icap}$. Podrían suponerse, por ende, los siguientes dos jugadores:

\[p_{ij}: [x_0, x_1, x_2, x_3, x_4, x_5, x_6, x_7, x_8, x_9]\]
\[p_{icap}: [y_0, y_1, y_2, y_3, y_4, y_5, y_6, y_7, y_8, y_9]\]

Donde $x_i$ e $y_i$ son los elementos que componen cada uno de los jugadores, es decir, los valores que toman las variables de la solución que representan a $p_{ij}$ y $p_{icap}$. Una posible combinación de estas características, y por lo tanto el jugador resultante, podría ser la siguiente:

\[p'_{ij}: [x_0, x_1, y_2, y_3, y_4, y_5, x_6, x_7, x_8, x_9]\]

El jugador $p'_{ij}$ creado a partir de este procedimiento reemplaza a $p_{ij}$, independiente de la diferencia de calidades entre ambas soluciones. Considerando el TSP como ejemplo, los diversos  $x_i$ e $y_i$ serían las diferentes ciudades que componen el entorno geográfico, mientras que una función que combinase las características de ambos jugadores podrían ser las famosas OX o HX (vistas en la Sección \ref{sec:crossover}). 

En el mundo real esta es una práctica ampliamente extendida, en la que los jugadores con mayor calidad, o mayor experiencia, ayudan a sus compañeros a mejorar aquellas cualidades que tienen aún por explotar.

\subsubsection{Partidos}
\label{sec:Matches}

En los partidos ligueros que se suceden en el GB, al igual que los que acontecen en el mundo real, dos equipos participan por llevarse la victoria. Como ya se ha mencionado en la introducción de esta Sección \ref{sec:GB}, en el GB se realizan tantos partidos como sean necesarios para completar una liga convencional, en la que cada equipo se enfrenta a todos y cada uno de sus rivales en, exactamente, dos ocasiones.

En la técnica propuesta el procedimiento que sigue cada partido es el siguiente: cada juego consta de $PT$ ocasiones de gol, la cuales se materializan en gol a través de un torneo entre un $p_{ij}$ de cada uno de los equipos. Los jugadores se enfrentan uno a uno en función de la posición que ocupan en sus correspondientes conjuntos. Cada ocasión se resuelve a favor del jugador con mayor $q_{ij}$ de cada enfrentamiento individual. 

Después de decidir el resultado de todas las ocasiones, se decide qué equipo es el vencedor del enfrentamiento. De esta forma, como resulta lógico, el conjunto que haya conseguido anotar un mayor número de goles se llevará tres puntos, mientras que el perdedor no sumará ningún punto. En caso de empate, ambos $t_i$ añadirán un punto a su casillero particular.

\begin{figure}[tb]
	\centering
		\includegraphics[width=0.8\textwidth]{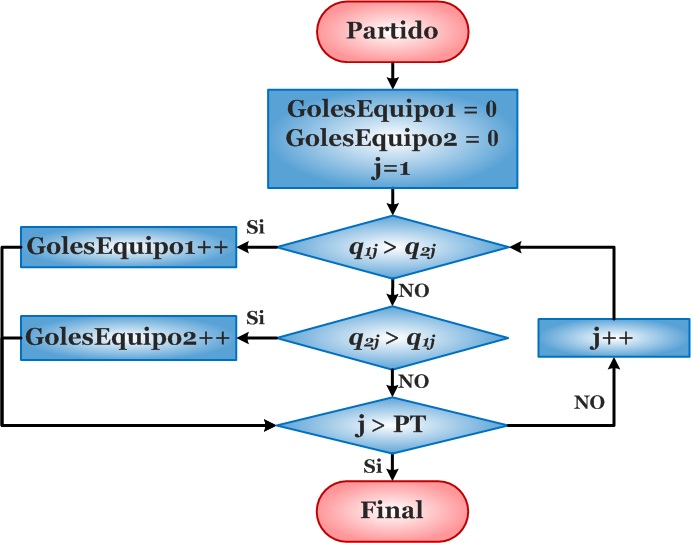}
	\caption{Representación gráfica de un partido}
	\label{fig:match}
\end{figure}

Los puntos logrados por cada conjunto serán utilizados para realizar una clasificación, ordenada en orden descendente. El procedimiento de un partido se muestra de forma esquemática en el algoritmo \ref{alg:match}. Por otro lado, la figura \ref{fig:match} muestra el flujo de trabajo de un partido de manera visual.  Es interesante mencionar que los jugadores de cada $t_i$ están ordenados en orden decreciente en función de su $q_{ij}$.  

 \begin{algorithm}[tb]
	 \SetAlgoLined
		$GolesEquipo1$=0\;
		$GolesEquipo2$=0\;
	 \For{cada jugador en un equipo ($PT$)}{
		\If{$q_{1i}>q_{2i}$ (El jugador del primer equipo es mejor)}{
			$GolesEquipo1++$\;
		\ElseIf{$q_{1i}<q_{2i}$ (El jugador del segundo equipo es mejor)}{
			$GolesEquipo2++$\;
		}}
	 }
   \caption{Pseudocódigo del procedimiento de un partido}
	 \label{alg:match}
 \end{algorithm}

\subsubsection{Periodo de transferencias}
\label{sec:PeriodTransfers}

El periodo de transferencias es un proceso en el que, principalmente, los equipos intercambian jugadores entre sí. Con esta iniciativa los conjuntos acogen a nuevos jugadores con la intención de verse reforzados, a cambio de dejar escapar alguno de sus efectivos. Este es un proceso común en el mundo del fútbol, existiendo anualmente dos ``mercados de fichajes": el mercado veraniego, y el mercado invernal. El primero de ellos acontece al comienzo de cada temporada, mientras que el segundo se realiza en la etapa central de la misma. Lo más normal en estos periodos de transferencias es que los mejores equipos logren la adquisición de los jugadores más cualificados, mientras que los equipos más mediocres se conforman con jugadores de categoría inferior.

Este hecho también ha sido implementado en el GB. En este caso, la puntuación liguera de cada $t_i$ y su posición en la tabla de clasificación son factores decisivos para decidir el tipo de transferencia que dicho $t_i$ recibe. Con todo esto, en el periodo central y final de cada temporada los equipos que se encuentran en la parte mitad superior de la clasificación son reforzados con los mejores $p_{ij}$ de los conjuntos clasificados en la mitad inferior. Por el contrario, los $t_i$ de la mitad baja han de conformarse con la adquisición de los jugadores menos aptos de los equipos punteros.

Ha de tenerse en cuenta que cuanta mejor sea la posición de un equipo, mejor será el jugador que recibirá en la transacción. Es decir, el $t_i$ posicionado en el primer lugar recibirá al mejor jugador del equipo peor posicionado. Por otro lado, el segundo clasificado acogerá al segundo mejor $p_{ij}$ del penúltimo equipo. Conviene aclarar que si $TN$ es un número impar, el equipo colocado en la mitad de la tabla no recibe ningún jugador. A continuación se muestra un ejemplo práctico de una liga compuesta por 4 equipos, teniendo en cuenta que los jugadores están ordenados por su $q_{ij}$ y los equipos por su orden clasificatorio:

\[Equipo \ t_1: \{a_0, a_1, a_2, a_3, a_4, a_5, a_6, a_7, a_8 \}\]
\[Equipo \ t_2: \{b_0, b_1, b_2, b_3, b_4, b_5, b_6, b_7, b_8 \}\]
\[Equipo \ t_3: \{c_0, c_1, c_2, c_3, c_4, c_5, c_6, c_7, c_8 \}\]
\[Equipo \ t_4: \{d_0, d_1, d_2, d_3, d_4, d_5, d_6, d_7, d_8 \}\]

Después del periodo de fichajes estos equipos quedarían formados del siguiente modo:

\[Equipo \ t_1: \{a_0, a_1, a_2, a_3, a_4, a_5, a_6, a_7, \boldsymbol{d_0} \}\]
\[Equipo \ t_2: \{b_0, b_1, b_2, b_3, b_4, b_5, b_6, \boldsymbol{c_1}, b_8 \}\]
\[Equipo \ t_3: \{c_0, \boldsymbol{b_7}, c_2, c_3, c_4, c_5, c_6, c_7, c_8 \}\]
\[Equipo \ t_4: \{\boldsymbol{a_8}, d_1, d_2, d_3, d_4, d_5, d_6, d_7, d_8 \}\]

Este intercambio de $p_{ij}$ ayuda al proceso de búsqueda, ya que permite tratar de diferente manera las soluciones durante la ejecución, evitando que estas caigan con facilidad en óptimos locales, y aumentando la capacidad de rastreo de la meta-heurística. En otras palabras, este proceso de intercambio de vecindario ayuda a la capacidad de exploración de la técnica, y especialmente, contribuye a mejorar la habilidad de explotación de regiones prometedoras del espacio de soluciones. 

Otro tipo de transferencia que también se tiene en cuenta en el GB son las denominadas \textit{transferencias especiales}. En el mundo del fútbol, los jugadores no sólo dejan sus equipos con el propósito de ir a uno de mayor calidad. Existen muchos casos en los que un jugador, después de haber permanecido durante años en un mismo equipo, pierde su ambición y cae en un estado en el que no consigue mejorar sus cualidades. Como consecuencia de esto, el jugador decide cambiar de equipo sin importar si el conjunto de destino es mejor o peor.

Este hecho es reflejado en la meta-heurística diseñada en esta tesis, pese a que se aplica de un modo distinto al que puede verse en el mundo real. En concreto, cuando un $p_{ij}$ cualquiera toma un número prefijado de entrenamientos sin experimentar ningún tipo de mejora en su $q_{ij}$, incluso habiendo recibido entrenamientos personalizados, realiza un cambio de equipo. Este cambio se produce entre el equipo $t_i$ y otro conjunto $t_k$ seleccionado al azar, sin importar si $TQ_k < TQ_i$. Este traspaso hace que $t_i$ pierda uno de sus efectivos en beneficio de $t_k$. Para afrontar esta perdida y para poder mantener el $PT$ de cada equipo, el conjunto $t_k$ entrega a $t_i$ un jugador seleccionado al azar.

Por otro lado, este tipo de transferencia puede suceder en cualquier momento de la temporada, por lo que, pese a que mantiene la filosofía principal que la adapta al mundo real, no es estricta con las reglas impuestas por las federaciones de fútbol profesionales. En relación al número de entrenamientos que tiene que experimentar el jugador en cuestión antes de ser transferido, conviene aclarar que es un parámetro de entrada, el cual depende del desarrollador de la técnica. En la experimentación llevada a cabo en este estudio este número ha sido establecido en 10. A continuación se muestra un pequeño ejemplo para una transferencia especial, en la que se asumen dos equipos $t_i$ y $t_k$:

\[Equipo \ t_i: \{a_0, a_1, a_2, a_3, a_4, a_5, a_6, a_7, a_8 \}\]
\[Equipo \ t_k: \{b_0, b_1, b_2, b_3, b_4, b_5, b_6, b_7, b_8 \}\]

Suponiendo que el jugador $a_3$ es el que no ha experimentado mejora en su $q_{ij}$, se produce un intercambio con un jugador seleccionado al azar proveniente de un equipo también elegido aleatoriamente. En este caso, el equipo escogido es $t_k$. Asumiendo que el jugador elegido para la transacción es $b_5$, la composición de los equipos resultaría de la siguiente manera:

\[Equipo \ t_i: \{a_0, a_1, a_2, \boldsymbol{b_5}, a_4, a_5, a_6, a_7, a_8 \} \]
\[Equipo \ t_k: \{b_0, b_1, b_2, b_3, b_4, \boldsymbol{a_3}, b_6, b_7, b_8 \} \]

Asimismo, en el mundo del fútbol, no sólo los jugadores son transferidos entre diferentes conjuntos. Los entrenadores también son reemplazados por otros frecuentemente, sobre todo cuando estos no están logrando los resultados esperados por la afición y por la directiva, o cuando encadenan una sucesión de derrotas demasiado larga.

Este suceso, realmente común en la vida real, también se refleja en el método propuesto en esta tesis doctoral en un proceso denominado \textit{Cese de entrenadores}. Este acontecimiento se realiza de la siguiente manera: en cada periodo de fichajes, todos los $t_i$ posicionados en la mitad inferior de la tabla clasificatoria sustituyen su método de entrenamiento por otro seleccionado de forma aleatoria. Este intercambio se realiza con la esperanza de obtener un nuevo procedimiento de entrenamiento, o entrenador, que consiga mejorar los resultados del anterior, aumentando con esto la $TQ_i$ del equipo.

Como ya se citado, este intercambio se realiza de forma aleatoria, seleccionando al azar uno de los tipos de entrenamiento existentes en el sistema, permitiendo repeticiones entre los diferentes $t_i$. Esta sustitución en el vecindario afecta a todos los $p_{ij}$ que componen el equipo, contribuyendo a la capacidad de exploración de la técnica.

\subsection{Fase final}
\label{sec:term}

La finalización de la ejecución de la meta-heurística depende directamente del cumplimiento simultáneo de tres factores. Estos tres criterios tienen que permitir que la búsqueda realizada por el algoritmo haya rastreado una amplia extensión del espacio de soluciones, evitando a su vez una excesiva pérdida de tiempo, la cual se produce cuando la técnica continúa su ejecución cuando la probabilidad de mejora en las soluciones es muy baja. Con esta intención, el criterio de finalización formulado para el GB se compone de las siguientes tres condiciones:

\begin{equation}
	\sum_{i=1}^{TN}{TQ'_i} \leq \sum_{i=1}^{TN}{TQ_i} \label{eq1}
\end{equation}
\begin{equation}
	\sum_{i=1}^{TN}{q'_{icap}} \leq \sum_{i=1}^{TN}{q_{icap}} \label{eq2}
\end{equation}
\begin{equation}
	BestSolution' \leq BestSolution \label{eq3}
\end{equation}

Dicho de otra manera, la ejecución del GB finaliza cuando la suma de las fuerzas $TQ'_i$ de todos los equipos no mejora respecto a la temporada anterior, (\ref{eq1}), a la vez que la suma de la calidad $q'_{icap}$ de todos los capitanes no experimenta mejora en relación a la temporada previa (\ref{eq2}), y no se mejora la mejor solución encontrada hasta el momento ($BestSolution'$) en comparación con la temporada precedente (\ref{eq3}).

Cuando este criterio de finalización se satisface, el algoritmo devuelve el $p_{ij}$ con la mejor $q_{ij}$ de todo el sistema. Este jugador será considerado la solución que la meta-heurística proporcionará al problema.
	

\section{Principales aportaciones del modelo propuesto}	
\label{sec:aportaciones}

Después de haber descrito en detalle el modelo propuesto en esta tesis doctoral, en esta sección se mencionarán las aportaciones y los aspectos originales de la contribución en comparación con varias de las técnicas multi-poblacionales más exitosas de la literatura. Para esta comparación se utilizarán las meta-heurísticas descritas en el apartado \ref{sec:multi}, todas ellas con un reconocimiento contrastado, y con una eficacia comprobada en la literatura.

Es interesante mencionar que existe una mayor cantidad de diferencias entre las distintas técnicas además de las listadas en este apartado. En esta pequeña sección se han querido mencionar tan solo las originalidades más significativas. De esta manera, y comenzando con los PABC, las principales aportaciones que presenta el GB podrían listarse de la siguiente manera:
\begin{enumerate}
		\item En los algoritmos PABC existen tres tipos diferentes de individuos en las distintas poblaciones, cada uno con su propio comportamiento. En el GB solo existe un tipo de individuo, y todos los elementos del sistema se comportan de la misma manera.
		\item En el GB se presenta una estrategia de intercambio de jugadores entre equipos basado en la calidad de las subpoblaciones en su conjunto. Un sistema de estas características no ha sido nunca aplicado a un ABC multi-poblacional. 
		\item En el PABC las abejas comparten información entre ellas. Por otro lado, en el GB los jugadores no sólo comparten información entre ellos (conociendo en cada momento la calidad de sus compañeros de equipo), además de esto, y en caso de ser necesario, también pueden compartir sus propias características durante el proceso de entrenamiento personalizado.	
\end{enumerate}

En relación a la meta-heurística multi-poblacional PPSO, las siguientes dos diferencias podrían ser fácilmente destacadas:

\begin{enumerate}
		\item En el caso de los PPSO, cada partícula realiza sus movimientos en función de su posición actual, su mejor posición encontrada y la mejor posición encontrada por todo el enjambre. En el caso del GB, cada jugador realiza sus movimientos de forma autónoma, sin la necesidad de utilizar información proveniente de otros jugadores o equipos (a excepción de los entrenamientos personalizados, cuya ejecución es ocasional)		
		\item En las versiones paralelas del PSO, cada subpoblación ejecuta sus movimientos de la misma manera, basándose en el parámetro ``velocidad". En el GB, por el contrario, cada equipo posee su propia estructura de vecindario, y cada uno evoluciona de forma diferente. Además de esto, cada jugador puede cambiar su método de entrenamiento dependiendo del equipo en el que se encuentre en cada momento.
\end{enumerate}

En los referente a los algoritmos genéticos paralelos, y teniendo en cuenta las distintas filosofías de los tres tipos de PGAs existentes (las cuales pueden consultarse en la sección \ref{sec:PGA}), puede concluirse que los PGAs basados en islas, o los PGAs distribuidos (DGA), son lo más similares al GB en cuanto a concepto y forma de trabajar. Con todo esto, las principales originalidades que aporta el GB frente a los DGAs son las siguientes:

\begin{enumerate}
		\item Ambos algoritmos hacen uso de dos tipos de operadores, el primero de ellos aplicado a un único individuo (mutación y entrenamiento convencional), y el segundo aplicado a más de un individuo de forma cooperativa (cruce y entrenamiento personalizado). Como puede leerse en \cite{DGAstudy}, la gran mayoría de DGAs dan mayor importancia al segundo operador, dejando el primero de ellos en un segundo plano. Esto no es así en el caso del GB, donde la optimización individual recibe mayor importancia, utilizando la cooperación entre individuos como un operador ocasional.
		\item En relación a los operadores cooperativos y su manera de funcionar, en el caso de los DGAs se hace uso de las clásicas funciones de selección, utilizadas también en las versiones uni-poblacionales de los GAs, para decidir qué individuos participan en cada cruce. En el GB, por el contrario, los entrenamientos personalizados se realizan rara vez, y siempre se materializan entre un individuo atrapado en un óptimo local y el mejor individuo de su subpoblación.
		\begin{sloppypar}
		\item Como bien puede leerse en la literatura, en trabajos como \cite{PGAVRP1,DGAstudy}, el diseño e implementación de sistemas eficientes de migración para un DGA es un proceso complejo, existiendo una gran cantidad de arquitecturas y topologías. Por el contrario, en el GB existe una estrategia de migración perfectamente definida, la cual se basa en la calidad de cada una de las poblaciones.
		\end{sloppypar}
		\item En los DGAs, las poblaciones cooperan entre sí, en el caso del GB, compiten. 
\end{enumerate}

Teniendo el ICA como objetivo de comparación, es recomendable destacar que la filosofía y manera de trabajar de ambos algoritmos distan mucho el uno del otro. Las principales originalidades del GB respecto al ICA podrían listarse del siguiente modo:

\begin{enumerate}
		\item En los ICA, todas las colonias realizan sus movimientos condicionadas por la posición del estado imperialista, mientras este último permanece inmóvil. En el caso del GB, cada individuo realizas sus movimientos de forma autónoma.
		\item En los ICA, el número de subpoblaciones va reduciéndose paulatinamente hasta que todos los individuos quedan concentrados en una única población. En el GB, por el contrario, el número de subpoblaciones y el tamaño de cada una de ellas se mantienen a lo largo de toda la ejecución.
		\item En lo que respecta al intercambio de individuos entre diferentes subpoblaciones, este se realiza de un modo diferente en cada meta-heurística. Mientras que en el caso del GB son intercambios de individuos, en los ICAs son simples transferencias, en las que una población envía un individuo, y la otra lo recibe.	
\end{enumerate}

Finalmente, las diferencias entre el SOA y el GB son también significativas, siendo la más notable el hecho de que en el SOA cada individuo realiza sus movimientos en función de su historial de posiciones, o los historiales de las posiciones de sus vecinos (dependiendo del comportamiento que tenga). En el GB, como ya se mencionado anteriormente, todos los individuos tienen el mismo comportamiento, y realizan sus movimientos de manera autónoma. Otra originalidad importante del GB frente al SOA es el sistema de migración. En el SOA, esta migración se realiza mediante el cruce de individuos de diferentes subpoblaciones, mientras que en el GB los individuos cambian de equipo en función de la calidad de estos.

Como puede comprobarse, el método propuesto en este trabajo ofrece numerosas originalidades en relación a las diferentes técnicas presentadas. Como ya se ha indicado, estas técnicas son ampliamente utilizadas hoy en día, gracias a su eficacia a la hora de abordar distintos problemas.

Analizando la filosofía y el modo de trabajo de los algoritmos anteriormente descritos, puede concluirse que el DGA es el que comparte más similitudes con el GB. Ambos utilizan dos tipos de operadores diferentes para la evolución de sus individuos, uno de ellos local y el otro cooperativo. Estos dos operadores se utilizan para encontrar un equilibrio aceptable entre la explotación y exploración del espacio de soluciones. Además de esto, ambas técnicas son fácilmente aplicables a problemas de asignación de rutas a vehículos, y son sencillas de parametrizar. Han sido estas las principales razones que han empujado a optar por el DGA y por el GA uni-poblacional como algoritmos de comparación en la fase experimentación. Todos los detalles de esta fase serán ampliamente explicados en el siguiente capítulo.	



\begin{savequote}[40mm]
No creo que haya alguna emoción más intensa para un inventor que ver una de sus creaciones funcionando
\qauthor{Nikola Tesla}
\end{savequote}

\chapter{Experimentación y resultados}
\label{cha:Exp}

\graphicspath{ {4_exp/figures/} }


\lettrine{C}{}{on} el propósito de comprobar que la meta-heurística planteada es una técnica prometedora y que los resultados ofrecidos por esta son competitivos se ha llevado a cabo una extensa experimentación, en la que se han utilizado seis problemas de optimización combinatoria. Todos estos problemas han sido descritos en capítulos anteriores de este documento, más concretamente en el Capítulo \ref{cha:optimizacion}, y son los siguientes: TSP, ATSP, CVRP, VRPB, NQP y BPP. Los cuatro primeros son problemas de asignación de rutas a vehículos, mientras que el NQP es un problema de satisfacción de restricciones, y el BPP pertenece a la categoría de diseño combinatorio óptimo. Estos dos últimos han sido empleados con la intención de demostrar que el GB no es solo efectivo frente a problemas de asignación de rutas a vehículos, si no que esta efectividad puede ser ampliada a otros tipos de problemas de distinta naturaleza.

Para contrastar los resultados obtenidos por el GB, estos serán comparados con los obtenidos por dos versiones diferentes del GA clásico y dos versiones adicionales del DGA. La razón por la cual se han escogido estas meta-heurísticas para el cotejo de resultados puede verse en la sección \ref{sec:aportaciones}. Con todo esto, las cinco técnicas serán aplicadas a un total de 98 instancias, repartidas entre los seis problemas anteriormente mencionados. Finalmente, con el deseo de extraer conclusiones rigurosas se han llevado a cabo dos exámenes estadísticos con los resultados obtenidos: el test de Friedman y el test normal $z$.

Todos estos aspectos relacionados con la experimentación serán minuciosamente detallados a lo largo de este capítulo, junto a otras cuestiones interesantes, como por ejemplo, la introducción de buenas prácticas para la comparación e implementación de meta-heurísticas (sección \ref{sec:practicas}). En la sección \ref{sec:problemasexp} se describirá la estrategia adoptada para abordar los problemas que se tratarán para validar el problema propuesto. A continuación, las técnicas utilizadas y su parametrización serán descritas en el apartado \ref{sec:parametrizacion}. Finalmente, los resultados se mostrarán en la sección \ref{sec:resultados}, junto al correspondiente análisis de los mismos y conclusiones extraídas (apartado \ref{sec:analisis}).


\section{Buenas prácticas para la implementación y comparación de meta-heurísticas}
\label{sec:practicas}

Antes de comenzar con los detalles de la experimentación llevada a cabo en esta tesis, se ha decidido dedicar un apartado a la introducción de un conjunto de buenas prácticas para la implementación y comparación de diferentes meta-heurísticas aplicadas a problemas de optimización combinatoria. Esta metodología ha sido escrupulosamente aplicada a lo largo de la fase experimental de esta tesis.

Con la intención de contextualizar estas buenas prácticas es conveniente mencionar que los investigadores que trabajan en cualquier campo de la computación algorítmica (ya sea diseñando nuevos algoritmos o mejorando los ya existentes) encuentran en muchas ocasiones grandes dificultades para evaluar sus trabajos. Frecuentemente, la comparación entre diferentes trabajos científicos en esta área es muy difícil, debido a la ambigüedad o falta de detalle en la presentación de técnicas o resultados. Esto hace que la tarea de replicar trabajos realizados por otros investigadores sea una labor engorrosa, produciéndose una gran pérdida de tiempo y un retraso en los avances científicos. Tras sufrir este problema en varias ocasiones, el autor de esta tesis doctoral ha desarrollado junto a varios investigadores más un procedimiento sencillo para la presentación e implementación de meta-heurísticas. En esta sección se introducirá tal procedimiento, el cual ha sido contrastado por la comunidad científica en diversos trabajos, habiendo sido presentados en congresos o publicados en revistas internacionales \cite{practicas1,practicas2,practicas3,practicas4}.

Como ya se ha comentado en la Sección \ref{sec:tecnicas_de_resolucion}, las técnicas más utilizadas para la resolución de problemas de optimización combinatoria son las heurísticas y las meta-heurísticas. Explicado brevemente, las heurísticas se centran en la resolución de problemas con una formulación simple, intentando encontrar la solución óptima de una forma rápida. Las meta-heurísticas, por el contrario, pueden ser aplicadas a un amplia variedad de problemas reales, con una complejidad elevada, para los cuales sería muy complicado implementar una heurística específica para ellos. En este sentido, la comparación entre heurísticas es más simple que la comparación entre meta-heurísticas, ya que están implementadas para un problema específico. No importa cuál sea la naturaleza de la heurística o los parámetros y características utilizadas, la mejor heurística será aquella que obtenga los mejores resultados en un tiempo razonable. Pese a esto, existen problemas a la hora de comparar heurísticas si sus resultados no se muestran de forma adecuada. Este hecho se explica en \cite{VRPTW3}.

La comparación entre meta-heurísticas, por otro lado, es mucho más compleja, ya que hay que tener en cuenta muchos más factores. Este hecho puede generar una gran controversia y puede llevar a confusiones y malas prácticas. A pesar de esto, no existe hoy en día una metodología o procedimiento que ayude a los investigadores a describir y comparar sus meta-heurísticas de una forma justa y objetiva. Es esta la razón principal que ha impulsado el desarrollo del presente conjunto de buenas prácticas, el cual podría dividirse en dos sub-apartados perfectamente diferenciables. El primero de ellos está relacionado con la implementación y presentación de meta-heurísticas (Sección \ref{sec:ImplementationOfTheMetaHeuristic}), mientras que el segundo se centra en la adecuada comparación de diversas técnicas (Sección \ref{sec:ShowingTheResults}). 

\subsection{Buenas prácticas para la implementación y presentación de una meta-heurística}
\label{sec:ImplementationOfTheMetaHeuristic}

Un procedimiento adecuado para la implementación y presentación de una meta-heurística podría estar confeccionado por los siguientes puntos:

\begin{itemize}
	\item Especificar detalladamente las restricciones del problema, clasificadas en restricciones estrictas y ligeras.
	\item Especificar la función objetivo utilizada, la cual debería incluir las restricciones ligeras, en caso de existir.
	\item A la hora de presentar un trabajo, el tipo de meta-heurística utilizada tiene que ser especificada en el título o en el resumen, mencionando la utilización de heurísticas, en caso de hacer uso de ellas. 
	\item Describir en detalle todos y cada uno de los operadores empleados para la implementación de la técnica. Si estos han sido desarrollados por el propio autor tienen que ser minuciosamente detallados. En caso contrario, tienen que ser correctamente enunciados y referenciados. Si los operadores utilizados no se describen o referencian, la replicabilidad de los resultados se hace imposible.
\end{itemize}

El primer paso en el diseño de una técnica para la resolución de un problema de optimización es definir claramente las restricciones que el problema a resolver tendrá. Estas restricciones tienen que estar sujetas a una función objetivo, la cual se tiene que mostrar y explicar de la forma más detallada posible. Este último punto tiene una importancia vital a la hora de implementar una técnica de esta índole. Para problemas como el TSP, por ejemplo, los cuales tienen una formulación simple, este hecho tiene menos importancia, ya que se da por hecho que la función objetivo es la distancia total recorrida y el objetivo principal minimizarla. Para problemas con una complejidad más elevada, como el CVRP o el VRPTW, por ejemplo, este hecho puede convertirse en una fuente de confusión. En estos casos la función objetivo puede variar dependiendo de las necesidades de los desarrolladores. Para el CVRP, por ejemplo, hay estudios que priorizan la minimización de los vehículos utilizados \cite{vehiculos}, mientras que otros se centran únicamente en la reducción de las distancias recorridas \cite{distancia}. Es por esta razón por la que, con la intención de evitar confusiones, describir detalladamente la función objetivo utilizada se considera una buena práctica. El caso contrario podría considerarse una mala práctica.

Una vez introducido el problema y sus características es importante presentar de forma adecuada la meta-heurística que se va a emplear para su resolución. Una práctica que tiene que ser evitada es la confusa denominación de las técnicas. Un ejemplo de esta denominación confuso puede encontrarse en \cite{malaPractica2}. En este trabajo, los autores presentan un enfoque para resolver un problema de asignación de rutas a vehículos heterogéneos con restricciones de carga bidimensionales como una meta-heurística, cuando, realmente, la técnica utilizada es un recocido simulado basado en una búsqueda local heurística. Con todo esto, los autores deberían precisar en mayor detalle esta denominación, presentando la técnica propuesta como una \textit{meta-heurística adaptada al problema}.

Después de definir en detalle la función objetivo y de elegir el tipo de meta-heurística a emplear, el siguiente paso es decidir cómo implementar la técnica y qué tipo de operadores utilizar. Pese a que parece simple, este hecho puede acarrear una gran controversia. Como ya se ha mencionado en capítulos anteriores, las meta-heurísticas utilizan diferentes operadores para modificar y transformar las soluciones disponibles, con el propósito principal de mejorarlas. El primer punto a tener en cuenta en este aspecto es el siguiente: si la finalidad del estudio es comparar una meta-heurística con otra u otras preexistentes para validar su capacidad genérica de optimización, es necesaria la utilización de operadores neutros a lo largo de toda la implementación. En otras palabras, los operadores que utilicen información específica del problema y optimicen por sí solos han de ser evitados. Por el contrario, si la finalidad es resolver un problema específico encontrando la mejor solución en tiempo razonable, los operadores podrán ser todo lo específicos que el autor de dicho estudio considere necesario, y la comparación se deberá hacer con el mejor algoritmo conocido hasta el momento.

Como ejemplo de este hecho se podría utilizar el proceso de inicialización de un GA. La forma más apropiada para probar la capacidad de optimización de una meta-heurística es utilizar un proceso de inicialización 100\% aleatorio, en el que los individuos se generen de forma aleatoria, en lugar de utilizar funciones de inicialización como las propuestas, por ejemplo, en \cite{solomon}, para el VRPTW. En caso de hacer uso de alguna de estas funciones heurísticas, los individuos ya habrán pasado por un proceso de optimización antes de comenzar la ejecución central del algoritmo. De esta forma, no sería posible determinar con exactitud cuál es la capacidad de optimización de la técnica. En este caso concreto, se tendría que decir que se está generando una meta-heurística adaptada al problema en lugar de una meta-heurística, ya que se hace uso de información específica del problema para inicializar la población.

\begin{sloppypar}
Continuando con el GA, y poniendo en este caso el TSP como ejemplo, una operador de cruce heurístico podría ser el very greddy crossover (VGX) \cite{VGX}. El VGX es un operador para el TSP que utiliza las distancias entre las ciudades para generar los hijos resultantes de un cruce. Es lógico pensar que haciendo uso de este operador el GA obtendrá unos resultados prometedores para el TSP, ya que realiza una pequeña optimización en los individuos por sí mismo. Para generar una meta-heurística pura habría que evitar operadores de este tipo, utilizando en su lugar operadores como los descritos en la sección \ref{sec:crossover}. Funciones como el OX o HX solo se preocupan de cumplir las restricciones del problema, y no utilizan ningún tipo de información específica del problema.
\end{sloppypar}

Por lo tanto, tiene que evitarse en cualquier trabajo de validación de la capacidad genérica de optimización de una técnicas la comparación entre técnicas meta-heurísticas con operadores neutros y métodos heurísticos con operadores específicos. De lo contrario, la comparación no será válida dada la diferencia de la naturaleza de las técnicas. Un ejemplo de esta mala práctica se puede ver en \cite{malaPractica1}, donde se comparan tres procedimientos para la resolución de un problema de \textit{clustering}. En este trabajo dos de las técnicas comparadas son meta-heurísticas puras, mientras que la última es una meta-heurística adaptada al problema. Otro ejemplo de esta mala práctica puede ser encontrada en \cite{malaPractica3}, donde se introduce un nuevo operador de cruce heurístico aplicable al TSP, llamado cruce constructivo secuencial (SCX). Para comprobar la calidad de este nuevo operador los resultados obtenidos por el SCX se comparan con los obtenidos por otros dos GA, utilizando cada uno de ellos operadores de cruce neutros. Lógicamente, el SCX obtiene resultados mejores, pero la diferencia en la naturaleza de los operadores hace que la comparación no sea válida.

Relacionado con este aspecto, hay que tener en cuenta que, para realizar una comparación completamente fiable entre diversas meta-heurísticas, es obligatorio utilizar los mismos operadores y parámetros, siempre que la naturaleza de las técnicas lo haga posible. Si las características de los algoritmos lo impiden, hay que intentar que los operadores tengan una filosofía similar. Es por esa razón por lo que los puntos explicados en este apartado son de vital importancia para obtener unos resultados fácilmente reutilizables y dar credibilidad a las comparaciones que se hagan en un mismo trabajo.

\subsection{Buenas prácticas para la presentación de resultados ofrecidos por una meta-heurística}
\label{sec:ShowingTheResults}

Una vez descritas las características de una meta-heurística, es apropiado mostrar los resultados que esta puede obtener. Este es un hecho importante, ya que de la forma en la que se muestren las pruebas realizadas depende la replicabilidad e impacto que pueda tener una técnica. Con todo esto, en términos de realización de experimentos y demostración de resultados, un procedimiento de buenas prácticas constaría de los siguientes puntos:

\begin{itemize}
	\item Si el problema lo permite, las pruebas tienen que ser realizadas con instancias obtenidas de un banco de pruebas, o benchmark. Obviamente, cuantas más instancias se utilicen, más rico será el estudio. Todas las instancias empleadas han de ser referenciadas, con su nombre y el benchmark al que pertenecen. Es interesante mencionar que el uso de instancias particulares puede ser correcto siempre y cuando queden perfectamente descritas, y puedan replicarse.
	\item Es de vital importancia mostrar los tiempos de ejecución, acompañados por la unidad temporal y una explicación de las características del ordenador en el que las pruebas se han llevado a cabo.
	\item Aparte de mostrar los resultados, y con la intención de realizar una comparación justa entre técnicas presentadas en diferentes estudios, es aconsejable mostrar el número de iteraciones necesarias por la meta-heurística para obtener los resultados de cada ejecución.
	\item Es recomendable mostrar la mayor cantidad de datos posibles. De esta forma, las comparaciones entre diferentes técnicas pueden ser más fiables. Ejemplos de esta información podrían ser el número de ejecuciones, la mejor y peor solución, la media, desviación típica, tiempos de ejecución\dots
\end{itemize}

Para comprobar la calidad de una nueva técnica, esta tiene que ser aplicada a varias instancias del problema que se esté tratando. La mejor opción para realizar estas pruebas es con uno de los muchos benchmarks que pueden encontrarse en la literatura. Estos conjuntos de pruebas están compuestos por instancias de problemas particulares, las cuales pueden ser utilizadas por los investigadores para aplicar sus técnicas sobre ellas. Muchas de estas instancias tienen una solución óptima conocida, de forma que se puede determinar con mayor facilidad cómo de buena es una meta-heurística comparando sus resultados obtenidos con los ofrecidos por el benchmark. Utilizando como ejemplo los problemas de asignación de rutas, existen varios benchmark los cuales poseen una gran cantidad de instancias de diferentes problemas, como el TSPLIB \cite{TSPLibrary} o el VRPWeb\footnote{http://neo.lcc.uma.es/vrp}. De esta forma, hay que evitar realizar pruebas con instancias desconocidas, siempre y cuando estas no sean perfectamente descritas, como puede verse en \cite{memetico2} o \cite{noBenchmark2}.

Otro punto importante que no puede pasarse por alto son los tiempos de ejecución. Puede considerarse una mala práctica mostrar los resultados de una técnica sin mostrar los tiempos de ejecución necesitados, como ocurre, por ejemplo, en \cite{noTimes} o \cite{noTime2}. Aunque pueda parecer un asunto lógico, junto a los tiempos han de mostrarse también las unidades de tiempo, es decir, segundos, minutos\dots De lo contrario se consideraría una mala práctica, como ocurre en \cite{imperialistTSP1}. Aparte de mostrar los tiempos de ejecución, es importante describir también las características del ordenador en el que las técnicas han sido ejecutadas, con el objetivo de contextualizar dichos tiempos.

Aunque los tiempos de ejecución ayudan a la comparación de métodos mostrados en el mismo trabajo, para la comparación entre técnicas descritas en diferentes estudios se hace necesaria la utilización de otro parámetro adicional. Este hecho se da porque la comparación entre diferentes algoritmos implementados en diferentes ordenadores puede resultar injusta. Es lógico asumir que cuanto más potente sea el ordenador con el que se realizan las pruebas, menos tiempo requerirá la ejecución de una meta-heurística. Por esta razón, una buena medida para comparar diferentes algoritmos es la de mostrar el número de iteraciones necesitadas para obtener la solución resultante, o dicho de otra manera, el comportamiento de convergencia de la meta-heurística. Este valor puede variar dependiendo de la técnica que se esté utilizando. Por ejemplo, para una búsqueda tabú o un recocido simulado, este valor podría ser la cantidad de iteraciones necesarias para obtener la solución. Para un GA, por el contrario, este valor podría ser el número de generaciones ejecutadas. Otra alternativa igualmente adecuada sería la de plasmar la cantidad de evaluaciones que se han realizado a lo largo de la ejecución. Pese a que esto se considera una buena práctica, hoy en día son pocos los estudios que muestran este parámetro, \cite{iterations1} and \cite{iterations2} son dos ejemplos de esta buena práctica.

Otro factor que aumenta la riqueza y replicabilidad de los resultados de un estudio es mostrar un completo set de resultados, con datos como la media, el mejor y peor resultado o la desviación típica. Los resultados pueden ser mostrados de diferentes maneras, como la desviación en porcentajes respecto al óptimo, o el coste de la solución aplicada a la función objetivo. Como se menciona en \cite{VRPTW3}, mostrar tan solo la mejor solución obtenida por una heurística, algo que sucede en muchas ocasiones en la literatura (\cite{NoAVG,noTimes}), puede crear una falsa imagen sobre su calidad. Esta afirmación puede ser perfectamente extrapolable a las meta-heurísticas, y puede ser considerado como una mala práctica. La base para una comparación justa entre diferentes meta-heurísticas se fundamenta en el resultado medio obtenido por múltiples ejecuciones, acompañado por otro parámetros relacionados, como por ejemplo, la varianza o la desviación típica.

Para terminar con este apartado de buenas prácticas es conveniente aconsejar el uso de pruebas estadísticas a la hora de comparar resultados obtenidos por diferentes técnicas. Esta tendencia es considerada una buena práctica para lograr realizar comparaciones estrictas y objetivas. Algunas de estas pruebas podrían ser la famosa prueba $t$ de Student, o el conocido test de Friedman.


\section{Codificaciones empleadas para la representación de soluciones}
\label{sec:problemasexp}

Un aspecto importante a la hora de abordar un problema de optimización es la codificación empleada para representar sus soluciones, tanto parciales como finales. Dependiendo de la manera en que se codifiquen estas soluciones podrán hacerse uso de distintos operadores, ya sea de cruce, mutación, generadores de sucesores, o de otro tipo que implique el tratamiento o modificación de los mismos.

En esta sección se van a definir las diferentes codificaciones empleadas para representar las soluciones de los seis problemas utilizados en la experimentación. 

\begin{sloppypar}
En primer lugar, en relación a las soluciones para los problemas TSP y ATSP, se ha hecho uso de codificación basada en la permutación de elementos, o \textit{path encoding}, tal como se recoge en el estudio sobre AG y el TSP de \cite{TSP2}. De acuerdo a esta representación, cada solución es representada por una permutación de números, la cual simboliza el orden en el que los clientes tienen que ser visitados. Por ejemplo, en el caso de la instancia compuesta por 10 nodos que se ha podido observar en la figura \ref{fig:TSP}, correspondiente a la sección \ref{sec:TSP}, la solución propuesta podría ser codificada como $X=(0, 5, 2, 4, 3, 1, 6, 8, 9, 7)$, y su distancia total se calcularía de la siguiente forma:  $f(X)$ = $d_{05}$+$d_{52}$+$d_{24}$+$d_{43}$+ $d_{31}$+$d_{16}$+$d_{68}$+$d_{89}$+$d_{97}$+$d_{70}$. Pese a que existen otros tipos de codificación, ésta es la que más se utiliza en la literatura.
\end{sloppypar}

\begin{figure}
	\centering
		\includegraphics[width=1.0\textwidth]{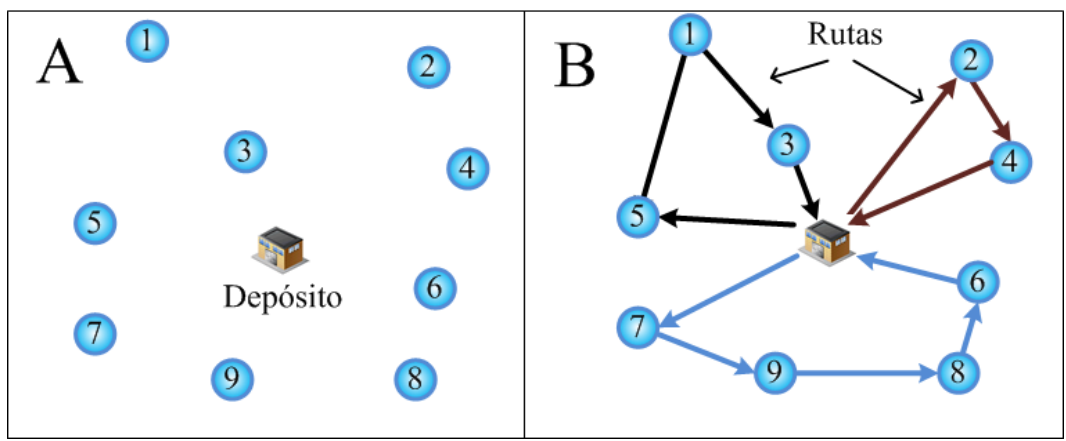}
		\caption{Ejemplo de una pequeña instancia del CVRP y posible solución}
	\label{fig:CVRP}
\end{figure}

Por otro lado, en lo relacionado al CVRP, al igual que en el caso anterior una de las codificaciones más utilizadas para la representación de soluciones es la permutación de elementos \cite{CVRP4}, la cual ha sido empleada en este estudio. En esta codificación las rutas son también presentadas como permutaciones de nodos, los cuales representan los clientes que componen la ruta, organizados por orden de visita. Además de esto, para distinguir las diferentes rutas de una solución, se utiliza el carácter ``0". Con todo esto, la solución propuesta en la figura \ref{fig:CVRP}.B para la instancia presentada en \ref{fig:CVRP}.A se codificaría como $X=(5,1,3,\textbf{0},2,4,\textbf{0},7,9,8,6)$, y su función objetivo se calcularía de la siguiente forma: $f(X)$ = $d_{05}$+$d_{51}$+$d_{13}$+$d_{30}$+ $d_{02}$+$d_{24}$+ $d_{40}$+$d_{07}$+$d_{79}$+$d_{98}$+$d_{86}$+$d_{60}$. Este tipo de representación ha sido frecuentemente utilizada en la literatura en estudios relacionados con el VRP \cite{CVRP5}.

En lo relativo al problema VRPB, repasado en el apartado \ref{sec:VRPB}, el conjunto de clientes V$\backslash \{0\}$ puede separarse en dos subconjuntos \cite{VRPB6}. El primero de ellos, $L$, llamado \textit{Linehaul customers}, son aquellos consumidores que demandan la entrega de mercancías. Por otro lado, el segundo subconjunto, $B$, llamado \textit{Backhaul customers}, son los que demandan la recogida de cierta cantidad de materiales. De esta forma, y como las entregas son servidas antes que las recogidas, los clientes podrían listarse de la siguiente forma: $L={1,\dots,n}$ y $B={n+1,\dots,n+m}$.

Para este problema, al igual que se ha visto en casos anteriores, una de las codificaciones más empleadas para la representación de soluciones es la permutación de elementos. En este caso, la manera de codificar un individuo se realizará de igual manera que para el CVRP, con la excepción de que hay que tener en cuenta los distintos tipos de clientes.

\begin{figure}[tb]
	\centering
		\includegraphics[width=0.6\textwidth]{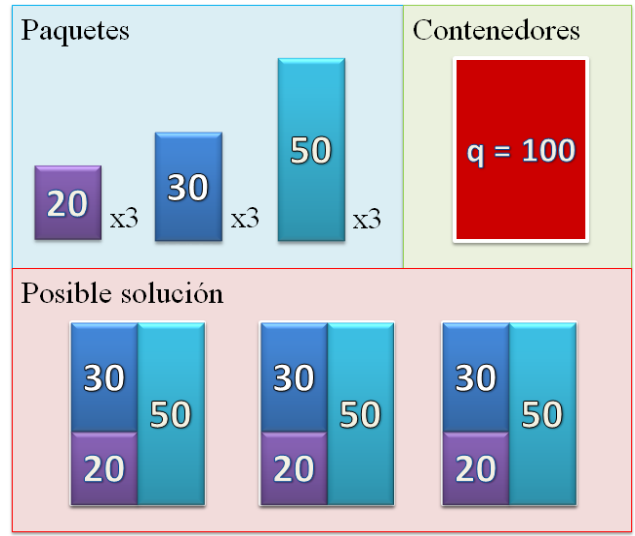}
	\caption{Ejemplo de una instancia de BPP y posible solución}
	\label{fig:BPP}
\end{figure}

Por otro lado, para el problema NQP se ha hecho uso de la representación descrita en la sección \ref{sec:NQP}. Finalmente, un método sencillo para codificar las soluciones del problema BPP es mediante la permutación de ítems. Para contabilizar el número de recipientes utilizados en una solución concreta, los tamaños de los paquetes son acumulados en una variable ($sumSize$). Cuando $sumSize$ excede el valor de $q$, el número de contenedores se incrementa en 1, y $sumSize$ es reseteado a 0. Con todo esto, suponiendo una instancia de 9 paquetes $I=\{i_1, i_2, \dots , i_{9}\}$, con tres diferentes tamaños $s_{1-3} = 20$, $s_{4-6} = 30$, y $s_{7-9} = 50$, y una capacidad máxima de $q$=100, una posible solución al problema podría ser $X=(i_1,i_4,i_7,i_2,i_5,i_8,i_3,i_6,i_9)$, cuyo número de contenedores utilizados sería 3. Este ejemplo es representado en la figura \ref{fig:BPP}.

\section{Configuración de las técnicas usadas}
\label{sec:parametrizacion}

En esta sección se detallará la parametrización adoptada por cada una de las técnicas empleadas para la experimentación llevada a cabo en esta tesis. Como ya se ha mencionado con anterioridad, los resultados obtenidos por el GB serán comparados con los logrados por dos algoritmos genéticos uni-poblacionales básicos ($GA_1$ y $GA_2$) y dos algoritmos genéticos distribuidos ($DGA_1$ and $DGA_2$). Las características principales de cada una de las meta-heurísticas implementadas son las siguiente:

\begin{itemize}

	\item $GA_1$: \textit{Algoritmo genético clásico con parámetros convencionales}. Algoritmo genético clásico con la estructura básica, previamente descrita en el Algoritmo \ref{alg:GA} perteneciente a la Sección \ref{sec:genetico}, con operadores convencionales y parámetros convencionales, es decir, probabilidad alta de cruce, y probabilidad baja de mutación. Esta parametrización es la más extendida en la literatura, y su filosofía ha sido descrita en una gran cantidad de trabajos \cite{DGAstudy,CXandMut2,CXandMut3}.
	
	\item $GA_2$: \textit{Algoritmo genético clásico con parámetros ajustados al GB}. La estructura adoptada por esta técnica es la misma que la empleada para el $GA_1$. En este caso, los operadores y los parámetros se han ajustado a los utilizados por el GB. Con todo esto, el número de movimientos individuales (mutación y entrenamiento individual) y movimientos cooperativos (cruce y entrenamiento personalizado) es el mismo para ambas meta-heurísticas. Además de esto, las funciones utilizadas son las mismas en ambos casos, de manera que la función de mutación es una de las implementadas para el GB como método de entrenamiento convencional, mientras que la función de cruce es la misma que se hace uso en el GB como entrenamiento personalizado. Para realizar este ajuste las probabilidades de cruce y mutación han sido adaptadas. Esta técnica se utiliza con la intención de realizar una comparación fiable, en la que la diferencia entre los algoritmos comparados tan solo resida en su estructura, y no en otros factores. De esta forma, podrá deducirse qué meta-heurística obtiene los mejores resultados haciendo uso de las mismas funciones el mismo número de veces.
	
	\item $DGA_1$: \textit{Algoritmo genético distribuido con parámetros convencionales}. Se trata de un algoritmo genético multi-poblacional en el que las distintas subpoblaciones evolucionan independientemente intercambiando individuos entre sí de manera ocasional. Esta técnica sigue la filosofía de los PGAs basados en islas, explicada en la sección \ref{sec:PGA}. En cada una de las subpoblaciones del $DGA_1$ se ejecuta en paralelo un $GA_1$ diferente, cada cual con su propia probabilidad de cruce y mutación. La topología implementada para esta técnica es dinámica, lo que quiere decir que una subpoblación no está restringida a comunicarse con un conjunto fijo de subpoblaciones. De esta forma, cuando una población mejora la mejor solución encontrada hasta el momento, esta es compartida con el resto de poblaciones sustituyendo el peor individuo de cada una de ellas. 
			
	\item $DGA_2$: \textit{Algoritmo genético distribuido con parámetros ajustados al GB}. Este último método es un híbrido entre el $DGA_1$ y el $GA_2$, ya que su estructura es exactamente igual a la del $DGA_1$, mientras que en cada subpoblación se ejecuta un $GA_2$. En este caso, cada población tiene su propia función de mutación, de esta manera, las similitudes con el GB, en términos de operadores y parámetros, som más fiables.
		
\end{itemize}

El tamaño de la población total para todas las meta-heurísticas se ha establecido en 48 individuos, los cuales son generados de forma aleatoria. Para el $DGA_1$ y el $DGA_2$, esta población es dividida en 4 diferentes subpoblaciones, compuestas cada una de ellas por 12 individuos. Este esquema se repite también para el GB, donde los 48 jugadores son divididos en 4 equipos de igual tamaño. En relación a la probabilidad de cruce ($p_c$) y de mutación ($p_m$), para el $GA_1$ estas son del 95\% y 5\%, respectivamente. Por otro lado, los valores de $p_c$ empleados para las diferentes subpoblaciones del $DGA_1$ han sido 95\%, 90\%, 80\%, y 75\%, mientras que los valores de $p_m$ se han establecido en 5\%, 10\%, 20\%, y 25\%, respectivamente. Finalmente, tanto para el $GA_2$ como para el $DGA_2$ se ha utilizado un valor de 0.03\% para $p_c$ y 100\% para $p_m$, con la intención de adaptarse a los parámetros del GB. 

En lo referente al criterio de selección de reproductores, la técnica del torneo binario ha sido empleada en todos los casos ($GA_1$, $GA_2$, $DGA_1$ y $DGA_2$). Por otro lado, una función 100\% elitista ha sido empleada como función de supervivencia para $GA_2$ y $DGA_2$, mientras que una función 50\% elitista- 50\% aleatoria se ha implementado para $GA_1$ y $DGA_1$. Finalmente, la ejecución de todos estos algoritmos finaliza cuando se cumplen $n+\sum_{k=1}^n{k}$ generaciones sin encontrar una solución que supere a la mejor solución encontrada hasta el momento, donde $n$ es el tamaño del problema. Este tamaño es igual al número de clientes en el caso de los problemas de asignación de rutas a vehículos, mientras que para el NQP es el número de reinas, y para el BPP el número de ítems a almacenar.

Las funciones utilizadas como métodos de entrenamiento convencionales para el GB para los problemas TSP, ATSP, NQP y BPP son los anteriormente vistos en la sección \ref{sec:gensuc} 2-opt, 3-opt, Vertex Insertion y Swapping. Estas funciones de sucesores han sido utilizadas tambien como funciones de mutación en el caso del $DGA_2$ (una función para cada población). Por otro lado, para el $GA_1$, el $GA_2$, y el $DGA_1$ la función utilizada para este próposito ha sido la 2-opt, ya que es la que obtiene mejores resultados.

El operador utilizado por el GB como función de entrenamiento personalizado para los mismos problemas es el HX, visto en la sección \ref{sec:crossover}. Esta función ha sido también utilizada en los algoritmos $GA_2$ y $DGA_2$ como función de cruce, mientras que para $GA_1$ y $DGA_1$ el OX ha sido implementado con esta finalidad.  

Finalmente, la tabla \ref{tab:summary1} ofrece un resumen de las características específicas de ambos $GA$s y $DGA$s.

\begin{table}[tb]	
	\centering
	\renewcommand{\arraystretch}{1}
	\scalebox{0.80}{
		\begin{tabular}{| m{1.2cm} || m{2.6cm} | m{2.6cm} | m{2.5cm} | m{1.5cm} | m{2.3cm} |}
		  \hline Alg. & Población & $p_c$ y $p_m$ & Superv. & Cruce & Mutación\\ 
			\hline\hline $GA_1$ &  48 individuos, 1 población & 80\% y 20\% & 50\% elitista - 50\% aleatoria & OX & 2-opt\\
			\hline $GA_2$ &  48 individuos, 1 población & 0.03\% y 100\% & 100\% elitista & HX & 2-opt\\
			\hline $DGA_1$ & 48 individuos, 4 poblaciones de 12 individuos & En este orden: 95\% y 5\%, 90\% y 10\%, 75\% y 25\%, 80\% y 20\% & 50\% elitista - 50\% aleatoria & OX & 2-opt\\
			\hline $DGA_2$ & 48 individuos, 4 poblaciones de 12 individuos & 0.03\% y 100\% & 100\% elitista & HX & 2-opt, 3-opt, Swapping \& Insertion \\
			\hline
		\end{tabular}
		}
	\caption{Resumen de las características de $GA_1$, $GA_2$, $DGA_1$ y $DGA_2$ para el TSP, ATSP, NQP, BPP}
	\label{tab:summary1}
\end{table}

En relación a los problemas CVRP y VRPB, los operadores intra-ruta 2-opt y Vertex Insertion han sido utilizados como funciones de entrenamiento convencional por el GB, junto a la versión inter-ruta del Vertex Insertion y Swapping. El método de trabajo de estos operadores puede consultarse en la sección \ref{sec:gensuc}. Siguiendo la misma conducta que en los problemas anteriores, estos operadores han sido utilizados como funciones de mutación para el $DGA_2$ (una función para cada población). Por otro lado, el operador inter-ruta Vertex Insertion ha sido empleado con el mismo objetivo en el $GA_1$, el $GA_2$, y el $DGA1$. 

Por otra parte, la función SRX ha sido implementada como función de entrenamiento personalizado para el GB. Esta función ha sido empleada a su vez en el resto de algoritmos desarrollados. Finalmente, las parametrizaciones concretas de los cuatro algoritmos que se compararán con el GB pueden verse sintetizadas en la tabla \ref{tab:summary2}, mientras que las características del GB pueden observarse resumidas en la tabla \ref{tab:summary3}.

\begin{table}[tb]	
	\centering
	\renewcommand{\arraystretch}{1}
	\scalebox{0.8}{
		\begin{tabular}{| m{1.2cm} || m{2.6cm} | m{2.6cm} | m{2.5cm} | m{1.5cm} | m{2.6cm} |}
		  \hline Alg. & Población & $p_c$ y $p_m$ & Superv. & Cruce & Mutación\\  
			\hline\hline $GA_1$ &  48 individuos, 1 población & 80\% y 20\% & 50\% elitista - 50\% aleatorio & SRX & Vertex Insertion \\
			\hline $GA_2$ &  48 individuos, 1 población & 0.03\% y 100\% & 100\% elitista & SRX & Vertex Insertion \\
			\hline $DGA_1$ & 48 individuos, 4 poblaciones de 12 individuos & En este orden: 95\% y 5\%, 90\% y 10\%, 75\% y 25\%, 80\% y 20\% & 50\% elitista - 50\% aleatorio & SRX & Vertex Insertion \\
			\hline $DGA_2$ & 48 individuos, 4 poblaciones de 12 individuos & 0.03\% y 100\% & 100\% elitista & SRX & 2-opt, Vertex Insertion (intra-ruta e inter-ruta) y Swapping \\
			\hline
		\end{tabular}
		}
	\caption{Resumen de las características de $GA_1$, $GA_2$, $DGA_1$ y $DGA_2$ para el CVRP y VRPB}
	\label{tab:summary2}
\end{table}

\begin{table}[tb]	
	\centering
	\renewcommand{\arraystretch}{1}
	\scalebox{0.8}{
		\begin{tabular}{| m{8cm}| m{7cm} |}
		  \hline 
			Número de equipos (TN) & 4\\[3pt]
			\hline 
			Número de jugadores por equipo (PT) & 12\\[3pt]
			\hline
			Número de entrenamientos sin mejora para entrenamiento personalizado & 6\\[3pt]
			\hline 
			Número de entrenamientos sin mejora para transferencia especial & 12\\[3pt]
			\hline
			Función de entrenamiento personalizado para el TSP, ATSP, NQP y BPP & HX\\[3pt]
			\hline 
			Función de entrenamiento personalizado para el CVRP y el VRPB & HRX\\[3pt]
			\hline
			Función de entrenamiento convencional para el TSP, ATSP, NQP y BPP & 2-opt, 3-opt, Swapping, y Vertex Insertion \\[3pt]
			\hline 
			Función de entrenamiento convencional para el CVRP y VRPB & 2-opt, Vertex Insertion (intra-ruta e inter-ruta) y Swapping \\[3pt]
			\hline
		\end{tabular}
		}
	\caption{Resumen de las características del GB}
	\label{tab:summary3}
\end{table}


\section{Resultados y pruebas estadísticos}
\label{sec:resultados}

En esta sección se presentarán los resultados obtenidos por el GB y por el resto de técnicas implementadas para los problemas anteriormente descritos. Además de esto, también se realizarán dos prueba estadísticas, el test normal $z$ y el test de Friedman, con el objetivo de proporcionar una comparación justa y rigurosa. Antes de todo esto, conviene presentar en un primer momento los aspectos básicos de esta experimentación.   

Para empezar, es interesante mencionar que toda la experimentación ha sido ejecutada en un ordenador Intel Core i7 3930, con 3.20 GHz y una memoria RAM de 16 GB. Se ha utilizado el sistema Microsoft Windows 7, y todos los algoritmos han sido implementados utilizando el lenguaje de programación Java. 

Para el TSP se han empleado 18 instancias diferentes, mientras que para el ATSP se han hecho uso de 19, todas ellas obtenidas del conocido banco de pruebas TSPLib \cite{TSPLibrary}. Por otro lado, se han utilizado 11 instancias del problema CVRP, pertenecientes al famoso benchmark de Christofides/Elion, el cual está disponible en la VRPWeb\footnote{\textit{http://neo.lcc.uma.es/vrp}}. El nombre de cada uno de los casos denota la cantidad de nodos existentes, de esta manera, la instancia Berlin52, por ejemplo, se compone de 52 de nodos.

Además de estos, el número de ejemplos empleados para el VRPB ha sido 12. Estos 12 casos han sido creados por el autor de este trabajo, y están disponibles en su página personal\footnote{\textit{http://paginaspersonales.deusto.es/e.osaba}}. Las primeras 6 instancias pertenecen al benchmark VRPTW de Solomon\footnote{\textit{http://w.cba.neu.edu/msolomon/problems.htm}}, con la particularidad de haber sido eliminadas las restricciones de tiempo y haber modificado el tipo de las demandas. Las instancias restantes, por otro lado, han sido obtenidas del banco de pruebas Christofides/Elion arriba mencionado, habiendo modificado también para este caso los tipos de las demandas. Al tratarse de un banco de pruebas adaptado a esta experimentación, las soluciones óptimas de las instancias no son conocidas. Este benchmark ha sido utilizado por el autor de este trabajo en varias publicaciones, ya sea para congresos internacionales, como para revistas científicas \cite{osaba2013GB,osaba2014GB2}.

Respecto al NQP, se han utilizado 15 casos diferentes. El nombre de cada uno de ellos describe el número de reinas a tener en cuenta y el tamaño del tablero. Es decir, la instancia 20-Queens consiste en colocar 20 reinas diferentes en un tablero de 20x20. Como es lógico, las soluciones óptimas para cada uno de los casos es 0, por esta razón este valor no es mostrado en las tablas posteriores.
\begin{sloppypar}
Finalmente, se han empleado 16 instancias para el BPP, obtenidas todas ellas del famoso benchmark Scholl/Klein\footnote{\textit{http://www.wiwi.uni-jena.de/entscheidung/binpp/index.htm.}}. Estos ejemplos son nombrados como $NxCyWz\_a$, donde $x$ es 1 (50 ítems), 2 (100 ítems), 3 (200 ítems) o 4 (500 ítems); $y$ es 1 (capacidad de 100), 2 (capacidad de 120) o 3 (capacidad de 150); $z$ es 1 (tamaño de los paquetes entre 1 y 100) o 2 (tamaño de los paquetes entre 20 y 100); y $a$ es A,B o C a modo de parámetro de indexación dentro del propio benchmark.
\end{sloppypar}

Una vez explicados estos conceptos, permítase un breve paréntesis para llevar a cabo un pequeño estudio acerca de la parametrización del algoritmo GB. Es conveniente tener en cuenta que realizar un estudio exhaustivo de este tipo sería un trabajo arduo y extenso. Este tipo de investigación ha sido considerada como trabajo futuro, ya que el objetivo de esta tesis es el de presentar la nueva técnica propuesta y demostrar que es una alternativa prometedora para resolver problemas de asignación de rutas a vehículos. Con todo esto, a continuación se muestra una pequeña experimentación mediante la que se justifica la elección del número de equipos (4) utilizados en el GB en la comparativa con las restantes meta-heurísticas poblacionales. Para llevar a cabo esta experimentación se han comparado 4 diferentes versiones del GB, todas ellas compuestas por una población inicial confeccionada por 48 jugadores, y en las que la única diferencia es la distribución de esta población inicial. Esta distribución es la siguiente: 

\begin{itemize}
	\item Versión 1: 2 equipos compuestos por 24 jugadores.
	\item Versión 2: 4 equipos compuestos por 12 jugadores.
	\item Versión 3: 6 equipos compuestos por 8 jugadores.
	\item Versión 4: 8 equipos compuestos por 6 jugadores.
\end{itemize}

Este estudio ha sido llevado a cabo exclusivamente con el TSP, y las características adoptadas por las cuatro versiones del GB son las mismas que se han presentado en la tabla \ref{tab:summary3} (a excepción de la población). Los resultados obtenidos por las cuatro alternativas pueden verse en la tabla \ref{tab:GBVersions}. Debido a que es una comparación entre diferentes versiones del GB, se ha decidido exponer únicamente los resultados promedio y las medias aritméticas de los tiempos de ejecución (en segundos). Las instancias empleadas han sido las mismas que se han introducido al inicio de esta sección. Finalmente, el número de ejecuciones para cada una de ellas ha sido de 40.

\begin{table}[tb]
	\centering
	\scalebox{0.75}{
		\begin{tabular}{| l || l l || l l || l l || l l |}
		  \hline  
		  \multicolumn{1}{|c||}{Instancia} & \multicolumn{2}{c||}{Versión 1} & \multicolumn{2}{c||}{Versión 2} & \multicolumn{2}{c||}{Versión 3} & \multicolumn{2}{c|}{Version 4}\\
			\hline Nombre & Media & T. & Media & T. & Media & T. & Media & T.\\ 
			\hline Oliver30 & 420.30 & 0.12 & \textbf{420.00} & 0.18 & \textbf{420.00} & 0.32 & \textbf{420.00} & 0.47 \\
			Eilon50 & 431.30 & 0.46 & \textbf{427.00} & 0.85 & 427.40 & 1.23 & 427.40 & 1.64\\
			Eil51 & 433.30 & 0.54 & 428.50 & 0.91 & 429.20 & 1.43 & \textbf{427.90} & 1.76\\
			Berlin52 & 7599.80 & 0.75 & \textbf{7542.00} & 1.25 & \textbf{7542.00} & 2.18 & \textbf{7542.00} & 2.84\\
			St70 & 685.85 & 1.55 & 679.45 & 2.09 & 678.20 & 4.08 & \textbf{678.00} & 5.75\\
			Eilon75 & 553.55 & 1.63 & 544.35 & 3.37 & \textbf{541.50} & 5.23 & 541.70 & 6.53\\
			Eil76 & 553.10 & 2.09 & 545.30 & 3.85 & 545.60 & 5.52 & \textbf{544.45} & 6.99 \\
			KroA100 & 21549.60 & 5.54 & 21386.70 & 8.12 & \textbf{21318.70} & 13.98 & 21325.70 & 17.76 \\
			KroB100 & 22729.20 & 4.58 & 22311.05 & 7.51 & 22337.50 & 11.95 & \textbf{22284.7}0 & 21.55 \\
			KroC100 & 21055.30 & 5.76 & 20968.25 & 8.05 & 20846.60 & 16.84 & \textbf{20840.25} & 18.14\\
			KroD100 & 21602.00 & 6.44 & 21485.80 & 7.75 & \textbf{21481.60} & 13.93 & 21510.60 & 17.48 \\
			KroE100 & 22385.00 & 6.78 & 22266.80 & 7.95 & 22211.90 & 16.79 & \textbf{22157.80} & 19.08 \\
			Eil101 & 648.70 & 6.08 & 643.70 & 7.97 & 641.20 & 12.76 & \textbf{640.55} & 19.68 \\
			Pr107 & 45049.00 & 6.46 & 44693.00 & 9.45 & \textbf{44492.90} & 16.18 & 44561.85 & 24.26\\
			Pr124 & 59664.10 & 10.41 & 59348.20 & 16.67 & 59402.10 & 21.45 & \textbf{59288.10} & 36.25\\
			Pr136 & 99215.45 & 18.33 & 98906.50 & 24.25 & 98356.85 & 32.53 & \textbf{98155.80} & 65.36 \\
			Pr144 & 59120.40 & 19.92 & 58712.00 & 29.76 & 58698.50 & 55.63 & \textbf{58656.55} & 73.54 \\
			Pr152 & 74952.20 & 21.56 & 74320.70 & 32.45 & 74275.35 & 59.30 & \textbf{74190.00} & 68.25\\
			\hline
		\end{tabular}
	}
	\caption{Resultados obtenidos por las cuatro versiones del GB. Por cada instancia se muestran los resultados medios y los tiempos de ejecución medios.}
	\label{tab:GBVersions}
\end{table}

Varias conclusiones pueden extraerse analizando los datos presentados en la tabla \ref{tab:GBVersions}. De inicio, puede observarse una ligera tendencia de mejora a medida que el número de equipos incrementa. Esta mejora puede explicarse de manera simple: cuanto mayor sea el número de subpoblaciones, mayor será la capacidad de explotación y exploración de la técnica. Esto es así ya que el número de sesiones de entrenamiento y de interacciones entre los equipos aumenta. 

Pese a esto, este hecho envuelve un incremento significativo en los tiempos de ejecución, el cual no es directamente proporcional con la mejora de los resultados. En este trabajo se ha optado por escoger la opción que mayor equilibrio demuestra respecto a calidad de resultados y tiempos de ejecución. La versión que mejor encaja con este propósito es la número 2. Esta opción ofrece unos tiempos de ejecución aceptables, y es la alternativa que muestra una mayor capacidad de mejora respecto a su versión anterior (versión 1). Las opciones 3 y 4, por otro lado, exigen un tiempo computacional muy elevado respecto a la alternativa 2, sin llegar a ofrecer mejoras destacables en la calidad de los resultados.


Una vez finalizado este pequeño paréntesis, a continuación se mostrarán los resultados obtenidos por cada técnica para los 6 problemas presentados. En este caso cada instancia también ha sido ejecutada un total de 40 ocasiones. Con esto, los resultados mostrados para cada instancia son los siguientes: media aritmética, desviación típica, mejor resultado y tiempo promedio por ejecución (en segundos). Además de esto, se realizará un pequeño estudio en relación al comportamiento de convergencia del GB, realizando una comparación con el $DGA_2$ para los problemas TSP y CVRP. Esta investigación se mostrará en la sección \ref{sec:analisis}

\begin{figure}[tb]
	\centering
		\includegraphics[width=1.0\textwidth]{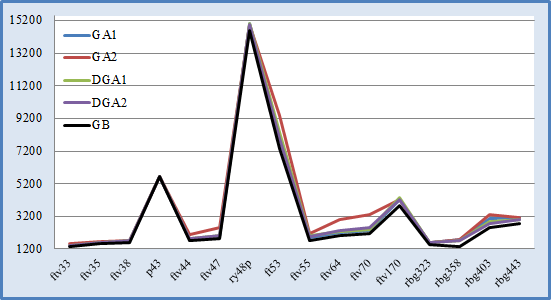}
		\caption{Resumen de los resultados promedio obtenidos para el ATSP. Conviene mencionar que, con la intención de facilitar la visualización del gráfico, 3 instancias (br17, ft70 y pro124p) han quedado fuera debido a la diferencia de los resultados promedio}
	\label{fig:ATSPgraf}
\end{figure}

\begin{figure}[tb]
	\centering
		\includegraphics[width=1.0\textwidth]{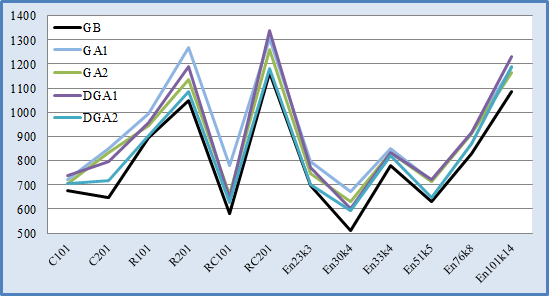}
		\caption{Resumen de los resultados promedio obtenidos para el VRPB.}
	\label{fig:VRPBgraf}
\end{figure}

\begin{figure}[tb]
	\centering
		\includegraphics[width=1.0\textwidth]{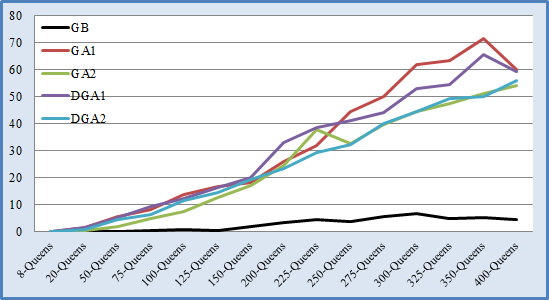}
		\caption{Resumen de los resultados promedio obtenidos para el NQP.}
	\label{fig:NQPgraf}
\end{figure}

\begin{figure}[tb]
	\centering
		\includegraphics[width=1.0\textwidth]{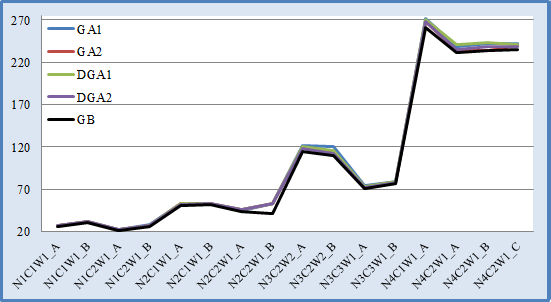}
		\caption{Resumen de los resultados promedio obtenidos para el BPP.}
	\label{fig:BPPgraf}
\end{figure}

Es conveniente aclarar que en esta sección, con la intención de facilitar la lectura del documento, tan sólo se mostrarán los resultados completos obtenidos para los problemas TSP y CVRP. Estos pueden encontrarse en las tablas \ref{tab:TSP1} y \ref{tab:TSP2} y la tabla \ref{tab:CVRP}. Para representar en este apartado los resultados logrados para el resto de problemas se hará uso de diferentes gráficas resumen (Figuras \ref{fig:ATSPgraf}, \ref{fig:VRPBgraf}, \ref{fig:NQPgraf} y \ref{fig:BPPgraf}). Estas gráficas sintetizan la información mostrando una comparación de los resultados promedio de cada técnica para cada instancia del problema correspondiente. Finalmente, los resultados completos del ATSP, VRPB, NQP o BPP se encuentran en el apéndice de este trabajo (Apéndice \ref{app:resultados}).

\begin{table}[tb]
	\centering
	\setlength{\tabcolsep}{2.5pt}
	\scalebox{0.7}{
		\begin{tabular}{| l || l l || l l | l l || l l | l l |}
		  \hline  
		  \multicolumn{1}{|c||}{Instancia} & \multicolumn{2}{c||}{Golden Ball} & \multicolumn{2}{c|}{$GA_1$} & \multicolumn{2}{c||}{$GA_2$} & \multicolumn{2}{c|}{$DGA_1$} &\multicolumn{2}{c|}{$DGA_2$}\\
			\hline\hline Oliver30 (420) & \textbf{420.0} & ($\pm$0.0) & 425.9 & ($\pm$7.1) & 426.0 & ($\pm$9.9) & 428.0 & ($\pm$4.8) & 424.1 & ($\pm$6.8)\\
			$\dot{x}$ \& T & 420 & 0.2 & 420 & 0.3 & 420 & 0.2 & 420 & 0.3 & 420 & 0.2\\
			\hline Eilon50 (425) & \textbf{427.0} & ($\pm$1.5) & 451.5 & ($\pm$15.1) & 442.5 & ($\pm$6.5) & 442.1 & ($\pm$7.4) & 435.8 & ($\pm$5.8)\\
			$\dot{x}$ \& T & 425 & 1.1 & 435 & 1.3 & 432 & 1.5 &  433 & 0.9 & 429 & 0.8\\
			\hline Eil51 (426) & \textbf{428.6} & ($\pm$1.3) & 453.6 & ($\pm$10.5) & 444.8 & ($\pm$9.8) & 442.5 & ($\pm$7.9) & 438.4 & ($\pm$4.8)\\
			$\dot{x}$ \& T & 427 & 1.4 & 439 & 1.8 & 434 & 1.8 & 432 & 1.1 & 430 & 1.3 \\
			\hline Berlin52 (7542) & \textbf{7542.0} & ($\pm$0.0) & 7945.6 & ($\pm$168.6) & 7841.3 & ($\pm$256.7) & 7914.3 & ($\pm$186.9) & 7866.0 & ($\pm$296.4)\\
			$\dot{x}$ \& T & 7542 & 2.1 & 7542 & 1.4 & 7542  & 1.8 & 7542  & 1.2 &  7542 & 1.9\\
			\hline St70 (675) & \textbf{679.4} & ($\pm$3.5) & 711.8 & ($\pm$33.6) & 716.4 & ($\pm$44.3) & 719.6 & ($\pm$18.8) & 699.0 & ($\pm$11.9)\\
			$\dot{x}$ \& T & 675 & 4.2 & 682 & 5.2 & 684 & 4.2 & 705 & 3.8 & 683 & 3.9 \\
			\hline Eilon75 (535) & \textbf{544.3} & ($\pm$3.3) & 582.0 & ($\pm$14.3) & 565.6 & ($\pm$12.0) & 570.4 & ($\pm$10.6) & 557.1 & ($\pm$8.9)\\
			$\dot{x}$ \& T & 536 & 5.4 & 570 & 5.8 & 550 & 5.5 & 556 & 5.1 & 544 & 4.5 \\
			\hline Eil76 (538) & \textbf{545.3} & ($\pm$3.7) & 582.8 & ($\pm$15.0) & 569.7 & ($\pm$11.5) & 574.8 & ($\pm$15.1) & 563.5 & ($\pm$6.4)\\
			$\dot{x}$ \& T & 539 & 5.5 & 560 & 6.0 & 545 & 5.7 & 556 & 5.1 & 552 & 5.1 \\
			\hline Eil101 (629) & \textbf{643.7} & ($\pm$4.3) & 696.0 & ($\pm$16.8) & 676.6 & ($\pm$11.2) & 678.7 & ($\pm$13.7) & 665.6 & ($\pm$10.3)\\
			$\dot{x}$ \& T & 636 & 8.9 & 676 & 17.2 & 657 & 10.7 & 657 & 12.9 & 643 & 8.4 \\
			\hline
		\end{tabular}
	}
	\caption{Primera parte de los resultados de los algoritmos $GB$, $GA_1$, $GA_2$, $DGA_1$ y $DGA_2$ para el TSP. Para cada instancia se muestra el resultado promedio, desviación típica, mejor resultado obtenido y tiempo de ejecución promedio.}
	\label{tab:TSP1}
\end{table}

\begin{table}[tb]
	\centering
	\setlength{\tabcolsep}{2.5pt}
	\scalebox{0.63}{
		\begin{tabular}{| l || l l || l l | l l || l l | l l |}
		  \hline  
		  \multicolumn{1}{|c||}{Instancia} & \multicolumn{2}{c||}{Golden Ball} & \multicolumn{2}{c|}{$GA_1$} & \multicolumn{2}{c||}{$GA_2$} & \multicolumn{2}{c|}{$DGA_1$} &\multicolumn{2}{c|}{$DGA_2$}\\
			\hline\hline KroA100 (21282) & \textbf{21386.7} & ($\pm$99.7) & 22559.3 & ($\pm$538.4) & 21838.8 & ($\pm$419.1) & 22757.1 & ($\pm$433.6) & 21940.6 & ($\pm$313.1)\\
			$\dot{x}$ \& T & 21282 & 9.5 & 21679 & 12.4 & 21376 & 9.3 & 22206 & 13.7 & 21478 & 10.4 \\
			\hline KroB100 (22140) & \textbf{22311.0} & ($\pm$139.6) & 23342.4 & ($\pm$468.6) & 22896.3 & ($\pm$424.8) & 23323.1 & ($\pm$375.8) & 22815.3 & ($\pm$329.7)\\
			$\dot{x}$ \& T & 22140 & 9.7 & 22574 & 12.4 & 22178 & 10.5 & 22763 & 13.5 & 22264 & 10.9 \\
			\hline KroC100 (20749) & \textbf{20968.2} & ($\pm$111.3) & 22010.3 & ($\pm$607.1) & 21536.1 & ($\pm$396.1) & 22311.5 & ($\pm$582.6) & 21472.3 & ($\pm$321.9)\\
			$\dot{x}$ \& T & 20769 & 9.3 & 21348 &  12.8 & 20880 & 9.8 & 21454 & 13.6 & 21039 & 9.5 \\
			\hline KroD100 (21294) & \textbf{21485.8} & ($\pm$188.2) & 22642.2 & ($\pm$543.2) & 22205.6 & ($\pm$401.6) & 22719.0 & ($\pm$616.2) & 22065.0 & ($\pm$406.2)\\
			$\dot{x}$ \& T & 21294 & 9.7 & 21696 & 12.2 & 21495 & 9.9 & 22013 & 12.7 & 21459 & 10.8\\
			\hline KroE100 (22068) & \textbf{22266.8} & ($\pm$158.1) & 23228.3 & ($\pm$416.4) & 22752.7 & ($\pm$304.3) & 23062.8 & ($\pm$443.9) & 22819.0 & ($\pm$312.2)\\
			$\dot{x}$ \& T & 22068 & 9.8 & 22418 & 11.1 & 22147 & 9.5 & 22299 & 13.5 & 22819 & 10.4 \\
			\hline\hline Pr107 (44303) & \textbf{44693.0} & ($\pm$210.7) & 47356.1 & ($\pm$1210.1) & 45614.4 & ($\pm$1389.4) & 46533.1 & ($\pm$1507.9) & 45506.8 & ($\pm$1323.4)\\
			$\dot{x}$ \& T & 44391 & 10.1 & 45512  & 16.8 &  44387 & 10.6 & 44872 & 16.1 & 44438 & 12.4 \\
			\hline Pr124 (59030) & \textbf{59348.2} & ($\pm$190.3) & 60871.8 & ($\pm$694.0) & 59943.7 & ($\pm$544.7) & 61149.0 & ($\pm$888.2) & 60193.8 & ($\pm$569.4)\\
			$\dot{x}$ \& T & 59030 & 16.2 & 59953 & 24.5 & 59030 & 17.1 & 59490 & 30.2 & 59076 & 14.3 \\
			\hline Pr136 (96772) & \textbf{98906.5} & ($\pm$1296.2) & 102819.1 & ($\pm$1929.7) & 100610.5 & ($\pm$1230.7) & 102585.2 & ($\pm$3241.8) & 100949.0 & ($\pm$1706.6)\\
			$\dot{x}$ \& T & 97439 & 23.5 & 99468 & 38.4 & 98137 & 24.1 & 98973 & 40.9 & 98797 & 26.2 \\
			\hline Pr144 (58537) & \textbf{58712.0} & ($\pm$247.7) & 60715.4 & ($\pm$1753.4) & 60662.3 & ($\pm$2330.9) & 61447.3 & ($\pm$1763.7) & 59470.5 & ($\pm$641.1)\\
			$\dot{x}$ \& T & 58537 & 34.1 & 58922 & 53.6 & 58599 & 32.5 & 59143 & 58.3 & 58538 & 33.5\\
			\hline Pr152 (73682) & \textbf{74320.7} & ($\pm$390.3) & 76819.0 & ($\pm$2038.8) & 75699.1 & ($\pm$912.0) & 76563.5 & ($\pm$904.8) & 75663.9 & ($\pm$1253.3)\\
			$\dot{x}$ \& T & 73818 & 36.7 & 74268 & 68.1 & 74526 & 37.5 & 74613 & 80.0 & 74249 & 35.8\\
			\hline
		\end{tabular}
	}
	\caption{Segunda parte de los resultados de los algoritmos $GB$, $GA_1$, $GA_2$, $DGA_1$ y $DGA_2$ para el TSP. Para cada instancia se muestra el resultado promedio, desviación típica, mejor resultado obtenido y tiempo de ejecución promedio.}
	\label{tab:TSP2}
\end{table}

\begin{table}[tb]
	\centering
	\setlength{\tabcolsep}{1pt}
	\scalebox{0.7}{
		\begin{tabular}{| l || l l || l l | l l || l l | l l |}
		  \hline  
		  \multicolumn{1}{|c||}{Instancia} & \multicolumn{2}{c||}{Golden Ball} & \multicolumn{2}{c|}{$GA_1$} & \multicolumn{2}{c||}{$GA_2$} & \multicolumn{2}{c|}{$DGA_1$} &\multicolumn{2}{c|}{$DGA_2$}\\ 
			\hline En22k4 (375) & \textbf{376.0} & ($\pm$2.2) & 390.3 & ($\pm$13.4) & 410.0  & ($\pm$14.4) & 401.0 & ($\pm$15.3) & 393.5 & ($\pm$15.0)\\
			$\dot{x}$ \& T & 375 & 0.9 & 375 & 2.4 & 390 & 1.2 & 375 & 3.8 & 375 & 1.8\\
			
			\hline En23k3 (569) & 589.7 & ($\pm$17.7) & 625.1 & ($\pm$31.5) & 660.0 & ($\pm$27.8) & 655.1 & ($\pm$23.1) & \textbf{587.6} & ($\pm$23.5) \\
			$\dot{x}$ \& T & 569 & 0.7 & 569 & 3.8 & 602 & 1.1 & 601 & 3.1 & 569 & 1.8\\
			
			\hline En30k3 (503) & \textbf{517.4} & ($\pm$15.6) & 574.3 & ($\pm$28.4) & 597.4 & ($\pm$49.3) & 560.4 & ($\pm$49.5) & 577.7 & ($\pm$71.5) \\
			$\dot{x}$ \& T & 503 & 2.2 & 521 & 5.1 & 529 & 2.8 & 503 & 4.1 & 503 & 2.2\\
			
			\hline En33k4 (835) & \textbf{857.8} & ($\pm$9.6) & 917.0 & ($\pm$35.1) & 947.7 & ($\pm$26.5) & 921.8 & ($\pm$27.1) & 919.8 & ($\pm$23.4)\\
			$\dot{x}$ \& T & 844 & 2.8 & 862 & 7.6 & 902 & 2.8 & 888 & 6.1 & 862 & 2.3\\
			
			\hline En51k5 (521) & \textbf{578.1} & ($\pm$10.9) & 681.6 & ($\pm$51.4) & 677.9 & ($\pm$81.7) & 635.4 & ($\pm$31.6) & 624.4 & ($\pm$43.7) \\
			$\dot{x}$ \& T & 561 & 9.8 & 574 & 17.6 & 589 & 10.1 & 572 & 23.1 & 568 & 7.6\\
			
			\hline\hline En76k7 (682) & \textbf{755.8} & ($\pm$13.1) & 852.5 & ($\pm$47.5) & 849.9 & ($\pm$53.5) & 819.7 & ($\pm$31.6) & 799.9 & ($\pm$43.7) \\
			$\dot{x}$ \& T & 736 & 25.5 & 755 & 59.6 & 753 & 24.8 & 722 & 67.0 & 750 & 23.4\\
			
			\hline En76k8 (735) & \textbf{816.9} & ($\pm$14.8) & 923.2 & ($\pm$37.0) & 908.9 & ($\pm$38.4) & 883.3 & ($\pm$53.5) & 873.9 & ($\pm$43.7) \\
			$\dot{x}$ \& T & 795 & 31.5 & 851 & 64.6 & 859 & 32.5 & 801 & 69.2 & 808 & 28.5\\
			
			\hline En76k10 (830) & \textbf{913.6} & ($\pm$15.6) & 1002.8 & ($\pm$32.4) & 995.2 & ($\pm$58.7) & 962.6 & ($\pm$41.5) & 959.0 & ($\pm$50.6) \\
			$\dot{x}$ \& T & 888 & 37.8 & 932 & 65.4 & 928 & 37.6 & 906 & 55.2 & 888 & 27.2\\
			
			\hline En76k14 (1021) & \textbf{1124.6} & ($\pm$11.5) & 1198.8 & ($\pm$20.0) & 1186.7 & ($\pm$47.9) & 1177.3 & ($\pm$52.8) & 1172.2 & ($\pm$37.1) \\
			$\dot{x}$ \& T & 1107 & 28.8 & 1142 & 60.1 & 1117 & 32.7 & 1104 & 46.4 & 1110 & 29.4\\
			
			\hline\hline En101k8 (815) & \textbf{906.4} & ($\pm$16.4) & 1104.4 & ($\pm$44.8) & 999.9 & ($\pm$46.0) & 971.7 & ($\pm$69.1) & 991.1 & ($\pm$41.1) \\
			$\dot{x}$ \& T & 867 & 69.8 & 1042 & 124.5 & 908 & 67.5 & 893 & 134.2 & 933 & 67.5\\
			
			\hline En101k14 (1071) & \textbf{1191.5} & ($\pm$26.1) & 1298.0 & ($\pm$112.4) & 1288.8 & ($\pm$52.8) & 1249.9 & ($\pm$56.5) & 1273.5 & ($\pm$50.7) \\
			$\dot{x}$ \& T & 1155 & 77.9 & 1175 & 119.4 & 1187 & 78.5 & 1182 & 134.4 & 1194 & 75.4\\
			
			\hline
		\end{tabular}
	}
	\caption{Resultados de los algoritmos $GB$, $GA_1$, $GA_2$, $DGA_1$ y $DGA_2$ para el CVRP. Para cada instancia se muestra el resultado promedio, desviación típica, mejor resultado obtenido y tiempo de ejecución promedio.}
	\label{tab:CVRP}
\end{table}

Como ya se ha mencionado en la introducción de esta sección, y con la intención de realizar un análisis de resultados riguroso, se han llevado a cabo dos tests estadísticos. La primera de las pruebas realizadas es el test normal $z$. Con esta prueba los resultados obtenidos por el GB son directamente comparados con los obtenidos por todas y cada una de las técnicas restantes. De esta manera, con este test puede concluirse si las diferencias en los resultados obtenidos por el GB y el resto de algoritmos son significativas o no. El estadístico $z$ tiene la siguiente forma:

\[z = \frac{\overline{X_{GB}} - \overline{X_{i}}}{\sqrt{\frac{\sigma_{GB}^2}{n_{GB}} + \frac{\sigma_{i}^2}{n_{i}}}}\]

en la que:

\

$\overline{X_{GB}}$: Media aritmética de los resultados obtenidos por GB,

$\sigma_{GB}$: Desviación típica de los resultados obtenidos por GB,

$n_{GB}$: Número de ejecuciones realizadas por GB,

$\overline{X_i}$: Media aritmética de los resultados obtenidos por la técnica $i$,

$\sigma_i$: Desviación típica de los resultados obtenidos por la técnica $i$,

$n_i$: Número de ejecuciones realizadas por el algoritmo $i$,

\

Es interesante mencionar que en este test el GB se enfrenta al resto de meta-heurísticas implementadas por separado. Por lo tanto, el parámetro $i$ puede tomar los valores $GA_1$, $GA_2$, $DGA_1$ and $DGA_2$. El intervalo de confianza ha sido establecido en el 95\% ($z_{0.05}$ = 1.96). De esta forma, el resultado del test puede ser positivo (+) si $z\geq1.96$; negativo (-) si $z\leq-1.96$; o neutro (*) si $-1.96<z<1.96$. Un valor + indica que el GB es significativamente mejor que su adversario. El caso contrario implica que el GB es significativamente inferior, mientras que * demuestra que la diferencia en los resultados no es relevante. En este estudio se mostrarán también los valores del estadístico $z$ para cada uno de los casos. Con esto, las diferencias pueden concretarse con mayor minuciosidad. En las tablas \ref{tab:ztestATSPTSP}, \ref{tab:ztestVRPBCVRP} y \ref{tab:ztestBPPNQP} se muestran los resultados obtenidos al realizar este test.

\begin{table}[tb]
	\centering
	\setlength{\tabcolsep}{1.7pt}
	\scalebox{0.75}{
		\begin{tabular}{| l || c | c | c | c || l || c | c | c | c |}
			\hline Instancia & vs. $GA_1$ & vs. $GA_2$ & vs. $DGA_1$ & vs. $DGA_2$ & Instancia & vs. $GA_1$ & vs. $GA_2$ & vs. $DGA_1$ & vs. $DGA_2$\\ 
			\hline\hline 
			br17 & + (3.16) & + (3.16) & * (0.00) & * (0.00) & Oliver30 & + (3.71)& + (2.71)& + (7.45) & + (2.69)\\
			ftv33 & + (5.97)& + (14.88)& + (6.76)& + (5.74)& Eilon50 & + (7.22)& + (10.39)& + (8.94) & + (6.56)\\
			ftv35 & + (7.67)& + (6.35)& + (7.68)& + (8.79)& Eil51 & + (10.56) & + (7.32)& + (7.76) & + (8.81)\\
			ftv38 & + (7.48)& + (5.40)& + (7.86)& + (9.03)& Berlin52 & + (10.70)& + (5.21)& + (8.90) & + (4.88)\\
			p43 & + (8.07)& + (10.26)& + (8.85)& + (7.60)& St70 & + (4.28)& + (3.72) & + (9.40) & + (7.12)\\
			ftv44 & + (5.67)& + (15.48)& + (6.27)& + (7.61)& Eilon75 & + (11.48)& + (7.61)& + (10.51) & + (6.03)\\
			ftv47 & + (6.10)& + (5.93)& + (6.67)& + (7.84)& Eil76 & + (10.85) & + (9.03)& + (8.48) & + (11.01)\\
			ry48p & + (6.46)& + (7.45)& + (6.45)& + (8.61)& Eil101  & + (13.48)& + (12.26)& + (10.90) & + (8.77)\\
			ft53 & + (11.70)& + (19.65)& + (12.45)& + (13.53)& KroA100 & + (9.57) & + (4.68)& + (13.77) & + (7.53)\\
			ftv55 & + (6.77)& + (8.18)& + (10.57)& + (12.19)& KroB100 & + (9.43)& + (5.85)& + (11.29) & + (6.29)\\
			ftv64 & + (7.12)& + (26.02)& + (8.77)& + (11.13)& KroC100 & + (7.55)& + (6.17)& + (10.47) & + (6.61)\\
			ftv70 & + (6.10)& + (21.97)& + (9.89)& + (13.83)& KroD100 & + (8.99)& + (7.25)& + (8.55) & + (5.78)\\
			ft70 & + (2.28)& + (25.62)& + (4.88)& + (6.65)& KroE100 & + (9.65)& + (6.33)& + (7.55) & + (7.05)\\
			kro124p & + (9.80)& + (14.48)& + (12.17)& + (8.89)& Pr107 & + (9.69) & + (2.93) & + (5.40) & + (2.71)\\
			ftv170 & + (3.64)& + (3.16)& + (4.69)& + (4.79)& Pr124 & + (9.46)& + (4.61)& + (8.86) & + (6.29)\\
			rbg323 & + (7.97)& + (7.56)& + (6.98)& + (3.85)& Pr136 & + (7.52)& + (4.26)& + (4.71) & + (4.26)\\
			rbg358 & + (35.52)& + (35.54)& + (12.54)& + (11.73)& Pr144 & + (5.05)& + (3.72)& + (6.86) & + (4.93)\\
			rbg403 & + (18.11)& + (13.14)& + (11.56)& + (11.43)& Pr152 & + (5.38)& + (6.21)& + (10.17) & + (4.57)\\
			rbg443 & + (10.76)& + (10.05)& + (10.96)& + (10.84)&  & & & & \\
			\hline
		\end{tabular}
	}
	\caption{Test normal $z$ para los problemas ATSP y TSP. + implica una mejora significativa. - denota que es sustancialmente peor. * indica que las diferencias no son significativas (con un nivel de confianza del 95\%).}
	\label{tab:ztestATSPTSP}
\end{table}

\begin{table}[tb]
	\centering
	\setlength{\tabcolsep}{1.7pt}
	\scalebox{0.75}{
		\begin{tabular}{| l || c | c | c | c || l || c | c | c | c |}
			\hline Instancia & vs. $GA_1$ & vs. $GA_2$ & vs. $DGA_1$ & vs. $DGA_2$ & Instancia & vs. $GA_1$ & vs. $GA_2$ & vs. $DGA_1$ & vs. $DGA_2$\\ 
			\hline\hline 
			C101 & + (3.80)& + (3.46)& + (6.54)& + (2.62)& En22k4 & + (4.70) & + (10.43) & + (7.23) & + (5.16)\\
			C201 & + (9.75)& + (13.49)& + (13.93)& + (3.08)& En23k3 & + (4.38) & + (9.53) & + (10.03) & $\ast$ (-1.03)\\
			R101 & + (7.46)& + (5.79)& + (7.96)& * (0.89)& En30k3 & + (7.85) & + (6.91) & + (3.70) & + (3.68) \\
			R201 & + (21.16)& + (8.18)& + (11.39)& + (5.29)& En33k4 & + (7.27) & + (14.26) & + (9.95) & + (10.96) \\
			RC101 & + (10.32)& + (7.96)& + (5.74)& + (5.83)& En51k5 & + (8.80) & + (5.41) & + (7.66) & + (4.59)\\
			RC201 & + (10.16)& + (6.26)& + (14.93)& * (1.45)& En76k7 & + (8.77) & + (7.64) & + (8.35) & + (4.32)\\
			En23k3 & + (9.16)& + (8.94)& + (9.18)& * (1.32)& En76k8 & + (11.92) & + (9.99) & + (5.34) & + (5.52) \\
			En30k4 & + (18.97)& + (17.84)& + (9.77)& + (7.44)& En76k10 & + (11.09) & + (6.00) & + (4.94) & + (3.83) \\
			En33k4 & + (8.98)& + (6.48)& + (7.53)& + (6.29)& En76k14 & + (14.38) & + (5.63) & + (4.36) & + (5.48)\\
			En51k5 & + (9.70)& + (10.49)& + (14.61)& + (2.38)& En101k8 & + (18.56) & + (8.56) & + (4.11) & + (8.55)\\
			En76k8 & + (10.57)& + (8.96)& + (7.06)& + (5.44)& En101k14 & + (4.12) & + (7.38) & + (4.19) & + (6.43)\\
			En101k14 & + (13.24)& + (7.93)& + (18.47)& + (10.24)& & & & & \\
			\hline
		\end{tabular}
	}
	\caption{Test normal $z$ para los problemas VRPB y CVRP. + implica una mejora significativa. - denota que es sustancialmente peor. * indica que las diferencias no son significativas (con un nivel de confianza del 95\%).}
	\label{tab:ztestVRPBCVRP}
\end{table}

\begin{table}[tb]
	\centering
	\setlength{\tabcolsep}{1.7pt}
	\scalebox{0.73}{
		\begin{tabular}{| l || c | c | c | c || l || c | c | c | c |}
			\hline Instancia & vs. $GA_1$ & vs. $GA_2$ & vs. $DGA_1$ & vs. $DGA_2$ & Instancia & vs. $GA_1$ & vs. $GA_2$ & vs. $DGA_1$ & vs. $DGA_2$\\ 
			\hline\hline 
			N1C1W1\_A & + (6.32)& + (11.06)& + (10.11)& + (8.85) & 8-Queens & * (0.00)& * (0.00)& * (0.00)& * (0.00)\\
			N1C1W1\_B & + (14.23)& + (6.32)& + (6.32)& + (6.32) & 20-Queens & + (13.00)& * (0.00)& + (7.91)& + (6.08)\\
			N1C2W1\_A & + (9.39)& + (9.39)& + (9.89)& + (12.72) & 50-Queens & + (19.71)& + (17.16)& + (28.74)& + (16.99)\\
			N1C2W1\_B & + (17.61)& + (14.14)& + (9.89)& + (8.22) & 75-Queens & + (31.37)& + (15.71)& + (33.25)& + (22.16)\\
			N2C1W1\_A & + (19.79)& + (13.19)& + (17.91)& + (9.96) & 100-Queens & + (37.42)& + (23.04)& + (34.32)& + (22.37)\\
			N2C1W1\_B & + (9.71)& + (10.72)& + (12.73)& + (11.33) & 125-Queens & + (31.57)& + (31.97)& + (39.71)& + (36.39)\\
			N2C2W1\_A & + (7.00)& + (17.61)& + (18.00)& + (14.00) & 150-Queens & + (29.69)& + (30.04)& + (33.13)& + (46.36)\\
			N2C2W1\_B & + (8.09)& + (12.49)& + (16.19)& + (14.57) & 200-Queens & + (33.09)& + (33.66)& + (36.14)& + (34.96)\\
			N3C2W2\_A & + (28.59)& + (15.64)& + (20.95)& + (14.48) & 225-Queens & + (33.05)& + (58.64)& + (55.42)& + (34.05)\\
			N3C2W2\_B & + (40.08)& + (19.89)& + (19.29)& + (16.90) & 250-Queens & + (61.21)& + (26.80)& + (43.06)& + (51.66)\\
			N3C3W1\_A & + (32.34)& + (9.55)& + (14.74)& + (8.04) & 275-Queens & + (24.21)& + (37.31)& + (30.14)& + (37.75)\\
			N3C3W1\_B & + (20.57)& + (7.98)& + (14.14)& + (8.48) & 300-Queens & + (60.37)& + (40.71)& + (45.51)& + (37.27)\\
			N4C1W1\_A & + (24.07)& + (12.23)& + (21.45)& + (17.64) & 325-Queens & + (60.93)& + (39.41)& + (72.50)& + (58.97)\\
			N4C2W1\_A & + (24.98)& + (2.48)& + (27.58)& + (15.01) & 350-Queens & + (66.00)& + (52.06)& + (59.30)& + (43.39)\\
			N4C2W1\_B & + (17.97)& * (1.27)& + (26.22)& + (21.00) & 400-Queens & + (34.01)& + (31.60)& + (41.51)& + (41.40)\\
			N4C2W1\_C & + (18.90)& + (4.70)& + (16.79)& + (9.16) &  & & & & \\
			\hline
		\end{tabular}
	}
	\caption{Test normal $z$ para los problemas BPP y NQP. + implica una mejora significativa. - denota que es sustancialmente peor. * indica que las diferencias no son significativas (con un nivel de confianza del 95\%).}
	\label{tab:ztestBPPNQP}
\end{table}


El segundo test estadístico llevado a cabo es el famoso test de Friedman. En la tabla \ref{tab:friedman} se exponen los resultados medios obtenidos mediante esta prueba. Es interesante tener en cuenta que cuanto menor sea el número logrado mejor será el ranking del algoritmo. Este ranking se ha obtenido considerando los resultados medios de cada técnica y comparándolos instancia por instancia. Además de esto, el valor $X^2_r$ también se ha representado en la tabla \ref{tab:friedman}, con la intención de comprobar si las diferencias entre las técnicas son estadísticamente significativas. Este valor se ha obtenido mediante la fórmula que se presenta a continuación:

	\[ X^2_r = \frac{12}{H K (K+1)} \sum (H Rc)^2 - 3H(K+1)\]
	
Siendo $H$ el número de instancias utilizadas para el problema (por ejemplo, para el ATSP $H$=19), $K$ el número de técnicas empleadas (5), y $Rc$ el valor conseguido en el ranking de Friedman. El intervalo de confianza se ha establecido en 99\%. Con esto, el punto crítico en una distribución $\chi^2$ con 4 grados de libertad es 13.277. Por lo tanto, si $X^2_r > 13.277$, puede concluirse que las diferencias entre los algoritmos son estadísticamente significativas. De cualquier otra manera, las diferencias son consideradas insignificantes.

\begin{table}[tb]
	\renewcommand{\arraystretch}{1}
	\centering
	\scalebox{0.8}{
		\begin{tabular}{| c || c | c | c | c | c || c |}
		  \hline  
		  Problema & $GB$ & $GA_{1}$ & $GA_{2}$ & $DGA_{1}$ & $DGA_{2}$ & $X^2_r$\\  
			\hline 
			TSP & \textbf{1.00} & 4.61 & 2.94 & 4.16 & 2.27 & 60.15\\
			ATSP & \textbf{1.01} & 3.16 & 4.52 & 3.17 & 3.01 & 42.14\\
			CVRP & \textbf{1.09} & 4.19 & 4.26 & 3.00 & 2.45 & 52.39\\
			VRPB & \textbf{1.00} & 4.33 & 3.25 & 4.00 & 2.25 & 30.59\\
			NQP & \textbf{1.04} & 4.21 & 2.28 & 4.21 & 2.64 & 22.18\\
			BPP & \textbf{1.00} & 3.96 & 3.03 & 4.06 & 2.81 & 33.55\\
			\hline
		\end{tabular}
	}
	\caption{Resultados del test de Friedman (Cuanto menor sea el resultado, mejor será el ranking). La última columna representa el valor $X^2_r$}
	\label{tab:friedman}
\end{table}


\section{Análisis de los resultados}
\label{sec:analisis}

Continuando con la filosofía adoptada en el apartado anterior, y con la intención de no extender en exceso la presente sección con información redundante, se ha adoptado la siguiente estrategia: en un primer momento se realizará un profundo análisis de los resultados obtenidos para el problema TSP. Después de esto, se analizarán de la misma forma los resultados logrados para el CVRP. Finalmente, se llevará a cabo un análisis general respecto a los cuatro problemas restantes (ATSP, VRPB, NQP y BPP). 

Tras observar los datos representados en las Tablas \ref{tab:TSP1} y \ref{tab:TSP2}, a simple vista podría decirse que la técnica propuesta en este trabajo ofrece mejores resultados que todos y cada uno de los algoritmos utilizados para la comparación, en términos tanto de mejor solución obtenida por instancia, como resultado promedio. Esto es debido a que el GB aventaja al resto de alternativas en el 100\% de los casos (18 de 18) en relación al resultado medio, mientras que la mejor solución obtenida en cada instancia por el mismo supera las del resto en el 94\% de los casos (17 de 18). Además de esto, si se presta atención a los resultados logrados al llevar a cabo el test estadístico $z$ normal (Tabla \ref{tab:ztestATSPTSP}) puede deducirse que las diferencias entre los resultados obtenidos por el GB y el resto de técnicas son estadísticamente significativas el 100\% de las ocasiones. Esta mejora es apoyada también por los resultados ofrecidos por el Test de Friedman en la Tabla \ref{tab:friedman}. El único caso en el que el GB se ve superado por alguno de los algoritmos implementados es en la instancia Pr107, en la que el $GA_2$ proporciona una mejor solución encontrada (44391 ofrecida por el GB, frente a 44387 por $GA_2$). Aun así, este hecho no supone un problema mayor, ya que el GB aventaja al $GA_2$ en relación al resultado promedio y desviación estándar.

La tendencia del GB de obtener mejores resultados en esta experimentación podría darse por diversos factores \cite{osaba2014GB1}. Como ya se ha pormenorizado en secciones anteriores, el GB es una técnica que combina la optimización individual de los jugadores (entrenamientos convencionales), con procesos de competición (partidos) y cooperación entre los mismos (entrenamientos personalizados). Mientras que la meta-heurística propuesta en este trabajo dota de mayor importancia a la mejora individual de los jugadores, el resto de técnicas centra su esfuerzo en los mecanismos cooperativos. Este hecho es muy común en la comunidad de técnicas poblacionales y multi-poblacionales, como bien se ha comprobado en la Sección \ref{sec:poblacionales} y \ref{sec:multi}. De todos modos, es adecuado recalcar que el GB posee un mecanismo de cooperación de individuos, por medio del entrenamiento personalizado (Sección \ref{sec:TrainingPhase}). Este recurso se emplea en contadas ocasiones, siempre y cuando se considere presumiblemente beneficioso para el proceso de búsqueda y la exploración del espacio de soluciones. Este mecanismo ayuda a evitar los óptimos locales y a explorar el espacio de manera exhaustiva. 

Otra ventaja respecto al $GA_1$, $GA_2$ y $DGA_1$, en este caso, y respecto a muchas otras técnicas de la literatura, es la opción de un jugador de explorar diferentes estructuras de vecindario. Este hecho sucede gracias a que los individuos pueden alternar sus equipos, obteniendo con esto diferentes tipos de entrenamiento a lo largo de la ejecución del algoritmo. Esta herramienta es otro camino para evadir los óptimos locales y contribuye a que los jugadores rastreen el espacio de diversas maneras. Además de esto, los jugadores pueden ahondar en aquellas regiones del espacio que resulten más interesantes para el proceso de búsqueda. Por otro lado, los algoritmos $GA_1$, $GA_2$, $DGA_1$ y $DGA_2$ cuentan también con varios mecanismos para intentar esquivar óptimos locales, pero no son tan efectivos como los ofrecidos por el GB.

En lo relativo a los tiempos de ejecución, tanto el $GA_1$ como el $DGA_1$ requieren de un esfuerzo mayor que el GB, mientras que el $GA_2$ y el $DGA_2$ demandan tiempos similares a los del GB. Este hecho proporciona una nueva ventaja para el GB, debido a que utilizando unos tiempos semejantes logra encontrar mejores soluciones que el resto de meta-heurísticas.

Las razones por las que el algoritmo propuesto precisa de un esfuerzo computacional menor que el $GA_1$ y $DGA_1$ son lógicas. Si se comparan los operadores individuales (mutaciones y entrenamientos convencionales) y los operadores cooperativos (cruces y entrenamientos personalizados), los segundos necesitan un mayor tiempo de ejecución, ya que se trata de operaciones complejas entre dos soluciones diferentes. Por el contrario, los operadores individuales trabajan con una sola solución, y consisten en pequeños cambios realizadas en un tiempo mínimo. El GB realiza un menor número de movimientos cooperativos en comparación con el $GA_1$ y el $DGA_1$, y este hecho se refleja perfectamente en los tiempos de ejecución. Finalmente, el GB, el $GA_2$ y el $DGA_2$ obtienen tiempos similares debido a que utilizan sus operadores de manera semejante.

Otro factor importante que es obligatorio destacar es la robustez propia de la meta-heurística GB. Si se fija la atención en las desviaciones típicas de cada una de las técnicas empleadas, puede comprobarse como en el 100\% de los casos (18 sobre 18) el GB ofrece una menor dispersión de resultados. Este hecho implica que la diferencia entre la mejor y la peor solución encontrada para cada instancia es pequeña. Esta característica es un claro indicativo de la robustez y la fiabilidad inherentes del GB, algo muy apreciado en caso de querer utilizar la meta-heurística en entornos reales, en los que una mala solución puede ser muy perjudicial.

Para terminar con este primer análisis de resultados, y con la intención de realizar un examen con mayor profundidad, se realizará una comparación en relación al comportamiento de convergencia del GB y del $DGA_2$. Cabe destacar que se ha seleccionado el algoritmo $DGA_2$ debido a que es el que guarda mayor similitud con el GB, ya sea en cuanto a concepto, como en resultados obtenidos. En la tabla \ref{tab:convergenceTSP} se exhibe el promedio de evaluaciones de la función objetivo necesitadas por cada técnica para alcanzar la solución ofrecida como solución final a cada instancia del TSP. Este número se expresa en miles.

\begin{table}[htb]
	\centering
	\scalebox{0.8}{
		\begin{tabular}{| l || c | c || l || c | c |}
			\hline Instancia & $GB$ & $DGA_2$ & Instancia & $GB$ & $DGA_2$\\ 
			\hline\hline Oliver30 & \textbf{10.72} & 17.06  & KroB100 & 285.29 &  \textbf{242.91}\\
			Eilon50 &  52.86 &  \textbf{49.74} & KroC100 & \textbf{277.70} &  308.12\\
			Eil51 &  \textbf{51.46} &  54.27 & KroD100 & \textbf{199.56} &  248.74\\
			Berlin52 &  \textbf{53.04} &  54.00 & KroE100 & 294.94 & \textbf{287.33}\\
			St70 & 127.62 &  \textbf{104.61} & Pr107 & \textbf{338.33} & 362.50\\
			Eilon75 & 138.40 &  \textbf{128.78} & Pr124 & \textbf{408.16} & 464.77\\
			Eil76 & 144.30 &  \textbf{137.77} & Pr136 & \textbf{616.52} & 706.00\\
			Eil101 & 312.89 &  \textbf{311.03} & Pr144 & \textbf{771.01} & 867.56\\
			KroA100 & \textbf{232.86} &  281.11 & Pr152 & 1195.98 & \textbf{996.85}\\
			\hline
		\end{tabular}
	}
	\caption{Comportamiento de convergencia del $GB$ y $DGA_2$ para el TSP, expresada en miles de evaluaciones de la función objetivo}
	\label{tab:convergenceTSP}
\end{table}

La tabla \ref{tab:convergenceTSP} muestra cómo ambos algoritmos presentan un comportamiento parecido, destacando ligeramente el GB, ya que obtiene una convergencia más rápida en 10 de las 18 instancias. Este hecho también supone una ventaja para la meta-heurística presentada en este trabajo, debido a su capacidad para encontrar mejores soluciones haciendo uso de un número similar de evaluaciones de la función objetivo.

Como conclusión podría decirse que, haciendo uso de los mismos operadores y los mismos parámetros, la técnica propuesta en este documento es más eficiente que los algoritmos $GA_1$, $GA_2$, $DGA_1$ y $DGA_2$ para el TSP, en términos de calidad de soluciones, tiempos de ejecución, robustez y comportamiento de convergencia.  


Respecto al problema CVRP, al igual que sucede en el caso del TSP, la técnica propuesta obtiene mejores resultados en la mayoría de las instancias utilizadas (10 de 11). El único caso en el que el GB se ha visto superado por otra técnica en este aspecto es en la instancia En23k3, en la que $DGA_2$ ha obtenido un resultado promedio mejor. De todas formas, como bien puede comprobarse en la tabla \ref{tab:ztestVRPBCVRP} correspondiente al test estadístico $z$ normal, esta mejora no es estadísticamente significativa. Observando la misma tabla, por otro lado, puede verse cómo GB ofrece mejoras estadísticamente reseñables en todos los enfrentamientos (43 de 44), a excepción del arriba mencionado.

En relación a la mejor solución encontrada, el GB supera al resto de técnicas en 5 de las 11 instancias (En33k4, En51k5, En76k8, En101k8 y En101k14), mientras que se ve rebasado en 2 de ellas (En76k7 y En76k14). En las 4 instancias restantes el GB comparte el privilegio de ser la meta-heurística que mejor solución ha encontrado al problema con alguna de las alternativas empleadas.

Dando paso a los tiempos de ejecución, el GB mejora ampliamente a los algoritmos $GA_1$ y $DGA_1$, de la misma forma y por las mismas razones que ocurre en el TSP. Comparado por el $GA_2$, por otro lado, los tiempos empleados por cada una de las técnicas guardan un gran parecido. Finalmente, debido a la utilización de la misma parametrización y operadores, los tiempos del GB y $DGA_2$ son también similares entre sí. Aun así, el GB es una técnica más compleja que el $DGA_2$, con una capacidad de exploración y explotación mayor. Es por esta razón por la que el GB efectúa una búsqueda más exhaustiva del espacio de soluciones, acarreando con ello unos tiempos ligeramente mayores.

Además de esto, las desviaciones típicas presentadas por el GB son menores que las dadas por el resto de alternativas también para el CVRP, lo que conlleva a una mayor regularidad a la hora de ofrecer soluciones para el GB. Como bien se ha dicho anteriormente, esta característica proporciona robustez y fiabilidad a la meta-heurística.  

Finalmente, siguiendo los pasos realizados con el problema anterior, en la tabla \ref{tab:convergenceVRP} pueden observarse los comportamientos de convergencia del GB y $DGA_2$ para el CVRP.

\begin{table}[htb]
	\centering
	\scalebox{0.8}{
		\begin{tabular}{| l || c | c || l || c | c |}
			\hline Instancia & $GB$ & $DGA_2$ & Instance & $GB$ & $DGA_2$\\ 
			\hline\hline En22k4 & \textbf{29.56} & 33.16  & En76k8 & \textbf{242.37} & 284.69\\
			En23k3 &  \textbf{16.65} & 24.04 & En76k10 & \textbf{216.19} & 262.75\\
			En30k3 &  \textbf{57.51} & 72.39 & En76k14 & 232.66 &  \textbf{227.74}\\
			En33k4 &  \textbf{40.99} & 63.09 & En101k8 & \textbf{595.84} & 796.99\\
			En51k5 & \textbf{122.76} & 138.45 & En101k14 & \textbf{502.24} & 591.75\\
			En76k7 & \textbf{227.84} & 307.19 & & & \\
			\hline
		\end{tabular}
	}
	\caption{Comportamiento de convergencia del $GB$ y $DGA_2$ para el TSP, expresada en miles de evaluaciones de la función objetivo}
	\label{tab:convergenceVRP}
\end{table}

Al contrario que sucede con el TSP, en el presente problema el GB muestra un comportamiento de convergencia mejor que el $DGA_2$, necesitando un menor número de evaluaciones para casi todas las instancias (10 de 11). En este caso, esto reporta una gran ventaja para el GB, demostrando que en un número menor de evaluaciones puede obtener mejores resultados que su competidor.

Con todo esto, podría decirse que, en términos generales, el GB es mejor que los algoritmos $GA_1$, $GA_2$, $DGA_1$ y $DGA_2$ también para el CVRP. La técnica propuesta es mejor en términos de calidad, robustez y comportamiento de convergencia. En lo referente a los tiempos de ejecución, el GB actúa de forma similar a las técnicas $GA_2$ y $DGA_2$, siendo ampliamente mejor que los métodos $GA_1$ y $DGA_1$.


En lo que respecta a los problemas restantes, las conclusiones que pueden obtenerse al analizar los resultados obtenidos para dichos problemas son las mismas que las relatadas con anterioridad. En general podría decirse que el GB supera al resto de técnicas en términos de calidad de soluciones. Observando los datos ofrecidos por las tablas \ref{tab:ATSP}, \ref{tab:VRPB}, \ref{tab:NQP} y \ref{tab:BPP}, disponibles en el apéndice \ref{app:resultados}, puede verse como el GB demuestra un rendimiento mejor en el 95.16\% de las instancias (59 de 62). En las 3 instancias restantes el GB obtiene los mejores registros junto a una o más técnicas. En todo caso, la meta-heurística propuesta en este trabajo no es superada en ninguna de estas 62 instancias. Además de esto, como bien se demuestra en las tablas \ref{tab:ztestVRPBCVRP} y \ref{tab:ztestBPPNQP}, las diferencias entre los resultados del GB y el resto de alternativas son estadísticamente significativas en el 95.96\% de los enfrentamientos (238 de 248), siendo las diferencias insignificantes en el 4.04\% restante.  

En relación al ATSP, el GB logra imponerse a sus competidores en el 94.73\% de los casos (18 de 19). En la instancia restante el GB alcanza el mismo resultado promedio que el $DGA_1$ y $DGA_2$. Estas diferencias son significativas en el 97.36\% de las confrontaciones (74 de 76). Por otro lado, centrando la atención en los problemas VRPB y BPP, el GB supera al resto de alternativas en el 100\% de los casos, siendo estas mejoras significativas en el 93.75\% de los enfrentamientos (45 de 48) para el VRPB, y en el 100\% de los enfrentamientos (64 de 64) en el caso del BPP. Finalmente, en lo que respecta al NQP, la meta-heurística presentada en esta tesis demuestra ser mejor en el 86.66\% de las instancias (13 de 15). En las instancias restantes el GB obtiene los mejores registros en conjunto con alguna de las otras alternativas. En cuanto al apartado estadístico, estas mejoras son reseñables en el 91\% de los casos  (55 de 60).

Asimismo, observando los datos ofrecidos por el test estadístico de Friedman (Tabla \ref{tab:friedman}), podría decirse de nuevo que el GB es la técnica que mejores resultados ha logrado, tanto para el ATSP, como para el VRPB, NQP y BPP. Por otro lado, todos los valores de $X^2_r$ que pueden contemplarse en dicha tabla son mayores que el punto crítico, 13.227. Esto indica que las diferencias entre los resultados de los diferentes algoritmos son estadísticamente significativas, al igual que se ha concluido al realizar el test $z$ normal.

En lo relativo a los tiempos de ejecución, podría decirse que los rendimientos observados son realmente similares a los presentados para el TSP y CVRP, ya que el GB es más rápido que el $GA_1$ y $DGA_1$, mientras que requiere unos tiempos parecidos a los ofrecidos por el $GA_2$ y $DGA_2$. Como ya se ha explicado con anterioridad, este hecho proporciona una gran ventaja al GB, debido a que logra alcanzar unos resultados superiores en cuanto a calidad haciendo uso del mismo esfuerzo computacional. 

En lo que respecta a la robustez de las técnicas, el GB muestra unas desviaciones estándar inferiores a las del resto de algoritmos en el 93.54\% de las instancias (58 de 62), pudiendo, de esta manera, extender a estos problemas las conclusiones vertidas para el TSP y el CVRP.

Como conclusión final, puede destacarse la capacidad del GB para superar al resto de las alternativas también para los problemas ATSP, VRPB, NQP y BPP. Añadiendo estas conclusiones a las expuestas para los problemas anteriores, es prudente confirmar que el GB se posiciona como una técnica prometedora para la resolución tanto de problemas de asignación de rutas a vehículos, en particular, como de problemas de optimización combinatoria en general.
	


\begin{savequote}[40mm]
La inteligencia consiste no sólo en el conocimiento, sino también en la destreza de aplicar los conocimientos en la práctica.
\qauthor{Aristóteles}
\end{savequote}

\chapter{Aplicación práctica del Golden Ball}
\label{cha:aplicacion}

\graphicspath{ {5_caso_practico/figures/} }


\lettrine{C}{}{omo} bien se ha ido mencionando a lo largo de esta tesis, una de las principales características de los problemas de asignación de rutas a vehículos es su capacidad para adaptarse y resolver problemas surgidos en el mundo real. De hecho, estos problemas, normalmente, son creados a partir de una situación de transporte veraz, a la cual se le pretende dar una solución eficiente. En relación a esto, y como ya se ha mencionado en la sección \ref{sec:RVRP}, los problemas ricos de asignación de rutas a vehículos, R-VPR, también llamados problemas de asignación de rutas a vehículos multi-atributo, son una clase de problemas especialmente concebidos para ser aplicados a situaciones complejas propias del mundo real.

La principal característica de los RVRP es la gran cantidad de restricciones con las que cuentan, lo que hace que su tratamiento y resolución sean unas tareas extraordinariamente complejas. Es por esto por lo que una cualidad como la de ser capaz de hacer frente a este tipo de situaciones es realmente apreciada en un algoritmo. En esta sección se describirá, con sendos ejemplos prácticos, cómo la meta-heurística propuesta en esta tesis cuenta con dicha propiedad.

Con todo esto, este capítulo constará de dos diferentes apartados. En el primero de ellos se recapitularán los trabajos llevados a cabo por el autor de esta tesis en relación a este tema concreto (sección \ref{sec:AplicPrevios}). Después de esto, en la sección \ref{sec:AplicNueva} se presentará una nueva situación de transporte real, la cual se modelizará como un RVRP, para ser abordada posteriormente por la meta-heurística presentada en este trabajo. Con este esfuerzo se pretende demostrar la capacidad de la técnica propuesta para hacer frente a problemas de esta índole.

\section{Trabajos previos relacionados con los problemas RVRP}
\label{sec:AplicPrevios}

El autor de esta tesis cuenta con cierta experiencia en el campo de los problemas de asignación de rutas a vehículos multi-atributo. Hasta la fecha se han realizados dos trabajos distintos, ambos enfocados al problema rico del viajante comercial, RTSP, o problema del viajante comercial multi-atributo. En el primero de ellos se diseñó un problema múltiple del viajante comercial asimétrico y con recogidas, el cual fue resuelto con una meta-heurística poblacional y adaptativa. Los detalles de este trabajo pueden encontrarse en \cite{osaba2015SAMI}. En el segundo de los trabajos se desarrolló un problema múltiple del viajante comercial asimétrico con entregas y recogidas simultáneas, o \textit{Multiple Asymmetric Traveling Salesman Problem with Simultaneous Pickups and Deliveries, MA-TSP-SPD} (MA-TSP-SPD) para hacer frente a un problema de transporte bajo demanda \cite{osaba2015HAIS}. En este trabajo no solo se plantea la problemática que se intenta cubrir con el problema planteado, sino que se presenta un banco de pruebas y una primera solución a este, proporcionada por una meta-heurística Golden Ball. En esta sección se detallarán los principales aspectos de este trabajo.  


Como bien se ha mencionado en la introducción de esta tesis, el transporte público cuenta con una gran importancia en la sociedad actual, debido a que es utilizado por casi toda la población mundial, y afecta directa e indirectamente a la calidad de vida de las personas. Existen muchos tipos de transporte público, cada uno con sus características, pero todos ellos comparten los mismos contratiempos. Con la intención de hacer frente a esos contratiempos nació el concepto del transporte bajo demanda (\textit{Transport On Demand}, TOD)).

El TOD se refiere, principalmente, al transporte de pasajeros y bienes entre un punto de origen y otro de destino. La gran mayoría de problemas relacionados con el TOD se caracterizan por la presencia de tres objetivos enfrentados: maximizar el número de peticiones de servicio atendidas, minimizar el costo operativo, y maximizar la satisfacción de los usuarios.

\begin{sloppypar}
Existen diferentes tipos de problemas TOD, siendo el Dial-a-Ride (DAR) \cite{DRP} uno de los más conocidos. El DAR se caracteriza por su flexibilidad a la hora de diseñar las rutas y por la programación de rutas llevadas a cabo por vehículos pequeños/medianos en modo de viaje compartido entre varios puntos de recogida y entrega impuestos por las necesidades de los pasajeros. Una aplicación común para este tipo de sistemas es el servicio de transporte en áreas de baja demanda, en las que una línea de transporte regular no resulta económicamente viable. Otra aplicación típica para los sistemas TOD es el servicio puerta-a-puerta para personas discapacitadas o personas ancianas. En este contexto, los usuarios formulan dos tipos de peticiones, una de salida para llegar su lugar de destino, y otra de retorno. Este tipo de aplicación cuenta con un amplio interés social, debido a que, por encima de todo, ayuda a garantizar el bienestar de las personas con necesidades especiales.
\end{sloppypar}

Los sistemas DAR y otro tipo de problemas TOD son el foco de muchos trabajos hoy en día \cite{DRPej1,DRPej2}. Además de esto, varios sistemas DAR sofisticados han sido implementados en diversas ciudades a lo largo del mundo, como en Bristol (Reino Unido)\footnote{http://www.bristoldialaride.org.uk/}, Ciudad del Cabo (Sudáfrica)\footnote{https://www.capetown.gov.za/en/Transport/Pages/AboutDialaRide.aspx}, Riverside (California)\footnote{http://www.riversidetransit.com/home/index.php/dial-a-ride}, o Londres\footnote{https://www.tfl.gov.uk/modes/dial-a-ride/}. 


El autor de esta tesis presenta, junto con varios investigadores más, una aproximación para conseguir abordar diferentes problemas del tipo DAR. Para esto, el problema DAR es modelizado como un RTSP, concretamente como el arriba mencionado MA-TSP-SPD. Las principales características de este problema son las siguientes:

\begin{enumerate}
	\item \textit{Múltiples vehículos}: Esta es una característica típica del m-TSP visto en la Sección \ref{sec:mTSP}. De esta manera, se cuenta con una flota compuesta por un número finito y concreto de $k$ vehículos. Con estos $k$ vehículos ha de satisfacerse la demanda de todos los clientes del sistema. Además de esto, existe un depósito central, en el que las unidades móviles han de comenzar y finalizar sus rutas de manera obligatoria. Esta propiedad requiere la planificación de, exactamente, $k$ rutas, cada una de las cuales realizada por una sola unidad móvil. Finalmente, cada ruta no puede estar compuesta por más de un número fijo de $q$ nodos.
	
	\item \textit{Asimetría}: Los costes de viaje en el MA-TSP-SPD son asimétricos, lo que quiere decir que el viajar de un cliente $i$ a un cliente $j$ requiere un coste diferente al trayecto inverso. Esta particularidad es propia del ATSP visto a lo largo de este documento y aporta realismo y complejidad al problema.
			
	\item \textit{Recogidas y entregas simultáneas}: Esta propiedad es una adaptación de la característica típica de los problemas VRPSPD vistos en la sección \ref{sec:VRPSPD}. Básicamente, consiste en la existencia de dos tipos de nodos, los \textit{puntos de entrega} y \textit{puntos de recogida}. Como sus propios nombres indican, los primeros son aquellos puntos en los que los clientes abandonan la unidad móvil, mientras que los primeros son aquellos en los que los clientes acceden al vehículo.
	
Asimismo, es importante destacar que, debido a la naturaleza simultánea del problema, un nodo puede pertenecer a ambas categorías. Finalmente, el depósito, haciendo las veces de central de autobuses, puede actuar también como punto de recogida y/o entrega.

Esta propiedad es especialmente importante en muchos problemas del tipo DAR, como por ejemplo, el transporte puerta-a-puerta de personas con necesidades especiales.
\end{enumerate}

Por lo tanto, el MA-TSP-SPD diseñado se trata de un problema de asignación de rutas rico, en el que los costes son asimétricos y en el que el objetivo es encontrar exactamente $k$ rutas, cada una de las cuales con una longitud máxima de $q$ nodos, minimizando el coste total de la solución completa. Al tratarse de un problema del tipo RTSP, el MA-TSP-SPD cuenta con diversas restricciones, aumentando con ello la complejidad de la formulación del problema. Esta característica conduce directamente a una mayor dificultad para resolverlo de forma eficiente, conllevando a un importante reto científico al mismo tiempo. 

\begin{figure}[tb]
	\centering
		\includegraphics[width=1.0\textwidth]{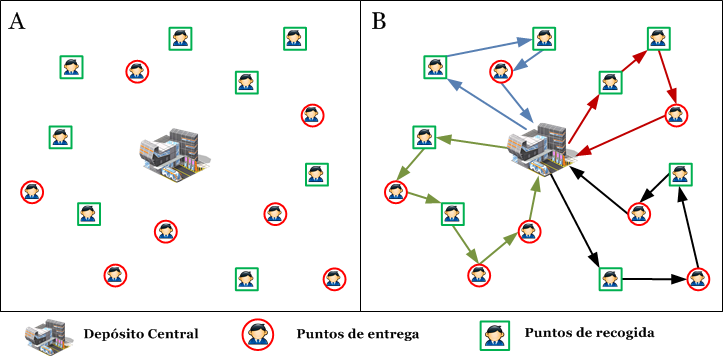}
	\caption{Posible instancia del MA-TSP-SPD compuesta por 15 nodos, y con unos parametros $k$=4 y $q$=5, y posible solución}
	\label{fig:MATSPSPD}
\end{figure}

En la figura \ref{fig:MATSPSPD}(a) se muestra un ejemplo visual de una posible instancia del MA-TSP-SPD propuesto compuesto por 15 nodos y con unos parámetros $k$=4 y $q$=5. Igualmente, en la figura \ref{fig:MATSPSPD}(b) se representa una posible solución factible a dicha instancia.

De esta forma, el MA-TSP-SPD propuesto por el autor puede ser definido como una grafo completo $G= (V,A)$ donde $V= \{v_0,v_1,v_2,\dots,v_p,\}$ es el conjunto de vértices que representan los nodos del sistema. Por otra parte, $A= \{(v_i,v_j): v_i,v_j  \in V,i\neq j \}$ es el conjunto de arcos que simbolizan las interconexiones entre los diversos nodos. Cada arco tiene asociada una distancia $d_{ij}$. A causa de la naturaleza asimétrica del problema, $d_{ij} \neq d_{ji}$. Asimismo, el vértice $v_0$ representa el depósito central, mientras que el resto de vértices representan los puntos a visitar. Para concluir, y con la intención de facilitar la formulación del problema, el conjunto $V$ de puntos puede ser desglosado en dos subconjuntos diferentes, el primero de ellos para los \textit{puntos de recogida} $PN = \{c_1, c_2, \dots , c_n\}$, y el segundo para los \textit{puntos de entrega} $DN = \{c_{n+1}, c_{n+2}, \dots , c_{n+m}\}$, siendo $n+m$ $\geq$ $p$.

Además de esto, se ha utilizado la codificación por permutación para la representación de las soluciones. De esta manera, cada solución es codificada mediante un conjunto de números, los cuales representan las diferentes rutas que componen la solución. Asimismo, se ha utilizado el separador 0 con la intención de distinguir las diversas rutas dentro de una misma solución.

Finalmente, el MA-TSP-SPD propuesto puede formularse matemáticamente de la siguiente manera:

Minimizar:
\begin{equation}
	 \sum_{i = 0}^{p} \ \sum_{j = 0}^{p} \ \sum_{r = 1}^{k} {d_{ij}x^r_{ij}} \label{MA-TSP-SPDeq1}
\end{equation}

Donde:
\begin{equation}
	x^r_{ij} \in \{0,1\}, \ \ \ i,j = 0 \dots p, i \neq j; r = 1 \dots k  \label{MA-TSP-SPDeq2}
\end{equation}

Sujeto a las siguientes restricciones:
\begin{equation}
	\sum_{i = 0}^{p} \ \sum_{r = 1}^{k} {x^r_{ij}=1}, \ \ \ j = 0 \dots p; i \neq j \label{MA-TSP-SPDeq3}
\end{equation}
\begin{equation}
	\sum_{j = 0}^{p} \ \sum_{r = 1}^{k} {x^r_{ij}=1}, \ \ \ i = 0 \dots p; j \neq i  \label{MA-TSP-SPDeq4}
\end{equation}
\begin{equation}
	\sum_{i = 0}^{p} \ \sum_{j = 0}^{p} {x^r_{ij} \leq q}, \ \ \ r = 1 \dots k  \label{MA-TSP-SPDeq7}
\end{equation}
\begin{equation}
	\sum_{j = 0}^{p} \ \sum_{r = 1}^{k} {x^r_{0j}=k} \label{MA-TSP-SPDeq5}
\end{equation}
\begin{equation}
	\sum_{i = 0}^{p} \ \sum_{r = 1}^{k} {x^r_{i0}=k} \label{MA-TSP-SPDeq6}
\end{equation}
\begin{equation}
	\sum_{i = 0}^{p} {x^r_{ij}} - \sum_{l = 0}^{p} {x^r_{jl}} = 0, \ j = 0 \dots p; r = 1 \dots k \label{MA-TSP-SPDeq8}
\end{equation}

\begin{equation}
	\sum_{j = 0}^{p} {x^r_{ij}} - \sum_{l = 0}^{p} {x^r_{li}} = 0, \ i = 0 \dots p; r = 1 \dots k \label{MA-TSP-SPDeq9}
\end{equation}

La primera de las fórmulas simboliza la función objetivo, la cual hay que minimizar, y que representa la suma del coste de todas y cada una de las rutas de la solución. La fórmula \ref{MA-TSP-SPDeq2} describe la naturaleza de la variable binaria $x_{ij}^k$, cuyo valor es 1 si el vehículo $k$ emplea el arco $(c_i, c_j)$, y 0 en caso contrario. Las funciones \ref{MA-TSP-SPDeq3} y \ref{MA-TSP-SPDeq4} garantizan que todos los nodos son visitados, exactamente, en una ocasión. Asimismo, la sentencia \ref{MA-TSP-SPDeq7} asegura que todas las rutas cumplen con la restricción de longitud $q$. Por otro lado, las restricciones \ref{MA-TSP-SPDeq5} y \ref{MA-TSP-SPDeq6} certifican que el número de vehículos que abandonan y vuelven al depósito es el mismo. Igualmente, este valor ha de ser $k$, es decir, la cuantía total de unidades móviles disponibles. Para finalizar, el correcto flujo de cada ruta se garantiza gracias a las funciones \ref{MA-TSP-SPDeq8} y \ref{MA-TSP-SPDeq9}.


Como ya se ha especificado en la sección \ref{sec:practicas} de esta tesis, la existencia de un conjunto de pruebas, o benchmark, para resolver un problema de optimización es un factor crucial. En el aludido trabajo \cite{osaba2015HAIS} se presenta un conjunto de pruebas para el MA-TSP-SPD, el cual es una modificación del ATSP Benchmark que puede encontrarse en la librería TSPLib \cite{TSPLibrary}. 

En total se han diseñado 19 instancias diferentes para el MA-TSP-SPD propuesto, las cuales tienen una suma de nodos que oscila entre 17 y 443. El primer nodo de cada instancia corresponde al depósito central, mientras que cada punto restante cuenta con un parámetro extra denominado $tipo_i$ cuyo valor indica si el punto en cuestión es de entrega o recogida. Este parámetro ha sido establecido de la siguiente forma:
\[ tipo_i = \textrm{punto de entrega}, \ \ \ \forall i \in \{1, 3, 5, \dots, n\} \] 
\[ tipo_i = \textrm{punto de recogida}, \ \ \ \forall i \in \{2, 4, 6, \dots, n\} \]
Además de esto, el número de vehículos disponibles se ha fijado en  $k$=4. Por otro lado, se ha decidido que la longitud máxima de cada ruta sea $q = int(p/4)+1$, donde $p$ es la suma total de nodos de la instancia. Finalmente, la localización geográfica y los costes de los trayectos permanecen igual que en la instancia original.

Con la intención de aumentar la replicabilidad de este estudio, el banco de pruebas descrito se encuentra a disposición en la página personal del autor de esta tesis y bajo demanda.


Como bien se ha indicado al comienzo de esta sección, el banco de pruebas propuesto para este nuevo problema ha sido tratado por la meta-heurística presentada en esta tesis, el Golden Ball. La parametrización utilizada para el GB se representa en la tabla \ref{tab:summary}. Con esta experimentación se ha querido demostrar cómo el GB es una técnica perfectamente capaz de afrontar problemas de alta complejidad, los cuales son formulados para hacer frente a situaciones del transporte propias del mundo real.

\begin{table}[tb]	
	\centering
	\scalebox{0.8}{
		\begin{tabular}{| m{8cm}| m{5cm} |}
		  \hline 
			Número de equipos & 4\\[3pt]
			\hline 
			Número de jugadores por equipo & 12\\[3pt]
			\hline
			Número de entrenamientos sin mejora para \textit{entrenamiento personalizado} & 6\\[3pt]
			\hline 
			Número de entrenamientos sin mejora para \textit{transferencia especial} & 12\\[3pt]
			\hline
			Funciones de entrenamiento convencionales & 2-opt, Vertex Insertion (intra-ruta e inter-ruta) y Swapping \\[3pt]
			\hline 
			Funcion de entrenamiento personalizado & HRX\\[3pt]
			\hline
		\end{tabular}
		}
	\caption{Resumen de la parametrización empleada para el $GB$}
	\label{tab:summary}
\end{table}

Todas las pruebas se realizaron en un ordenador portátil Intel Core i5 – 2410, con 2.30 GHz y una memoria RAM de 4 GB. Se utilizaron todas las instancias descritas anteriormente, cuyos nombres portan un número que representa la cantidad de nodos que poseen. Se realizaron 30 ejecuciones por cada instancia. En la tabla \ref{tab:MATSPSPD} se muestran los resultados obtenidos, haciendo uso para este propósito de cinco parámetros diferentes: los resultados promedio, con sus desviaciones típicas relacionadas, las medianas, los rangos intercuartílicos, y los tiempos promedio (en segundos).
\begin{table}[tb]
	\centering
	\scalebox{0.75}{
		\begin{tabular}{| l || r | r || l | r || r |}
			\hline Instancia & Media & Desv. T. & Mediana & R. I. & Tiempo\\  
			\hline 
			MA-TSP-SPD\_br17 & 66.1 & 1.4 & 65 & 2.2 & 0.71\\
			MA-TSP-SPD\_ftv33 & 1652.2 & 88.4 & 1575 & 149.2 & 1.35 \\
			MA-TSP-SPD\_ftv35 & 1828.8 & 93.3 & 1765 & 123.5 & 1.46 \\
			MA-TSP-SPD\_ftv38 & 1883.7 & 77.5 & 1813 & 120.7 & 1.64 \\
			MA-TSP-SPD\_p43 & 5888.5 & 19.2 & 5873 & 24.7 & 1.69 \\
			MA-TSP-SPD\_ftv44 & 2063.5 & 142.2 & 1896 & 260.7 & 1.83\\
			MA-TSP-SPD\_ftv47 & 2214.3 & 517.2 & 2235 & 186.0 & 2.23\\
			MA-TSP-SPD\_ry48p & 18160.2 & 604.7 & 17784 & 601.5 & 3.09\\
			MA-TSP-SPD\_ft53 & 8614.5 & 444.8 & 8303 & 655.0 & 4.21\\
			MA-TSP-SPD\_ftv55 & 2239.6 & 141.6 & 2204 & 156.7 & 3.77\\
			MA-TSP-SPD\_ftv64 & 2505.9 & 145.1 & 2385 & 196.0 & 3.31\\
			MA-TSP-SPD\_ftv70 & 2720.5 & 136.7 & 2598 & 256.7 & 3.75\\
			MA-TSP-SPD\_ft70 & 44460.3 & 809.9 & 43717 & 1199.0 & 4.49\\
			MA-TSP-SPD\_kro124p & 48277.6 & 2036.4 & 46407 & 3028.0 & 13.41\\
			MA-TSP-SPD\_ftv170 & 5482.5 & 309.6 & 5261 & 320.7 & 21.12\\
			MA-TSP-SPD\_rbg323 & 1851.3 & 59.8 & 1797 & 112.7 & 72.54\\
			MA-TSP-SPD\_rbg358 & 1856.3 & 72.6 & 1800 & 122.2 & 81.29\\
			MA-TSP-SPD\_rbg403 & 2859.2 & 50.6 & 2807 & 88.0 & 87.25\\
			MA-TSP-SPD\_rbg443 & 3121.4 & 55.7 & 3110 & 91.2 & 136.59 \\
			\hline
		\end{tabular}
		}
		\caption{Resultados obtenidos por el GB para el MA-TSP-SPD propuesto. Por cada instancia se muestran la media aritmética, desviación típica, mediana, rango intercuartílico y tiempo medio.}
	\label{tab:MATSPSPD}
\end{table}

Por otra parte, en la tabla \ref{tab:best} se muestra el fitness de la mejor solución encontrada por el GB para cada instancia. Igualmente, se representa la cantidad de evaluaciones realizadas por la función de evaluación para alcanzar dicha solución, y el tiempo de ejecución empleado. Debido a que es la primera vez que se trata el problema MA-TSP-SPD en la literatura, estas soluciones son consideradas las mejores encontradas hasta el momento.

\begin{table}[tb]
	\centering
	\scalebox{0.78}{
		\begin{tabular}{| l | r | r | r || l | r | r | r |}
			\hline Instancia & Fitness & Eval. & T. & Instancia & Fitness & Eval. & T.\\ 
			\hline 
			MA-TSP-SPD\_br17 & 65 & 40 & 0.65  & MA-TSP-SPD\_ftv64 & 2323 & 7867 & 3.82 \\
			MA-TSP-SPD\_ftv33 & 1515 & 3783 & 1.72 & MA-TSP-SPD\_ftv70 & 2540 & 16719 & 6.35\\
			MA-TSP-SPD\_ftv35 & 1703 & 1939 & 1.71  & MA-TSP-SPD\_ft70 & 43563 & 14190 & 5.26\\
			MA-TSP-SPD\_ftv38 & 1800 & 6351 & 2.28  & MA-TSP-SPD\_kro124p & 45991 & 31664 & 13.85\\
			MA-TSP-SPD\_p43 & 5850 & 3421 & 1.53 & MA-TSP-SPD\_ftv170 & 5054 & 30779 & 21.20\\
			MA-TSP-SPD\_ftv44 & 1872 & 8887 & 3.57 & MA-TSP-SPD\_rbg323 & 1795 & 62850 & 80.72\\
			MA-TSP-SPD\_ftv47 & 2202 & 5047 & 2.07  & MA-TSP-SPD\_rbg358 & 1773 & 87121 & 112.56\\
			MA-TSP-SPD\_ry48p & 17394 & 5157 & 2.14 & MA-TSP-SPD\_rbg403 & 2801 & 44375 & 81.67\\
			MA-TSP-SPD\_ft53 & 7901 & 8028 & 3.97  & MA-TSP-SPD\_rbg443 & 3044 & 86175 & 163.57\\
			MA-TSP-SPD\_ftv55 & 2001 & 5757 & 2.92 & & & & \\
			\hline
		\end{tabular}
		}
		\caption{Mejor soluciones encontradas por el GB para el problema MA-TSP-SPD.}
	\label{tab:best}
\end{table}

\section{Aplicación del Golden Ball a un nuevo RVRP}
\label{sec:AplicNueva}

Los problemas con los que se ha trabajado en el apartado anterior son problemas del tipo RTSP. Estos problemas cuentan con cierto atractivo científico, como bien se ha destacado, pero son inferiores en este aspecto a los problemas RVRP. Estos últimos tienen la capacidad de hacer frente a situaciones reales con mayor eficacia, debido a su mayor facilidad para adaptarse a situaciones complejas de transporte. Con todo esto, en esta sección se pretende describir una situación de transporte real, la cual se modelizará como un RVRP. Tras esto, se presentará un benchmark adaptado al problema propuesto, abordándolo después con la meta-heurística presentada en esta tesis.


La situación real que se ha decidido tratar en este apartado está relacionada con la distribución de prensa. Concretamente, el objeto de estudio será una empresa mediana de distribución de periódicos de ámbito provincial. La empresa en cuestión cuenta con ciertos principios, sobre los que basan su planificación logística. El primero de estos principios es el de tratar las ciudades, o pueblos, como unidades separadas, obligando a cada vehículo que entra en una ciudad, a servir a todos y cada uno de los clientes de dicha ciudad, o dicho pueblo. Si el vehículo no tiene la capacidad suficiente para satisfacer la demanda de todos los clientes emplazados en un mismo centro urbano, no podrá entrar en él. Por otro lado y debido al compromiso medioambiental de la empresa, ésta cuenta con un sistema de reciclaje de papel, consistente en reciclar los periódicos que no se hayan vendido el día anterior. De esta manera, y como puede predecirse, los vehículos no solo tendrán que atender demandas de entrega, si no que tendrán que recoger en cada punto de visita aquellos periódicos a los que no se les ha dado salida.

Además de esto, a la hora de planificar las rutas que las unidades móviles van a completar, la empresa tiene en cuenta ciertos factores de obligatoria consideración. El primero de ellos está relacionado con las horas en las que se realizan los trayectos. La distribución se realizará diariamente, durante las mañanas, de 6:00am a 14:00pm. Entre este horizonte temporal existirá cierto rango, entre las 8:00am y 10:00am, considerado ``hora punta", en el que los costes de los trayectos de un punto a otro serán mayores que si se realizasen fuera de dicho rango. Además de esto, y con la intención de respetar todas y cada una de las normas de tráfico, no se podrán realizar trayectos por vías que lo prohíban.


Con todas estas premisas tenidas en cuenta, se ha propuesto un problema RVRP que cuenta con las siguientes características:

\begin{enumerate}
		
	\item \textit{Asimetría}: Esta propiedad es la misma que se ha visto en el apartado anterior para el MA-TSP-SPD propuesto. 
	
	\item \textit{Clusterizado}: Este atributo hace que los distintos puntos de visita que conforman el sistema sean agrupados en diferentes conjuntos, o clústeres. En este caso, cada clúster representara un núcleo urbano, es decir, una misma ciudad, o un mismo pueblo. El requisito que debe cumplir una ruta es el siguiente: si un vehículo atiende la demanda de un cliente perteneciente a un conjunto cualquiera, dicho vehículo deberá servir a todos y cada uno de los clientes de tal clúster. En caso de que una unidad móvil no tenga la capacidad suficiente para atender a todos los clientes de una ciudad, o pueblo, no podrá utilizarse para el reparto en ese núcleo urbano. Esta particularidad ya ha sido anteriormente utilizada en una gran variedad de estudios \cite{clustered1,clustered2}
			
	\item \textit{Recogidas y entregas simultaneas}: Al igual que la primera de las propiedades, esta característica tiene la misma naturaleza que la descrita en la Sección \ref{sec:AplicPrevios} para el MA-TSP-SPD presentado.
	
	\item \textit{Costes de trayecto variables}: Como puede resultar lógico, en situaciones reales de transporte el trayecto entre un mismo punto y otro no conlleva siempre el mismo coste, ya sea temporal o económico. En muchas ocasiones este coste está sujeto a variables externas, como la hora del día, el tráfico o el clima. En este problema se ha intentado recrear dicha situación para añadirle mayor realismo al mismo. Para ello, se ha establecido una jornada laboral comprendida entre las 6:00am, y las 14:00pm, y se han asignado dos periodos temporales dentro de dicha jornada, llamados \textit{hora punta} y \textit{hora valle}. La hora punta será un periodo de dos horas, entre las 8:00am y las 10:00am, y todos los trayectos realizados en dicha ventana temporal requerirán un esfuerzo mayor, suponiendo un tráfico menos fluido que en las horas valle. Este mismo atributo ha sido previamente utilizado en la literatura en contadas ocasiones \cite{variableCosts}.
	
	\item \textit{Caminos prohibidos}: En el mundo real es común encontrarse con vías de un solo sentido, en las que la circulación en determinada dirección está prohibida. Igualmente, existen travesías peatonales, en las que los vehículos tienen vetada la entrada. Con la intención de recrear esta habitual situación, el problema contará con ciertos arcos $(i,j)$ que no podrán ser utilizados en la solución final. Una filosofía similar a esta se ha utilizado con anterioridad en algunos estudios de la literatura \cite{forbidden}.
\end{enumerate}

Con todo esto, se podría decir que se trata de un problema de asignación de rutas a vehículos asimétrico y clusterizado, con recogidas y entregas simultaneas, costes de trayecto variables y caminos prohibidos (\textit{Asymmetric Clustered Vehicle Routing Problem with Simultaneous Pickups and Deliveries, Variable Costs and Forbidden Paths}, AC-VRP-SPDVCFP), el cual se trata de un problema de asignación de rutas rico, en el que los costes son asimétricos, los nodos están congregados en conjuntos, y en el que el objetivo es encontrar un número de rutas que respeten la naturaleza de los clústeres y no viajen por ningún camino prohibido, minimizando, además, el coste total de la solución completa. Al tratarse de un problema RVRP, el AC-VRP-SPDVCFP cuenta con múltiples restricciones, lo que hace que la complejidad del problema aumente de forma extraordinaria, como ya se ha mencionado anteriormente. 

\begin{figure}[tb]
	\centering
		\includegraphics[width=0.8\textwidth]{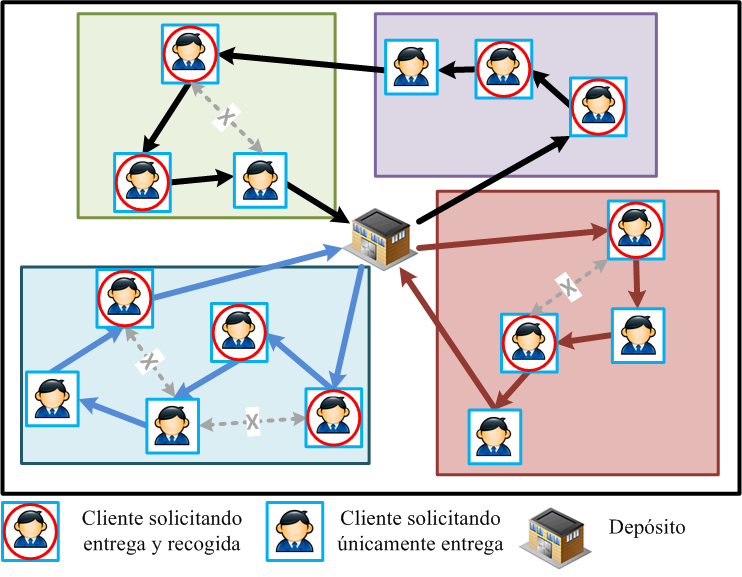}
	\caption{Posible instancia del AC-VRP-SPDVCFP compuesta por 14 nodos y posible solución a la misma}
	\label{fig:ACVRPSPDVCFP}
\end{figure}

A modo ilustrativo, en la figura \ref{fig:ACVRPSPDVCFP} se representa un ejemplo de una posible instancia del AC-VRP-SPDVCFP compuesta por 16 nodos. Asimismo, en esta misma imagen se muestra una posible solución factible a tal hipotética situación.

De esta forma, el AC-VRP-SPDVCFP puede ser definido como un grafo completo $G= (V,A)$ donde $V= \{v_0,v_1,v_2,\dots,v_n,\}$ representa el conjunto de nodos del sistema. Por otro lado $A= \{(v_i,v_j): v_i,v_j  \in V,i\neq j \}$ es el conjunto de arcos que representan las conexiones entre nodos. Cada arco tiene un coste $d_{ij}$ asignado. Debido a la asimetría $d_{ij} \neq d_{ji}$. Además de esto, el coste de los caminos prohibidos se establecerá en $10^{10}$, de esta manera es seguro que no aparecerán en la solución final. Asimismo, el vértice $v_0$ representa el depósito, mientras que el resto de nodos son los clientes a visitar. Adicionalmente, $V$ está dividido en $k+1$ subconjuntos mutuamente excluyentes, $C=\{V_0,V_1,...,V_k\}$, uno por cada clúster. Estos subconjuntos cumplen las siguientes condiciones:

\begin{equation}
	 V = V_0 \cup V_1 \cup ... \cup V_k \label{clustered1}
\end{equation}
\begin{equation}
	 V_a \cap V_b = \emptyset, \ \ \ a,b \in {0,1,...,k}, a \neq b \label{clustered2}
\end{equation}

Es aconsejable indicar que $V_0$ contiene únicamente el nodo $v_0$, el cual representa el depósito. Los $n$ nodos restantes son divididos proporcionalmente en los $k$ clústeres restantes. Además de esto, cada nodo $i$ tiene asignadas dos demandas, una relacionada con la entrega $d_i$ $>$ 0, y la otra con la recogida $p_i$ $\geq$ 0. Finalmente, al igual que se ha empleado en otra ocasiones, la codificación por permutación ha sido utilizada para representar las soluciones del problema.


Con todo esto, el AC-VRP-SPDVCFP puede ser matemáticamente formulado de la siguiente manera. Es importante tener en cuenta que la variable $y_{ij}$ denota la demanda de recogida en los clientes hasta el nodo $i$ (incluyendo el mismo $i$) y transportado en el arco $(i,j)$. Por otro lado, $z_{ij}$ representa la cantidad de productos que hay que transportar después del nodo $i$ y transportados en el camino $(i,j)$ \cite{montane2006tabu}. Igualmente, la variable binaria $w_o^r$ adquiere el valor 1 si un vehículo $r$ entra en el clúster $o$, y 0 en otro caso. Finalmente, la variable binaria $x_{ij}^r$ toma el valor 1 si la unidad móvil $r$ utiliza el arco $(i, j)$, y 0 en caso contrario.

Minimizar:
\begin{equation}
	 \sum_{i = 0}^{n} \ \sum_{j = 0}^{n} \ \sum_{r = 1}^{k} {d_{ij}x^r_{ij}} \label{ACVRPSPDVCFPeq1}
\end{equation}

Donde:
\begin{equation}
	x^r_{ij} \in \{0,1\}, \ \ \ i,j = 0, \dots ,n, i \neq j; r = 1 \dots k  \label{ACVRPSPDVCFPeq2}
\end{equation}
\begin{equation}
	w^r_o \in \{0,1\}, \ \ \ r = 1, \dots ,k, o = 1,...,c \label{ACVRPSPDVCFPeq1_1}
\end{equation}
\begin{equation}
	y_{ij} \geq 0, \ \ \ i,j = 0, \dots ,n \label{ACVRPSPDVCFPeq1_2}
\end{equation}
\begin{equation}
	z_{ij} \geq 0, \ \ \ i,j = 0, \dots ,n  \label{ACVRPSPDVCFPeq1_3}
\end{equation}

Sujeto a:
\begin{equation}
	\sum_{i = 0}^{n} \ \sum_{r = 1}^{k} {x^r_{ij}=1}, \ \ \ j = 0, \dots ,n; i \neq j \label{ACVRPSPDVCFPeq3}
\end{equation}
\begin{equation}
	\sum_{j = 0}^{n} \ \sum_{r = 1}^{k} {x^r_{ij}=1}, \ \ \ i = 0, \dots ,n; j \neq i  \label{ACVRPSPDVCFPeq4}
\end{equation}
\begin{equation}
	\sum_{j = 0}^{n} \ \sum_{r = 1}^{k} {x^r_{0j}=k} \label{ACVRPSPDVCFPeq5}
\end{equation}
\begin{equation}
	\sum_{i = 0}^{n} \ \sum_{r = 1}^{k} {x^r_{i0}=k} \label{ACVRPSPDVCFPeq6}
\end{equation}
\begin{equation}
	\sum_{i = 0}^{n} {x^r_{ij}} - \sum_{l = 0}^{n} {x^r_{jl}}, \ j = 0, \dots ,n; r = 1 \dots k \label{ACVRPSPDVCFPeq7}
\end{equation}
\begin{equation}
	\sum_{j = 0}^{n} {x^r_{ij}} - \sum_{l = 0}^{n} {x^r_{li}}, \ i = 0, \dots ,n; r = 1 \dots k \label{ACVRPSPDVCFPeq8}
\end{equation}

\begin{equation}
	\sum_{i = 0}^{n} {y_{ji}} - \sum_{i = 0}^{n} {y_{ij}} = p_j, \ \ \ j = 0, \dots ,n  \label{ACVRPSPDVCFPeq9}
\end{equation}
\begin{equation}
	\sum_{i = 0}^{n} {z_{ji}} - \sum_{i = 0}^{n} {z_{ij}} = d_j, \ \ \ j = 0, \dots ,n  \label{ACVRPSPDVCFPeq10}
\end{equation}
\begin{equation}
	y_{ij} + z_{ij} \leq Q \sum_{r = 1}^{k} {x^r_{ij}}, \ \ \ i,j = 0, ..., n  \label{ACVRPSPDVCFPeq11}
\end{equation}


\begin{equation}
	\sum_{i = 0}^{n} \ \sum_{r = 1}^{k} {d_{ij}x^r_{ij}<10^{10}}, \ \ \ j = 0, \dots ,n; i \neq j \label{ACVRPSPDVCFPeq12}
\end{equation}
\begin{equation}
	\sum_{j = 0}^{n} \ \sum_{r = 1}^{k} {d_{ij}x^r_{ij}<10^{10}}, \ \ \ i = 0, \dots ,n; j \neq i  \label{ACVRPSPDVCFPeq13}
\end{equation}

\begin{equation}
\sum_{r = 1}^{k} {w^r_{o}} = 1 \ \ \ o = 1,...,c  \label{ACVRPSPDVCFPeq14}
\end{equation}

La primera fórmula representa la función objetivo, la cual es la suma de los costes de todas las rutas de la solución y cuyo valor debe ser minimizado. Las fórmulas (\ref{ACVRPSPDVCFPeq2}), (\ref{ACVRPSPDVCFPeq1_1}), (\ref{ACVRPSPDVCFPeq1_2}) y (\ref{ACVRPSPDVCFPeq1_3}) denotan la naturaleza de las variables $x_{ij}^r$, $w^r_o$, $y_{ij}$ y $z_{ij}$, respectivamente. Las condiciones (\ref{ACVRPSPDVCFPeq3}) y (\ref{ACVRPSPDVCFPeq4}) aseguran que todos los nodos son visitados exactamente una vez. Por otro lado, las restricciones (\ref{ACVRPSPDVCFPeq5}) y (\ref{ACVRPSPDVCFPeq6}) garantizan que el número total de vehículos que abandonan el depósito es igual que el número de vehículos que acceden a él. Además de esto, el correcto flujo de cada ruta está seguro gracias a las restricciones (\ref{ACVRPSPDVCFPeq7}) y (\ref{ACVRPSPDVCFPeq8}).

Adicionalmente, las restricciones (\ref{ACVRPSPDVCFPeq9}) y (\ref{ACVRPSPDVCFPeq10}) certifican que el flujo de entregas y recogidas, respectivamente, son llevadas a cabo de manera correcta. Ambas fórmulas garantizan que las demandas son satisfechas para cada cliente. Asimismo, la fórmula (\ref{ACVRPSPDVCFPeq11}) asegura que la capacidad de cada unidad móvil no será nunca sobrepasada, y establece a su vez que tanto las recogidas como las entregas serán realizadas utilizando arcos incluidos en la solución \cite{montane2006tabu}.

De igual manera, las restricciones (\ref{ACVRPSPDVCFPeq12}) y (\ref{ACVRPSPDVCFPeq13}) avalan que todo trayecto desde un nodo $i$ a un nodo $j$ tendrá un coste menor que $10^{10}$. De esta manera, queda asegurado que los caminos prohibidos no formarán parte de la solución. Finalmente, la ecuación (\ref{ACVRPSPDVCFPeq14}) garantiza que cada clúster es visitado únicamente por un vehículo. Esta restricción, unida a las anteriormente mencionadas (\ref{ACVRPSPDVCFPeq3}) y (\ref{ACVRPSPDVCFPeq4}), asegura que todos los puntos del mismo clúster son visitados por la misma unidad móvil.

\begin{figure}[tb]
	\centering
		\includegraphics[width=0.8\textwidth]{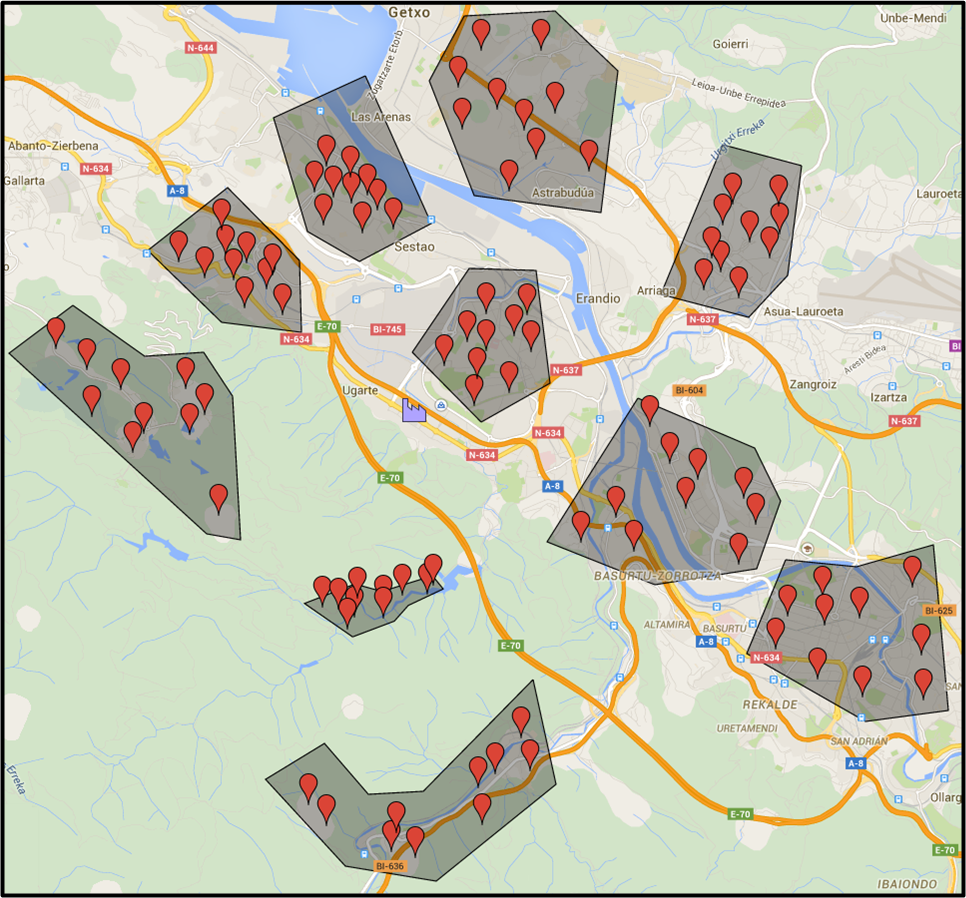}
	\caption{Localización geográfica del depósito, clientes y clústeres en la provincia de Bizkaia. Fuente: \textit{Google Maps}, accedido en abril del 2015.}
	\label{fig:map}
\end{figure}

Este RVRP propuesto por el autor de esta tesis nunca antes ha sido tratado en la literatura. Es por esto por lo que no existen conjuntos de prueba disponibles en la comunidad científica específicamente para él. Para hacer frente a este inconveniente, en este documento se propone un benchmark comprendido por 15 instancias. Estos casos de prueba están compuestos por 50-100 nodos, representando cada uno de ellos un cliente. Cada cliente estará colocado en una posición geográfica real, localizada en la provincia de Bizkaia. Además de esto, el número máximo de clústeres se ha establecido en diez, existiendo también instancias con cinco y ocho. En la figura \ref{fig:map} se muestra un mapa con la localización del depósito, los clientes y los clústeres. Este mapa se ha realizado haciendo uso de la tecnología \textit{Google Maps}.

Los clústeres han sido organizados en orden de aparición. Es decir, los nodos 1-10 componen el conjunto 1, los clientes 11-20 completan el grupo 2, y así sucesivamente. Es recomendable apuntar que todos los grupos tienen el mismo número de nodos. Además de esto, cada cliente tiene asociadas dos tipos de demandas, una relacionada con la entrega de periódicos $d_i$, y la otra con la recogida $p_i$. La asignación de estas demandas se ha llevado a cabo siguiendo el método que se describe a continuación:

\begin{equation}
	  d_i = 10, p_i = 5, \ \ \ \forall i \in \{1, 5, 9, \dots , 97\} \label{demands1}
\end{equation}
\begin{equation}
	  d_i = 10, p_i = 0, \ \ \ \forall i \in \{2, 6, 10, \dots , 98\} \label{demands2}
\end{equation}
\begin{equation}
	  d_i = 5, p_i = 3, \ \ \ \forall i \in \{3, 7, 11, \dots , 99\} \label{demands3}
\end{equation}
\begin{equation}
	  d_i = 5, p_i = 0, \ \ \ \forall i \in \{4, 8, 12, \dots , 100\} \label{demands4}
\end{equation}
donde $d_0$=0 y $p_0$=0, debido a que $v_0$ está considerado como el depósito. Además de esto, el coste de viajar desde un cliente $i$ cualquiera a otro cliente $j$ se ha establecido siguiendo el procedimiento propuesto en el algoritmo \ref{alg:distances}. Es importante indicar que estos costes han sido asignados al periodo ``valle". Estos gastos se ven aumentados en caso de ser realizados en periodo de ``hora punta", siguiendo el proceso representado en el algoritmo \ref{alg:distances2}.

\begin{algorithm}[htb]
	 \SetAlgoLined
		\For{$\forall i \in \{1,2, \dots , 99\}$}{
			\For{$\forall j \in \{i+1, \dots , 100\}$}{
				
				$d_{ij}$ = \textrm{DistanciaEuclídea}(i,j)\;
				
				\eIf{$j$ \textrm{es un número par}}{
					$d_{ji}$ = \textrm{DistanciaEuclídea}(j,i) * 1.2 \;
				}{
				  $d_{ji}$ = \textrm{DistanciaEuclídea}(j,i) * 0.8 \;
				}
		}}
   \caption{Procedimiento para la asignación de los costes en ``hora valle".}
	 \label{alg:distances}
\end{algorithm}

\begin{algorithm}[htb]
	 \SetAlgoLined
		\For{$\forall i \in \{1,2, \dots , 99\}$}{
			\For{$\forall j \in \{i+1, \dots , 100\}$}{
				
				$d_{ij}$ = $d_{ij}$ * 1.3\;
				\eIf{$j$ \textrm{es un número par}}{
					$d_{ji}$ = $d_{ji}$ * 1.2 \;
				}{
				  $d_{ji}$ = $d_{ji}$ * 1.4 \;
				}
		}}
   \caption{Procedimiento para la asignación de los costes en ``hora punta".}
	 \label{alg:distances2}
\end{algorithm}

Finalmente, dependiendo de la instancia, algunos caminos serán escogidos en cada clúster para catalogarlos como ``prohibidos". En la tabla \ref{tab:instances} se muestra un resumen de las características de cada de uno de los casos elaborados para este benchmark. Con la intención de facilitar la compresión de dicha tabla, es recomendable tener en cuenta las siguiente premisas: Los casos Osaba\_50\_1\_1 y Osaba\_50\_1\_2 están comprendidos por cinco grupos, en este caso los \{1, 3,..., 9\}.  De la misma forma, Osaba\_50\_2\_1 y Osaba\_50\_2\_2 están compuestos por los clústeres \{2, 4,..., 10\}. Por otro lado, los conjuntos que completan los casos Osaba\_50\_1\_3 y Osaba\_50\_1\_4 están constituidos por cinco nodos. En estos dos ejemplos, estos nodos son los cinco primeros de cada clúster. El caso contrario sucede en las instancias Osaba\_50\_2\_3 y Osaba\_50\_2\_4, donde los diez grupos están compuestos por los cinco últimos clientes de cada uno de ellos. Para concluir, en la elaboración de todos los ejemplos Osaba\_80\_X se ha hecho uso de los primeros ocho clústeres, u ocho primeros nodos (dependiendo de la instancia).

\begin{table}[htb]
	\centering
	\setlength{\tabcolsep}{2pt}
	\scalebox{0.8}{
		\begin{tabular}{| l || r | r | r | r |}
			\hline Instancia & Nodos & Clústeres & Capacidad  & Caminos \\ 
			\hline Osaba\_50\_1\_1 & 50 & 5 & 240 & 5\\
			Osaba\_50\_1\_2 & 50 & 5 & 160 & 10 \\
			Osaba\_50\_1\_3 & 50 & 10 & 240 & 5 \\
			Osaba\_50\_1\_4 & 50 & 10 & 160 & 10 \\
			\hline
			Osaba\_50\_2\_1 & 50 & 5 & 240 & 5\\
			Osaba\_50\_2\_2 & 50 & 5 & 160 & 10 \\
			Osaba\_50\_2\_3 & 50 & 10 & 240 & 5 \\
			Osaba\_50\_2\_4 & 50 & 10 & 160 & 10 \\
			\hline
			Osaba\_80\_1 & 80 & 8 & 240 & 5 \\
			Osaba\_80\_2 & 80 & 8 & 160 & 10 \\
			Osaba\_80\_3 & 80 & 10 & 240 & 5 \\
			Osaba\_80\_4 & 80 & 10 & 160 & 10 \\
			\hline
			Osaba\_100\_1 & 100 & 10 & 140 & 5 \\
			Osaba\_100\_2 & 100 & 10 & 260 & 10 \\
			Osaba\_100\_3 & 100 & 10 & 320 & 10 \\
			\hline 
		\end{tabular}
	}
	\caption{Resumen del benchmark propuesto para el problema AC-VRP-SPDVCFP. \textit{Caminos} representa el número de arcos prohibidos por cada clúster.}
	\label{tab:instances}
\end{table}

Al igual que en el aparatado anterior, el problema planteado para este caso real ha sido tratado por el método desarrollado en este trabajo. La parametrización utilizada para el GB se representa en la tabla \ref{tab:summaryAC-VRP-SPDVCFP}. Es conveniente aclarar que para este problema las funciones de entrenamiento han sido adaptas al problema, respetando en todo momento las capacidades y la naturaleza de los clústeres, y nunca generando soluciones no factibles. En este caso, y con la intención de reforzar la hipótesis de que el GB es una técnica prometedora también para los problemas del tipo RVRP, se van a comparar los resultados obtenidos por el GB con los logrados por dos diferentes técnicas, un algoritmo evolutivo basado en mutaciones (EA), y un algoritmo de recocido simulado evolutivo (ESA) \cite{ESA}. La parametrización de estos dos algoritmos puede observarse en la tabla \ref{tab:ParametrizationAC-VRP-SPDVCFP}.

\begin{table}[htb]	
	\centering
	\scalebox{0.8}{
		\begin{tabular}{| m{8cm}| m{5cm} |}
		  \hline 
			Número de equipos & 5\\[3pt]
			\hline 
			Número de jugadores por equipo & 20\\[3pt]
			\hline
			Número de entrenamientos sin mejora para \textit{entrenamiento personalizado} & 6\\[3pt]
			\hline 
			Número de entrenamientos sin mejora para \textit{transferencia especial} & 12\\[3pt]
			\hline
			Funciones de entrenamiento convencionales & Vertex Insertion (intra-ruta) y Swapping (intra-ruta)\\[3pt]
			\hline 
			Función de entrenamiento personalizado & HRX (a nivel de cluster)\\[3pt]
			\hline
		\end{tabular}
		}
	\caption{Resumen de la parametrización empleada para el $GB$}
	\label{tab:summaryAC-VRP-SPDVCFP}
\end{table}

\begin{table}[htb]
	\centering
	\setlength{\tabcolsep}{2pt}
	\scalebox{0.7}{
		\begin{tabular}{| l | r || l | r |}
		  \hline  
		  \multicolumn{2}{|c||}{EA} & \multicolumn{2}{c|}{ESA}\\
			\hline
			\hline Parámetro & Valor & Parámetro & Valor \\ 
			\hline
			Tamaño de población & 100 & Tamaño de población & 100 \\
			Función de mutación & Vertex Insertion & Función de sucesores & Vertex Insertion \\ 
			Prob. de mutación & 1.0  & Temperatura & $-sup \Delta f/ ln(p) $\\
		  Func. de supervivientes & 70\% Elitista - 30\% Aleatoria & Constante de enfriamiento & 0.95 \\ 
			\hline 
		\end{tabular}
	}
	\caption{Parametrización del EA y ESA para el AC-VRP-SPDVCFP propuesto, donde $-sup \Delta f$ es la diferencia de la función objetivo entre la mejor y la peor solución en la poblacion initial, y $p$=0.95.}
	\label{tab:ParametrizationAC-VRP-SPDVCFP}
\end{table}

Para llevar a cabo la experimentación se han utilizado todas las instancias descritas anteriormente, y se han realizado 20 ejecuciones para cada una de ellas. En la tabla \ref{tab:ResultsAC-VRP-SPDVCFP} se muestran los resultados obtenidos, haciendo uso para este propósito de tres parámetros diferentes: los resultados promedio, con sus desviaciones típicas relacionadas, y los tiempos promedio (en segundos).

\begin{table}[htb]
	\centering
	\setlength{\tabcolsep}{2pt}
	\scalebox{0.7}{
		\begin{tabular}{| l || r r r || r r r || r r r |}
		  \hline  
		  \multicolumn{1}{|c||}{Instancia} & \multicolumn{3}{c||}{GB} & \multicolumn{3}{c||}{ESA} & \multicolumn{3}{c|}{EA}\\
			\hline
			\hline Nombre & Media & Desv. Est. & T. & Media & Desv. Est. & T. & Media & Desv. Est. & T. \\ 
			\hline Osaba\_50\_1\_1 & \textbf{50659.3} & 230.4 & 49.2 &  51615.9 & 485.3 & 46.3 & 51574.8 & 643.8 & 46.9 \\
			Osaba\_50\_1\_2 & \textbf{56250.5} & 210.6 & 46.2 & 56944.8 & 455.7 &  45.9 & 57002.5  & 537.4 & 47.2\\
			Osaba\_50\_1\_3 & \textbf{71539.2} & 1639.2 & 48.1 & 72306.7 & 1464.3 & 43.5  & 72493.4 & 1396.5 & 44.6\\
			Osaba\_50\_1\_4 & \textbf{78630.5} & 1068.3 & 47.5 & 79156.5 & 1776.3 & 44.7 & 79206.7 & 1582.5 & 42.3\\
			\hline
			Osaba\_50\_2\_1 & \textbf{49358.4} & 368.3 & 50.0 & 49745.7 & 544.6 & 45.3 & 48142.8 & 745.3 & 46.9\\
			Osaba\_50\_2\_2 & \textbf{54369.2} & 599.3 & 48.3 & 54863.4 & 616.2 &  47.5 & 55012.5 & 827.9 & 47.1\\
			Osaba\_50\_2\_3 & \textbf{69474.0} & 2512.3 & 48.2 & 71366.3 & 2902.7 & 45.6 & 71279.3 & 3012.4 & 46.8\\
			Osaba\_50\_2\_4 & \textbf{80740.2} & 1430.2 & 47.2 & 81126.2 & 1680.4 & 42.1 & 81288.8 & 1884.0 & 44.2\\
			\hline
			Osaba\_80\_1 & 81739.5 & 1149.2 & 84.8 & \textbf{80839.0} & 1814.9 &  80.0 & 81786.4 & 2011.1 & 83.5\\
			Osaba\_80\_2 & \textbf{89163.6} & 900.2 & 85.6 & 89997.6 & 1122.6 &  83.7 & 90093.5 & 1000.5 & 84.7\\
			Osaba\_80\_3 & 89500.2 & 1390.0 & 87.1 & \textbf{89429.2} & 2686.9 & 85.8 & 89892.4 & 2942.0 & 85.1\\
			Osaba\_80\_4 & \textbf{104403.5} & 1299.3 & 86.8 & 105135.7 & 1945.0 &  85.8 & 106692.7 & 1839.2 & 85.9\\
			\hline
			Osaba\_100\_1 & \textbf{107502.4} & 1489.2 & 176.3 & 109178.5 & 1536.6 & 174.6 & 109603.9 & 1690.6 & 173.3\\
			Osaba\_100\_2 & \textbf{100879.2} & 1639.1 & 175.0 & 101719.7 & 1639.4 & 173.6 & 101916.4 & 1719.8 & 174.4\\
			Osaba\_100\ 3 & \textbf{95327.4} & 1502.7 & 177.5& 95635.9 & 2706.8 & 172.0 & 95883.5 & 2449.0 & 172.4\\
			\hline
		\end{tabular}
	}
	\caption{Resultados del GB, ESA y EA para el AC-VRP-SPDVCFP propuesto.}
	\label{tab:ResultsAC-VRP-SPDVCFP}
\end{table}

Analizando los resultados representados en la tabla \ref{tab:ResultsAC-VRP-SPDVCFP} la primera conclusión lógica que puede extraerse es la siguiente: el GB supera claramente al resto de algoritmos en términos de resultados. Concretamente, el GB rinde mejor que el ESA en el 86.66\% de las instancias (13 de 15), y en el 100\% de los casos en comparación con el EA. Otro factor que merece la pena mencionar es la robustez propia del GB. Esta característica, sobre la cual ya se ha hablado en secciones anteriores, es de vital importancia si el algoritmo es aplicado en entornos reales. En este aspecto, el GB también demuestra ser el algoritmo más robusto y fiable en este caso, mostrando unas desviaciones típicas inferiores en la mayoría de las instancias.

Además de esto, con la intención de obtener conclusiones rigurosas y justas, se han llevado a cabo dos exámenes estadísticos diferentes. Las pautas para realizar esta valoración estadística han sido tomadas del trabajo realizado por Derrac et al. \cite{derrac2011practical}. En primer lugar, se ha llevado a cabo el test no paramétrico de Friedman para comparación múltiple. Con este test se ha querido comprobar si existen diferencias significativas entre las 3 técnicas. En la tabla \ref{tab:results_friedman} se muestra el ranking promedio obtenido por cada técnica para este este test (cuanto menor sea el valor, mejor el rendimiento). El estadístico resultante de este examen ha sido 17.73. Teniendo en cuenta un nivel de confianza del 99\%, el punto crítico de una distribución $\chi^2$ con dos grados de libertad es 9.21. Con todo esto, y debido a que 17.73$>$9.21, se puede concluir que existen diferencias sustanciales entre las tres meta-heurísticas, siendo el GB la que ha obtenido el mejor ranking. Finalmente, el valor $p$ computado en este test ha sido 0.000141.

Para evaluar la significancia estadística de esta mejora por parte del GB se ha llevado a cabo el test post-hoc de Holm, utilizando el GB como algoritmo de control. Los valores $p$ ajustados y no ajustados de este examen pueden verse en la tabla \ref{tab:results_holms}. Analizando estos datos, y teniendo en cuenta que todos los valores $p$ son menores que 0.05, se puede confirmar que el GB obtiene resultados significativamente mejores que el ESA y el EA con un nivel de confianza del 99\%.

\begin{table}[tbh]
	\centering
	\scalebox{0.7}{
		\begin{tabular}{|c|c|}\hline
			Algoritmo&Ranking promedio\\\hline
			GB&1.2\\
			ESA&2.0667\\
			EA&2.7333\\
			\hline
		\end{tabular}
	}
	\caption{Rankings promedio obtenidos por el test no-paramétrico de Friedman para los algoritmo GB, ESA y EA.}
	\label{tab:results_friedman}
\end{table}

\begin{table}[tbh]
	\centering
	\scalebox{0.7}{
		\begin{tabular}{|c|c c|}\hline
		Algoritmo & $p$ no ajustado& $p$ ajustado\\
		\hline 
		ESA&0.01762&0.017622\\ 
		EA&0.000027&0.000054\\\hline
		\end{tabular}
	}
	\caption{Valores $p$ ajustados y no ajustados obtenidos mediante la realización post-hoc del test de Holm. Se ha utilizado al GB como algoritmo de control.}
	\label{tab:results_holms}
\end{table}



\begin{savequote}[40mm]
El éxito no es definitivo, el fracaso no es fatídico. Lo que cuenta es el valor para continuar.
\qauthor{Winston Churchill}
\end{savequote}

\chapter{Conclusiones y líneas futuras}
\label{cha:Conclusiones}

\graphicspath{ {6_conclusion/figures/} }


\lettrine{A}{}{} lo largo del presente documento se han descrito paso a paso y de manera detallada las diferentes actividades realizadas durante el desarrollo de esta tesis doctoral. En el primer capítulo se ha realizado una breve introducción en la que se han afincado los detalles más básicos: contextualización, objetivos, hipótesis, metodología de la investigación, aportaciones... En los apartados siguientes se ha tratado de introducir al lector en el complejo mundo de la optimización combinatoria y los problemas de asignación de rutas a vehículos, en un primer momento, y en la vasta área de las técnicas de resolución posteriormente. Tras el minucioso análisis de la literatura relacionada, se ha procedido a la descripción del modelo propuesto para verificar la hipótesis planteada. Posteriormente, se ha dado paso a la experimentación correspondiente, cuyas conclusiones han derivado en la consecución de los objetivos propuestos y la justificación de la hipótesis impuesta.

Después de haber realizado todo este trabajo, en este último capítulo se detallarán las conclusiones y las líneas futuras que se han podido extraer de la investigación llevada a cabo. En un primer momento se pormenorizarán las conclusiones generales (sección \ref{sec:conc}), tanto las relacionadas con la técnica propuesta, como las vinculadas a la tesis como concepto genérico. Tras esto, se detallarán las líneas futuras surgidas (sección \ref{sec:lineas}), todas ellas relacionadas con la meta-heurística presentada, las cuales ofrecen un horizonte investigador realmente prometedor.

\section{Aportaciones y conclusiones generales}
\label{sec:conc}

La principal contribución de esta tesis es la meta-heurística denominada Golden-Ball. Como se ha explicado en apartados anteriores, el GB es una meta-heurística multi-poblacional, cuyo funcionamiento se cimienta en diversos conceptos inspirados en el mundo del fútbol. Al tratarse de una técnica multi-poblacional, el GB trabaja con un conjunto de soluciones, llamadas ``jugadores", las cuales son distribuidas en diferentes subgrupos, o subpoblaciones, llamadas ``equipos". Cada uno de estos equipos evolucionan de manera autónoma, en un proceso calificado como ``entrenamiento", en el que los jugadores son modificados de forma individual con la intención de mejorar progresivamente. Es conveniente destacar que cada equipo tiene una función de entrenamiento diferente, por lo que cada jugador evolucionará de diferente forma dependiendo del equipo en el que se encuentre.

A su vez, los equipos se enfrentan entre sí en unos procesos denominados ``partidos", conformando de esta manera una liga convencional. Dependiendo del resultado obtenido en estos partidos, cada equipo ocupará una posición u otra en la tabla clasificatoria. Esta tabla será clave para el procedimiento llamado ``periodo de fichajes", en el que los equipos intercambian sus jugadores, partiendo con ventaja aquellos equipos posicionados en la parte alta de la tabla. Finalmente, los equipos que obtengan malos resultados de manera sucesiva cambiarán su entrenador. Dicho de otra manera, cambiarán la manera en la que modifican sus soluciones.

Respecto a la experimentación llevada a cabo, se han utilizados 6 problemas diferentes de optimización combinatoria, cuatro de los cuales pertenecen a la familia de problemas de asignación de rutas a vehículos, mientras que los dos restantes son problemas clásicos de la optimización combinatoria, el primero de ellos de satisfacción de restricciones (el NQP) y el segundo de diseño combinatorio óptimo (el BPP). En total, se han utilizado 91 instancias, en las que el rendimiento del GB ha sido comparado con el demostrado por cuatro técnicas diferentes: dos algoritmos genéticos uni-poblacionales, y dos algoritmos genéticos distribuidos. Además de esto, aparte de la comparación basada en parámetros convencionales de la estadística descriptiva, también se han llevado a cabo dos tests estadísticos: el $z$-test y el test de Friedman.

Es importante destacar que la comparación de resultados y la descripción de la técnica se han hecho siguiendo los pasos dictados por un conjunto de buenas prácticas propuestas por el autor de esta tesis, las cuales han sido recogidas en el apartado \ref{sec:practicas}.

Además de presentar la técnica y comprobar su eficiencia en problemas académicos típicos de la comunidad científica, en esta tesis se ha demostrado cómo el GB es una técnica apta para ser aplicada a problemas más complejos, directamente extraídos de situaciones pertenecientes al mundo real. En la sección \ref{cha:aplicacion} se han recogido dos casos de transporte reales, modelados ambos como problemas de asignación de rutas complejos, en los que el GB ha mostrado un rendimiento aceptable. Estos dos problemas están relacionados con el transporte bajo demanda de pasajeros y la logística empresarial de una compañía de distribución de periódicos. En este aspecto, la principal aportación no consiste sólo en la aplicación del GB a estos casos concretos, si no que a esto habría que añadir la formulación y tratamiento de los problemas, ya que para esta empresa se han diseñado dos novedosos problemas de asignación a rutas nunca antes empleados en la literatura.

Una vez detalladas todas las aportaciones presentadas en esta tesis, conviene recordar el primer objetivo personal expuesto en la sección \ref{sec:Hipotesis} y analizar su cumplimiento. En concreto, este objetivo no es otro que el de maximizar, en la mayor medida posible, la contribución a la comunidad científica. En este aspecto, a lo largo del desarrollo de esta tesis doctoral se ha conseguido contribuir con la publicación de más de una treintena de artículos en revistas de carácter científico y divulgativo y comunicaciones en congresos nacionales e internacionales. En las tablas \ref{tab:congresos} y \ref{tab:revistas} se muestra un resumen cuantitativo de estas publicaciones, tanto en congresos como en revistas\footnote{En este trabajo se consideran revistas de factor de impacto aquellas que en el momento de la publicación del artículo contaban con este}.

\begin{table}[tb]
	\centering
	\scalebox{0.9}{
		\begin{tabular}{| l || r | r || r |}
		    
			\hline Tipo de congreso & Como primer autor & Como coautor & \textbf{Total}\\
			\hline
			\hline
			Congresos internacionales & 13 & 3 & 16\\
			Congresos nacionales & 0 & 2 & 2\\
			\hline
			\hline
			\textbf{Total} & 13 & 5 & \textbf{18} \\
			\hline
		\end{tabular}
	}
	\caption{Resumen cuantitativo de la contribución científica en congresos.}
	\label{tab:congresos}
\end{table}

\begin{table}[tb]
	\centering
	\scalebox{0.9}{
		\begin{tabular}{| l || r | r || r|}
		  \hline  
			Tipo de revista & Como primer autor & Como coautor & \textbf{Total}\\
			\hline
			\hline
			Revistas con factor de impacto & 6 & 3 & 9\\
			Revistas sin factor de impacto & 3 & 1 & 4\\
			Revistas de carácter divulgativo & 1 & 0 & 1\\
			\hline
			\hline
			\textbf{Total} & 10 & 4 & \textbf{14} \\
			\hline
		\end{tabular}
	}
	\caption{Resumen cuantitativo de la contribución científica en revistas}
	\label{tab:revistas}
\end{table}

Como puede resultar lógico, pese a que todas ellas se centran en el campo de la optimización combinatoria y los problemas de asignación de rutas, las publicaciones conseguidas han girado en torno a diversos temas. Es por eso por lo que, además de los resúmenes cuantitativos arriba expuestos, conviene clasificar las temáticas tratadas y listar el número de contribuciones llevadas a cabo en cada una de ellas. En la Tabla \ref{tab:articuloportema} se muestra un resumen de estas características para todas la publicaciones realizadas como primer autor.

\newcolumntype{L}[1]{>{\raggedright\let\newline\\\arraybackslash\hspace{0pt}}m{#1}}
\newcolumntype{R}[1]{>{\raggedleft\let\newline\\\arraybackslash\hspace{0pt}}m{#1}}

\begin{table}[tb]
	\centering
	\scalebox{0.89}{
		\begin{tabular}{| L{8.9cm} || R{1.8cm} | R{1.8cm}  || R{1cm}  |}
		  \hline  
			Temática central de la publicación & Pub. en revistas & Pub. en congresos & \textbf{Total}\\
			\hline
			\hline
			Presentación de nuevas técnicas u operadores & 6 & 2 & 8\\
			\hline Análisis teórico-prácticos sobre meta-heurísticas & 2 & 5 & 7\\
			\hline Aplicaciones prácticas de meta-heurísticas & 0 & 4 & 4 \\ 
			\hline Buenas prácticas en el campo de las meta-heurísticas & 2 & 2 & 4\\
			\hline
			\hline
			\textbf{Total} & 10 & 13 & \textbf{23} \\
			\hline
		\end{tabular}
	}
	\caption{Resumen de la contribución científica clasificado por temática}
	\label{tab:articuloportema}
\end{table}

En lo referente al segundo objetivo personal, el relacionado con la elaboración de algoritmos transparentes con la intención de ayudar a su estudio y aplicación posterior, se han realizado algoritmos de sencilla compresión y en un lenguaje de programación accesible (Java). Para contribuir a la replicabilidad del estudio, el autor de esta tesis ofrece una versión básica del Golden Ball bajo demanda (vía mail), o bien a través de su página personal\footnote{http://paginaspersonales.deusto.es/e.osaba}. Con todo esto, puede decirse que este segundo objetivo personal también ha sido satisfecho.

Después de haber mencionado las principales aportaciones de esta tesis, las conclusiones que pueden extraerse en relación al trabajo realizado son las siguientes. Conviene aclarar que estas conclusiones han conducido al cumplimiento de la hipótesis y los objetivos planteados en la sección \ref{sec:Hipotesis}:

\begin{itemize}
	
	\item Como se ha razonado en la sección \ref{sec:analisis}, la experimentación realizada ha demostrado que la particular estrategia de migración de jugadores empleada, la estrategia utilizada para hacer que las soluciones modifiquen su vecindario, y el enfoque invertido respecto a los operadores de carácter individual y cooperativo son elementos que mejoran la calidad media de las soluciones obtenidas frente a las logradas por el resto de técnicas utilizadas. 

	\item De acuerdo con la experimentación llevada a cabo en este estudio, puede observarse cómo el GB mejora de forma amplia los resultados obtenidos por las demás técnicas empleadas. Esta mejora se manifiesta en los 6 problemas utilizados, siendo, además, estadísticamente significativa en la gran mayoría de los casos.
	
	\item En lo relacionado a la robustez, el GB destaca sobre el resto de alternativas, presentando unas desviaciones típicas menores. Dicho de otra manera, las soluciones propuestas por el GB se mueven en un rango pequeño, lo que aporta cierta fiabilidad a la técnica. Como bien se ha remarcado en apartados anteriores, esta característica es especialmente apreciada en situaciones del mundo real.
	
	\item En términos de tiempos de ejecución y convergencia, el GB ofrece un comportamiento prometedor, igualando o superando a las técnicas con las que se ha realizado la comparación. Este hecho resulta ser una ventaja para el GB, debido a que obtiene resultados mejores que el resto de meta-heurísticas realizando un esfuerzo computacional similar o menor.
	
	\item Según se ha podido ver en la sección \ref{cha:aplicacion}, el GB ha demostrado ser una técnica prometedora no sólo para problemas básicos y académicos de optimización combinatoria y asignación de rutas, sino que es una excelente alternativa para afrontar problemas más complejos. Estos problemas han sido extraídos de situaciones de transporte complejas, pertenecientes al mundo real.
	
	\item Se ha logrado contribuir de una manera notable a la comunidad científica mediante la publicación de más de una treintena de artículos de revista y comunicaciones de congreso.
	
\end{itemize}

Con todo esto, se puede afirmarse que la meta-heurística propuesta en esta tesis es una técnica competitiva en cuanto a rendimiento se refiere y original conceptualmente hablando, la cual ofrece unos resultados prometedores, presentando unos tiempos de ejecución admisibles y un comportamiento de convergencia y robustez destacables.

Finalmente, conviene mencionar que el GB ya ha tenido cierta repercusión en la comunidad científica. Prueba de ello es el artículo publicado recientemente por varios investigadores de la universidad de Mahanakorn, en Bangkok, Tailandia, en el que se presenta una versión del GB con un proceso de inicialización heurístico para la resolución de CVRP \cite{golTai}. Adicionalmente, el doctor Pichpibul ha publicado posteriormente otro trabajo en el que se aplica el GB a un problema real de planificación propuesto por una agencia de viajes de Chonburi, Tailandia \cite{pichpibul2015improving}. Además de estas dos publicaciones, se tiene constancia del interés en el GB por parte de varios grupos de investigación procedentes de varias universidades como la Xian University, en China, o la Karamanoglu Mehmetbey Üniversitesi, en Turquía. 

El hecho de que investigadores de diversas partes del mundo estén centrando sus investigaciones en el GB, aún siendo esta una meta-heurística de corta vida, no solo demuestra el atractivo que genera la propia técnica, si no que supone una inyección de moral para que el autor de esta tesis plantee un ambicioso plan de trabajo futuro.

\section{Líneas futuras de trabajo}
\label{sec:lineas}

Como ya se ha comentado a lo largo de este documento, el campo de la optimización combinatoria y los problemas de asignación de rutas a vehículos son un tema candente y en constante expansión dentro de la comunidad científica. Por esto, y debido al potencial de la técnica propuesta y a la posible contribución al avance científico, se ha trazado un plan de futuro, el cual va a ser descrito en esta sección. 

Es preciso comenzar mencionando que las líneas futuras de trabajo surgidas de esta tesis pueden dividirse en dos campos de diferente naturaleza. El primer campo se centra en el Golden Ball como meta-heurística, en su análisis y sus posibles mejoras. En relación a esto se han identificado las siguientes líneas de trabajo:

\begin{itemize}
	\item Como bien se ha mencionado en la sección \ref{sec:resultados}, el siguiente paso a la investigación expuesta en este trabajo podría ser la realización de un extenso estudio acerca de la parametrización del GB. Para llevar a cabo un estudio de esta naturaleza se ejecutará un elevado número de pruebas sobre un conjunto pequeño de problemas (como bien pueden ser el TSP y el CVRP), alterando en cada ocasión el valor asignado a cada uno de los parámetros del algoritmo, como el número de equipos o el número de jugadores por equipo. Con este estudio se pretende adquirir un mayor conocimiento del comportamiento de la técnica, pudiendo intuir en investigaciones posteriores la parametrización que obtendrá mejores resultados, o la configuración ideal para consumir el menor tiempo de ejecución posible.
	
	Asimismo, mediante la realización de este estudio podrá concluirse qué partes del algoritmo resultan más efectivas para el proceso de optimización y cuáles colaboran en menor medida, lo que podría conducir al siguiente hito en el trabajo futuro.		
	
	\item Como bien podrá intuir un lector experimentado en el campo, pese a que la meta-heurística presentada en este trabajo cuenta con varios puntos favorables, también cuenta con varios aspectos mejorables. Uno de ellos es la complejidad de la técnica. En comparación con otras técnicas, tanto trayectoriales de búsqueda simple (como el recocido simulado o la búsqueda tabú) como poblacionales (como el algoritmo genético o la optimización por partículas), el GB tiene un nivel de complejidad que podría ser catalogado como alto. Pese a ser fácilmente comprensible, como bien se ha justificado en la sección \ref{sec:futbol}, un programador poco familiarizado en el campo podría encontrarse con dificultades a la hora de implementar la técnica. Esto es debido a dos factores: el primero de ellos es la cantidad de operadores que hay que desarrollar para poder sacar mayor rendimiento de la técnica, mientras que el otro es el elevado número de pasos dentro del flujo de ejecución.
	
	Es por esto por lo que uno de los objetivos principales del trabajo futuro consiste en reducir la complejidad inherente al GB. Para ello, habrá que analizar los resultados obtenidos en el examen descrito en el punto anterior, y decidir qué apartados pueden ser, o bien modificados, o simplemente eliminados.

	\item En la presente tesis se ha comparado el rendimiento de la técnica propuesta con dos tipos de meta-heurísticas diferentes: dos algoritmos genéticos uni-poblacionales, y dos algoritmos genéticos distribuidos. El cotejo del GB con estas técnicas ha de considerarse justo, como bien se ha razonado en la sección \ref{sec:aportaciones}. Aun así, en la literatura actual existen multitud de meta-heurísticas de diferente naturaleza, muchas de las cuales distan mucho de la filosofía adoptada por el GB. Pese a esto, un trabajo futuro con un gran potencial científico es la comparación cualitativa del GB con diversas técnicas de este estilo. Varios de los métodos que podrían utilizarse para este propósito son el PPSO, el ICA o el PABC.
	
\end{itemize}

El otro campo que conforma el plan de trabajo futuro se centra en la posible aplicación del Golden Ball. En este aspecto se han detectado las siguientes líneas de investigación:

\begin{itemize}

	\item Como bien se ha apuntado en apartados previos, un área de investigación que goza de mucha atención por parte de la comunidad científica es la aplicación de meta-heurísticas a problemas de asignación de rutas a vehículos adaptados a situaciones del mundo real, dando como resultado los anteriormente descritos problemas multi-atributo. Hasta ahora se ha realizado una primera aproximación a este campo, como se ha recogido en el Capítulo \ref{cha:aplicacion}, habiendo modelizado y abordado con el GB varios problemas de estas características, y habiendo producido las primeras publicaciones en este ámbito. Aun así, y debido al creciente atractivo de este campo, resultaría interesante hacer un mayor énfasis en esta área. Para ello, habría que identificar situaciones reales relacionadas con el transporte y dotadas de un gran interés social, las cuales habría que modelizar como problemas multi-atributo de asignación de rutas a vehículos, y darles después un tratamiento adecuado haciendo uso del GB.
	
	\item A día de hoy, como bien puede comprobarse en este documento, el GB ha sido aplicado tan solo a problemas de asignación de rutas a vehículos, como son el TSP, ATSP, CVRP y VRPB, y a dos problemas clásicos de la optimización combinatoria, el NQP y el BPP. Como reto futuro se plantea abrir los horizontes de aplicación del Golden Ball en cuanto a tipos de problemas y tipos de optimización se refiere. Quizá la ampliación más ``cercana", por tratarse también de un problema de optimización combinatoria, podría ser la del problema de la programación de la producción discreta y sus múltiple variantes. Por otro lado, sería interesante desde el punto de vista científico plantear a largo plazo una aplicación de la técnica propuesta a problemas de optimización continua, campo que genera multitud de producción científica año tras año.
	
\end{itemize}


\backmatter


\appendix
\selectlanguage{spanish}

\chapter{Resultados completos para los problemas ATSP, VRPB, NQP y BPP}
\label{app:resultados}

\lettrine{C}{}{omo} bien se ha mencionado en la Sección \ref{sec:resultados} de este documento, en este apartado se mostrarán los resultados completos obtenidos por cada una de las técnicas para los problemas ATSP, VRPB, NQP y BPP. Con todo esto, en la Tabla \ref{tab:ATSP} se muestran los resultados obtenidos para el ATSP, mientras que los logrados para el VRPB son revelados en la Tabla \ref{tab:VRPB}. Por otro lado, el rendimiento de los algoritmos en relación al problema NQP pueden ser observado en la Tabla \ref{tab:NQP}. Finalmente, la Tabla \ref{tab:BPP} exhibe los resultados conseguidos para el BPP.

\begin{table}[htb]
	\centering
	\setlength{\tabcolsep}{0pt}
	\scalebox{0.7}{
		\begin{tabular}{| l || l l || l l | l l || l l | l l |}
		  \hline  
		  \multicolumn{1}{|c||}{Instancia} & \multicolumn{2}{c||}{Golden Ball} & \multicolumn{2}{c|}{$GA_1$} & \multicolumn{2}{c||}{$GA_2$} & \multicolumn{2}{c|}{$DGA_1$} &\multicolumn{2}{c|}{$DGA_2$}\\
			
			\hline\hline br17 (39) & \textbf{39.0} & ($\pm$0.0) & 39.2 & ($\pm$0.4) & 39.1 & ($\pm$0.2) & \textbf{39.0} & ($\pm$0.0) & \textbf{39.0} & ($\pm$0.0)\\
			$\dot{x}$ \& T & 39 & 0.1 & 39 & 0.1 & 39 & 0.1 & 39 & 0.1 & 39 & 0.1\\
			
			\hline ftv33 (1286) & \textbf{1329.2} & ($\pm$33.7) & 1412.5 & ($\pm$81.5) & 1540.3 & ($\pm$83.1) & 1403.71 & ($\pm$60.9) & 1416.8 & ($\pm$90.4)\\
			$\dot{x}$ \& T & 1286 & 0.2 & 1298 & 0.4 & 1407 & 0.2 & 1329 & 0.4 & 1286 & 0.4\\
			
			\hline ftv35 (1473) & \textbf{1509.5} & ($\pm$28.8) & 1609.1 & ($\pm$76.9) & 1678.3 & ($\pm$165.3) & 1606.8 & ($\pm$74.7) & 1598.3 & ($\pm$57.0)\\
			$\dot{x}$ \& T & 1473 & 0.2 & 1493 & 0.4 & 1509 & 0.2 & 1473 & 0.4 & 1533 & 0.4\\
			
			\hline ftv38 (1530) & \textbf{1580.4} & ($\pm$37.3) & 1676.1 & ($\pm$71.7) & 1709.1 & ($\pm$145.8) & 1703.6 & ($\pm$91.8) & 1699.4 & ($\pm$74.5)\\
			$\dot{x}$ \& T & 1530 & 0.3 & 1573 & 0.5 & 1575 & 0.3 & 1582 & 0.5 & 1580 & 0.4\\
			
			\hline p43 (5620) & \textbf{5620.6} & ($\pm$0.8) & 5627.7 & ($\pm$5.5) & 5626.9 & ($\pm$3.8) & 5625.9 & ($\pm$3.7) & 5624.8 & ($\pm$3.4)\\
			$\dot{x}$ \& T & 5620 & 0.3 & 5622 & 0.9 & 5622 & 0.4 & 5621 & 0.8 & 5622 & 0.4\\
			
			\hline ftv44 (1613) & \textbf{1695.1} & ($\pm$42.7) & 1787.1 & ($\pm$93.2) & 2071.5 & ($\pm$147.7) & 1832.6 & ($\pm$131.9) & 1835.0 & ($\pm$108.0)\\
			$\dot{x}$ \& T & 1634 & 0.4 & 1652 & 1.0 & 1814 & 0.4 & 1652 & 0.9 & 1659 & 0.6\\
			
			\hline ftv47 (1776) & \textbf{1862.2} & ($\pm$55.2) & 1961.4 & ($\pm$86.7) & 2526.2 & ($\pm$705.5) & 2020.2 & ($\pm$139.1) & 2038.2 & ($\pm$130.7)\\
			$\dot{x}$ \& T & 1776 & 0.5 & 1826 & 1.4 & 2131 & 0.6 & 1806 & 1.0 & 1854 & 0.8\\
			
			\hline ry48p (14422) & \textbf{14614.2} & ($\pm$164.5) & 15008.2 & ($\pm$348.6) & 14976.5 & ($\pm$259.7) & 15038.8 & ($\pm$381.9) & 14945.2 & ($\pm$178.8)\\
			$\dot{x}$ \& T & 14082 & 0.6 & 14556 & 1.6 & 14547 & 0.8 & 14544 & 1.8 & 14596 & 0.7\\
			
			\hline\hline ft53 (6905) & \textbf{7335.0} & ($\pm$204.7) & 8077.2 & ($\pm$344.9) & 9401.1 & ($\pm$632.6) & 8331.5 & ($\pm$462.9) & 7997.4 & ($\pm$232.2)\\
			$\dot{x}$ \& T & 6909 & 0.8 & 7503 & 1.8 & 8805 & 0.9 & 7476 & 1.7 & 7642 & 0.9\\
			
			\hline ftv55 (1608) & \textbf{1737.1} & ($\pm$73.2) & 1879.3 & ($\pm$110.7) & 2152.4 & ($\pm$312.5) & 2021.2 & ($\pm$153.4) & 1990.9 & ($\pm$109.4)\\
			$\dot{x}$ \& T & 1632 & 0.8 & 1704 & 1.4 & 1704 & 1.4 & 1725 & 1.7 & 1859 & 1.4\\
			
			\hline ftv64 (1839) & \textbf{2023.5} & ($\pm$93.4) & 2203.5 & ($\pm$129.5) & 3032.9 & ($\pm$226.8) & 2284.3 & ($\pm$163.2) & 2321.8 & ($\pm$141.3)\\
			$\dot{x}$ \& T & 1871 & 1.6 & 2012 & 2.1 & 2604 & 1.8 & 2083 & 3.2 & 2135 & 1.7\\
			
			\hline ftv70 (1950) & \textbf{2151.9} & ($\pm$83.9) & 2313.7 & ($\pm$145.2) & 3335.5 & ($\pm$330.2) & 2390.0 & ($\pm$127.0) & 2509.6 & ($\pm$140.4)\\
			$\dot{x}$ \& T & 2017 & 1.8 & 2105 & 2.7 & 2680 & 2.1 & 2133 & 2.5 & 2221 & 2.1\\
			
			\hline ft70 (38673) & \textbf{40135.9} & ($\pm$461.4) & 40416.0 & ($\pm$623.4) & 47067.0 & ($\pm$1647.2) & 40813.1 & ($\pm$746.0) & 41129.9 & ($\pm$823.5)\\
			$\dot{x}$ \& T & 39547 & 2.1 & 39409 & 3.2 & 44172 & 2.1 & 39467 & 2.6 & 40156 & 2.3\\
			
			\hline\hline kro124p (36230) & \textbf{38924.6} & ($\pm$1157.4) & 42259.0 & ($\pm$1813.8) & 44084.0 & ($\pm$1932.5) & 43408.1 & ($\pm$2020.3) & 41116.5 & ($\pm$1044.9)\\
			$\dot{x}$ \& T & 36547 & 7.4 & 39270 & 9.4 & 40078 & 8.8 & 39529 & 11.4 & 39106 & 7.8\\
			
			\hline ftv170 (2755) & \textbf{3873.4} & ($\pm$468.7) & 4214.8 & ($\pm$361.8) & 4210.1 & ($\pm$481.3) & 4367.0 & ($\pm$470.7) & 4252.4 & ($\pm$174.2)\\
			$\dot{x}$ \& T & 2755 & 41.2 & 3546 & 49.8 & 3618 & 43.5 & 4031 & 51.7 & 4012 & 39.8\\
			
			\hline rbg323 (1326) & \textbf{1494.2} & ($\pm$35.7) & 1601.0 & ($\pm$76.8) & 1596.1 & ($\pm$77.3) & 1584.7 & ($\pm$73.7) & 1614.7 & ($\pm$194.4)\\
			$\dot{x}$ \& T & 1436 & 120.3 & 1514 & 130.7 & 1501 & 124.9 & 1496 & 130.7 & 1496 & 124.9\\
			
			\hline rbg358 (1163) & \textbf{1364.8} & ($\pm$40.1) & 1781.9 & ($\pm$62.5) & 1799.8 & ($\pm$66.2) & 1720.8 & ($\pm$175.0) & 1724.7 & ($\pm$189.7)\\
			$\dot{x}$ \& T & 1302 & 147.7 & 1667 & 158.1 & 1667 & 150.4 & 1369 & 164.8 & 1624 & 159.4\\
			
			\hline rbg403 (2465) & \textbf{2510.4} & ($\pm$29.6) & 3088.4 & ($\pm$199.6) & 3298.8 & ($\pm$378.1) & 2870.2 & ($\pm$194.5) & 2766.2 & ($\pm$138.4)\\
			$\dot{x}$ \& T & 2465 & 222.0 & 2908 & 227.4 & 2994 & 224.2 & 3501 & 235.1 & 2721 & 220.4\\
			
			\hline rbg443 (2720) & \textbf{2767.9} & ($\pm$17.5) & 3142.5 & ($\pm$219.3) & 3154.4 & ($\pm$242.5) & 2992.2 & ($\pm$125.6) & 2989.6 & ($\pm$128.1)\\
			$\dot{x}$ \& T & 2748 & 324.5 & 2960 & 335.9 & 2960 & 321.0 & 2960 & 335.9 & 2960 & 329.0\\
			
			\hline
		\end{tabular}
	}
	\caption{Resultados de los algoritmos $GB$, $GA_1$, $GA_2$, $DGA_1$ y $DGA_2$ para el ATSP. Para cada instancia se muestra el resultados promedio, desviación típica, mejor resultado obtenido y tiempo de ejecución promedio.}
	\label{tab:ATSP}
\end{table}

\begin{table}[htb]
	\centering
	\setlength{\tabcolsep}{2.5pt}
	\scalebox{0.7}{
		\begin{tabular}{| l || l l || l l | l l || l l | l l |}
		  \hline  
		  \multicolumn{1}{|c||}{Instancia} & \multicolumn{2}{c||}{Golden Ball} & \multicolumn{2}{c|}{$GA_1$} & \multicolumn{2}{c||}{$GA_2$} & \multicolumn{2}{c|}{$DGA_1$} &\multicolumn{2}{c|}{$DGA_2$}\\ 
			
			\hline C101 & \textbf{675.3} & ($\pm$39.1) & 722.3 & ($\pm$67.7) & 706.0  & ($\pm$40.2) & 739.1 & ($\pm$47.7) & 707.0 & ($\pm$65.6)\\
			$\dot{x}$ \& T & 589 & 3.3 & 613 & 10.4 & 640 & 2.9 & 640 & 12.6 & 629 & 3.4\\
			
			\hline C201 & \textbf{648.6} & ($\pm$44.1) & 852.3 & ($\pm$124.5) & 834.8 & ($\pm$75.3) & 795.8 & ($\pm$50.2) & 717.2 & ($\pm$133.7) \\
			$\dot{x}$ \& T & 564 & 1.1 & 678 & 1.2& 711 & 1.4& 722 & 2.4& 536 & 1.1\\
			
			\hline R101 & \textbf{895.8} & ($\pm$25.1) & 995.8 & ($\pm$80.9) & 946.1 & ($\pm$48.8) & 959.5 & ($\pm$43.9) & 903.9 & ($\pm$51.5) \\
			$\dot{x}$ \& T & 851 & 3.1 & 922 & 7.8& 873 & 2.5& 888 & 9.1& 811 & 3.1\\
			
			\hline R201 & \textbf{1047.6} & ($\pm$22.6) & 1270.0 & ($\pm$62.5) & 1137.2 & ($\pm$65.4) & 1188.9 & ($\pm$75.1) & 1085.7 & ($\pm$39.5)\\
			$\dot{x}$ \& T & 1007 & 7.0 & 1144 & 13.0& 1038 & 6.5& 1056 & 12.4& 908 & 6.8\\
			
			\hline RC101 & \textbf{583.3} & ($\pm$15.1) & 778.9 & ($\pm$118.9) & 660.2 & ($\pm$59.2) & 645.1 & ($\pm$66.3) & 626.3 & ($\pm$44.1) \\
			$\dot{x}$ \& T & 551 & 0.5 & 582 & 0.9& 555 & 0.9& 554 & 1.1& 549 & 1.1\\
			
			\hline RC201 & \textbf{1164.6} & ($\pm$41.6) & 1304.5 & ($\pm$76.5) & 1261.0 & ($\pm$87.9) & 1337.2 & ($\pm$60.1) & 1182.1 & ($\pm$63.8) \\
			$\dot{x}$ \& T & 1124 & 6.2 & 1147 & 13.2& 1127 & 5.9& 1242 & 12.4& 1072 & 6.1\\
			
			\hline\hline En23k3 & \textbf{696.8} & ($\pm$13.5) & 797.0 & ($\pm$67.8) & 748.4 & ($\pm$33.9) & 771.0 & ($\pm$49.3) & 702.6 & ($\pm$24.1) \\
		  $\dot{x}$ \& T & 676 & 0.5 & 639 & 0.9& 693 & 0.8& 703 & 0.8& 673 & 0.8\\
			
			\hline En30k4 & \textbf{509.6} & ($\pm$16.3) & 672.2 & ($\pm$51.7) & 630.7 & ($\pm$39.7) & 600.6 & ($\pm$56.6) & 593.3 & ($\pm$69.2) \\
			$\dot{x}$ \& T & 492 & 0.5 & 600 & 1.5& 557 & 1.3& 510 & 1.4& 498 & 0.8\\
			
			\hline En33k4 & \textbf{777.9} & ($\pm$30.7) & 851.7 & ($\pm$41.9) & 835.7 & ($\pm$47.3) & 833.6 & ($\pm$35.3) & 819.7 & ($\pm$28.7) \\
			$\dot{x}$ \& T & 725 & 0.6 & 791 & 1.7& 791 & 1.1& 768 & 1.2& 776 & 0.9\\
			
			\hline\hline En51k5 & \textbf{630.5} & ($\pm$20.7) & 716.8 & ($\pm$52.3) & 715.0 & ($\pm$46.5) & 721.5 & ($\pm$33.5) & 646.0 & ($\pm$35.6) \\
			$\dot{x}$ \& T & 592 & 2.0 & 645 & 2.6& 627 & 2.3& 637 & 2.5& 583 & 1.9\\
			
			\hline En76k8 & \textbf{830.7} & ($\pm$26.4) & 915.2 & ($\pm$43.1) & 913.3 & ($\pm$54.3) & 918.5 & ($\pm$74.0) & 871.4 & ($\pm$39.2) \\
			$\dot{x}$ \& T & 783 & 6.3 & 851 & 10.7& 848 & 6.1& 826 & 9.7& 814 & 5.8\\
			
			\hline En101k14 & \textbf{1088.0} & ($\pm$24.2) & 1183.8 & ($\pm$38.8) & 1164.8 & ($\pm$56.2) & 1231.9 & ($\pm$42.9) & 1191.4 & ($\pm$19.8) \\
			$\dot{x}$ \& T & 1048 & 22.0 & 1129 & 26.3& 1089 & 20.8& 1185 & 24.8& 1296 & 19.8\\
			
			\hline
		\end{tabular}
	}
	\caption{Resultados de los algoritmos $GB$, $GA_1$, $GA_2$, $DGA_1$ y $DGA_2$ para el VRPB. Para cada instancia se muestra el resultados promedio, desviación típica, mejor resultado obtenido y tiempo de ejecución promedio.}
	\label{tab:VRPB}
\end{table}

\begin{table}[htb]
	\centering
	\setlength{\tabcolsep}{2.5pt}
	\scalebox{0.7}{
		\begin{tabular}{| l || l l || l l | l l || l l | l l |}
		  \hline  
		  \multicolumn{1}{|c||}{Instancia} & \multicolumn{2}{c||}{Golden Ball} & \multicolumn{2}{c|}{$GA_1$} & \multicolumn{2}{c||}{$GA_2$} & \multicolumn{2}{c|}{$DGA_1$} &\multicolumn{2}{c|}{$DGA_2$}\\ 
			
			\hline 8-Queens & \textbf{0.0} & ($\pm$0.0) & \textbf{0.0} & ($\pm$0.0) & \textbf{0.0}  & ($\pm$0.0) & \textbf{0.0} & ($\pm$0.0) & \textbf{0.0} & ($\pm$0.0)\\
			$\dot{x}$ \& T & 0 & 0.1 & 0 & 0.1 & 0 & 0.1 & 0 & 0.1 & 0 & 0.1\\
			
			\hline 20-Queens & \textbf{0.1} & ($\pm$0.2) & 1.4 & ($\pm$0.6) & \textbf{0.1} & ($\pm$0.3) & 1.5 & ($\pm$1.1) & 0.8 & ($\pm$0.7) \\
			$\dot{x}$ \& T & 0 & 0.1  & 1 & 0.1 & 0 & 0.1 & 0 & 0.2 & 0 & 0.1\\
			
			\hline 50-Queens & \textbf{0.0} & ($\pm$0.0) & 5.3 & ($\pm$1.7) & 1.9 & ($\pm$0.7) & 5.0 & ($\pm$1.1) & 4.3 & ($\pm$1.6) \\
			$\dot{x}$ \& T & 0 & 0.7  & 2 & 0.8 & 1 & 0.8 & 3 & 1.1 & 2 & 0.8\\
			
			\hline 75-Queens & \textbf{0.1} & ($\pm$0.2) & 8.1 & ($\pm$1.6) & 4.6 & ($\pm$1.8) & 9.1 & ($\pm$1.7) & 6.1 & ($\pm$1.7)\\
			$\dot{x}$ \& T & 0 & 4.1  & 5 & 4.1 & 1 & 4.6 & 6 & 5.4 & 4 & 4.8\\
			
			\hline 100-Queens & \textbf{0.5} & ($\pm$0.7) & 13.6 & ($\pm$2.1) & 7.2 & ($\pm$1.7) & 12.0 & ($\pm$2.0) & 11.4 & ($\pm$3.0) \\
			$\dot{x}$ \& T & 0 & 5.8  & 9 & 6.8 & 4 & 7.2 & 9 & 10.1 & 4 & 11.0\\
			
			\hline\hline 125-Queens & \textbf{0.3} & ($\pm$0.4) & 16.4 & ($\pm$3.2) & 12.6 & ($\pm$2.4) & 16.2 & ($\pm$2.5) & 14.3 & ($\pm$2.4) \\
			$\dot{x}$ \& T & 0 & 13.4  & 11 & 15.8 & 8 & 14.8 & 12 & 18.4 & 10 & 14.8\\
			
			\hline 150-Queens & \textbf{1.7} & ($\pm$1.4) & 18.1 & ($\pm$3.2) & 17.0 & ($\pm$2.9) & 20.0 & ($\pm$3.2) & 19.0 & ($\pm$1.9) \\
		  $\dot{x}$ \& T & 0 & 16.7  & 10 & 18.4 & 11 & 16.5 & 13 & 20.6 & 16 & 16.5\\
			
			\hline 200-Queens & \textbf{3.3} & ($\pm$1.9) & 26.0 & ($\pm$3.9) & 24.5 & ($\pm$3.5) & 32.8 & ($\pm$4.8) & 23.4 & ($\pm$3.1) \\
			$\dot{x}$ \& T & 0 & 23.1  & 18 & 26.1 & 20 & 26.1 & 25 & 31.1 & 18 & 26.2\\
			
			\hline 225-Queens & \textbf{4.3} & ($\pm$1.7) & 31.9 & ($\pm$5.0) & 37.9 & ($\pm$3.2) & 38.4 & ($\pm$3.5) & 29.2 & ($\pm$4.3) \\
			$\dot{x}$ \& T & 1 & 35.4  & 19 & 41.5 & 34 & 31.2 & 31 & 31.2 & 23 & 35.8\\
			
			\hline 250-Queens & \textbf{3.5} & ($\pm$1.6) & 44.3 & ($\pm$3.9) & 32.7 & ($\pm$6.7) & 41.2 & ($\pm$5.3) & 32.0 & ($\pm$3.1) \\
			$\dot{x}$ \& T & 1 & 72.4  & 41 & 83.1 & 19 & 78.1 & 31 & 78.1 & 28 & 78.3\\
			
			275-Queens & \textbf{5.6} & ($\pm$3.0) & 50.0 & ($\pm$11.2) & 39.5 & ($\pm$4.9) & 44.1 & ($\pm$7.5) & 39.9 & ($\pm$4.9) \\
			$\dot{x}$ \& T & 1 & 101.6  & 36 & 104.2 & 29 & 102.5 & 31 & 107.6 & 32 & 104.7\\
			
			\hline\hline 300-Queens & \textbf{6.4} & ($\pm$2.6) & 61.9 & ($\pm$5.2) & 44.4 & ($\pm$5.3) & 52.8 & ($\pm$5.9) & 44.4 & ($\pm$5.9) \\
			$\dot{x}$ \& T & 2 & 131.0  & 52 & 132.9 & 37 & 130.9 & 45 & 134.5 & 34 & 128.4\\
			
			\hline 325-Queens & \textbf{4.8} & ($\pm$2.4) & 63.5 & ($\pm$5.6) & 47.4 & ($\pm$6.4) & 54.4 & ($\pm$3.6) & 49.1 & ($\pm$4.1) \\
			$\dot{x}$ \& T & 1 & 215.6  & 56 & 225.3 & 37 & 220.7 & 46 & 228.7 & 44 & 218.1\\
			
			\hline 350-Queens & \textbf{5.1} & ($\pm$3.0) & 71.4 & ($\pm$5.6) & 51.0 & ($\pm$4.7) & 65.5 & ($\pm$5.7) & 49.9 & ($\pm$5.8) \\
			$\dot{x}$ \& T & 2 & 275.3  & 63 & 286.7 & 44 & 281.2 & 59 & 289.6 & 40 & 278.5\\
			
			\hline 400-Queens & \textbf{4.3} & ($\pm$2.2) & 59.9 & ($\pm$10.1) & 54.0 & ($\pm$9.7) & 59.4 & ($\pm$8.1) & 56.1 & ($\pm$7.6) \\
			$\dot{x}$ \& T & 1 & 359.7  & 42 & 371.8 & 42 & 365.7 & 50 & 379.5 & 44 & 357.8\\
			
			\hline
		\end{tabular}
	}
	\caption{Resultados de los algoritmos $GB$, $GA_1$, $GA_2$, $DGA_1$ y $DGA_2$ para el NQP. Para cada instancia se muestra el resultados promedio, desviación típica, mejor resultado obtenido y tiempo de ejecución promedio.}
	\label{tab:NQP}
\end{table}

\begin{table}[htb]
	\centering
	\setlength{\tabcolsep}{2.5pt}
	\scalebox{0.7}{
		\begin{tabular}{| l || l l || l l | l l || l l | l l |}
		  \hline  
		  \multicolumn{1}{|c||}{Instancia} & \multicolumn{2}{c||}{Golden Ball} & \multicolumn{2}{c|}{$GA_1$} & \multicolumn{2}{c||}{$GA_2$} & \multicolumn{2}{c|}{$DGA_1$} &\multicolumn{2}{c|}{$DGA_2$}\\ 
			
			\hline N1C1W1\_A (25) & \textbf{26.0} & ($\pm$0.0) & 26.5 & ($\pm$0.5) & 26.7 & ($\pm$0.4) & 26.8 & ($\pm$0.5) & 26.7 & ($\pm$0.5)\\
			$\dot{x}$ \& T & 26 & 0.2 & 26 & 0.2 & 26 & 0.1 & 26 & 0.3 & 26 & 0.3 \\
			
			\hline N1C1W1\_B (31) & \textbf{31.0} & ($\pm$0.0) & 31.9 & ($\pm$0.4) & 31.5 & ($\pm$0.5) & 31.5 & ($\pm$0.5) & 31.6 & ($\pm$0.6) \\
			$\dot{x}$ \& T & 31 & 0.2 & 31 & 0.2 & 31 & 0.2 & 31 & 0.4 & 31 & 0.3 \\
			
			\hline  N1C2W1\_A (21) & \textbf{21.1} & ($\pm$0.2) & 21.9 & ($\pm$0.5) & 21.9 & ($\pm$0.5) & 21.8 & ($\pm$0.4) & 22.0 & ($\pm$0.4) \\
			$\dot{x}$ \& T & 21 & 0.2 & 21 & 0.3 & 21 & 0.2 & 21 & 0.4 & 21 & 0.3 \\
			
			\hline N1C2W1\_B (26) & \textbf{26.1} & ($\pm$0.2) & 27.6 & ($\pm$0.5) & 27.1 & ($\pm$0.4) & 26.8 & ($\pm$0.4) & 26.8 & ($\pm$0.5)\\
			$\dot{x}$ \& T & 26 & 0.3 & 26 & 0.3 & 26 & 0.3 & 26 & 0.3 & 26 & 0.3 \\
			
			\hline\hline N2C1W1\_A (48) & \textbf{51.0} & ($\pm$0.3) & 53.1 & ($\pm$0.6) & 52.4 & ($\pm$0.6) & 52.9 & ($\pm$0.6) & 52.2 & ($\pm$0.7) \\
			$\dot{x}$ \& T & 50 & 1.8 & 51 & 1.7 & 51 & 1.4 & 52 & 1.8 & 51 & 1.4 \\
			
			\hline N2C1W1\_B (49) & \textbf{51.4} & ($\pm$0.5) & 52.6 & ($\pm$0.6) & 53.0 & ($\pm$0.8) & 53.3 & ($\pm$0.8) & 52.8 & ($\pm$0.6) \\
			$\dot{x}$ \& T & 51 & 1.8 & 51 & 1.9 & 52 & 1.5 & 52 & 1.8 & 52 & 1.4 \\
			
			\hline\hline N2C2W1\_A (42) & \textbf{43.9} & ($\pm$0.2) & 44.6 & ($\pm$0.6) & 45.4 & ($\pm$0.5) & 45.7 & ($\pm$0.6) & 45.3 & ($\pm$0.6) \\
		  $\dot{x}$ \& T & 43 & 1.8 & 44 & 1.8 & 45 & 1.7 & 44 & 1.9 & 44 & 1.7 \\
			
			\hline N2C2W1\_B (50) & \textbf{51.4} & ($\pm$0.5) & 52.4 & ($\pm$0.6) & 53.1 & ($\pm$0.7) & 53.4 & ($\pm$0.6) & 53.2 & ($\pm$0.6) \\
			$\dot{x}$ \& T & 51 & 2.1 & 51 & 1.9 & 52 & 1.8 & 53 & 1.9 & 52 & 1.5 \\
			
			\hline\hline N3C2W2\_A (107) & \textbf{114.1} & ($\pm$1.1) & 121.8 & ($\pm$1.3) & 118.7 & ($\pm$1.5) & 120.0 & ($\pm$1.4) & 118.0 & ($\pm$1.3) \\
			$\dot{x}$ \& T & 112 & 15.0 & 115 & 14.8 & 117 & 13.5 & 117 & 15.2 & 115 & 14.1 \\
			
			\hline N3C2W2\_B (105) & \textbf{109.6} & ($\pm$0.5) & 119.8 & ($\pm$1.5) & 113.4 & ($\pm$1.1) & 115.3 & ($\pm$1.8) & 111.9 & ($\pm$0.7) \\
			$\dot{x}$ \& T & 109 & 17.1 & 109 & 16.5 & 111 & 16.1 & 112 & 15.4 & 111 & 14.9 \\
			
			\hline N3C3W1\_A (66)  & \textbf{70.2} & ($\pm$0.5) & 74.6 & ($\pm$0.7) & 71.5 & ($\pm$0.7) & 72.6 & ($\pm$0.9) & 71.4 & ($\pm$0.8) \\
			$\dot{x}$ \& T & 69 & 12.2 & 70 & 12.9 & 70 & 12.1 & 71 & 14.8 & 70 & 13.8 \\
			
			\hline N3C3W1\_B (71) & \textbf{76.1} & ($\pm$0.5) & 78.4 & ($\pm$0.6) & 77.4 & ($\pm$0.9) & 78.6 & ($\pm$1.0) & 77.6 & ($\pm$1.0) \\
			$\dot{x}$ \& T & 75 & 12.1 & 76 & 13.1 & 76 & 12.7 & 77 & 15.7 & 76 & 14.5 \\
			
			\hline\hline N4C1W1\_A (240) & \textbf{260.5} & ($\pm$1.5) & 271.6 & ($\pm$2.5) & 268.4 & ($\pm$3.8) & 270.1 & ($\pm$2.4) & 267.7 & ($\pm$2.1) \\
			$\dot{x}$ \& T & 258 & 194.7 & 268 & 187.4 & 265 & 181.0 & 268 & 200.7 & 263 & 199.9 \\
			
			\hline N4C2W1\_A (210) & \textbf{231.2} & ($\pm$1.2) & 239.1 & ($\pm$1.6) & 233.3 & ($\pm$5.2) & 241.0 & ($\pm$1.9) & 235.4 & ($\pm$1.3) \\
			$\dot{x}$ \& T & 229 & 195.8 & 235 & 188.5 & 219 & 186.4 & 238 & 203.2 & 233 & 200.1 \\
			
			\hline N4C2W1\_B (213) & \textbf{233.3} & ($\pm$1.6) & 241.5 & ($\pm$2.4) & 234.3 & ($\pm$4.7) & 243.6 & ($\pm$1.9) & 239.1 & ($\pm$0.7) \\
			$\dot{x}$ \& T & 230 & 190.5 & 237 & 186.2 & 224 & 184.2 & 241 & 198.6 & 238 & 195.4 \\
			
			\hline N4C2W1\_C (213) & \textbf{234.5} & ($\pm$1.6) & 241.7 & ($\pm$1.8) & 239.7 & ($\pm$6.8) & 241.3 & ($\pm$2.0) & 238.1 & ($\pm$1.9) \\
			$\dot{x}$ \& T & 231 & 199.8 & 238 & 194.2 & 218 & 191.4 & 241 & 201.5 & 235 & 198.3 \\
			
			\hline
		\end{tabular}
	}
	\caption{Resultados de los algoritmos $GB$, $GA_1$, $GA_2$, $DGA_1$ y $DGA_2$ para el BPP. Para cada instancia se muestra el resultados promedio, desviación típica, mejor resultado obtenido y tiempo de ejecución promedio.}
	\label{tab:BPP}
\end{table}	


\selectlanguage{british}

\begin{savequote}[40mm]
Success is not final, failure is not fatal. It is the courage to continue that counts.
\qauthor{Winston Churchill}
\end{savequote}

\hyphenation{lo-ca-tion com-mu-ni-ca-tions prob-a-bil-i-ty in-for-ma-tion syn-chro-nization re-sults ad-van-tage the-o-ret-i-cally}

\chapter{Conclusions and further work}
\label{cha:Conclusions}

\graphicspath{ {6_conclusion/figures/} }


\lettrine{T}{}{hroughout} this whole document the different activities performed during the development of this thesis have been described in detail. In the first chapter a brief introduction has been done. In this introduction the basic aspects of this thesis have been established: contextualization, objectives, hypothesis, research methodology, main contributions... In following sections, at first, an extensive state of the art in combinatorial optimization and vehicle routing problems has been performed. After that, another state of the art has been developed. In this case, regarding the solving techniques related to these combinatorial and routing problems. After this careful analysis of the related literature, it has proceeded with the description of proposed model. Subsequently, the experimentation carried out has been detailed. The conclusions drawn from this experimentation have led to the achievement of the main objectives of this thesis and the ratification of the outlined hypothesis.

Having done all this work, the main conclusions and possible future works are detailed in this last chapter. At first, the general conclusions are specified (Section \ref{sec:concIng}), both those connected with the proposed meta-heuristic, such as those related to the thesis as a generic concept. After that, the possible future lines are described (Section \ref{sec:lines}), all of them related to the presented technique. These lines offer a very promising research horizon.

\section{General conclusions}
\label{sec:concIng}

The main contribution of this thesis is the meta-heuristic called Golden-Ball. As has been explained in previous sections, the GB is a multi-population meta-heuristic, whose operation is based on several soccer concepts. Being a multi-populational technique, GB works with a set of solutions, each one called ``player", which are distributed in different subgroups, or subpopulations, referred as ``team". Each of these teams evolves autonomously, in a process named ``training", in which players are modified individually with the intention of improving them gradually. It is noteworthy that each team has a different training function, which means that each player evolves differently depending on its team.

At the same time, teams face each other in a process called ``matches", forming a conventional league. Depending on the results obtained in these matches, each team will be in one position or another in the league standing. This classification is crucial to the procedure named ``transfer period", in which teams exchange their players. Finally, the teams that perform poorly change their coach, in other words, they change the way they modify their solutions.

Regarding the experimentation carried out, 6 different combinatorial optimization problems have been used. Four of these are routing problems, while the remaining two are classic combinatorial optimization problems, the first one a constraint satisfaction problem (NQP) and the second an optimal combinatorial design one (BPP). In total, 91 instances have been used. The performance of the GB has been compared with the one shown by four different techniques: two uni-population genetic algorithms, and two distributed genetic algorithms. Besides this, with the aim of obtaining rigorous conclusions, two different statistical tests have been performed: the $z$ -test and the Friedman test.

It is important to highlight that the results comparison and the description of the technique have been performed following the steps dictated by a set of good practices proposed by the author of this thesis. These good practices have been introduced in section \ref{sec:practicas}.

Besides this, in this thesis how the GB can be applied to more complex problems has been also shown. These complex problems have been directly extracted from real world situations. In section \ref{cha:aplicacion} two different transportation problems have been described. Both of them have been modeled as complex routing problems, for which the GB has shown acceptable performance. These two problems are related to passenger on-demand transportation and the logistics of a newspaper distribution company. In this aspect, the main contribution is not only the application of the GB to these problems. It should be added the formulation and the treatment of them, because it is the first time that such problems are addressed in the literature.

Once the main contributions of this work have been detailed, it is appropriate to analyze the compliance of the first personal objective proposed in Section \ref{sec:Hipotesis}. Specifically, the objective is to maximize the contribution to the scientific community. Throughout the development of this doctoral thesis, the author has been able to contribute significantly to the scientific community, with the publication of around thirty papers in national and international journals and conferences. A quantitative summary of these publications is shown in Table \ref{tab:conferences} for conferences, and in Table \ref{tab:journals} for journals.

\begin{table}[tb]
	\centering
	\scalebox{0.9}{
		\begin{tabular}{| l || r | r || r |}
		    
			\hline Kind of conference & as corresponding author & as coauthor & \textbf{Total}\\
			\hline
			\hline
			International conferences & 13 & 3 & 16\\
			National conferences & 0 & 2 & 2\\
			\hline
			\hline
			\textbf{Total} & 13 & 5 & \textbf{18} \\
			\hline
		\end{tabular}
	}
	\caption{Quantitative summary of the scientific contribution in conferences.}
	\label{tab:conferences}
\end{table}

\begin{table}[tb]
	\centering
	\scalebox{0.9}{
		\begin{tabular}{| l || r | r || r|}
		  \hline  
			Kind of journal & as corresponding author & as coauthor & \textbf{Total}\\
			\hline
			\hline
			Journals with IF & 6 & 3 & 9\\
			Journals without IF & 3 & 1 & 4\\
			Informative journals & 1 & 0 & 1\\
			\hline
			\hline
			\textbf{Total} & 10 & 4 & \textbf{14} \\
			\hline
		\end{tabular}
	}
	\caption{Quantitative summary of the scientific contribution in journals.}
	\label{tab:journals}
\end{table}

As can be logical, even though they all are focused on the field of combinatorial optimization and routing problems, the published papers move around various topics. It is for this reason that, in addition to the quantitative summaries outlined above, it is convenient to classify the topics that has been addressed in those works, and list the number of publicatons carried out in each of them. In Table \ref{tab:paperTopic} a summary is shown for all the papers published as corresponding author.

\begin{table}[t]
	\centering
	\scalebox{0.89}{
		\begin{tabular}{| L{8.9cm} || R{1.8cm} | R{1.8cm}  || R{1cm}  |}
		  \hline  
			Main topic of the paper & P. in journals & P. in conferences & \textbf{Total}\\
			\hline
			\hline
			Presenting new techniques or operators & 6 & 2 & 8\\
			\hline Theoretical and practical analysis of meta-heuristics & 2 & 5 & 7\\
			\hline Practical applications of meta-heuristics & 0 & 4 & 4 \\ 
			\hline Good practices in the field of meta-heuristics & 2 & 2 & 4\\
			\hline
			\hline
			\textbf{Total} & 10 & 13 & \textbf{23} \\
			\hline
		\end{tabular}
	}
	\caption{Summary of the scientific contribution classified by topic}
	\label{tab:paperTopic}
\end{table}

Regarding the second personal objective, which is related to the development of transparent algorithms in order to facilitate their replicability, it can be said that simple compression algorithms have been developed throughout all this work. In addition, the programming language used for the development of these algorithms is the well-known Java. With the aim of contributing to the replication of this study, the author of this thesis provides a basic version of the Golden Ball on demand (via mail), or through his personal page\footnote{http://paginaspersonales.deusto.es/e.osaba}.

Having mentioned the main contributions of this thesis, the conclusions drawn in relation to the work performed are the following.

\begin{itemize}
	\item The GB provides several originalies compared with the existing techniques. Among them, the original migration strategy or the strategy of neighborhood changing process could be highlighted. It can also emphasize the way in which the GB prioritizes the individual improvement, leaving the cooperative one in a background
\end{itemize}

\begin{itemize}
	\item As has been argued in section \ref{sec:analisis}, the experimentation has shown that the particular migration strategy, the procedure used to make solutions change their neighborhood, and the reverse approach about individual and cooperative operators help to improve the average quality of the solutions obtained.

	\item According to the experimentation carried out in this study, it can be seen how the GB broadly improves the results obtained by other techniques used. This improvement is manifested in all the 6 problems used, and it is statistically significant in most of the cases.
	
	\item In regard to the robustness, the GB algorithm stands above the rest of alternatives, presenting lower standard deviations. In other words, the quality of the solutions proposed by the GB moves in a narrow range. This characteristic gives robustness and reliability to the algorithm, something crucial if the technique is applied a real environment.
	
	\item In terms of runtimes and convergence, GB offers a promising performance, matching or beating the other techniques used in the experimentation. This is an advantage for the GB, because it can get better results needing a similar or lower computational effort.
	
	\item As has been seen in section \ref{cha:aplicacion}, the GB has proven to be a promising technique not only for basic and academic combinatorial optimization and routing problems, but also for more complex problems. These problems have been drawn from complex transportation situations, belonging to the real world.
	
	\item It can be said that the contribution to the scientific community has been significant, publishing more than thirty journal papers and conference communications.	
\end{itemize}

With all this, it is prudent to conclude that the meta-heuristic proposed in this work is a competitive technique in terms of conceptual originality and performance. The algorithm offers promising results, needing admissible runtimes and showing a remarkable convergence behavior and robustness.

Finally, it is worth mentioning that the GB has already had some impact on the scientific community. Proof of this is the paper published recently by some researchers of the University of Mahanakorn, in Bangkok, Thailand. In this work a version of the GB is presented for solving the CVRP using a heuristic initialization process \cite{golTai}. Additionally, Pichpibul published recently another paper in which the GB is used to address a real-world problem proposed by a travel agency operating in Chonburi, Thailand \cite{pichpibul2015improving}. Besides this publication, there are evidences of interest in the GB by several research groups from various universities of the world, such as Xi'an University in China, or Karamanoglu Mehmetbey University, in Turkey.

The fact that some researchers around the world are focusing their research in the GB meta-heuristics not only demonstrates the attractiveness generated by the technique itself, but it is also a boost for the author of this thesis. In this way, it has been designed an ambitious plan for future work, which is going to be described in the following section.

\section{Future lines of work}
\label{sec:lines}

As has been mentioned throughout this document, the research fields of combinatorial optimization and routing problems are a hot topic in the scientific community, and they are in constant expansion. That is the reason why it could be considered a mistake the fact of putting an end to the work developed in this thesis. Therefore, and because of the potential of the presented meta-heuristic, it has been drawn up a plan for the future work, which will be described in this section.

In this way, future lines arising from this thesis can be divided into two different areas. The first area is focused on the Golden Ball as a meta-heuristic, in its analysis and possible improvements. In this context, the following activities have been planned:

\begin{itemize}
	\item As has been mentioned in Section \ref{sec:resultados}, the next step in the research presented in this thesis could be to conduct an extensive study on the parameterization of GB.To perform such study a high number of tests will be carried out on a small set of problems (as the TSP or the CVRP), altering the values of some parameters of the algorithm, as the number of teams or the number of players per team. The objective of this study is to perfectly understand the behavior of the meta-heuristic. In this way, it could be intuited which is the best configuration to get good results in future developments, or it could be predicted the ideal configuration to consume the least possible runtime.
		
	Furthermore, by conducting this study it may be concluded what parts of the algorithm are more effective, and which ones collaborate lesser in the optimization process. These conclusions lead directly to the next milestone in the planned future work.	
	
	\item Any experienced reader in this field can guess that, although the GB has several strong points, it also has several areas for improvement. One of them is the complexity of the technique. Compared with other techniques, the GB has a level of complexity that could be classified as high. Despite being easily understandable, as has been justified in Section \ref{sec:futbol}, a novice programmer could have some difficulties implementing the technique. This is due to two factors, the first is the large number of operators needed to achieve the best performance of the technique, while the other is the high number of steps in the execution flow.
	
	It is for this reason that one of the main objectives of future work is to reduce the inherent complexity of the GB. To do this, the results obtained from the experimentations described in the previous section have to be analyzed, in order to decide which steps can be either modified or eliminated.	
	
	\item In this thesis the performance of the proposed technique has been compared with two sorts of meta-heuristics: two basic genetic algorithms, and two distributed genetic algorithms. The comparison of GB with these techniques should be considered fair, as have been reasoned in Section \ref{sec:aportaciones}. Still, many different meta-heuristics can be found in the literature, many of which are very far from the philosophy adopted by the GB. Despite this, one possible future work with a great scientific potential could be the qualitative comparison between the GB and some other techniques. Several methods could be used for this purpose, such as the PPSO, ICA or PABC.
		
\end{itemize}

The other field that shapes the future work plan is focused on the application of the Golden Ball. In this aspect, the following research lines have been detected:

\begin{itemize}

	\item As have been referred in previous sections, one interesting research area is the application of meta-heuristics to complex routing problems adapted to real world situations. A first approach to this field has been made, as have been described in Chapter \ref{cha:aplicacion}, having modeled and addressed with the GB some problems of this kind. Additionally, the first publications in this field have been produced. Nevertheless, due to its growing scientific interest, it would be interesting to make a greater emphasis on this area. To reach this goal, some real situations related to transport and logistic should be identified, which must be modeled as multi-attribute vehicle routing problems, in order to give them an adequate treatment with the GB.
	
	\item Until now, as can be seen in this thesis, the GB has been applied to some vehicle routing problems, such as TSP, ATSP, CVRP and VRPB, and two classical combinatorial optimization problems, the first one a constraint satisfaction problem (the NQP), and the other one a combinatorial design problem (BPP).
	
	In this way, the application of GB to some other optimization problems has been planned as future challenge. Perhaps, the following step,  since it is also a combinatorial optimization problem, could be the application of the GB to the Job Shop Scheduling Problem problem. On the other hand, it would be interesting from the scientific point of view the application of the proposed technique to continuous optimization problems. This concrete field generates many scientific production every year.
	
\end{itemize}

\selectlanguage{spanish}

\cleardoublepage

\begin{dedication} 

\raggedleft
\textit{<<El peor mal del hombre es la irreflexión.>>}
\newline
\textit{Sófocles}
	
\end{dedication}


\bibliography{7_referencias/tesis_refs} 




\end{document}